\documentclass[lettersize,journal]{IEEEtran}
\usepackage{amsmath,amsfonts}
\usepackage{algorithmic}
\usepackage{array}
\usepackage{textcomp}
\usepackage{stfloats}
\usepackage{url}
\usepackage{verbatim}
\usepackage{graphicx}

\usepackage{microtype}
\usepackage{graphicx}
\usepackage{booktabs} 
\usepackage{hyperref}
\usepackage[table]{xcolor}
\usepackage{rotating}
\usepackage{xcolor}
\definecolor{tab_others}{RGB}{225, 235, 246}
\definecolor{tab_ours}{RGB}{240, 240, 240}
\definecolor{mygreen}{RGB}{0,90,0}
\definecolor{myyellow}{RGB}{208,158,0}

\usepackage{amsmath}
\usepackage{amssymb}
\usepackage{mathtools}
\usepackage{amsthm}
\usepackage{diagbox}
\usepackage[capitalize,noabbrev]{cleveref}
\usepackage{multirow}
\usepackage{graphicx}
\usepackage{placeins}

\usepackage{algorithm}
\usepackage[caption=false,font=scriptsize,labelfont=rm,textfont=rm]{subfig}
\usepackage{cite}
\usepackage{makecell}
\usepackage[mathscr]{eucal}
\usepackage{float}   
\usepackage{subfloat}
\usepackage{ragged2e} 
\usepackage{bbm}
\usepackage{bm}
\usepackage{hyperref}
\hypersetup{hypertex=true,
	colorlinks=true,
	linkcolor=blue,
	anchorcolor=blue,
	citecolor=blue}
\usepackage{orcidlink}
\usepackage{tikz}
\usepackage{mathrsfs}
\usepackage{dsfont}

\definecolor{cvprblue}{RGB}{232, 241, 250}
\definecolor{cvprgreen}{RGB}{238, 250, 242}
\definecolor{cvprgold}{RGB}{255, 248, 220}
\definecolor{lightgreen}{RGB}{227, 239, 219}
\definecolor{cvprpurple}{RGB}{237, 234, 252}

\definecolor{cvprgray}{RGB}{235, 235, 235}
\definecolor{lightblue}{RGB}{240, 246, 255}

\def\BibTeX{{\rm B\kern-.05em{\sc i\kern-.025em b}\kern-.08em
    T\kern-.1667em\lower.7ex\hbox{E}\kern-.125emX}}
\usepackage{balance}
\begin{document}
\title{Unified Unsupervised Anomaly Detection via Matching Cost Filtering}

\author{Zhe Zhang\hspace{-1.2mm}$^{~\orcidlink{0009-0000-0298-5497}}$, Mingxiu Cai\hspace{-1.2mm}$^{~\orcidlink{0000-0003-2769-1916}}$, Gaochang Wu\hspace{-1.2mm}$^{~\orcidlink{0000-0002-5149-2995}}$,~\IEEEmembership{Member,~IEEE}, Jing Zhang\hspace{-1.2mm}$^{~\orcidlink{0000-0001-6595-7661}}$,~\IEEEmembership{Senior Member,~IEEE}, Lingqiao Liu\hspace{-1.2mm}$^{~\orcidlink{0000-0003-3584-795X}}$,  Dacheng~Tao\hspace{-1.2mm}$^{~\orcidlink{0000-0001-7225-5449}}$,~\IEEEmembership{Fellow,~IEEE,} Tianyou Chai\hspace{-1.2mm}$^{~\orcidlink{0000-0002-4623-1483}}$,~\IEEEmembership{Life Fellow,~IEEE}, Xiatian Zhu\hspace{-1.2mm}$^{~\orcidlink{0000-0002-9284-2955}}$
\thanks{
\IEEEcompsocthanksitem Zhe Zhang, Mingxiu Cai, Gaochang Wu, and Tianyou Chai are with the State Key Laboratory of Synthetical Automation for Process Industries, Northeastern University, Shenyang, China. Email: zhangzhe17@stumail.neu.edu.cn, 2410285@stu.neu.edu.cn, wugc@mail.neu.edu.cn, tychai@mail.neu.edu.cn. Zhe Zhang is also a visiting student at the University of Surrey.
\IEEEcompsocthanksitem Jing Zhang is with the School of Computer Science, Wuhan University, China. E-mail: jingzhang.cv@gmail.com.
\IEEEcompsocthanksitem Lingqiao Liu is with the School of Computer Science, The University of Adelaide, Australia. E-mail: lingqiao.liu@adelaide.edu.au.
\IEEEcompsocthanksitem Dacheng Tao is with the College of Computing \& Data Science, Nanyang Technological University, Singapore. E-mail: dacheng.tao@gmail.com.
 \IEEEcompsocthanksitem Xiatian Zhu is with the Surrey Institute for People-Centred Artificial Intelligence, and Centre for Vision, Speech and Signal Processing, University of Surrey, Guildford, UK. E-mail: Eddy.zhuxt@gmail.com.
 \IEEEcompsocthanksitem Corresponding author: Tianyou Chai, Gaochang Wu.
}
}

\markboth{Journal of \LaTeX\ Class Files,~Vol.~18, No.~9, September~2025}%
{Unified Unsupervised Anomaly Detection via Matching Cost Filtering}

\maketitle

\begin{abstract}
Unsupervised anomaly detection (UAD) aims to identify image- and pixel-level anomalies using only normal training data, with wide applications such as industrial inspection and medical analysis, where anomalies are scarce due to privacy concerns and cold-start constraints.
Existing methods, whether reconstruction-based (restoring normal counterparts) or embedding-based (pretrained representations), fundamentally conduct image- or feature-level {\em matching} to generate anomaly maps. Nonetheless, matching noise has been largely overlooked, limiting their detection ability. 
Beyond earlier focus on unimodal RGB-based UAD, recent advances expand to multimodal scenarios, e.g., RGB–3D and RGB–Text, enabled by point cloud sensing and vision–language models.
Despite shared challenges, these lines remain largely isolated, hindering a comprehensive understanding and knowledge transfer. 
In this paper, we advocate unified UAD for both unimodal and multimodal settings {\em in the matching perspective}.
Under this insight, we present Unified Cost Filtering ({\em UCF}), 
a generic post-hoc refinement framework
for refining anomaly cost volume of any UAD model.
The cost volume is constructed by matching a test sample against normal samples from the same or different modalities, followed by a learnable filtering module with multi-layer attention guidance from the test sample, mitigating matching noise and highlighting subtle anomalies.
Comprehensive experiments on 22 diverse benchmarks demonstrate the efficacy of UCF in enhancing a variety of UAD methods, consistently achieving new state-of-the-art results in both unimodal (RGB) and multimodal (RGB–3D, RGB–Text) UAD scenarios. Code and models will be released at \url{https://github.com/ZHE-SAPI/CostFilter-AD}.
\end{abstract}


\section{Introduction}

\begin{figure}
\setlength{\abovecaptionskip}{2pt}  
\setlength{\belowcaptionskip}{0pt} 
\begin{center}
\centerline{\includegraphics[width=0.5\textwidth]{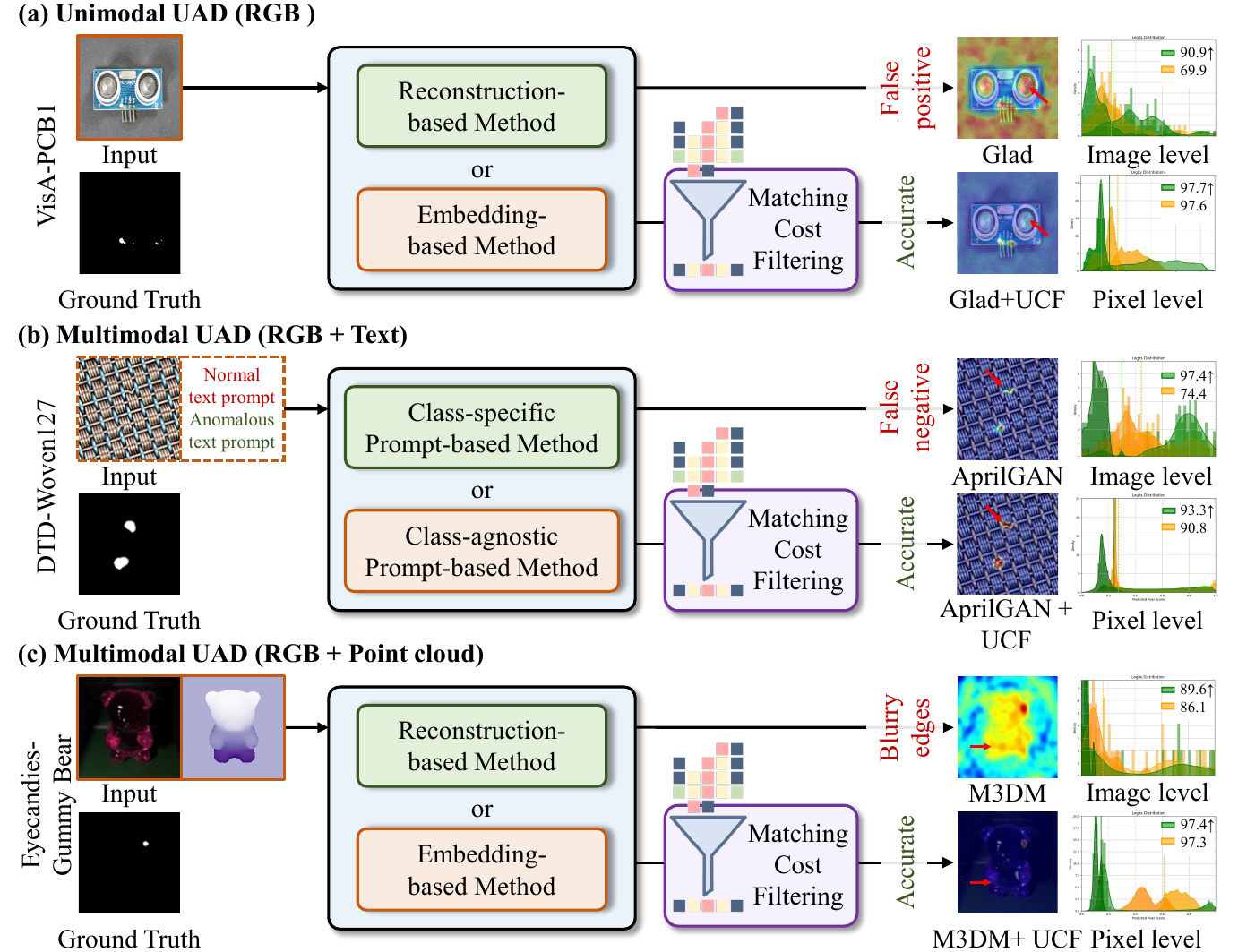}}  
\vspace{-1mm}
\caption{We advocate a unified UAD perspective and introduce UCF, a generic matching-cost filtering method that plugs seamlessly into unimodal RGB~\cite{glad}, RGB–Text~\cite{aprilgan}, and RGB–3D~\cite{m3dm} scenarios. For each scenario, we present anomaly heatmaps and kernel density estimates (KDE)~\cite{kde} of detection logits. Baselines are shown in \textcolor{myyellow}{yellow} and ours (+UCF) in \textcolor{mygreen}{green}. UCF suppresses matching noise, reduces false positives and negatives, sharpens separability between anomalies and normals, and consistently improves performance.}
\label{fig1}
\end{center}
\vspace{-7mm}
\end{figure}

\IEEEPARstart{U}{nsupervised} anomaly detection (UAD) is a practical yet challenging task in domains such as industrial quality inspection~\cite{fmf} and medical diagnosis~\cite{pami4}, where models are typically trained only on normal samples to identify anomalies at both the image and pixel levels, without using anomaly labels from the target datasets~\cite{pami1,pami3,glad,winclip,anomalyclip,cfm,pami2}. Among the earliest and most studied forms, unimodal RGB UAD addresses anomaly scarcity and diversity, often via anomaly synthesis~\cite{jnld,glass}. In practice, inspectors also rely on complementary cues such as 3D shape and surface texture, motivating multimodal RGB–3D UAD~\cite{mvtec3d,eyecan} enabled by advances in 3D sensing. Many RGB and RGB–3D UAD methods adopt a pipeline of one model per category, achieving strong performance~\cite{pami1,liu2023simplenet,diffad,oneclass,ast,cfm} but incurring high training costs and poor scalability as categories grow. To address this, multi-class UAD~\cite{hvqtrans,diad,glad,btf,m3dm} uses a unified model to improve category scalability but often struggles with highly heterogeneous anomalies. In parallel, vision–language models such as CLIP~\cite{clipac} have been adapted to RGB–Text UAD by lightly calibrating pre-trained encoders on small auxiliary datasets (e.g., industrial defects), enabling anomaly detection on unseen datasets (e.g., medical lesions~\cite{pami4}) without using target anomaly labels. Despite these advances, the three lines above are often pursued in isolation, obscuring shared challenges and limiting cross-line knowledge transfer, particularly for subtle anomalies in small, low-contrast, or near-normal regions~\cite{pami4}. We therefore advocate higher-level unification across modalities and task formulations to reveal shared challenges and motivate a unified method.

Existing UAD approaches fall into two broad paradigms. \emph{Reconstruction}-based methods, used primarily for RGB and RGB–3D, detect anomalies by comparing inputs with unimodal~\cite{m3dm} or cross-modal~\cite{cfm} reconstructions and scoring residuals or similarities. 
Architectures such as autoencoder~\cite{eyecan}, U-Net~\cite{zhao2023omnial, das3d}, Transformer~\cite{hvqtrans}, and diffusion model~\cite{glad, pami1} are used to reconstruct normal counterparts of the input, often trained with synthetic anomalies~\cite{jnld, realnet, glass} to mimic real defects. Nonetheless, reconstructions often retain anomalies or misaligned structures due to the ``identical shortcut" issue~\cite{uniad}, undermining input-to-reconstruction matching.

\emph{Embedding}-based methods instead leverage pre-trained models to extract modality-specific features~\cite{patchcore, anomalydino}. For unimodal RGB UAD, they commonly assume models trained only on normal data cannot effectively encode features deviating from the normal distribution, enabling separation of anomalous from normal clusters~\cite{dinomaly,glad,hvqtrans}. Multimodal variants extend this principle via cross-modal comparison~\cite{winclip}, fusion~\cite{m3dm}, or mapping~\cite{cfm}, using similarity between RGB features and class-specific~\cite{aprilgan, adaclip} or class-agnostic~\cite{anomalyclip} normal/abnormal text embeddings, or between RGB and point-cloud features~\cite{m3dm, pami2}. In essence, both paradigms detect anomalies via input-to-template \textbf{matching} over reconstructions, normals, text embeddings, or point-cloud features at global or local scales, where anomalies emerge as regions of high matching cost.

From a matching perspective, current unimodal and multimodal UAD methods often emphasize high-fidelity reconstruction, representation learning, or extensive memory banks, yet devote limited attention to intrinsic noise within the matching process. In practice, anomaly maps are often generated via distance-based matching, for example, using L2 norms in unimodal RGB~\cite{uniad,hvqtrans,dinomaly,vpdm} and RGB–3D settings~\cite{ast}, or cosine similarity in RGB~\cite{mambaad,diad,anomalydino,glad}, RGB–Text UAD~\cite{winclip,aprilgan,adaclip,anomalyclip}, and multimodal RGB–3D UAD~\cite{m3dm, cfm}. Earlier designs, such as DRAEM~\cite{draem} and JNLD~\cite{jnld}, fuse paired image features with discriminative networks. However, as shown in Fig.~\ref{fig1}, these strategies often yield \textbf{matching noise} that blurs boundaries between normal and anomalous regions, an issue that has been largely underexplored. Such noise can arise from factors including the ``identical shortcut”, imperfect templates, and cross-modal feature misalignment~\cite{survey0,survey1}, leading to blurred edges, false positives, and false negatives, especially for subtle defects, low-contrast conditions, and even unseen categories. Addressing this overlooked issue calls for a formulation that effectively represents matching while suppressing noise for more reliable anomaly detection.

Inspired by the concept of matching cost filtering (also known as cost volume filtering) from fields like stereo matching~\cite{stereo}, depth estimation~\cite{depth}, flow estimation~\cite{optical2},  and light field rendering~\cite{lightfield}, we reformulate unsupervised anomaly detection as a three-step paradigm: feature extraction, anomaly cost volume construction, and anomaly cost volume filtering. Building on this view, we present \emph{Unified Cost Filtering (UCF)}, a model-agnostic post-hoc plug-in that unifies anomaly detection across unimodal and multimodal methods. Conceptually, we introduce a matching cost volume to address ``what to match" and a cost volume filtering network to address ``how to refine," enabling adaptive noise suppression and more accurate matching between the input image and its reference templates, including RGB exemplars, text embeddings covering normal and anomalous cues, or 3D structural features.

Specifically, we use modality-specific pre-trained encoders to extract hierarchical features for unimodal UAD (from inputs and image templates) or multimodal UAD (from each modality). Then, we perform patch-wise intra- or cross-modal matching to construct a multi-layer cost volume with two spatial dimensions indexing anomalous locations and one matching dimension encoding correspondence scores.
To refine this cost volume, we introduce a filtering network that progressively aggregates evidence from multiple templates in a coarse-to-fine manner. The refinement employs dual-stream attention guidance where the input features and an initial anomaly map serve as attention queries, suppressing matching noise while preserving edge structures and revealing subtle anomalies. To further enhance detection, we enlarge the matching range by incorporating multiple templates, such as reconstructed normals, multi-view exemplars, or diverse text prompts. In addition, we design a class-aware adaptor that dynamically adjusts the segmentation loss using soft classification logits, prioritizing challenging samples and improving generalization.

Our contributions are as follows. (i) We reconceptualize unimodal and multimodal UAD from the matching perspective, explicitly addressing intrinsic matching noise, an overlooked yet critical factor in existing methods. Under this perspective, we reformulate UAD with a three-step pipeline: feature extraction, matching cost volume construction, and cost volume filtering. 
(ii) We propose UCF, which is characterized by employing multi-layer input observations as attention queries to guide match denoising while preserving edge structures and details of subtle or unseen anomalies. 
(iii) Serving as a general plug-in, our method flexibly constructs and filters matching cost volumes from RGB features along with reconstruction- or embedding-based RGB, text, and point-cloud representations, enabling seamless integration into diverse anomaly detection paradigms.
(iv) We integrate UCF into 10 state-of-the-art uni- and multimodal methods, achieving consistent state-of-the-art performance on 4 RGB, 2 RGB–3D, and 16 RGB–Text UAD benchmarks spanning challenging and widely used datasets.

This work builds upon \emph{CostFilter-AD}~\cite{costfilter} with four advances. First, we extend UAD from unimodal RGB to multimodal RGB–3D and RGB–Text paradigms, fostering a unified view. Second, we introduce paradigm-specific strategies for anomaly cost-volume construction that are denoised by a unified filtering network, improving anomaly classification and localization. Third, we broaden the scope from industrial inspection to medical diagnosis, and from full-shot to zero-shot and few-shot settings, validated on 18 additional multimodal benchmarks. Finally, we integrate UCF into five additional reconstruction- and embedding-based baselines, delivering consistent gains and state-of-the-art results across datasets and modalities, exemplified by AnomalDF~\cite{anomalydino} on VisA~\cite{visa} (RGB), I-/P-AUROC $90.5\%/97.5\% \!\rightarrow\! 94.3\%/99.2\%$; M3DM~\cite{m3dm} on MVTec 3D-AD~\cite{mvtec3d} (RGB-3D), AUPRO@1\% $39.4\% \!\rightarrow\! 45.6\%$ \,(+6.2\%) and I-/P-AUROC $94.5\%/99.1\% \!\rightarrow\! 96.2\%/99.3\%$; and AdaCLIP~\cite{adaclip} on industrial datasets (RGB-Text), AUPRO $40.5\% \!\!\rightarrow\!\! 64.7\%$ \,(+24.2\%) with pixel-AUROC $94.5\% \!\rightarrow\! 94.8\%$.

\section{Related Work}

\subsection{Unimodal RGB Unsupervised Anomaly Detection}
Unimodal UAD methods are typically organized into three lines: embedding-, reconstruction-, and synthesis-based~\cite{survey1}. Embedding-based methods employ pre-trained backbones for feature extraction with knowledge distillation~\cite{rd4ad}, distribution modeling~\cite{kddliu, disicml, textgene}, or memory banks~\cite{patchcore, pni}. Although effective, their reliance on datasets such as ImageNet~\cite{imagenet} limits adaptability to rare or unseen anomalies. Reconstruction-based works, including autoencoders~\cite{tnnlsliu}, GANs~\cite{ocrgan, gan2icml, adgan}, transformers~\cite{uniad, hvqtrans, dinomaly}, diffusion models~\cite{diffad, glad, omiaddiffu}, and MoEs~\cite{moead}, seek to rebuild normal patterns yet often suffer from ``identical shortcut'' issue. Synthesis-based methods generate pixel- or feature-level pseudo anomalies~\cite{jnld, glass, dualsyn} to approximate real distributions, yet remain constrained by domain gaps~\cite{udass, Proto}. Additionally, discriminative pairwise models~\cite{draem, zhao2023omnial} also leave matching noise unresolved.

Recent progress with diffusion~\cite{diad} and foundation models~\cite{dino} has advanced multi-class UAD. GLAD~\cite{glad} adaptively selects denoising steps, VPDM~\cite{vpdm} reduces anomaly leakage using vague prototypes, HVQ-Trans~\cite{hvqtrans} introduces hierarchical vector quantization, and MambaAD~\cite{mambaad} strengthens reconstruction with a multiscale decoder. Nevertheless, matching noise from imperfect reconstructions or suboptimal embeddings persists and degrades localization. This observation motivates our feature-level matching cost volume filtering, which explicitly models and suppresses such noise with architectural generality.

\subsection{Multimodal RGB–3D Unsupervised Anomaly Detection}
Advances in point cloud sensing and RGB-3D UAD benchmarks~\cite{mvtec3d, eyecan} have spurred multimodal UAD that couples RGB appearance with 3D geometry~\cite{btf,ast,m3dm,pami2,cfm}. Two challenges dominate: designing modality-specific feature extractors that remain amenable to cross-modal fusion~\cite{m3dm, mmrd}, and devising fusion strategies that exploit complementarity while avoiding single-modality dominance~\cite{btf, cfm}.

Embedding-based methods typically adopt pre-trained RGB~\cite{dino} and 3D~\cite{pointmae} backbones for feature extraction, followed by cross-modal fusion. Representative designs include AST~\cite{ast}, which employs teacher–student distillation, and BTF~\cite{btf} that utilizes parameter-free fusion. More recent methods~\cite{m3dm, pami2} integrate contrastive objectives with cross-modal fusion and nearest-neighbour retrieval. While effective with diverse, high-quality template patches, their predictions can become noisy with limited templates, and large memory banks can incur substantial costs~\cite{survey0}.
Reconstruction-based methods encode modality features with dual branches (e.g., EasyNet~\cite{easynet}) or translate them through cross-modal mapping (e.g., CFM~\cite{cfm}). When one modality fails to faithfully represent normal or anomalous cues, ambiguity in modality selection will amplify the matching noise. These challenges motivate us to mitigate intrinsic noise in anomaly cost volumes from intra- or cross-modal RGB–3D features via cost filtering.

\subsection{Multimodal RGB–Text Unsupervised Anomaly Detection}
Vision-language models have recently been explored for zero-/few-shot UAD, where neither test-class images nor anomaly labels are available during training~\cite{survey0,aprilgan}. We thus unify this setting under \textit{unsupervised} UAD (dashed box in Fig.~\ref{fig1}). CLIP~\cite{clipac}, pre-trained on large-scale image–text pairs, is widely used to align visual features with text prompts~\cite{clipsurvey}. Early works~\cite{clipac,winclip} compute affinities between visual features and normal/abnormal text embeddings, and WinCLIP~\cite{winclip} enhances robustness via diversified prompts and pyramid aggregation.

Prompt learning in UAD further adapts textual or joint representations to encode normality and abnormality rather than semantic classes. CoOp~\cite{coop} replaces fixed text templates with learnable tokens; AprilGan~\cite{aprilgan} leverages a lightweight linear head for cross-modal alignment; AnomalyCLIP~\cite{anomalyclip} introduces object-agnostic prompts to capture domain-independent anomaly concepts; AdaCLIP~\cite{adaclip} jointly optimizes visual and textual prompts to calibrate embeddings and refine localization. Despite these advances, Fig.~\ref{fig1} reveals persistent false negatives or positives, blurry boundaries, as auxiliary data are limited, which weakens generalization to unseen anomaly categories. To address this, we first match RGB features against diverse normal and abnormal text embeddings to build a cost volume, then apply dual-stream guidance that queries multiple feature layers of the input. This design suppresses matching noise and enhances anomaly detection from coarse-to-fine granularity.

\subsection{Cost Volume Filtering in Vision Tasks}
Cost volume filtering is a crucial technique in vision tasks, widely used to improve local matching~\cite{filtering}. In stereo matching, cost volumes correlate left and right image features along the disparity dimension, capturing pixel-level similarities between views~\cite{stereo2,energy}. In depth estimation, they encode multi-view geometric relationships to produce accurate depth maps~\cite{depth2}. In motion analysis, they represent inter-frame pixel correspondences refined to improve motion estimates~\cite{optical1, optical2}. In light field rendering, cost volumes evaluate reconstruction quality across depth-sheared views, encoding geometry-aware cues to guide neural interpolation for high-angular-resolution synthesis~\cite{lightfield}. In all cases, filtering refines these matching correspondences to improve accuracy.

We review the anomaly cost volume from an energy perspective~\cite{energy}, where a higher energy signals weaker matching consistency with norms, and thus potential anomalies. Unlike prior methods~\cite{hvqtrans,glad,anomalyclip,adaclip,m3dm,cfm} that have been studied separately, we propose a unified method that models this energy across diverse templates and modalities. It comprises two components: (i) cost volume construction via hierarchical matching across multiple templates and modalities to encode anomaly cues; and (ii) a cost volume filtering network with dual-stream guidance from hierarchical queries, suppressing noise while preserving edges and revealing subtle anomalies.

\begin{figure*}
\setlength{\abovecaptionskip}{2pt}  
\setlength{\belowcaptionskip}{0pt} 
\begin{center}
\centerline{\includegraphics[width=0.98\textwidth]{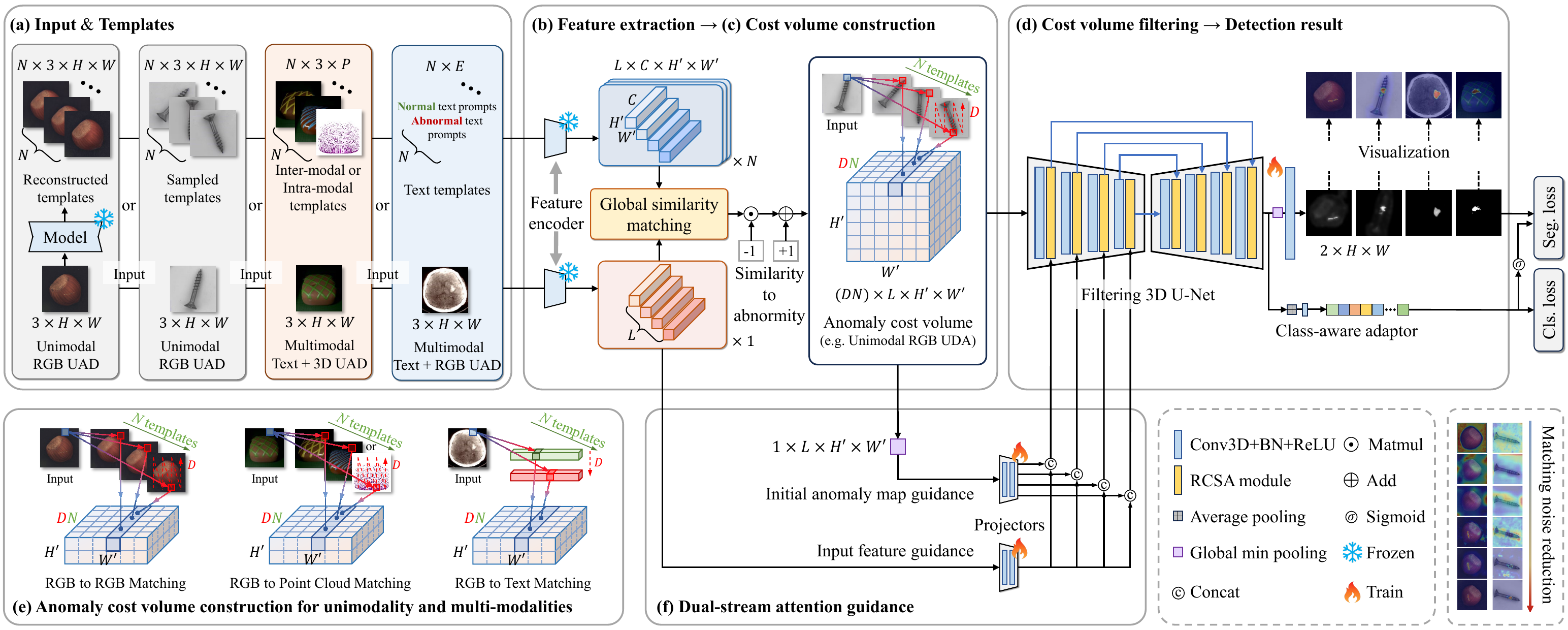}}
\caption{Overview of our UCF, a generic plug-in for UAD. We reformulate UAD as a matching cost filtering process applicable to both unimodal (RGB) and multimodal (RGB–3D, RGB–Text) scenarios.
(i) First, we employ baseline pre-trained encoders to extract features from the input and reference templates, which may be reconstructed normal samples, randomly sampled normal templates, or cross-modal counterparts.
(ii) Second, we construct an anomaly cost volume based on global similarity matching across or within modalities.
(iii) Lastly, we learn a matching cost filtering network, guided by attention queries derived from the input features and an initial anomaly map, to refine the volume and generate the final detection results. 
(iv) Further, we integrate a class-aware adaptor to tackle class imbalance and enhance the ability to deal with multiple anomaly classes simultaneously.}
\label{fig2}
\end{center}
\vspace{-5mm}
\end{figure*}

\section{Methodology}
\subsection{Overview}

Unsupervised anomaly detection fundamentally relies on intra-modal or cross-modal feature matching to reveal deviations from normal patterns. As shown in Fig.~\ref{fig1}, matching noise is pervasive yet often overlooked, and it substantially degrades accuracy. We propose \emph{Unified Cost Filtering (UCF)}, which recasts anomaly detection as a three-stage pipeline comprising feature extraction, anomaly cost volume construction, and anomaly cost volume filtering (Fig.~\ref{fig2}). Given features from training or test images with reconstructed, embedding, or cross-modal templates, we construct and filter cost volumes irrespective of the modality pairing. This unified formulation yields a generic framework for unimodal UAD (RGB) and multimodal UAD (RGB-3D, RGB-Text) with broad applicability.

Since anomalous samples are commonly unavailable in unsupervised training, we follow GLAD~\cite{glad} for anomalous image synthesis and extend it to pixel-registered 3D point clouds, used solely to train the cost volume filtering network across unified UAD scenarios. Inspired by stereo matching and optical flow estimation, we match an input RGB sample \(I_{\mathrm{rgb}}\in\mathbb{R}^{3\times H\times W}\) (channel, height, width) with reference templates drawn from reconstructions, random views, or other modalities. As shown in Fig.~\ref{fig2}(a), the input can be matched with normal-like image templates \(T_{\mathrm{rgb}}\in\mathbb{R}^{3\times H\times W}\), 3D point clouds \(T_{\mathrm{3D}}\in\mathbb{R}^{3\times P}\) (3D coordinates with $P$ points), and text embeddings \(T_{\mathrm{text}}\in\mathbb{R}^{N\times E}\) (prompts, dimension) from normal and abnormal prompts. Our method supports multi-template, intra-/cross-modal matching for a single image input and yields accurate anomaly reasoning across hierarchical features.

\subsection{Reference Templates in UAD}

As shown in Fig.~\ref{fig2}, we employ modality-specific reference templates to construct anomaly cost volumes across different UAD scenarios. In unimodal RGB UAD, the input image is matched with normal-like RGB templates \(T_{\mathrm{rgb}}\) for both image- and pixel-level detection. In multimodal UAD, the input can also be matched with 3D point clouds \(T_{\mathrm{3D}}\) for joint detection or with text prompts \(T_{\mathrm{text}}\) for zero-/few-shot RGB-Text UAD, providing a unified formulation across modalities.

\subsubsection{Templates in Unimodal RGB UAD}

\textbf{Templates for reconstruction-based methods.}
A reconstruction network can be viewed as learning the manifold of normal data, where reconstructions approximate the nearest normal neighbors of inputs and are thus regarded as templates. Recent approaches include transformer-based methods (e.g., UniAD~\cite{uniad}, HVQ-Trans~\cite{hvqtrans}, Dinomaly~\cite{dinomaly}) and diffusion-based methods (e.g., GLAD~\cite{glad}, DiAD~\cite{diad}) for generating high-fidelity normal counterparts. For transformer-based reconstruction, which has no intermediate outputs, we set \(N{=}1\). Diffusion reconstructions are effective when final-step denoised images are used as templates, but imperfect outputs may preserve anomalies through the ``identical shortcut'' (see Appendix for details). Frequency evolution~\cite{freq} shows that while final-step denoising preserves fine details, intermediate steps retain complementary low-frequency cues useful for capturing normal contours.

To exploit this property, we randomly sample $N$ templates from multiple denoising steps, including the final one, to enrich feature representations. The reconstruction at step $t$ is
\[
I_{t \to 0} = \frac{1}{\sqrt{\bar{\alpha}_t}} \left( I_t - \sqrt{1 - \bar{\alpha}_t}\,\epsilon_\theta(I_t, t) \right), \tag{1}
\]
where $\epsilon_\theta$ is the noise predictor of the frozen baseline diffusion model, and $\bar{\alpha}_t$ is predefined and inversely related to $t$.

\textbf{Templates for embedding-based methods.}
Embedding-based UAD is particularly sensitive to matching noise caused by input–template misalignments in scale, texture, or viewpoint~\cite{survey0}. Existing methods~\cite{patchcore, bank,anomalydino} typically address this issue by constructing large memory banks for exhaustive template search. In contrast, we explicitly reformulate it as a matching noise problem and resolve it through global matching combined with cost volume filtering. This strategy achieves accurate alignment with only a few ($N$) normal templates per image, while effectively suppressing noise and eliminating the reliance on extensive memory banks.

\subsubsection{Templates in Multimodal RGB-3D UAD}

Multimodal RGB-3D UAD typically exploits the complementarity between RGB and 3D features. Early approaches concatenate features along the channel dimension for discriminators (e.g., AST~\cite{ast}, CheatDepth~\cite{cheatdepth}). More recent embedding-based methods construct memory banks for unimodal features and fuse cross-modal representations (e.g., M3DM~\cite{m3dm}), whereas reconstruction-based methods employ lightweight networks such as MLPs to reconstruct normal multimodal features (e.g., CFM~\cite{cfm}). We reformulate these designs from a similarity-matching perspective: intra-modal matching covers RGB–RGB, 3D–3D, and fused–fused feature pairs~\cite{m3dm}, whereas cross-modal matching corresponds to RGB–3D feature pairs~\cite{cfm}. Both matching pairs can originate from RGB, 3D, or fused representations, yielding a unified and generic formulation.

\subsubsection{Templates in Multimodal RGB-Text UAD}

Advances in vision–language models such as CLIP~\cite{clipac} enable text-image alignment to improve semantic generalization in UAD. Following prior work, we use text prompts that describe normal or abnormal states, for example, \texttt{a photo of normal/abnormal [cls]}~\cite{winclip, coop, aprilgan, adaclip}, or class-agnostic templates such as \texttt{a photo of a (damaged) object}~\cite{anomalyclip}, and detect anomalies via their similarity to image features. However, CLIP mainly encodes global category semantics rather than pixel-level cues, limiting fine-grained anomaly localization~\cite{winclip, coop, aprilgan, adaclip}. Prompt learning partially alleviates this by fine-tuning encoders~\cite{adaclip, anomalyclip}, yet matching noise, especially false negatives, remains common in zero-shot and few-shot regimes. To address this limitation, we introduce a dual-stream feature-guided filtering network that reduces matching noise, recovers missed anomalies, and adapts seamlessly to class-specific and class-agnostic RGB-Text UAD.

\vspace{-2mm}
\subsection{Input Feature Extraction}

As a generic plug-in, we use each baseline’s modality-specific encoders to extract multi-layer features (Fig.~\ref{fig2}(b)).

For \textbf{RGB inputs}, the feature extractor can be DINO-v2~\cite{dino} pretrained in ImageNet~\cite{imagenet}, the CLIP image encoder~\cite{clipac}, or feature extractors used in prior baselines, including ViT-based encoders~\cite{uniad, hvqtrans, dinomaly} or fine-tuned variants~\cite{adaclip, cfm}.
Given an input image \(I_{\mathrm{rgb}}\) and templates \(T_{\mathrm{rgb}}\), we obtain a multi-layer tensor \(f_{I,{\mathrm{rgb}}}\in \mathbb{R}^{L \times C \times H' \times W'}\) and $N$ template tensors $f_{T,\mathrm{rgb}}$ of identical size, where $L$ is the number of layers, $C$ is the embedding dimension, and $H',W'$ are the spatial resolutions. 

For \textbf{text templates}, normal prompts \(T_{\mathrm{nor}}\) and abnormal prompts \(T_{\mathrm{abn}}\) are encoded by the text branch of CLIP or its fine-tuned variants~\cite{anomalyclip, adaclip}, yielding $N$ tensors $f_{T,\mathrm{nor}}\in \mathbb{R}^{L \times C \times (H'W')}$ and $f_{T,\mathrm{abn}}$ with the same dimensions.

For \textbf{3D point clouds}, we adopt PointMAE~\cite{pointmae} pre-trained on ShapeNet~\cite{shapenet}. Following~\cite{m3dm, cfm}, we extract center-point features, propagate them to all points by nearest-neighbor interpolation, and project them to the image plane via pixel registration. This produces $f_{T,\mathrm{3D}}\in \mathbb{R}^{L \times C \times H'\times W' }$ that is spatially aligned with image patch tokens for feature matching.

\subsection{Anomaly Cost Volume Construction}
Building on multimodal representations, we recast anomaly detection as feature matching and denoising, where the matching cost volume serves as an energy field: low values align with normal templates, high values indicate anomalies. As shown in Fig.~\ref{fig2}(e), matching can occur within or across modalities, and anomalies exhibit high energy with respect to normal templates and low energy relative to abnormal counterparts.

For \textbf{unimodal RGB UAD}, to ensure generality across reconstruction-based and embedding-based methods, we perform global cosine similarity matching over all spatial indices of each normal-like template feature:
\[
\mathcal{V}(j,n,l,i)
= \frac{f_{I,\mathrm{rgb}}^{\,i,l}\cdot f_{T,\mathrm{rgb}}^{\,n,j,l}}
{\|f_{I,\mathrm{rgb}}^{\,i,l}\|\,\|f_{T,\mathrm{rgb}}^{\,n,j,l}\|}, \tag{2}
\]
where $f_{I,\mathrm{rgb}}^{\,i,l}$ is the patch token feature at spatial index $i$ of the input image at layer $l\in\{1,2,\dots,L\}$, $f_{T,\mathrm{rgb}}^{\,n,j,l}$ is the feature at spatial index $j$ of the $n$-th template, and $\mathcal{V}\in\mathbb{R}^{D\times N\times L\times(H'W')}$ is the similarity volume with $D=H'\times W'$ as the matching dimension. Here, $D$ indexes template locations, while $(H'W')$ indexes spatial locations. In contrast to local matching with a single reference or nearest-neighbor search in a memory bank~\cite{anomalydino,patchcore}, exhaustive global comparison captures comprehensive feature correlations (Fig.~\ref{fig2}(c)).

For \textbf{multimodal RGB–3D UAD}, we construct similarity volumes across spatial indices of input and normal template features. For the intra-modal case~\cite{m3dm, cheatdepth}, matching is restricted to the same modality, including RGB–RGB, 3D–3D, and fused–fused feature pairs:
\[
\underset{k\to k'}{\mathcal{V}'}(j,n,l,i) = \frac{f_{I,k}^{i,l} \cdot f_{T,k'}^{n,j,l}}{\|f_{I,k}^{i,l}\| \cdot \|f_{T,k'}^{n,j,l}\|}, \tag{3}
\label{rgb3d}
\]
where $k, k' \in \{\mathrm{rgb}, \mathrm{3D}, \mathrm{fused}\}$. Inter-modal matching~\cite{cfm} is also supported by pairing features across modalities (e.g., $f_{I,\mathrm{rgb}}^{i,l}$ with $f_{T,\mathrm{3D}}^{n,j,l}$, or vice versa). To capture both forms of relationships comprehensively, we concatenate the similarity volumes along the modality dimension:
\[
\mathcal{V}(j,n,l,i) = \operatorname*{cat}_{k, k'\in\{\mathrm{rgb},\,\mathrm{3D},\,\mathrm{fused}\}} \underset{k\to k'}{\mathcal{V}'}(j,n,l,i). \tag{4}
\]
In this way, template features can be drawn from normal instances randomly sampled from embedding-based memory banks for intra-modal matching~\cite{m3dm}, or from reconstruction-based mappings across modalities, such as projecting point-cloud features onto RGB features or vice versa~\cite{cfm}. This formulation consolidates intra- and inter-modal relations into a generic scheme for multimodal similarity volume construction, providing the foundation for subsequent cost volume filtering.

For \textbf{multimodal RGB–Text UAD}, we adopt the RGB–3D UAD formulation in Eq.~\ref{rgb3d} with the input feature fixed to the RGB embedding $f_{I,\mathrm{rgb}}$, while template features are the text embeddings of normal (nor) and abnormal (abn) prompts. Since similarity to text encodes image–language consistency, anomalies are expected to correlate more with abnormal prompts and less with normal prompts. We therefore define the normal-related similarity volume as
\[
\mathcal{V}(j,n,l,i)
= \operatorname*{cat}\!\Big[
  1 - \underset{\mathrm{rgb}\to \mathrm{abn}}{\mathcal{V}'}(j,n,l,i),\;
  \underset{\mathrm{rgb}\to \mathrm{nor}}{\mathcal{V}'}(j,n,l,i)
\Big]. \tag{5}
\]

\textbf{Unified cost volume transformation.}
As discussed for unimodal RGB, multimodal RGB–3D, and RGB–Text UAD scenarios, anomalies are characterized by reduced similarity to normal references or elevated similarity to abnormal ones. Therefore, we transform the normal-related similarity volume into an anomaly cost volume $\mathcal{C}\in\mathbb{R}^{D\times N\times L\times(H'W')}$ via
\[
\mathcal{C}(j,n,l,i) = 1 - \mathcal{V}(j,n,l,i), \tag{6}
\]
where higher cost values indicate higher anomaly likelihood. Then, we collapse the matching and template axes by merging $D$ and $N$ into a single dimension, and reshape the spatial axis from $H'W'$ to $H'\times W'$, yielding $\mathcal{C}\in\mathbb{R}^{(DN)\times L\times H'\times W'}$. Finally, we obtain a multi-layer initial anomaly map $\bar{\mathcal{M}}$ by global min pooling for unimodal RGB UAD and by average pooling for multimodal UAD, which provides a coarse localization to guide subsequent filtering.

\subsection{Anomaly Cost Volume Filtering}
Existing UAD methods commonly smooth anomaly score maps with Gaussian filters~\cite{uniad, hvqtrans, anomalydino, glad, dinomaly, adaclip, anomalyclip, m3dm, cfm}. However, as shown in Fig.~\ref{fig1}, such post hoc processing blurs boundaries, preserves matching noise, and induces false negatives or false positives, implying that anomalies may go undetected and normal regions may be misclassified. To address these issues, we propose filtering the intermediate anomaly cost volume using a dedicated 3D U\mbox{-}Net~\cite{3dunet} guided by dual streams of features. This design attenuates matching noise from reconstruction shortcuts, limited or suboptimal templates in embedding methods, and cross-modal misalignment, while preserving the cues of subtle anomalies.

\textbf{Network input.} As illustrated in Fig.~\ref{fig2}(d), the input to our filtering network is the anomaly matching cost volume $\mathcal{C}\in\mathbb{R}^{(DN)\times L \times H'\times W'}$. Here, the matching dimension $DN$ is mapped to the channel dimension of the network, $L$ denotes the depth dimension capturing feature matching across multiple layers, and $H'$ and $W'$ specify the spatial resolution. In addition, the multi-layer input features $f_{I,{\mathrm{rgb}}}$ and the initial anomaly map $\bar{\mathcal{M}}$ are utilized as auxiliary guidance for the filtering.

\textbf{Dual-stream attention guidance.}
The anomaly cost volume encodes global matching information but may remain vulnerable to information loss and noise introduced by reconstruction artifacts or cross-modal misalignment (detailed analysis in Appendix). To address this, we propose a dual-stream attention guidance mechanism (Fig.~\ref{fig2}(f)). The input image feature \(f_{I,{\mathrm{rgb}}}\) provides spatial guidance (SG) to preserve fine details such as anomaly edges, while the initial anomaly map \(\bar{\mathcal{M}}\) supplies matching guidance (MG) to highlight dimensions most indicative of anomalies. Together, these streams enable the network to capture global matching patterns while retaining and further enriching fine-grained spatial cues.

The mechanism is instantiated through a residual channel–spatial attention (RCSA) module inspired by~\cite{cbam}, enhanced with residual connections to preserve anomaly details and strengthen discriminative sensitivity:  
\[
x_{\ell}' = \text{cat}(x_{\ell}, h(\bar{\mathcal{M}}), h(f_{I,{\mathrm{rgb}}}^{l})),
\]
\[
x_{\ell}^{ca} = \sigma\left(\text{conv}(\text{MP}(x_{\ell}')) + \text{conv}(\text{AP}(x_{\ell}'))\right) * x_{\ell}' + x_{\ell}', \tag{7}
\label{eq4}
\]
\[
x_{\ell}^{sa} = \sigma\left(\text{conv}(\text{cat}(\mu(x_{\ell}^{ca}), \text{max}(x_{\ell}^{ca})))\right) * x_{\ell}^{ca} + x_{\ell}^{ca},
\]
where $x_{\ell}$ denotes the anomaly cost volume feature at encoder layer $\ell$, and $h$ is the feature projectors for channel projection and spatial resolution adjustment. The dual-stream guidance features are concatenated (\text{cat}) with cost volume features along the channel dimension. Here, $\sigma$ is the sigmoid activation, $\text{conv}$ denotes 3D convolution, and $\text{MP}$, $\text{AP}$, $\mu$, and $\text{max}$ represent global max pooling, global average pooling, channel-wise mean, and channel-wise max, respectively.

Attention-guided features $x_{\ell}^{sa}$ are progressively propagated to the decoder via skip connections, where dual-stream attention guidance can further refine decoding. This mechanism reinforces global feature matching via residual channel attention and sharpens pixel-level anomaly localization through residual spatial attention, thereby enabling coarse-to-fine denoising and precise anomaly detection. More detailed RCSA design and progressive denoising visualizations are in the Appendix.

\textbf{Class-aware adaptor.} 
To improve generalization across heterogeneous anomaly settings, we introduce a class-aware adaptor that dynamically modulates the segmentation loss through sigmoid-activated soft logits. The adaptor aggregates deep cost volume features by spatial average pooling and projects them onto multi-class classification logits with a fully connected layer. This allows the segmentation head to prioritize challenging samples and adapt to diverse anomaly characteristics, encompassing both multi-class cases within a dataset (e.g., RGB or RGB–3D UAD) and zero-shot scenarios across datasets (e.g., RGB–Text UAD).

\subsection{Anomaly Detection Output Generation}
\label{sec3.4}
As illustrated in Fig.~\ref{fig2}(d), following stereo matching~\cite{stereo2} and light field rendering~\cite{lightfield}, the filtered anomaly volume is processed by global min pooling along the matching dimension, followed by a convolutional layer and softmax, to generate the normal–anomaly score map $\mathcal{M} = \text{softmax}(\text{conv}(\text{min}(x)))$. For image-level detection, the anomaly score is obtained by averaging the top $250$ values in the anomaly score map.  

\subsection{Training and Inference}
We present our method as a generic plug-in applicable to both reconstruction- and embedding-based pipelines across unimodal (RGB) and multimodal (RGB–3D, RGB–Text) UAD. Anomaly cost volumes are constructed by feature matching that integrates evidence from multiple templates. Cost filtering then leverages dual-stream attention to dynamically aggregate evidence and suppress matching noise. This process is formulated as a normal–abnormal segmentation task, where predicted anomaly maps $\mathcal{M}$ are trained to align with synthetic masks $\mathcal{M}_s$ and are expected to generalize to real anomalies at evaluation. The \textbf{training objective} is
\begin{align}
\mathcal{L} &= \mathcal{L}_\text{Focal}(\mathcal{M}, \mathcal{M}_s, \sigma(\hat{Y}_c)) + \mathcal{L}_\text{CE}(\hat{Y}_c, Y) \notag \\
 &\hspace{12pt}+ \alpha \cdot (\mathcal{L}_\text{Soft-Iou}(\mathcal{M}, \mathcal{M}_s) + \mathcal{L}_\text{SSIM}(\mathcal{M}, \mathcal{M}_s)), \tag{8}
 \label{loss}
\end{align}
where $\mathcal{L}_\text{Focal}$ is focal loss~\cite{focalloss} to mitigate the imbalance between normal and anomalous regions, $\mathcal{L}_\text{Soft-Iou}$~\cite{iouloss} improves anomaly localization, $\mathcal{L}_\text{SSIM}$~\cite{ssimloss} enforces structural consistency, and $\mathcal{L}_\text{CE}$~\cite{celoss} supports multi-class classification. The focal loss parameter $\gamma$ is adaptively modulated by the adaptor: $\gamma = \gamma_0 - \sigma(\hat{Y}_c)$ when the class is correctly identified, and $\gamma = \gamma_0$ otherwise. Thus, the class-aware adaptor leverages predicted logits $\hat{Y}_c$ to regulate $\gamma$ and enhance multi-class segmentation.  

The \textbf{inference process} similarly constructs and filters the anomaly cost volume, yielding anomaly map $\mathcal{M}$. To integrate with baseline responses, we compute a weighted sum $\lambda \cdot \mathcal{M} + (1-\lambda) \cdot \mathcal{M}_{\text{baseline}}$ for anomaly localization and detection, where $\lambda \in [0, 1]$ compensates for potential scale differences between the two components.

\begin{table*}[!t]
\vspace{-3mm}
  \caption{Quantitative comparison for \textbf{unimodal RGB UAD}. Multi-class anomaly detection and localization results (image-level AUROC/pixel-level AUROC) on \textbf{MVTec-AD}, evaluated across all categories without fine-tuning. \textbf{Best} results are in bold.}
    \label{table1}
  \vspace{-4mm}
\begin{center}
  \tabcolsep=0.05cm
  \resizebox{1\textwidth}{!}{
  \begin{tabular}{c|c|c|c|c|c|c|c>{\columncolor{lightblue}}c|c>{\columncolor{lightblue}}c|c>{\columncolor{lightblue}}c}
    \toprule
     \multicolumn{2}{c|}{Category} & PatchCore\cite{patchcore} & OmniAL\cite{zhao2023omnial} & DiAD\cite{diad} & VPDM\cite{vpdm} & MambaAD\cite{mambaad} & GLAD\cite{glad} &  + UCF  & HVQ-Trans\cite{hvqtrans} &  + UCF & AnomalDF\cite{anomalydino} &  + UCF\\
    \midrule
    \multirow{10}{*}{\rotatebox{90}{Object}} 
& Bottle   &\textbf{100}\phantom{.} / \textbf{99.2} &\textbf{100}\phantom{.} / \textbf{99.2} &99.7 / 98.4 &\textbf{100}\phantom{.} / 98.6 &\textbf{100}\phantom{.} / 98.7 &\textbf{100}\phantom{.} / 98.4 &99.8 / 97.8 &\textbf{100}\phantom{.} / 98.3 &\textbf{100}\phantom{.} / 98.8 &\textbf{100}\phantom{.} / 99.3 &\textbf{100}\phantom{.} / 99.1\\
& Cable    &95.3 / 93.6 &98.2 / 97.3 &94.8 / 96.8 &97.8 / 98.1 &98.8 / 95.8 &98.7 / 93.4 &98.0 / 96.3 &99.0 / 98.1 &\textbf{99.8} / 98.2 &99.6 / \textbf{98.3} &99.3 / 98.2\\
& Capsule  &96.8 / 98.0 &95.2 / 96.9 &89.0 / 97.1 &\textbf{97.0} / 98.8 &94.4 / 98.4 &96.5 / 99.1 &94.3 / \textbf{99.2} &95.4 / 98.8 &96.4 / 98.9 &89.7 / 99.1 &96.1 / \textbf{99.2}\\
& Hazelnut &99.3 / 97.6 &95.6 / 98.4 &99.5 / 98.3 &99.9 / 98.7 &\textbf{100}\phantom{.} / 99.0 &97.0 / 98.9 &99.4 / 99.1 &\textbf{100}\phantom{.} / 98.8 &\textbf{100}\phantom{.} / 99.2 &99.9 / \textbf{99.6} &\textbf{100}\phantom{.} / 99.5\\
& Metal Nut&99.1 / 96.3 &99.2 / 99.1 &99.1 / 97.3 &98.9 / 96.0 &99.9 / 96.7 &99.9 / 97.3 &\textbf{100}\phantom{.} / \textbf{99.2} &99.9 / 96.3 &\textbf{100}\phantom{.} / 97.9 &\textbf{100}\phantom{.} / 96.7 &\textbf{100}\phantom{.} / 99.0\\
& Pill     &86.4 / 90.8 &97.2 / \textbf{98.9} &95.7 / 95.7 &97.9 / 96.4 &97.0 / 97.4 &94.4 / 97.9 &97.9 / 97.8 &95.8 / 97.1 &96.9 / 96.5 &97.2 / 98.1 &\textbf{98.9} / 98.4\\
& Screw    &94.2 / 98.9 &88.0 / 98.0 &90.7 / 97.9 &95.5 / 99.3 &94.7 / 99.5 &93.4 / \textbf{99.6} &95.4 / \textbf{99.6} &\textbf{95.6} / 98.9 &95.3 / 99.0 &74.3 / 97.6 &88.5 / 99.0\\
& Toothbrush &\textbf{100}\phantom{.} / 98.8 &\textbf{100}\phantom{.} / 99.0 &99.7 / 99.0 &94.6 / 98.8 &98.3 / 99.0 &99.7 / \textbf{99.2} &99.7 / 99.1 &93.6 / 98.6 &\textbf{100}\phantom{.} / 98.9 &99.7 / \textbf{99.2} &99.7 / \textbf{99.2}\\
& Transistor&98.9 / 92.3 &93.8 / 93.3 &99.8 / 95.1 &99.7 / 97.9 &\textbf{100}\phantom{.} / 97.1 &99.4 / 90.9 &99.5 / 91.6 &99.7 / 99.1 &99.7 / \textbf{99.2} &96.5 / 95.8 &97.8 / 97.5\\
& Zipper   &97.1 / 95.7 &\textbf{100}\phantom{.} / \textbf{99.5} &95.1 / 96.2 &99.0 / 98.0 &99.3 / 98.4 &96.4 / 93.0 &99.2 / 97.7 &97.9 / 97.5 &98.9 / 98.3 &98.8 / 94.3 &98.9 / 96.7\\ \midrule
\multirow{5}{*}{\rotatebox{90}{Texture}} 
& Carpet   &97.0 / 98.1 &98.7 / 99.4 &99.4 / 98.6 &\textbf{100}\phantom{.} / 98.8 &99.8 / 99.2 &97.2 / 98.9 &\textbf{100}\phantom{.} / 99.1 &99.9 / 98.7 &\textbf{100}\phantom{.} / 98.5 &99.9 / 99.4 &99.9 / \textbf{99.6}\\
& Grid     &91.4 / 98.4 &99.9 / 99.4 &98.5 / 96.6 &98.6 / 98.0 &\textbf{100}\phantom{.} / 99.2 &95.1 / 98.1 &\textbf{100}\phantom{.} / \textbf{99.5} &97.0 / 97.0 &99.3 / 98.3 &98.2 / 97.8 &\textbf{100}\phantom{.} / \textbf{99.5}\\
& Leather  &\textbf{100}\phantom{.} / 99.2 &99.0 / 99.3 &99.8 / 98.8 &\textbf{100}\phantom{.} / 99.2 &\textbf{100}\phantom{.} / 99.4 &99.5 / \textbf{99.7} &\textbf{100}\phantom{.} / 99.6 &\textbf{100}\phantom{.} / 98.8 &\textbf{100}\phantom{.} / 99.3 &\textbf{100}\phantom{.} / \textbf{99.7} &\textbf{100}\phantom{.} / \textbf{99.7}\\
& Tile     &96.0 / 90.3 &99.6 / 99.0 &96.8 / 92.4 &\textbf{100}\phantom{.} / 94.5 &98.2 / 93.8 &\textbf{100}\phantom{.} / 97.8 &\textbf{100}\phantom{.} / 99.4 &99.2 / 92.2 &\textbf{100}\phantom{.} / 95.0 &\textbf{100}\phantom{.} / 98.5 &\textbf{100}\phantom{.} / \textbf{99.6}\\
& Wood     &93.8 / 90.8 &93.2 / 97.4 &\textbf{99.7} / 93.3 &98.2 / 95.3 &98.8 / 94.4 &95.4 / 96.8 &97.4 / 97.4 &97.2 / 92.4 &98.5 / 94.3 &97.9 / 97.6 & 98.9 / \textbf{98.2}\\ \midrule
\multicolumn{2}{c|}{Mean} &96.4 / 95.8 &97.2 / 98.3 &97.2 / 96.8 &98.4 / 97.8 &98.6 / 97.7 &97.5 / 97.3 &98.7 / 98.2 & 98.0 / 97.3 &\textbf{99.0} / 98.0 &96.8 / 98.1 &98.5 / \textbf{98.8}\\ \bottomrule
  \end{tabular}}
  \end{center}
   \vspace{-5mm}
\end{table*}

\section{Experimental Evaluation}

We evaluate UCF on unimodal RGB and multimodal RGB-3D and RGB-Text anomaly detection (Table~\ref{expoverview}). Following existing protocols, we adopt three settings: (i) \textit{full-shot} for \textbf{RGB} and \textbf{RGB-3D} UAD, where all normal samples from the target categories are available for training and no anomalies are used; (ii) \textit{zero-shot} for \textbf{RGB-Text} UAD, where neither normal nor abnormal samples from the target datasets are provided and the model is trained on auxiliary datasets that do not overlap with the target categories and contain a small fraction of anomalies; and (iii) \textit{few-shot} for \textbf{RGB-Text} UAD, where only a small number of randomly selected normal samples per target category are available during testing, following~\cite{aprilgan}.

As a generic post-hoc plug-in for UAD, UCF is integrated into 10 state-of-the-art baselines and evaluated across 22 industrial and medical benchmarks with comprehensive image- and pixel-level metrics.
Our study comprises: (i) quantitative and qualitative comparisons across diverse UAD scenarios; (ii) extensive ablation studies; and (iii) t-SNE~\cite{tsne} visualizations, computational efficiency analysis, KDE~\cite{kde} analysis, hybrid cost volumes, hyperparameter sensitivity, and analysis of failure cases. Additional details of the baselines, benchmarks, anomaly synthesis, and implementation, together with substantial quantitative and qualitative results, are provided in the Appendix.

\begin{table}[!t]
\centering
\caption{Overview of experiment settings. ${}^{*}$ indicates that abnormal training data are from auxiliary datasets (non-overlapping with test classes), following the protocol of~\cite{aprilgan,anomalyclip,adaclip}.}

\label{expoverview}
\vspace{-2mm}
\setlength\tabcolsep{3pt}
\renewcommand{\arraystretch}{1.2}
\begin{tabular}{@{}ccc|ccc|cc|c@{}}
\toprule
\multicolumn{3}{c|}{Modality} & \multicolumn{3}{c|}{Shot} & \multicolumn{2}{c|}{Train data} & \multirow{2}{*}{Section} \\
\cmidrule(lr){1-3} \cmidrule(lr){4-6} \cmidrule(lr){7-8}
RGB & 3D & Text  & Full & Zero & Few & Normal & Abnormal & \\
\midrule
$\checkmark$ & $\times$ & $\times$ & $\checkmark$ & $\times$ & $\times$ & $\checkmark$ & $\times$ & Sec.~\ref{sec_rgb_uad} \\
$\checkmark$ & $\checkmark$ & $\times$ & $\checkmark$ & $\times$ & $\times$ & $\checkmark$ & $\times$ & Sec.~\ref{sec_rgb3d_uad} \\
$\checkmark$ & $\times$ & $\checkmark$ & $\times$ & $\checkmark$ & $\times$ & $\checkmark$ & $\checkmark^{*}$ & Sec.~\ref{sec_rgbtext_uad} \\
$\checkmark$  & $\times$ & $\checkmark$ & $\times$ & $\times$ & $\checkmark$ & $\checkmark$ & $\checkmark^{*}$ & Sec.~\ref{sec_rgbtextfew_uad} \\
\bottomrule
\end{tabular}
\vspace{-5mm}
\end{table}

\begin{table*}[!t]
    \caption{Quantitative comparison for \textbf{unimodal RGB UAD}. Multi-class anomaly detection and localization results (image-level AUROC/pixel-level AUROC) on \textbf{VisA}, evaluated across all categories without fine-tuning. \textbf{Best} results are in bold.}
      \label{table2}
  \vspace{-4mm}
\begin{center}
  \tabcolsep=0.05cm
  \resizebox{1\textwidth}{!}{
  \begin{tabular}{c|c|c|c|c|c|c|c>{\columncolor{lightblue}}c|c>{\columncolor{lightblue}}c|c>{\columncolor{lightblue}}c}
    \toprule
     \multicolumn{2}{c|}{Category} & JNLD~\cite{jnld} & OmniAL~\cite{zhao2023omnial} & DiAD~\cite{diad} & VPDM~\cite{vpdm} & MambaAD~\cite{mambaad} & GLAD~\cite{glad} &  + UCF & HVQ-Trans~\cite{hvqtrans} & + UCF & AnomalDF~\cite{anomalydino} &  + UCF\\
    \midrule
\multirow{4}{*}{\rotatebox{90}{\shortstack{Complex\\Structure}}}
& PCB1   & 82.9 / 98.9 & 77.7 / 97.6 & 88.1 / 98.7 & \textbf{98.2} / 99.6 & 95.4 / \textbf{99.8} & 69.9 / 97.6 & 90.9 / 97.7 & 95.1 / 99.5 & 96.3 / 99.3 & 87.4 / 99.3 & 91.8 / 99.7\\
& PCB2   & 79.1 / 95.0 & 81.0 / 93.9 & 91.4 / 95.2 & \textbf{97.5} / 98.8 & 94.2 / \textbf{98.9} & 89.9 / 97.1 & 93.2 / 95.7 & 93.4 / 98.1 & 97.0 / 98.0 & 81.9 / 94.2 & 95.7 / 98.0\\
& PCB3   & 90.1 / 98.5 & 88.1 / 94.7 & 86.2 / 96.7 & \textbf{94.5} / 98.7 & 93.7 / \textbf{99.1} & 93.3 / 96.2 & 90.5 / 97.4 & 88.5 / 98.2 & 89.8 / 97.7 & 87.4 / 96.5 & 94.0 / 98.9 \\
& PCB4   & 96.2 / 97.5 & 95.3 / 97.1 & 99.6 / 97.0 & \textbf{99.9} / 97.8 & \textbf{99.9} / 98.6 & 99.0 / \textbf{99.4} & 99.4 / 99.3 & 99.3 / 98.1 & 98.7 / 97.8 & 96.7 / 97.3 & 98.1 / 98.9\\ \midrule
\multirow{4}{*}{\rotatebox{90}{\shortstack{Multiple\\Instances}}}
& Macaroni1   & 90.5 / 93.3 & 92.6 / 98.6 & 85.7 / 94.1 & \textbf{97.5} / 99.6 & 91.6 / 99.5 & 93.1 / \textbf{99.9} & 96.0 / \textbf{99.9} & 88.7 / 99.1 & 93.7 / 99.4 & 88.0 / 98.2 & 95.3 / \textbf{99.9}\\
& Macaroni2   & 71.3 / 92.1 & 75.2 / 97.9 & 62.5 / 93.6 & 85.7 / 99.0 & 81.6 / 99.5 & 74.5 / 99.5 & 79.7 / 99.6 & 84.6 / 98.1 & \textbf{88.3} / 98.5 & 75.9 / 96.9 & 82.2 / \textbf{99.7}\\
& Capsules   & 91.4 / \textbf{99.6} & 90.6 / 99.4 & 58.2 / 97.3 & 79.5 / 99.1 & 91.8 / 99.1 & 88.8 / 99.3 & 89.1 / 99.0 & 74.8 / 98.4 & 80.1 / 97.6 & \textbf{93.6} / 97.0 & 88.5 / 98.6\\
& Candles   & 85.4 / 94.5 & 86.8 / 95.8 & 92.8 / 97.3 & 97.2 / \textbf{99.4} & 96.8 / 99.0 & 86.4 / 98.8 & 90.5 / 98.8 & 95.6 / 99.1 & \textbf{97.8} / 99.2 & 90.3 / 96.1 & 95.1 / \textbf{99.4}\\ \midrule
\multirow{4}{*}{\rotatebox{90}{\shortstack{Single\\Instance}}}
& Cashew   & 82.5 / 94.1 & 88.6 / 95.0 & 91.5 / 90.9 & 90.0 / 98.0 & 94.5 / 94.3 & 92.6 / 86.2 & 95.7 / 93.5 & 92.2 / 98.7 & 94.1 / 99.3 & 95.1 / 99.2 & \textbf{96.0} / \textbf{99.6}\\ 
& Chewing gum   & 96.0 / 98.9 & 96.4 / 99.0 & 99.1 / 94.7 & 99.0 / 98.6 & 97.7 / 98.1 & 98.0 / 99.6 & \textbf{99.4} / \textbf{99.7} & 99.1 / 98.1 & 99.3 / 99.5 & 98.0 / 99.3 & 99.1 / \textbf{99.7}\\
& Fryum   & 91.9 / 90.0 & 94.6 / 92.1 & 89.8 / 97.6 & 92.0 / \textbf{98.6} & 95.2 / 96.9 & 97.2 / 96.8 & \textbf{97.7} / 97.3 & 87.1 / 97.7 & 88.9 / 97.8 & 93.4 / 96.1 & 96.9 / 97.9\\
& Pipe Fryum   & 87.5 / 92.5 & 86.1 / 98.2 & 96.2 / 99.4 & 98.8 / 99.4 & 98.7 / 99.1 & 98.0 / 98.9 & 95.8 / 99.3 & 97.5 / 99.4 & 96.6 / 99.5 & 98.0 / 99.1 & \textbf{99.1} / \textbf{99.7}\\
\midrule
\multicolumn{2}{c|}{Mean}   & 87.1 / 95.3 & 87.8 / 96.6 & 86.8 / 96.0 & 94.2 / 98.9 & \textbf{94.3} / 98.5 & 90.1 / 97.4 & 93.2 / 98.1 & 91.3 / 98.5 & 93.4 / 98.6 & 90.5 / 97.5 & \textbf{94.3} / \textbf{99.2}\\ \bottomrule
  \end{tabular}}
  \end{center}
   \vspace{-6mm}
\end{table*}

\begin{table}[!t]
     \centering
    \renewcommand{\arraystretch}{1.52}  
	\caption{Quantitative comparison for \textbf{unimodal RGB UAD}. Multi-class detection and localization results on \textbf{additional baselines and benchmarks} using seven metrics, reporting class-wise \textbf{mean results} without fine-tuning. \textbf{Best} results are in bold.
    }
    \label{table25}
	\huge
	\resizebox{1\linewidth}{!}{
        \begin{tabular}{c|c|ccc|cccc}
\toprule
    \multirow{2}[2]{*}{\Huge Benchmark} & 
    \multicolumn{1}{|c|}{\multirow{2}[2]{*}{\Huge Method}} & 
    \multicolumn{3}{c|}{\Huge Image-level} & 
    \multicolumn{4}{c}{\Huge Pixel-level} \\
    \cmidrule{3-9}          
    &  & {\Huge AUROC} & {\Huge AP} & {\Huge F1-max} & {\Huge AUROC} & {\Huge AP} & {\Huge F1-max} & {\Huge AUPRO}  \\
    \midrule
    
\multirow{4}{*}{{\Huge MVTec-AD}}
  & {\Huge UniAD~\cite{uniad}}              & {\Huge 97.5} & {\Huge 99.1} & {\Huge 97.0} & {\Huge 96.9} & {\Huge 44.5} & {\Huge 50.5} & {\Huge 90.6} \\
  & {\cellcolor{lightblue}{\Huge  + UCF}}         
& {\cellcolor{lightblue}{\Huge 99.0}} 
& {\cellcolor{lightblue}{\Huge 99.7}} 
& {\cellcolor{lightblue}{\Huge 98.1}} 
& {\cellcolor{lightblue}{\Huge 97.5}} 
& {\cellcolor{lightblue}{\Huge 60.5}} 
& {\cellcolor{lightblue}{\Huge 59.9}} 
& {\cellcolor{lightblue}{\Huge 91.8}} \\

  \cline{2-9}
  & {\Huge Dinomaly~\cite{dinomaly}}           & {\Huge 99.6} & {\Huge \textbf{99.8}} & {\Huge 99.0} & {\Huge 98.3} & {\Huge 69.8} & {\Huge 68.7} & {\Huge 94.8} \\
  & {\cellcolor{lightblue}{\Huge  + UCF}}      
& {\cellcolor{lightblue}{\Huge \textbf{99.7}}}
& {\cellcolor{lightblue}{\Huge \textbf{99.8}}}
& {\cellcolor{lightblue}{\Huge \textbf{99.1}}}
& {\cellcolor{lightblue}{\Huge \textbf{98.7}}}
& {\cellcolor{lightblue}{\Huge \textbf{75.6}}}
& {\cellcolor{lightblue}{\Huge \textbf{72.9}}}
& {\cellcolor{lightblue}{\Huge \textbf{95.6}}} \\

\midrule
\multirow{4}{*}{{\Huge VisA}}
  & {\Huge UniAD~\cite{uniad}}              & {\Huge 91.5} & {\Huge 93.6} & {\Huge 88.5} & {\Huge 98.0} & {\Huge 32.7} & {\Huge 38.4} & {\Huge 76.1} \\
  & {\cellcolor{lightblue}{\Huge  + UCF}}         
& {\cellcolor{lightblue}{\Huge 92.1}} 
& {\cellcolor{lightblue}{\Huge 94.0}} 
& {\cellcolor{lightblue}{\Huge 88.9}} 
& {\cellcolor{lightblue}{\Huge 98.6}} 
& {\cellcolor{lightblue}{\Huge 34.0}} 
& {\cellcolor{lightblue}{\Huge 39.1}} 
& {\cellcolor{lightblue}{\Huge 86.4}} \\

  \cline{2-9}
  & {\Huge Dinomaly~\cite{dinomaly}}           & {\Huge 98.7} & {\Huge 98.8} & {\Huge 96.1} & {\Huge 98.7} & {\Huge 52.5} & {\Huge 55.4} & {\Huge 94.5} \\
 & {\cellcolor{lightblue}{\Huge  + UCF}}      
& {\cellcolor{lightblue}{\Huge \textbf{98.8}}}
& {\cellcolor{lightblue}{\Huge \textbf{99.0}}}
& {\cellcolor{lightblue}{\Huge \textbf{96.5}}}
& {\cellcolor{lightblue}{\Huge \textbf{98.9}}}
& {\cellcolor{lightblue}{\Huge \textbf{59.9}}}
& {\cellcolor{lightblue}{\Huge \textbf{59.9}}}
& {\cellcolor{lightblue}{\Huge \textbf{94.7}}} \\

\midrule
\multirow{4}{*}{{\Huge MPDD}}
  & {\Huge HVQ-Trans~\cite{hvqtrans}}          & {\Huge 86.5} & {\Huge 87.9} & {\Huge 85.6} & {\Huge 96.9} & {\Huge 26.4} & {\Huge 30.5} & {\Huge 88.0} \\
  & {\cellcolor{lightblue}{\Huge  + UCF}}     
& {\cellcolor{lightblue}{\Huge 93.1}} 
& {\cellcolor{lightblue}{\Huge 95.4}} 
& {\cellcolor{lightblue}{\Huge 90.3}} 
& {\cellcolor{lightblue}{\Huge 97.5}} 
& {\cellcolor{lightblue}{\Huge 34.1}} 
& {\cellcolor{lightblue}{\Huge 37.0}} 
& {\cellcolor{lightblue}{\Huge 82.9}} \\

  \cline{2-9}
  & {\Huge Dinomaly~\cite{dinomaly}}           & {\Huge 97.3} & {\Huge \textbf{98.5}} & {\Huge 95.6} & {\Huge 99.1} & {\Huge 60.0} & {\Huge 59.8} & {\Huge \textbf{96.7}} \\
  & {\cellcolor{lightblue}{\Huge  + UCF}}     
& {\cellcolor{lightblue}{\Huge \textbf{97.4}}}
& {\cellcolor{lightblue}{\Huge \textbf{98.5}}}
& {\cellcolor{lightblue}{\Huge \textbf{96.0}}}
& {\cellcolor{lightblue}{\Huge \textbf{99.2}}}
& {\cellcolor{lightblue}{\Huge \textbf{60.2}}}
& {\cellcolor{lightblue}{\Huge \textbf{59.9}}}
& {\cellcolor{lightblue}{\Huge \textbf{96.7}}} \\

\midrule
\multirow{4}{*}{{\Huge BTAD}}
  & {\Huge HVQ-Trans~\cite{hvqtrans}}          & {\Huge 90.9} & {\Huge 97.8} & {\Huge 94.8} & {\Huge 96.7} & {\Huge 43.2} & {\Huge 48.7} & {\Huge 75.6} \\
  & {\cellcolor{lightblue}{\Huge  + UCF}}    
& {\cellcolor{lightblue}{\Huge 93.3}} 
& {\cellcolor{lightblue}{\Huge 98.6}} 
& {\cellcolor{lightblue}{\Huge 96.0}}
& {\cellcolor{lightblue}{\Huge 97.3}} 
& {\cellcolor{lightblue}{\Huge 47.0}} 
& {\cellcolor{lightblue}{\Huge 50.2}} 
& {\cellcolor{lightblue}{\Huge 76.2}} \\

  \cline{2-9}
  & {\Huge Dinomaly~\cite{dinomaly}}           & {\Huge 95.4} & {\Huge 98.5} & {\Huge 95.5} & {\Huge 97.9} & {\Huge 70.1} & {\Huge 68.0} & {\Huge 76.5} \\
  & {\cellcolor{lightblue}{\Huge + UCF}}      
& {\cellcolor{lightblue}{\Huge \textbf{96.2}}}
& {\cellcolor{lightblue}{\Huge \textbf{98.6}}}
& {\cellcolor{lightblue}{\Huge \textbf{96.3}}} 
& {\cellcolor{lightblue}{\Huge \textbf{98.2}}} 
& {\cellcolor{lightblue}{\Huge \textbf{74.8}}} 
& {\cellcolor{lightblue}{\Huge \textbf{70.0}}} 
& {\cellcolor{lightblue}{\Huge \textbf{81.0}}} \\
\bottomrule
\end{tabular}}
\vspace{-2mm}
\end{table}

\subsection{Unimodal RGB Unsupervised Anomaly Detection}
\label{sec_rgb_uad}
\subsubsection{Datasets and Evaluation Metrics}

\textbf{Datasets.}
(1) \textbf{MVTec-AD}~\cite{mvtec} is a challenging and widely used benchmark comprising 5{,}354 images across 10 object and 5 texture classes, with 3{,}629 normal training and 1{,}725 test images covering diverse defects.
(2) \textbf{VisA}~\cite{visa} contains 10{,}821 images over 12 subsets (9{,}621 normal, 1{,}200 anomalous) spanning surface and structural defects (e.g., dents, scratches, cracks, misalignment).
(3) \textbf{MPDD}~\cite{mpdd} provides 1{,}346 images from 6 metal part categories, including 888 normal training and 458 test samples.
(4) \textbf{BTAD}~\cite{btad} consists of 2{,}540 images from 3 product categories, with 1{,}799 normal training and 741 test samples.

\textbf{Evaluation metrics.}
Following established practice~\cite{glad,dinomaly}, we report image-level AUROC (I-AUROC), AUPRC (I-AP), and F1-max (I-F1-max) for detection, and pixel-level AUROC (P-AUROC), AUPRC (P-AP), F1-max (P-F1-max), and AUPRO for localization. The main text primarily presents I-AUROC and P-AUROC; complete results for all metrics are provided in the Appendix, ensuring a rigorous and balanced assessment.

\subsubsection{Implementation Details}
We validate our method by integrating it with five recent multi-class UAD approaches: GLAD~\cite{glad} (diffusion-based), UniAD~\cite{uniad}, HVQ-Trans~\cite{hvqtrans}, Dinomaly~\cite{dinomaly} (transformer-based), and AnomalDF~\cite{anomalydino} (memory bank-based), following their original configurations for fair comparison. For GLAD and AnomalDF, three templates ($N{=}3$) are randomly sampled, either from 25 diffusion denoising steps or the same-category training set, while other baselines use a single template ($N{=}1$) since they do not reconstruct intermediates. The AnomalDF is a variant that dynamically samples templates from the full training set per input, improving diversity without extra memory cost compared with the original static few-shot protocol~\cite{anomalydino}. Anomalies are synthesized via Perlin-noise guided textures and structural perturbations~\cite{glad} with corresponding masks. All models are trained from scratch for 40 epochs (batch size 8) using the Adam optimizer, with the loss weight $\alpha$ set to 0.1 by default.

\begin{table*}[ht]
    \centering
    \caption{Quantitative comparison for \textbf{multimodal RGB–3D UAD}. Unsupervised anomaly detection and localization results on \textbf{MVTec 3D-AD} using I-AUROC, P-AUROC, and AUPRO@1\%/5\%/10\%/30\%. \textbf{Best} results are in bold and \underline{runner-ups} are underlined.}
    \renewcommand{\arraystretch}{1.15}
    \vspace{-1mm}
             \huge
             \setlength{\tabcolsep}{2pt}  
    \resizebox{1.0\textwidth}{!}{%
        \begin{tabular}{c|c|cccccccccc|c}
            \toprule
            &Method & Bagel & Cable Gland & Carrot & Cookie & Dowel & Foam & Peach & Potato & Rope & Tire & \textbf{Mean} \\ 
            \hline
            \multirow{6}{*}{\rotatebox{90}{\LARGE I-AUROC/P-AUROC}} 
            & BTF~\cite{btf}        & 91.8 / - & 74.8 / - & 96.7 / - & 88.3 / - & 93.2 / - & 58.2 / - & 89.6 / - & 91.2 / - & 92.1 / - & 88.6 / - & 86.5 / - \\
            & AST~\cite{ast}             & 98.3 / - & 87.3 / - & 97.6 / - & 97.1 / - & 93.2 / - & 88.5 / - & 97.4 / - & 98.1 / - & \textbf{100} / - & 79.7 / - & 93.7 / - \\  
            & M3DM~\cite{m3dm}    &  \textbf{99.4} / 99.5  &  90.9 / \textbf{99.4}  &  97.2 / 99.7  &  97.6 / 97.1  &  96.0 / \textbf{99.7}  &  \textbf{94.2} / 98.1  &  97.3 / 99.6  & 89.9 / 99.4 & 97.2 / 99.5 &  85.0 / 99.3  &  94.5 / 99.1 \\
           & {\cellcolor{lightblue}M3DM+UCF}   
& {\cellcolor{lightblue}99.3 / 99.3} 
& {\cellcolor{lightblue}91.6 / 99.2} 
& {\cellcolor{lightblue}\textbf{98.5} / \textbf{99.9}} 
& {\cellcolor{lightblue}99.0 / 97.1} 
& {\cellcolor{lightblue}\textbf{99.2} / 98.7} 
& {\cellcolor{lightblue}90.1 / \textbf{99.4}} 
& {\cellcolor{lightblue}\textbf{97.8} / \textbf{99.8}} 
& {\cellcolor{lightblue}92.5 / \textbf{99.9}} 
& {\cellcolor{lightblue}98.9 / \textbf{99.8}} 
& {\cellcolor{lightblue}84.9 / \textbf{99.8}} 
& {\cellcolor{lightblue}\underline{96.2} / \underline{99.3}} \\

            & CFM~\cite{cfm}                              &  \textbf{99.4} / \textbf{99.7}  &  88.8 / 99.2  &  98.4 / \textbf{99.9}  &  \textbf{99.3} / \textbf{97.2}  &  98.0 / 98.7  &  88.8 / 99.3  & 94.1 / \textbf{99.8} &  94.3 / \textbf{99.9}  &  98.0 / \textbf{99.8}  &  95.3 / \textbf{99.8}  &  95.4 / \underline{99.3}  \\
            & {\cellcolor{lightblue}CFM+UCF}
& {\cellcolor{lightblue}99.3 / \textbf{99.7}} 
& {\cellcolor{lightblue}\textbf{91.7} / 99.2} 
& {\cellcolor{lightblue}98.2 / \textbf{99.9}} 
& {\cellcolor{lightblue}99.2 / \textbf{97.2}} 
& {\cellcolor{lightblue}98.9 / 99.1} 
& {\cellcolor{lightblue}91.8 / 99.3} 
& {\cellcolor{lightblue}94.6 / \textbf{99.8}} 
& {\cellcolor{lightblue}\textbf{94.8} / \textbf{99.9}} 
& {\cellcolor{lightblue} 99.2 / 99.7} 
& {\cellcolor{lightblue}\textbf{95.9} / \textbf{99.8}} 
& {\cellcolor{lightblue}\textbf{96.4} / \textbf{99.4}} \\

            \hline 
            
            \multirow{6}{*}{\rotatebox{90}{\LARGE AUPRO@1\% / @5\%}}
             &BTF~\cite{btf}        &42.8 / -   &36.5 / -    & 45.2 / -    & 43.1 / -    & 37.0 / -    & 24.4 / -    & 42.7 / -    & 47.0 / -    & 29.8 / -    & 34.5 / -    & 38.3 / -   \\
            &AST~\cite{ast}              & 38.8 / -  & 32.2 / -  &  47.0 / -  & 41.1 / -  & 32.8 / -  & 27.5 / -  &  47.4 / -   &  48.7 / -   & 36.0 / -  & 47.4 / -   &  39.8 / -  \\  
           & M3DM~\cite{m3dm}     & 41.4 / 85.4   & 39.5 / \textbf{85.8}   & 44.7 / 88.1   & 31.8 / 78.9   &  \textbf{42.2} / \textbf{87.7}    &  33.5 / 75.5    & 44.4 / 88.5   & 35.1 / 84.9   &  41.6 / 85.8    & 39.8 / 83.7   & 39.4 / 84.4  \\
            & {\cellcolor{lightblue}M3DM+UCF}   
& {\cellcolor{lightblue} \textbf{48.0} / \textbf{88.2} } 
& {\cellcolor{lightblue} 41.0 / 82.2 } 
& {\cellcolor{lightblue} 48.3 / 89.3 } 
& {\cellcolor{lightblue} 44.4 / 82.9 } 
& {\cellcolor{lightblue} 39.2 / 76.9 } 
& {\cellcolor{lightblue} \textbf{41.5} / 82.3 } 
& {\cellcolor{lightblue} 48.7 / 89.0 } 
& {\cellcolor{lightblue} 49.8 / 89.7 } 
& {\cellcolor{lightblue} \textbf{47.2} / \textbf{86.9} } 
& {\cellcolor{lightblue} \textbf{47.5} / \textbf{89.0} } 
& {\cellcolor{lightblue} \underline{45.6} / 85.6 } \\

           & CFM~\cite{cfm}                              &  45.9 / 87.7    &  \textbf{43.1} / 84.3    &  48.5 / 89.4    &  \textbf{46.9} / \textbf{84.0}    &  39.4 / 76.5    &  41.3 / \textbf{82.8}    &  46.8 / 88.4    &  48.7 / 89.4    &  46.4 / 86.5    &  47.4 / 88.9    &  45.5 / \underline{85.8}   \\
            & {\cellcolor{lightblue}CFM+UCF}                             
& {\cellcolor{lightblue} \textbf{48.0} / \textbf{88.2} } 
& {\cellcolor{lightblue} 42.3 / 83.7 } 
& {\cellcolor{lightblue} \textbf{49.4} / \textbf{89.6} } 
& {\cellcolor{lightblue} 45.2 / 83.7 } 
& {\cellcolor{lightblue} 40.0 / 77.9 } 
& {\cellcolor{lightblue} 41.0 / 82.5 } 
& {\cellcolor{lightblue} \textbf{50.9} / \textbf{89.7} } 
& {\cellcolor{lightblue} \textbf{51.3} / \textbf{90.2} } 
& {\cellcolor{lightblue} \textbf{47.2} / \textbf{86.9} } 
& {\cellcolor{lightblue} \textbf{47.5} / \textbf{89.0} } 
& {\cellcolor{lightblue} \textbf{46.3} / \textbf{86.1} } \\
            \hline
            
            \multirow{6}{*}{\rotatebox{90}{\LARGE AUPRO@10\% / @30\%}}
           & BTF~\cite{btf}         & - / 97.6   &  - / 96.9  &  - / 97.9  &  - / 97.3   &  - / 93.3  &  - / 88.8  &  - / 97.5  &  - / 98.1   &  - / 95.0  &  - / 97.1  &  - / 95.9 \\
            & AST~\cite{ast}   & - / 97.0  & - / 94.7  & - / 98.1   & - / 93.9  & - / 91.3  &  - / 90.6  &  - / 97.9   &  - / 98.2   &  - / 88.9  &  - / 94.0  &  - / 94.4 \\  
            & M3DM~\cite{m3dm}      & 92.2 / 97.0   & \textbf{92.6} / 97.1   & 94.0 / 97.9   & 86.8 / \textbf{95.0}   & \textbf{93.8} / 94.1    & 84.3 / 93.2    & 94.2 / 97.7   & 92.5 / 97.1   & 92.2 / 97.1    & 90.7 / 97.5    & 91.3 / 96.4   \\
             & {\cellcolor{lightblue}M3DM+UCF}   
& {\cellcolor{lightblue} 94.0 / \textbf{98.0} } 
& {\cellcolor{lightblue} 90.5 / 96.8 } 
& {\cellcolor{lightblue} 94.6 / 98.2 } 
& {\cellcolor{lightblue} 88.9 / 94.0 } 
& {\cellcolor{lightblue} 86.0 / 95.2 } 
& {\cellcolor{lightblue} 90.1 / 96.6 } 
& {\cellcolor{lightblue} 94.5 / 98.2 } 
& {\cellcolor{lightblue} 94.9 / 98.3 } 
& {\cellcolor{lightblue} \textbf{92.8} / \textbf{97.6} } 
& {\cellcolor{lightblue} 94.5 / \textbf{98.2} } 
& {\cellcolor{lightblue} \underline{92.1} / \underline{97.1} } \\

            & CFM~\cite{cfm}                               & \textbf{94.1} / 97.9    & 90.0 / \textbf{97.2}    & 94.7 / 98.2    & 85.9 / 94.5   & 89.3 / 95.0    & 90.4 / \textbf{96.8}    & 94.3 / 98.0    & 94.9 / 98.2    & 92.4 / 97.5    & \textbf{94.7} / 98.1    & \underline{92.1} / \underline{97.1}   \\
            & {\cellcolor{lightblue}CFM+UCF}
& {\cellcolor{lightblue} 94.0 / \textbf{98.0} } 
& {\cellcolor{lightblue} 91.4 / 97.1 } 
& {\cellcolor{lightblue} \textbf{94.8} / \textbf{98.3} } 
& {\cellcolor{lightblue} \textbf{89.6} / 94.5 } 
& {\cellcolor{lightblue} 86.7 / \textbf{95.5} } 
& {\cellcolor{lightblue} \textbf{90.5} / \textbf{96.8} } 
& {\cellcolor{lightblue} \textbf{94.8} / \textbf{98.3} } 
& {\cellcolor{lightblue} \textbf{95.1} / \textbf{98.4} } 
& {\cellcolor{lightblue} \textbf{92.8} / \textbf{97.6} } 
& {\cellcolor{lightblue} 94.5 / \textbf{98.2} } 
& {\cellcolor{lightblue} \textbf{92.4} / \textbf{97.3} } \\

            \bottomrule

        \end{tabular}}
    
    \label{mvtec3d}
    \vspace{-2mm}
\end{table*}

\begin{table*}[!h]
    \centering
    \caption{Quantitative comparison for \textbf{multimodal RGB–3D UAD}. Unsupervised anomaly detection and localization results on \textbf{Eyecandies} using I-AUROC, P-AUROC, and AUPRO@1\%/5\%/10\%/30\%. \textbf{Best} results are in bold and \underline{runner-ups} are underlined.}
    \renewcommand{\arraystretch}{1.15}
    \vspace{-1mm}
             \huge
    \resizebox{1.0\textwidth}{!}{%
        \begin{tabular}{c|c|cccccccccc|c}
            \toprule
            & Method  & Can. C.  & Cho. C.  & Cho. P.  & Conf.  & Gum. B.  & Haz. T.  & Lic. S.  & Lollip.  & Marsh.  & Pep. C.  & \textbf{Mean}  \\
            \hline
            \multirow{5}{*}{\rotatebox{90}{\Large I-AUROC/P-AUROC}} 
            & AST~\cite{ast} & 57.4 / 76.3 & 74.7 / 96.0 & 74.7 / 91.1 & 88.9 / 96.9 & 59.6 / 78.8 & 61.7 / 83.7 & 81.6 / 91.8 & 84.1 / 92.4 & 98.7 / 98.3 & 98.7 / 96.8 & 78.0 / 90.2 \\  
            & M3DM~\cite{m3dm} & 55.2 / 96.4 & 86.6 / 98.0 & \textbf{94.6} / 96.1 & 98.4 / 99.8 & 86.1 / 97.3 & 66.1 / 93.8 & 90.1 / \textbf{97.8} & 88.4 / \textbf{98.7} & 98.7 / \textbf{99.6} & \textbf{99.2} / 99.5 & 86.3 / \underline{97.7} \\
            & {\cellcolor{lightblue}M3DM+UCF} & {\cellcolor{lightblue}\textbf{69.1} / 97.7} & {\cellcolor{lightblue}\textbf{97.0} / 98.2} & {\cellcolor{lightblue}91.8 / \textbf{96.5}} & {\cellcolor{lightblue}\textbf{99.5} / \textbf{99.9}} & {\cellcolor{lightblue}\textbf{89.6} / \textbf{97.4}} & {\cellcolor{lightblue}68.6 / 94.4} & {\cellcolor{lightblue}\textbf{91.0} / 97.5} & {\cellcolor{lightblue}78.5 / 98.6} & {\cellcolor{lightblue}\textbf{99.2} / \textbf{99.6}} & {\cellcolor{lightblue}97.9 / \textbf{99.7}} & {\cellcolor{lightblue}\textbf{88.2} / \textbf{97.9}} \\
            & CFM~\cite{cfm} & 50.7 / 97.9 & 91.4 / 98.3 & 88.2 / 95.2 & 89.3 / 98.5 & 79.0 / 95.2 & 75.4 / 93.7 & 88.0 / 96.3 & 82.7 / 98.2 & 97.6 / 99.3 & 83.8 / 97.1 & 82.6 / 97.0 \\
            & {\cellcolor{lightblue}CFM+UCF} & {\cellcolor{lightblue}61.3 / \textbf{98.5}} & {\cellcolor{lightblue}92.6 / \textbf{98.4}} & {\cellcolor{lightblue}89.1 / 96.4} & {\cellcolor{lightblue}89.9 / 98.8} & {\cellcolor{lightblue}85.3 / 95.4} & {\cellcolor{lightblue}\textbf{82.1} / \textbf{95.1}} & {\cellcolor{lightblue}87.7 / 97.3} & {\cellcolor{lightblue}\textbf{89.9} / 98.4} & {\cellcolor{lightblue}98.4 / 99.5} & {\cellcolor{lightblue}85.0 / 98.8} & {\cellcolor{lightblue}\underline{86.1} / \underline{97.7}} \\
            \hline

            \multirow{5}{*}{\rotatebox{90}{\Large AUPRO@1\% / @5\%}}
            & AST~\cite{ast} & 3.5 / 17.3 & 23.0 / 59.2 & 12.9 / 42.1 & 23.4 / 63.5 & 9.2 / 28.8 & 6.9 / 24.2 & 13.9 / 46.1 & 9.0 / 37.8 & 25.5 / 63.4 & 22.4 / 61.7 & 14.9 / 44.4 \\  
            & M3DM~\cite{m3dm} & 16.6 / 53.7 & 38.8 / 76.3 & 32.9 / 65.3 & 48.6 / \textbf{89.6} & 31.5 / 67.7 & 13.1 / 34.7 & 32.3 / 63.4 & 25.8 / 55.1 & 46.2 / 84.7 & 45.4 / 87.2 & 33.1 / 67.8 \\
            & {\cellcolor{lightblue}M3DM+UCF} & {\cellcolor{lightblue}21.3 / 56.4} & {\cellcolor{lightblue}\textbf{40.8} / \textbf{76.6}} & {\cellcolor{lightblue}\textbf{36.2} / \textbf{65.5}} & {\cellcolor{lightblue}\textbf{49.3} / 89.5} & {\cellcolor{lightblue}\textbf{35.2} / \textbf{69.8}} & {\cellcolor{lightblue}15.0 / 36.4} & {\cellcolor{lightblue}\textbf{34.1} / 63.8} & {\cellcolor{lightblue}\textbf{27.4} / 55.7} & {\cellcolor{lightblue}46.1 / 84.6} & {\cellcolor{lightblue}\textbf{47.4} / \textbf{87.7}} & {\cellcolor{lightblue}\textbf{35.3} / \underline{68.6}} \\
            & CFM~\cite{cfm} & \textbf{24.8} / 68.0 & 39.1 / 74.4 & 32.5 / 62.7 & 41.0 / 78.8 & 33.3 / 64.0 & \textbf{23.8} / 49.0 & 31.9 / 59.6 & 25.2 / 53.8 & 46.3 / 84.5 & 38.9 / 76.3 & 33.7 / 67.1 \\
            & {\cellcolor{lightblue}CFM+UCF} & {\cellcolor{lightblue}24.2 / \textbf{69.4}} & {\cellcolor{lightblue}38.0 / 75.3} & {\cellcolor{lightblue}34.2 / \textbf{65.5}} & {\cellcolor{lightblue}42.4 / 82.9} & {\cellcolor{lightblue}34.0 / 65.9} & {\cellcolor{lightblue}20.9 / \textbf{52.1}} & {\cellcolor{lightblue}\textbf{34.1} / \textbf{64.7}} & {\cellcolor{lightblue}26.9 / \textbf{58.3}} & {\cellcolor{lightblue}\textbf{46.6} / \textbf{85.1}} & {\cellcolor{lightblue}43.7 / 82.0} & {\cellcolor{lightblue}\underline{34.5} / \textbf{70.1}} \\
            \hline

            \multirow{5}{*}{\rotatebox{90}{\Large AUPRO@10\% / @30\%}}
            & AST~\cite{ast} & 28.5 / 51.4 & 70.9 / 83.5 & 54.5 / 71.4 & 77.0 / 90.5 & 40.4 / 58.7 & 35.0 / 59.0 & 58.4 / 73.6 & 54.4 / 76.9 & 77.0 / 91.8 & 74.4 / 87.8 & 57.0 / 74.4 \\  
            & M3DM~\cite{m3dm} & 72.7 / 88.0 & 82.5 / 89.1 & 71.6 / 78.8 & \textbf{94.8} / 98.1 & 76.2 / 88.9 & 46.9 / 65.9 & 70.4 / \textbf{86.3} & 72.3 / 90.9 & 90.3 / \textbf{96.3} & 93.3 / 97.0 & 77.1 / 87.9 \\
            & {\cellcolor{lightblue}M3DM+UCF} & {\cellcolor{lightblue}76.1 / 92.0} & {\cellcolor{lightblue}\textbf{82.8} / 90.3} & {\cellcolor{lightblue}71.4 / 80.8} & {\cellcolor{lightblue}94.7 / \textbf{98.2}} & {\cellcolor{lightblue}\textbf{77.6} / \textbf{90.0}} & {\cellcolor{lightblue}48.3 / 68.4} & {\cellcolor{lightblue}71.5 / 84.5} & {\cellcolor{lightblue}72.3 / 90.1} & {\cellcolor{lightblue}90.2 / 95.4} & {\cellcolor{lightblue}\textbf{93.6} / \textbf{97.8}} & {\cellcolor{lightblue}\underline{77.9} / \underline{88.8}} \\
            & CFM~\cite{cfm} & 83.9 / 94.6 & 81.2 / 89.8 & 70.7 / 81.2 & 87.9 / 95.9 & 72.8 / 87.2 & 61.1 / 79.4 & 64.8 / 77.0 & 73.4 / 91.1 & 90.1 / 94.3 & 84.4 / 93.8 & 77.0 / 88.4 \\
            & {\cellcolor{lightblue}CFM+UCF} & {\cellcolor{lightblue}\textbf{84.7} / \textbf{94.9}} & {\cellcolor{lightblue}82.7 / \textbf{90.8}} & {\cellcolor{lightblue}\textbf{73.6} / \textbf{84.8}} & {\cellcolor{lightblue}91.2 / 97.1} & {\cellcolor{lightblue}75.8 / 88.4} & {\cellcolor{lightblue}\textbf{66.4} / \textbf{83.4}} & {\cellcolor{lightblue}\textbf{72.2} / 84.3} & {\cellcolor{lightblue}\textbf{76.0} / \textbf{92.0}} & {\cellcolor{lightblue}\textbf{90.5} / 95.1} & {\cellcolor{lightblue}89.4 / 96.3} & {\cellcolor{lightblue}\textbf{80.3} / \textbf{90.7}} \\

            \bottomrule
        \end{tabular}}
    \label{eyecan}
    \vspace{-2mm}
\end{table*}

\begin{table*}[!t]
\centering
\caption{Quantitative comparison for \textbf{multimodal RGB–Text UAD}. Zero-shot anomaly detection and localization results on seven \textbf{industrial-domain} datasets. \textbf{Best} results are in bold and \underline{runner-ups} are underlined.} 
\label{industrial}
\vspace{-1mm}
\setlength\tabcolsep{2pt} 
\renewcommand{\arraystretch}{1.2}
\begin{tabular}{c|c|c|c|c|c|c>{\columncolor{lightblue}}c|c>{\columncolor{lightblue}}c|c>{\columncolor{lightblue}}c}
\toprule
Metrics &  Datasets & $|  \mathbb{C} |$ & CLIP-AC~\cite{clipac} & WinCLIP~\cite{winclip} & CoOp~\cite{coop}  & AprilGan~\cite{aprilgan} & \shortstack{ + UCF} & AdaCLIP~\cite{adaclip} & \shortstack{ + UCF} & \multicolumn{2}{c}{AnomalyCLIP~\cite{anomalyclip} + UCF}  \\ \hline

\multirow{8}{*}{\rotatebox{90}{\shortstack{Image-level\\AUROC/AP}}}
&MVTec-AD &15 &    71.5 / 86.4  &   91.8 / 96.5    &   88.8 / 94.8  &   86.2 / 93.6   & 91.1 / 95.7   & 89.9 / 95.7   & 91.5 / \textbf{96.7}   & 91.6 / 96.4   & \textbf{92.9} / 96.5    \\
& VisA &12  &    65.0 / 70.1  &    78.1 / 81.2    &   62.8 / 68.1      &    77.5 / 80.9    & 83.9 / 87.0   & 86.3 / 88.2   & \textbf{87.2} / \textbf{89.5}   & 82.0 / 85.3   & 82.6 / 85.7  \\
& MPDD &6    &  56.2 / 66.0  &  63.6 / 69.9  &     55.1 / 64.2     &    76.6 / 82.6    & 76.5 / 82.1   & 68.8 / 74.1   & 69.3 / 73.9   & 77.5 / 82.5   & \textbf{80.2} / \textbf{83.5}   \\
& BTAD &3    &  51.0 / 62.1  &  68.2 / 70.9  &     66.8 / 77.4      &    73.8 / 69.5    & 81.9 / 79.7   & 90.1 / \textbf{94.0}   & 90.3 / 92.5   & 88.2 / 88.2   & \textbf{92.1} / 90.9 \\
& SDD  &1   &  65.2 / 45.7  &    84.3 / 77.4   &   74.9 / 65.1   &    96.8 / 92.3    & 95.2 / 87.2   & 95.6 / 88.4   & 95.8 / 89.2   & 97.8 / \textbf{94.2}   & \textbf{97.9} / 93.5  \\
& DAGM &10  &  82.5 / 63.7  &  91.8 / 79.5  &       87.5 / 74.6   &    94.8 / 94.9   & 92.6 / 93.3   & 97.0 / 96.6   & 95.0 / 95.2   & 97.9 / 97.8   & \textbf{98.9} / \textbf{99.0}  \\
& DTD-Synthetic &12    &  66.8 / 83.2  &    93.2 / 92.6  &   -  /  -   &    85.5 / 94.0     & 92.1 / 97.7   & 91.6 / 95.5   & 94.9 / 97.9   & 93.9 / 97.2   & \textbf{97.2} / \textbf{98.9} \\  \cline{2-12}
& \textbf{Mean} & -    &  65.5 / 68.2  &    81.6 / 81.1  &   72.7 / 74.0   &    84.5 / 86.8     & 87.6 / 89.0   & 88.5 / 90.4   & 89.1 / 90.7   & \underline{89.8} / \underline{91.7}   & \textbf{91.7} / \textbf{92.6} \\ 
\hline

\multirow{8}{*}{\rotatebox{90}{\shortstack{Pixel-level\\AUROC/AUPRO}}}
&MVTec-AD&15   &  38.2 / 11.6  & 85.1 / 64.6    &    33.3 / 6.7   &    87.6 / 44.0    & 89.3 / 54.0   & 89.9 / 44.1   & 89.1 / 75.0   & 91.1 / 81.4   & \textbf{91.3} / \textbf{83.7}  \\
& VisA &12   &  47.8 / 17.3  &  79.6 / 56.8   &    24.2 / 3.8 &    94.2 / 86.6   & 94.8 / \textbf{88.1}   & \textbf{95.9} / 51.3   & 94.0 / 77.8   & 95.5 / 86.7   & 95.7 / 85.8    \\
& MPDD &6   &  58.7 / 29.1  &  76.4 / 48.9  &     15.4 / 2.3   &    94.3 / 83.8    & 95.9 / 87.5   & 96.1 / 30.6   & 94.1 / 63.6   & 96.5 / 88.7   & \textbf{96.7} / \textbf{89.9}  \\
& BTAD &3  &  32.8 / 8.3  &    72.7 / 27.3  &   28.6 / 3.8  &    89.3 / 68.7     & 91.7 / 67.4   & 93.7 / 20.2   & \textbf{95.9} / 41.6   & 94.2 / \textbf{75.4}   & 93.1 / \textbf{75.4} \\
& SDD &1   &  32.5 / 5.8  &  68.8 / 24.2  &     28.9 / 7.1   &    92.8 / 84.3    & 94.5 / 88.7   & 96.0 / 33.8   & 97.6 / 60.9   & \textbf{98.1} / \textbf{94.9}   & 97.5 / 91.8  \\
& DAGM &10  &  32.7 / 4.8  &    87.6 / 65.7  &   17.5 / 2.1  &    83.2 / 67.6    & 84.8 / 69.8   & 93.3 / 35.6   & 95.9 / 49.9   & 95.4 / \textbf{90.9}   & \textbf{96.6} / \textbf{90.9}  \\
& DTD-Synthetic &12   &  23.7 / 5.5  &  83.9 / 57.8  &    -  /  -   &    95.2 / 87.3     & 96.4 / 90.5   & 96.9 / 68.1   & 97.0 / 84.0   & 97.9 / 92.0   & \textbf{98.0} / \textbf{92.8} \\    \cline{2-12}
& \textbf{Mean} &-   &  38.1 / 11.8  &  79.2 / 49.3  &     24.7 / 4.3   &    90.9 / 74.6     & 92.5 / 78.0   & 94.5 / 40.5   & 94.8 / 64.7   & \underline{95.5} / \underline{87.1}   & \textbf{95.6} / \textbf{87.2} \\
\bottomrule
\end{tabular}
\vspace{-3mm}
\end{table*}

\begin{table*}[!t]
\centering
\caption{Quantitative comparison for \textbf{multimodal RGB–Text UAD}. Zero-shot anomaly detection and localization results on nine \textbf{medical-domain} datasets. \textbf{Best} results are in bold and \underline{runner-ups are underlined}.}
\label{medical1}
\vspace{-1mm}
\setlength\tabcolsep{3pt} 
\renewcommand{\arraystretch}{1.2}
\begin{tabular}{c|c|c|c|c|c|c>{\columncolor{lightblue}}c|c>{\columncolor{lightblue}}c|c>{\columncolor{lightblue}}c}
\toprule
Metrics &  Datasets & $|  \mathbb{C} |$ & CLIP-AC~\cite{clipac} & WinCLIP~\cite{winclip} & CoOp~\cite{coop}  & AprilGan~\cite{aprilgan} & + UCF & \multicolumn{2}{c|}{AnomalyCLIP~\cite{anomalyclip} + UCF}
 & AdaCLIP~\cite{adaclip} & + UCF  

\\ \hline
\multirow{4}{*}{\rotatebox{90}{\shortstack{Image-level\\AUROC/AP}}}
& HeadCT &1  &  60.0 / 60.7  &  81.8 / 80.2  &     78.4 / 78.8     &    86.9 / 87.8    &    90.7 / 91.1    &    93.0 / 91.1    &    96.5 / 96.2    &    97.3 / 97.4    &    \textbf{98.7} / \textbf{98.8}   \\           
& BrainMRI &1  &  80.6 / 86.4  &  86.6 /   91.5   &    61.3 / 44.9   &    92.7 / 93.7    &    93.7 / 95.5    &    90.0 / 92.1    &    95.4 / 95.8    &    96.8 / 97.3    &    \textbf{97.3} / \textbf{98.2}   \\
& Br35H &1 &  82.7 / 81.3  &  80.5 / 82.2  &     86.0 / 87.5      &    93.2 / 93.9    &    96.8 / 96.9    &    94.2 / 94.2    &    97.8 / 97.7    &    98.7 / \textbf{98.7}    &    \textbf{98.8} / \textbf{98.7}   \\  \cline{2-12}
& \textbf{Mean} &-  &  74.4 / 76.1  &  83.0 / 84.6  &     75.2 / 70.4      &    90.9 / 91.8    &    93.7 / 94.5    &    92.4 / 92.5    &    96.6 / 96.6    &    97.6 / 97.8    &    \textbf{98.3} / \textbf{98.6} \\
\hline

\multirow{7}{*}{\rotatebox{90}{\shortstack{Pixel-level\\AUROC/AUPRO}}}
& ISIC &1 &  36.0 / 7.7  &  83.3 / 55.1  &     51.7 / 15.9  &    90.0 / 80.2   &    91.6 / 82.6    &    89.4 / 78.4    &    \textbf{93.3} / \textbf{85.9}    &    90.2 / 18.8    &    89.4 / 45.5    \\
& ColonDB &1  &  49.5 / 11.5   &  70.3 / 32.5  &     40.5 / 2.6   &    78.2 / 65.0    &    80.0 / 65.9    &    81.9 / 71.2    &    83.2 / 75.2    &    \textbf{89.8} / 81.0    &    \textbf{89.8} / \textbf{82.1}    \\
& ClinicDB &1  &  48.5 / 12.6   &  51.2 / 13.8  &     34.8 / 2.4   &    79.2 / 57.0     &    80.6 / 58.6    &    81.5 / 62.1    &    84.4 / 69.1    &    90.3 / 53.9    &    \textbf{92.6} / \textbf{72.8}   \\
& Kvasir &1  &  45.0 / 16.8   &  69.7 / 24.5  &     44.1 / 3.5   &    75.0 / 36.3    &    78.4 / 39.0    &    79.0 / 45.4    &    81.7 / 41.9    &    95.1 / 36.2    &    \textbf{95.2} / \textbf{46.6}    \\
& Endo &1  &  46.6 / 12.6   &  68.2 / 28.3  &     40.6 / 3.9   &    81.9 / 54.9    &    84.5 / 61.5    &    84.2 / 63.4    &    87.1 / 70.0    &    \textbf{96.7} / 79.2    &    96.5 / \textbf{87.5}    \\
& TN3K &1 &  35.6 / 5.2  &  70.7 /   39.8   &     34.0 / 9.5   &    73.2 / 36.1    &    77.4 / 37.7    &    81.4 / \textbf{50.5}    &    \textbf{84.1} / 48.1    &    80.5 / 8.41    &    82.0 / 37.2   \\  \cline{2-12}
& \textbf{Mean} &- &  43.5 / 11.1  &  68.9 / 32.3  &     41.0 / 6.3      &    79.6 / 54.9    &    82.1 / 57.6   &    82.9 / 61.8    &    85.6 / \textbf{65.0}    & \underline{90.4} / 46.2    &    \textbf{90.9} / \underline{61.9}   \\

\bottomrule
\end{tabular}%
\vspace{-2mm}
\end{table*}

\subsubsection{Quantitative Comparison}
We reproduce the results of the five aforementioned unimodal multi-class UAD baselines and integrate our method into them. We also compare with advanced methods with distinct paradigms: synthetic-based JNLD~\cite{jnld}, CNN-based OmniAL~\cite{zhao2023omnial}, diffusion-based DiAD~\cite{diad} and VPDM~\cite{vpdm}, and Mamba-based MambaAD~\cite{mambaad}.

\textbf{Multi-class UAD on MVTec-AD.}
As reported in Table~\ref{table1}, integrating our method consistently enhances AUROC, with gains of 1.2\%/0.9\% for GLAD, 1.0\%/0.7\% for HVQ-Trans, and 1.7\%/0.7\% for AnomalDF (image/pixel), while additional metrics reported in the Appendix and Fig.~\ref{lamda} show even larger gains (up to 8\%). Notably, for texture anomalies such as the \textit{grid} category, our method improves the baselines by 4.9\%/1.4\%, 2.3\%/1.3\%, and 1.8\%/1.7\%, respectively, indicating effective suppression of the matching noise and better generalization.

\textbf{Multi-class UAD on VisA.}
VisA is more challenging due to complex structures and heterogeneous anomaly distributions. As shown in Table~\ref{table2}, our method yields consistent improvements: GLAD by 3.1\%/0.7\%, HVQ-Trans by 2.1\%/0.1\%, and AnomalDF by 3.8\%/1.7\% in image/pixel AUROC. Gains are especially pronounced in multi-instance categories (e.g., \textit{Macaroni1}, \textit{Candles}) and fine-grained single-instance categories (e.g., \textit{Cashew}, \textit{Chewing gum}). The overall mean attains 94.3\%/99.2\%, substantially enhancing the baselines and validating the effectiveness of our method across multi-classes.

\textbf{Multi-class UAD across additional baselines and benchmarks.}
Table~\ref{table25} evaluates our method on further baselines and benchmarks with comprehensive image- and pixel-level metrics. Across all settings, our method consistently enhances baseline performance. For example, on MVTec-AD, we improve UniAD from 96.9\%/90.6\% (P-AUROC/P-AUPRO) to 97.5\%/91.8\%, while Dinomaly rises from 98.3\%/94.8\% to 98.7\%/95.6\%. On VisA, UniAD is improved from 98.0\%/76.1\% to 98.6\%/86.4\%, and Dinomaly from 98.7\%/94.5\% to 98.9\%/94.7\%. On MPDD and BTAD, our method yields gains of up to 7\% in image-level AP and F1-max for HVQ-Trans and elevates Dinomaly to 81.0\% AUPRO on BTAD. These consistent gains further underscore the generality and effectiveness of our method.

\begin{figure*}
\setlength{\abovecaptionskip}{2pt}  
\setlength{\belowcaptionskip}{0pt} 
\begin{center}
\centerline{\includegraphics[width=\textwidth]{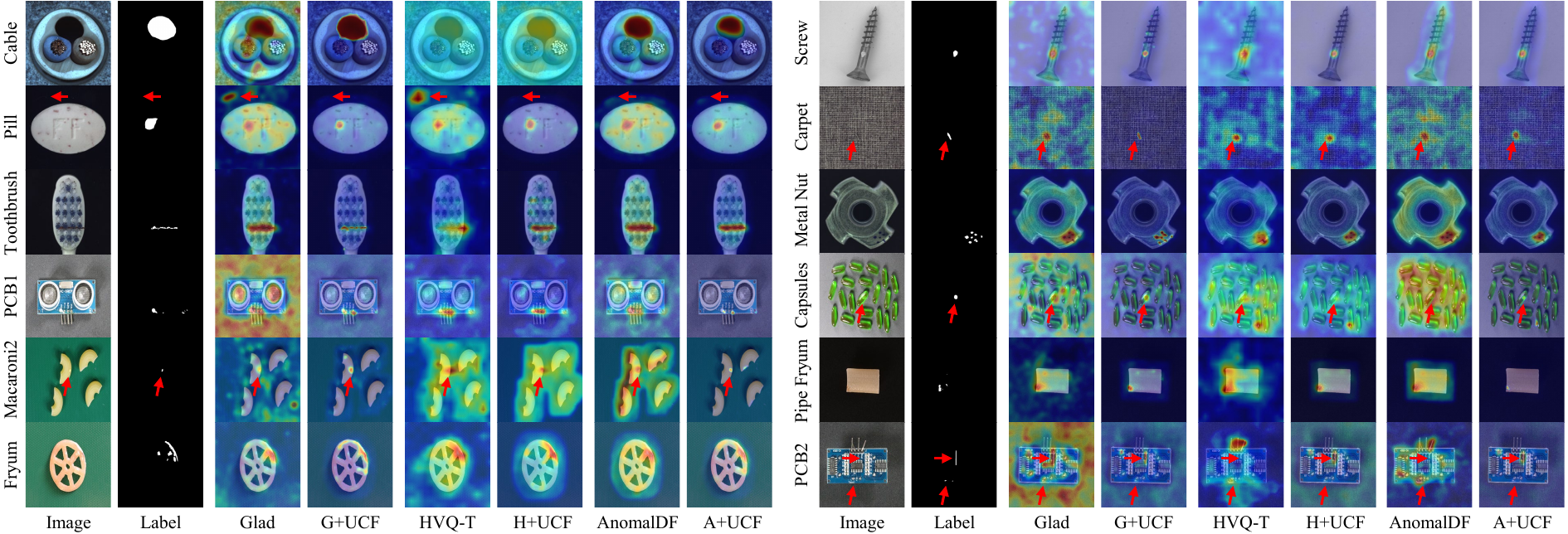}}
\vspace{-1mm}
\caption{Qualitative results of unimodal RGB UAD. We present a comparison of multi-class anomaly localization between our method and GLAD (G)~\cite{glad}, HVQ-Trans (H)~\cite{hvqtrans}, and AnomalDF (A)~\cite{anomalydino} on MVTec-AD~\cite{mvtec} (top 3 rows) and VisA~\cite{visa} (bottom 3 rows). By integrating with existing works, our method mitigates matching noise (e.g., false negatives in \textit{PCB2}, false positives in \textit{Pill}, and blurred boundaries in \textit{Carpet}), thus improving anomaly localization.}
\label{visualrgb}
\end{center}
\vspace{-6mm}
\end{figure*}

\subsubsection{Qualitative Comparison}
We further provide qualitative analyses on MVTec-AD and VisA to assess anomaly localization. As shown in Fig.~\ref{visualrgb}, baselines often exhibit severe matching noise, exhibited as blurred boundaries and spurious or missed regions that degrade segmentation accuracy. In contrast, our method suppresses these artifacts, yielding more refined and precise anomaly maps. Additional visualizations, such as localization comparisons and kernel density estimations of image- and pixel-level logits, are provided in the Appendix.

\subsection{Multimodal RGB-3D Unsupervised Anomaly Detection}
\label{sec_rgb3d_uad}

\subsubsection{Datasets and Evaluation Metrics}

We evaluate on two multimodal benchmarks with RGB and 3D data.
\textbf{MVTec 3D-AD}~\cite{mvtec3d} covers ten categories that include food items (e.g., \textit{peach}, \textit{carrot}) and industrial parts (e.g., \textit{cable gland}, \textit{rope}), with 2{,}656/294/1{,}197 train/val/test samples of paired RGB images and point clouds.
\textbf{Eyecandies}~\cite{eyecan} is a photorealistic synthetic conveyor-belt dataset with ten confectionery categories and 10{,}000/1{,}000/4{,}000 images, each with pixel-registered RGB–3D pairs.
Both datasets include image-level labels for anomaly detection and pixel-level masks for anomaly localization.

\textbf{Evaluation metrics.}
Following standard RGB–3D UAD protocols~\cite{mvtec3d}, we evaluate anomaly detection with image-level AUROC (I-AUROC) and localization with pixel-level AUROC (P-AUROC) and region-level AUPRO. While prior works~\cite{m3dm,mvtec3d,pami2} report AUPRO@30\% (calculated by integrating up to FPR=0.3), we also report AUPRO@10\%, @5\%, and @1\% to reflect stricter industrial tolerances, where smaller thresholds correspond to increasingly rigorous evaluation.
Given class imbalance and subtle anomalies in RGB-3D UAD benchmarks, AUPRO is a robust region-level metric and, together with I-/P-AUROC, provides a comprehensive evaluation protocol.

\subsubsection{Implementation Details}

We integrate UCF into two recent multimodal RGB-3D methods, M3DM~\cite{m3dm} and CFM~\cite{cfm}, following their original settings. Point clouds are pre-processed by fitting a background plane with RANSAC~\cite{btf,ast,pami2}, and the corresponding RGB pixels are masked to suppress background interference. The point clouds are then uniformly sampled with FPS~\cite{m3dm} into $1024$ groups of $32$ points, each embedded as a 1152-dimensional vector via Point-MAE~\cite{pointmae}. These 3D features are interpolated to full resolution and projected onto the 2D plane for pixel-level alignment with RGB features. Based on this alignment, UCF performs intra-modal matching for M3DM and cross-modal matching for CFM to construct and filter anomaly cost volumes. We adopt template selection following each baseline’s settings. Synthetic point-cloud anomalies are generated by projecting 2D anomaly masks and perturbing the corresponding regions, and the filtering network training matches that of unimodal RGB UAD.

\subsubsection {Quantitative Comparison}

We reproduce M3DM~\cite{m3dm} and CFM~\cite{cfm} and integrate our method into them to validate its effectiveness. We further compare with advanced methods, AST~\cite{ast} (distillation-based) and BTF~\cite{btf} (embedding-based).

\textbf{Unsupervised UAD on MVTec 3D-AD.}
Table~\ref{mvtec3d} reports image-level and pixel-level results. Our method consistently enhances both baselines. With M3DM~\cite{m3dm}, averages improve from 94.5\%/99.1\% to 96.2\%/99.3\% (I-AUROC/P-AUROC), with AUPRO@1\% rising from 39.4\% to 45.6\% and gains at AUPRO@5\% (+1.2\%). With CFM~\cite{cfm}, performance increases from 95.4\%/99.3\% to 96.4\%/99.4\%, with AUPRO@1\% growing from 45.5\% to 46.3\%, and moderate thresholds (AUPRO@10\%/@30\%) also improving. Gains are most pronounced for classes with cluttered geometry or fine structures (e.g., \textit{potato}, \textit{carrot}, \textit{peach}), where stricter region-level AUPRO and AUROC metrics both benefit substantially.

\textbf{Unsupervised UAD on Eyecandies.}
As indicated in Table~\ref{eyecan}, for M3DM~\cite{m3dm}, the mean results increase from 86.3\%/97.7\% to 88.2\%/97.9\% (I-AUROC/P-AUROC), and strict metrics AUPRO@1\%/@5\% improve from 33.1\%/67.8\% to 35.3\%/68.6\%, with further gains at @10\%/@30\%. Improvements are most evident in categories with severe color or shape ambiguity (e.g., \textit{Can.~C.} (+13.9\% I-AUROC, +4.7\% for AUPRO@1\%), \textit{Choc.~C.} (+10.4\%, +2.0\%)), where weak 3D geometry potentially misaligned with fine 2D cues. With CFM~\cite{cfm}, averages rise from 82.6\%/97.0\% to 86.1\%/97.7\% (I-AUROC/P-AUROC), accompanied by consistent gains across thresholds (e.g., AUPRO@1\%/@30\% from 33.7\%/88.4\% to 34.5\%/90.7\%). These results indicate that our method mitigates cross-modal matching noise and spurious correspondences, while preserving anomaly structures for more reliable detection.

\subsubsection{Qualitative Comparison} 

\begin{figure*}
\setlength{\abovecaptionskip}{2pt}  
\setlength{\belowcaptionskip}{0pt} 
\begin{center}
\centerline{\includegraphics[width=0.97\textwidth]{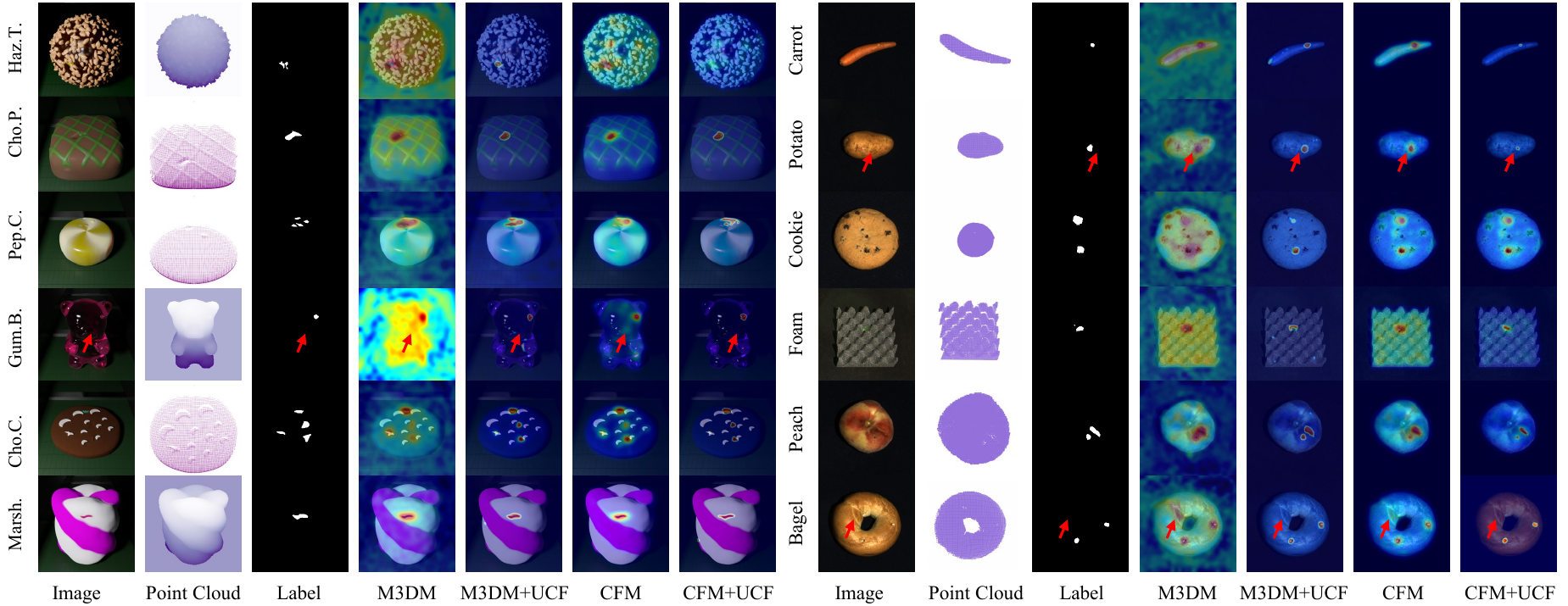}}
\vspace{-1mm}
\caption{Qualitative results of multimodal RGB–3D UAD. We compare our method against M3DM~\cite{m3dm} and CFM~\cite{cfm} on Eyecandies~\cite{eyecan} (left column) and MVTec 3D-AD~\cite{mvtec3d} (right column) for unsupervised anomaly localization. Our approach improves multimodal anomaly detection, effectively reducing noise and enhancing the localization of anomalies across both datasets.}
\label{visualrgb3d}
\end{center}
\vspace{-4mm}
\end{figure*}

We present qualitative comparisons on MVTec 3D-AD~\cite{mvtec3d} and Eyecandies~\cite{eyecan} for RGB-3D UAD. As indicated in Fig.~\ref{visualrgb3d}, M3DM and CFM often yield noisy or incomplete localization, with spurious activations on texture-rich regions (e.g., \textit{Potato}, \textit{Cookie} in RGB) and subtle defects on geometric structures (e.g., \textit{Gummy Bear (Gum. B.)}, \textit{Marshmallow (Marsh.)} in 3D). In contrast, our method produces cleaner anomaly maps by effectively leveraging RGB texture and 3D shape cues, capturing subtle defects while suppressing background noise. These results demonstrate suppressed intra-/cross-modal matching noise and enhanced anomaly cues.

\subsection{Multimodal RGB-Text Unsupervised Anomaly Detection}
\label{sec_rgbtext_uad}
\subsubsection{Datasets and Evaluation Metrics}  

\textbf{Industrial datasets.}
We evaluate RGB-Text UAD on seven industrial benchmarks: MVTec-AD~\cite{mvtec}, VisA~\cite{visa}, MPDD~\cite{mpdd}, BTAD~\cite{btad}, DAGM~\cite{dagm}, KSDD~\cite{ksdd}, and DTD-Synthetic~\cite{dtd}, spanning fine-grained surface scratches to large-scale structural defects.
\textbf{Medical datasets.}
To further assess cross-domain generalization, we evaluate on nine medical benchmarks: ISIC~\cite{isic} for skin-cancer (dermatology); CVC\mbox{-}ClinicDB~\cite{clidb}, CVC\mbox{-}ColonDB~\cite{clodb}, Kvasir~\cite{kvasir}, and Endo~\cite{endo} for colon\mbox{-}polyp (endoscopy); TN3K~\cite{tn3k} for thyroid\mbox{-}nodule (radiology); and HeadCT~\cite{headct}, BrainMRI~\cite{brainmri}, Br35H~\cite{br35h} for brain\mbox{-}lesion (CT/MRI). Together, they constitute a comprehensive testbed. Further details are provided in the Appendix.

\textbf{Evaluation metrics.}
We adopt the same image-level and pixel-level metrics as unimodal RGB UAD.

\subsubsection{Implementation Details}  

Following baselines~\cite{aprilgan,adaclip,anomalyclip}, we use CLIP (ViT-L/14@336px) as the common backbone, freezing the CLIP and baseline-specific fine-tuning layers. All images are resized to $518{\times}518$. In the zero-shot anomaly detection (ZSAD) setting, the categories of auxiliary and test datasets do not overlap, and we additionally synthesize anomalies on auxiliary images. Following the protocol of each baseline, (i) AnomalyCLIP/+UCF and AprilGAN/+UCF use MVTec-AD as the default auxiliary dataset to train the filtering network and are evaluated on the remaining 15 datasets; when evaluating on MVTec-AD, the auxiliary switches to VisA. (ii) AdaCLIP/+UCF uses MVTec-AD (industrial) and ClinicDB (medical) by default. When evaluating them, we set the auxiliary datasets to VisA and ColonDB. Our method is integrated without modifying baseline architectures. We report dataset-level results (category-wise mean) and provide category-level results in the Appendix. For AprilGAN~\cite{aprilgan} and AdaCLIP~\cite{adaclip}, category-specific text prompts are used for normal/abnormal states, whereas AnomalyCLIP~\cite{anomalyclip} employs category-agnostic text prompts (details in the Appendix). We train the filtering network as in our RGB/RGB–3D UAD setup.

\subsubsection {Quantitative Comparison}
We reproduce AprilGAN~\cite{aprilgan}, AdaCLIP~\cite{adaclip}, and AnomalyCLIP~\cite{anomalyclip}, integrate our method while keeping baseline settings unchanged, and also compare against CLIP-AC~\cite{clipac}, WinCLIP~\cite{winclip}, and CoOp~\cite{coop}.

\textbf{Zero-shot UAD on industrial datasets.}
Table~\ref{industrial} reports ZSAD results on seven industrial datasets with image-level (AUROC/AP) and pixel-level (AUROC/AUPRO) metrics following~\cite{anomalyclip}, where $|\mathbb{C}|$ is the number of classes per dataset. Class-level and other metrics are reported in the Appendix. At the \textbf{image} level, UCF consistently enhances baseline means: AprilGAN from 84.5\%/86.8\% to 87.6\%/89.0\%, AdaCLIP from 88.5\%/90.4\% to 89.1\%/90.7\%, and AnomalyCLIP from 89.8\%/91.7\% to 91.7\%/92.6\%. Gains are pronounced where anomalies are more subtle and diverse, e.g., VisA (+6.4\%/+6.1\% over AprilGAN), BTAD (+8.1\%/+10.2\%), and DTD-Synthetic (+6.6\%/+3.7\%). At the \textbf{pixel} level, AprilGAN rises from 90.9\%/74.6\% to 92.5\%/78.0\%, AnomalyCLIP from 95.5\%/87.1\% to 95.6\%/87.2\%, and AdaCLIP shows the largest effect, with AUPRO increasing from 40.5\% to 64.7\% at stable P-AUROC (94.5\%$\rightarrow$94.8\%). These gains reflect effective cost filtering that mitigates matching noise, while minor drops on some metric likely arise from the cross\mbox{-}dataset/domain heterogeneity of ZSAD, which exhibits higher variance than cross-class detection within a single dataset~\cite{diffad,zhao2023omnial,oneclass,liu2023simplenet}.

\textbf{Zero-shot UAD on medical datasets.}
Table~\ref{medical1} reports ZSAD on nine medical datasets. On datasets annotated only with \textbf{image}-level labels, AprilGAN is improved from 90.9\%/91.8\% to 93.7\%/94.5\%, and AnomalyCLIP from 92.4\%/92.5\% to 96.6\%/96.6\%; with AdaCLIP, the mean reaches 98.3\%/98.6\%. For datasets annotated only with \textbf{pixel}-level masks, localization gains are stronger: AprilGAN from 79.6\%/54.9\% to 82.1\%/57.6\%, AnomalyCLIP from 82.9\%/61.8\% to 85.6\%/65.0\%, and AdaCLIP with the largest jump in AUPRO (46.2\%$\rightarrow$61.9\%) at stable AUROC (90.4\%$\rightarrow$90.9\%). In particular, we reduce under-localization on ISIC (18.8\%$\rightarrow$45.5\%) (AUPRO), ClinicDB (53.9\%$\rightarrow$72.8\%), and Endo (79.2\%$\rightarrow$87.5\%), largely by curbing spurious activations and detecting anomalies that baselines fail to identify.

\begin{figure*}
\setlength{\abovecaptionskip}{2pt}  
\setlength{\belowcaptionskip}{0pt} 
\begin{center}
\centerline{\includegraphics[width=0.97\textwidth]{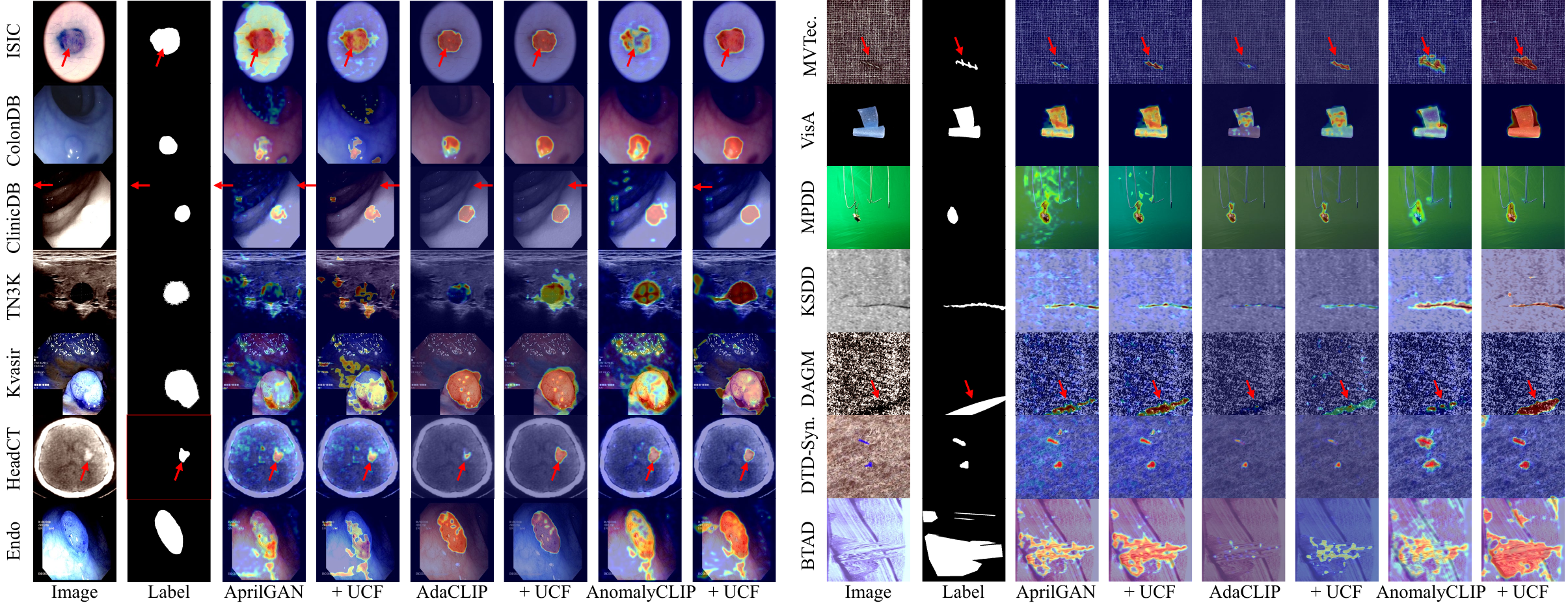}}
\vspace{-1mm}
\caption{Qualitative results of multimodal RGB–Text UAD. We compare our method with AprilGAN~\cite{aprilgan}, AdaCLIP~\cite{adaclip}, and AnomalyCLIP~\cite{anomalyclip} on representative categories from medical datasets (left column) and industrial datasets (right column). By integrating our cross-modal matching cost filtering with existing RGB–Text baselines, our method yields more precise and robust anomaly localization.}
\label{visualrgbtext}
\end{center}
\vspace{-6mm}
\end{figure*}

\subsubsection {Qualitative Comparison}

Fig.~\ref{visualrgbtext} shows qualitative results across medical (left) and industrial (right) datasets. Text-guided baselines (AprilGAN, AdaCLIP, AnomalyCLIP) frequently over-activate irrelevant regions (e.g., spurious highlights in MPDD and ISIC) or miss subtle defects (e.g., boundary losses in Kvasir and DAGM). With our method, anomaly maps become cleaner and more discriminative: medical lesions exhibit sharper contours, and industrial defects (scratches, texture inconsistencies) are localized without redundant noise. The results demonstrate that refining and denoising cross-modal matching cost enhances anomaly–text correspondence and leads to more reliable localization across heterogeneous datasets.

\subsection{Ablation Studies}
\vspace{-1mm}
\subsubsection{Unimodal RGB UAD}

We ablate \textit{components and losses} on MVTec-AD using the GLAD+UCF setting (Table~\ref{table4}). (i) Directly adopting stereo matching that maps correspondences to a \textit{depth dimension} built for local per-pixel disparity yields 87.8\%/89.0\%, likely because global multi-template matching may introduce cross-template contamination that inflates background responses and dilutes true matches. Mapping into the \textit{channel dimension}, which is largely spatially invariant and captures semantic subspaces, restores robust detection accuracy. (ii) Using \(\mathcal{C}_0\) (the final denoised step as template) achieves 96.2\%/96.8\%, and augmenting with $N{-}1$ randomly sampled intermediate denoised images brings +0.5\%/0.5\%, validating evidence aggregation across multiple template locations. (iii) The dual-stream attention guidance strengthens filtering: spatial guidance (SG) increases spatial attention to 97.8\%/97.5\%, and matching guidance (MG) improves channel matching to 98.3\%/97.8\%, illustrating that attention mechanisms dynamically select or suppress features as they adapt to the input. (iv) Focal loss serves as the base criterion; adding $\mathcal{L}_\text{CE}$ lifts performance to 98.5\%/98.0\%. Structural similarity and soft-IoU enforce structural consistency, and joint optimization reaches the highest 98.7\%/98.2\%. Overall, these ablations demonstrate that each component and loss contributes complementary gains, culminating in consistently stronger detection and localization.

\begin{table}[!t]
\centering
 \caption{Ablation studies on \textbf{unimodal RGB UAD} using GLAD+UCF on MVTec-AD. ``$DN$$\rightarrow$ depth/channel'' refers to mapping the matching dimension into the depth/channel dimension of the 3D U-Net. $\mathcal{C}_0$ denotes the volume uisng the final diffusion denoising step, $\mathcal{C}_{N-1}$ indicates uisng $N-1$ intermediate steps. SG and MG denote dual-stream attention guidance. $\mathcal{L}_\text{F}$ is focal loss~\cite{focalloss}, $\mathcal{L}_\text{CE}$ corresponds to the class-aware adaptor, and $\mathcal{L}_\text{S}$ is the combination of $\mathcal{L}_\text{SSIM}$~\cite{ssimloss} and $\mathcal{L}_\text{Soft-Iou}$~\cite{iouloss}.}
   \label{table4}
   \vspace{-1mm}
 \renewcommand{\arraystretch}{1.2}
 \resizebox{1.0\linewidth}{!}{
  \begin{tabular}{>{\centering\arraybackslash}p{0.7cm}|>{\centering\arraybackslash}p{0.3cm}|>{\centering\arraybackslash}p{0.8cm}|>{\centering\arraybackslash}p{0.55cm}|>{\centering\arraybackslash}p{0.55cm}|>{\centering\arraybackslash}p{0.6cm}|>{\centering\arraybackslash}p{0.7cm}|>{\centering\arraybackslash}p{0.6cm}|>{\centering\arraybackslash}p{1.2cm}}\toprule
  \multirow{1.5}{*}{$DN$$\rightarrow$} & \multicolumn{4}{c|}{$DN$$\rightarrow$ channel} & \multirow{2.5}{*}{\centering $\mathcal{L}_\text{F}$} &\multirow{2.5}{*}{\centering $\mathcal{L}_\text{CE}$} &\multirow{2.5}{*}{\centering $\mathcal{L}_\text{S}$}&\multirow{2.5}{*}{\centering Results}\\ \cmidrule{2-5}
  depth & \centering $\mathcal{C}_0$ & \centering $\mathcal{C}_{N-1}$ & \centering SG & \centering MG & & & \\ \midrule
  \checkmark & - & - & - & - & \checkmark & - & - & 87.8/89.0\\ 
  - & \checkmark & - & - & - & \checkmark & - & - & 96.2/96.8\\ 
  - & \checkmark & \checkmark & - & - & \checkmark & - & - & 96.7/97.3\\  
  - & \checkmark & \checkmark & \checkmark & - & \checkmark & - & - & 97.8/97.5\\ 
  - & \checkmark & \checkmark & \checkmark & \checkmark & \checkmark & - & - & 98.3/97.8\\ 
  - & \checkmark & \checkmark & \checkmark & \checkmark & \checkmark & \checkmark & - & 98.5/98.0\\ 
  - & \checkmark & - & \checkmark & \checkmark & \checkmark & \checkmark & \checkmark & 98.4/97.6\\ 
  - & \checkmark & \checkmark & \checkmark & \checkmark & \checkmark & \checkmark & \checkmark & \textbf{98.7}/\textbf{98.2}\\ 
  \bottomrule
  \end{tabular}}
  \vspace{-3mm}
\end{table}

\begin{table}[!t]
\centering
\caption{Ablation studies on \textbf{multimodal RGB-3D UAD} with M3DM (gray-shaded)/+UCF, assessing anomaly volume construction under missing modalities (ROC=AUROC, PRO=AUPRO).}
\label{ablationrgb3d}
\vspace{-1mm}
\setlength\tabcolsep{1.3pt}
\renewcommand{\arraystretch}{1.3}
\begin{tabular}{cc|cccc|cccc}
\toprule
\multicolumn{2}{c|}{Modality} & 
\multicolumn{4}{c|}{MVTec 3D-AD} & 
\multicolumn{4}{c}{Eyecandies} \\
\cmidrule{1-2} \cmidrule{3-6} \cmidrule{7-10}
RGB & 3D & I-ROC & P-ROC & PRO@1\% & @30\% & I-ROC & P-ROC & PRO@1\% & @30\% \\
\midrule
\checkmark &  \checkmark & {\cellcolor{cvprgray}{94.47}} & {\cellcolor{cvprgray}{99.13}} & {\cellcolor{cvprgray}{39.40}} & {\cellcolor{cvprgray}{96.37}} & {\cellcolor{cvprgray}{86.34}} & {\cellcolor{cvprgray}{97.70}} & {\cellcolor{cvprgray}{33.12}} & {\cellcolor{cvprgray}{87.93}} \\
\midrule
\checkmark &            & 94.68 & 99.07 & 44.06 & 96.56 & 87.20 & 97.83 & 34.50 & 88.49 \\
           & \checkmark & 94.70 & 99.14 & 44.40 & 96.90 & 85.99 & 97.87 & 34.52 & 88.58 \\
\checkmark & \checkmark & \textbf{96.18} & \textbf{99.29} & \textbf{45.56} & \textbf{97.11} & \textbf{88.22} & \textbf{97.94} & \textbf{35.28} & \textbf{88.75} \\
\bottomrule
\end{tabular}
  \vspace{-3mm}
\end{table}

\subsubsection{Multimodal RGB-3D UAD}

Beyond unimodal ablation as a representative case, we further evaluate robustness to missing modalities by building cost volumes from \textit{RGB-only, 3D-only, and joint RGB-3D features} (Table~\ref{ablationrgb3d}). Gray shading denotes the results of M3DM~\cite{m3dm} baseline, and our rows are unshaded. On \textbf{MVTec~3D-AD}, joint modeling raises I-AUROC from 94.68\% (RGB-only)/94.70\% (3D-only) to 96.18\%, P-AUROC from 99.07\%/99.14\% to 99.29\%, and AUPRO@1\% from 44.06\%/44.40\% to 45.56\%. On \textbf{Eyecandies}, I-AUROC improves from 87.20\%/85.99\% to 88.22\%, AUPRO@1\% from 34.50\%/34.52\% to 35.28\%. Notably, even with unimodal cost volumes, our method matches or surpasses the multimodal baseline (gray-shaded) on most metrics, while with multimodal inputs, it achieves further gains. These results demonstrate the complementarity of RGB texture and 3D geometry, and highlight the effectiveness of multimodal cost volumes and matching noise suppression via matching cost filtering.

\begin{table}[!t]
\centering
\caption{Ablation studies on \textbf{multimodal RGB-Text UAD} with AdaCLIP (gray-shaded)/+UCF, evaluating normal/abnormal prompts (single- and combined) for anomaly volume construction.
}
\label{ablationrgbtext}
\vspace{-1mm}
\setlength\tabcolsep{1.2pt}
\renewcommand{\arraystretch}{1.3}
\begin{tabular}{cc|cc|cc|cc}
\toprule

\multicolumn{2}{c|}{Prompts} & MVTecAD & VisA & HeadCT & Brain-MRI & TN3K & ClinicDB\\
\cmidrule{1-8}

Nor.  & Abn. & 
\multicolumn{2}{c|}{I-AUROC / AUPRO} & 
\multicolumn{2}{c|}{I-AUROC / I-AP} & 
\multicolumn{2}{c}{P-AUROC / P-AP}  \\

\midrule
\checkmark & \checkmark & {\cellcolor{cvprgray}{89.9/44.1}} & {\cellcolor{cvprgray}{86.3/51.3}} & {\cellcolor{cvprgray}{97.3/97.4}} & {\cellcolor{cvprgray}{96.8/97.3}} & {\cellcolor{cvprgray}{80.5/39.1}} & {\cellcolor{cvprgray}{90.3/69.0}}  \\
\midrule
\checkmark &            & 89.3/64.8 & 86.7/72.5 & 97.4/97.8 & 96.6/97.8 & 79.9/36.9 & 89.3/68.2  \\
           & \checkmark & 91.0/68.9 & 86.9/75.3 & 98.5/98.6 & 97.1/98.1 & 81.9/37.8 & 89.9/68.7  \\
\checkmark & \checkmark & \textbf{91.5/75.0} & \textbf{87.2/77.8} & \textbf{98.7/98.8} & \textbf{97.3/98.2} & \textbf{82.0/40.3} & \textbf{92.6/72.8}  \\
\bottomrule
\end{tabular}
  \vspace{-4mm}
\end{table}

\subsubsection{Multimodal RGB-Text UAD}

We further analyze template utilization for cost volume construction in RGB–Text UAD with AdaCLIP~\cite{adaclip}+UCF, comparing image feature matches against \textit{normal-only}, \textit{abnormal-only}, and \textit{joint (normal+abnormal)} templates on industrial and medical datasets (Table~\ref{ablationrgbtext}). Gray shading denotes the results of AdaCLIP, and our rows are unshaded. \textbf{Industrial:} on MVTec-AD, performance improves from 89.3\%/64.8\% (normal only) and 91.0\%/68.9\% (abnormal only) to 91.5\%/75.0\% with both. On VisA, P-AUPRO notably increases from 72.5\% and 75.3\% to 77.8\%. \textbf{Medical:} HeadCT and BrainMRI reach 98.7\%/98.8\% and 97.3\%/98.2\%, slightly exceeding single-prompt settings, while TN3K and ClinicDB gain more substantially, with P-AUROC/P-AP rising to 82.0\%/40.3\% and 92.6\%/72.8\%. Taken together, the results show that combining normal and abnormal prompts yields more discriminative matching cost volumes, thereby strengthening both anomaly detection and localization.

\subsection{Further Analysis and Discussion}

\begin{figure*}
\setlength{\abovecaptionskip}{2pt}  
\setlength{\belowcaptionskip}{0pt} 
\begin{center}
\centerline{\includegraphics[width=\textwidth]{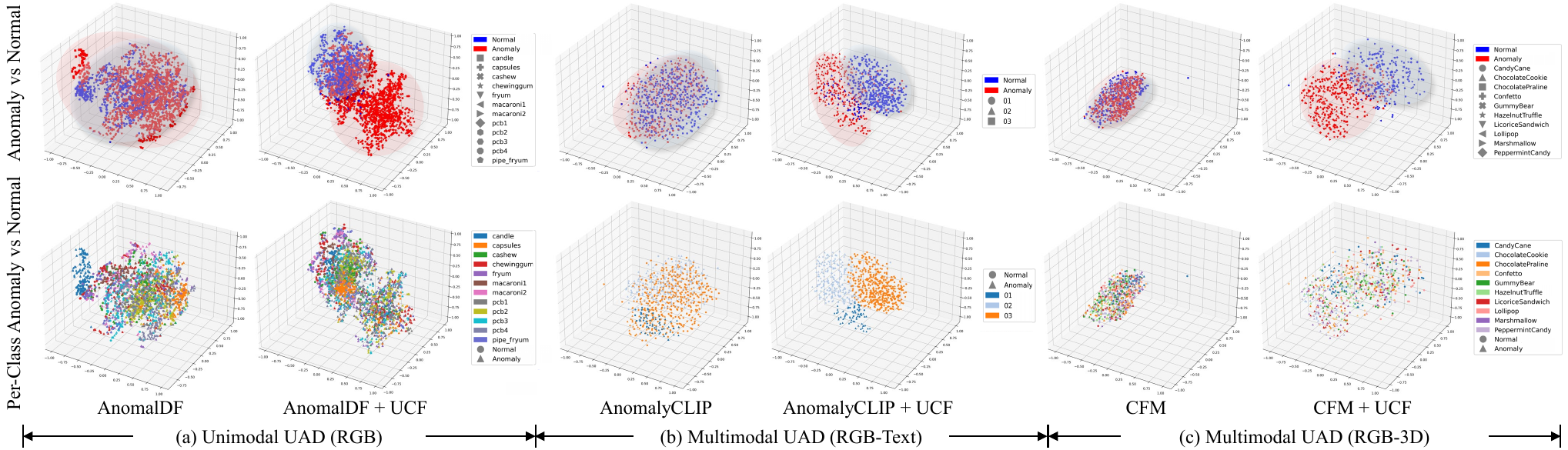}}
\vspace{-1mm}
\caption{T-SNE~\cite{tsne} visualization of cost volume features in (a) unimodal RGB, (b) multimodal RGB–Text, and (c) multimodal RGB–3D UAD. Top row: colors denote normal/abnormal patterns, shapes denote categories. Bottom row: shapes denote patterns, colors denote categories. Our filtering produces cleaner manifolds with sharper separation between anomalies and normals across multiple classes compared to the baselines.
}
\label{figtnse}
\end{center}
\vspace{-5mm}
\end{figure*}

\subsubsection{T-SNE Feature Visualization}

We visualize T\mbox{-}SNE~\cite{tsne} embeddings of cost volume features in three UAD scenarios (RGB, RGB-Text, RGB-3D), each before and after applying our method (Fig.~\ref{figtnse}). In row one, colors denote status (blue: normal; red: anomaly) and shapes encode categories; in row two, the encodings are swapped to enable direct cross\mbox{-}condition comparison. Baselines commonly yield entangled manifolds with mixed normals and anomalies, elongated category lobes, and narrow cross\mbox{-}class bridges that signal weak anomaly separability. After filtering, anomalies separate from normal cores, bridges recede, decision boundaries sharpen, and category clouds become compact and distinct. These changes indicate that our filter suppresses noisy correspondences while preserving anomaly cues, yielding cleaner and more separable manifolds consistent with gains across unimodal and multimodal UAD.

\begin{table}[!t]
\centering
 \Huge  
\caption{Comprehensive comparison of baselines and \mbox{+~UCF} in terms of memory usage (Mem.), per-image inference time (Inf.), parameter size (\#Params), and computational cost (FLOPs).}
\label{efficiency}
\renewcommand{\arraystretch}{1.5}
 \resizebox{\linewidth}{!}{
\begin{tabular}{c|c|c|c|c}
\toprule
{\Huge Method}  & {\Huge Mem. (GB)} & {\Huge Inf. (s/image)} & {\Huge \#Params} & {\Huge FLOPs} \\
\midrule
{\Huge UniAD\cite{uniad} / + UCF}         & {\Huge 4.53 / +0.56}   & {\Huge 0.01 / +0.04}  & {\Huge \phantom{.}\phantom{.}7.7M / +43.0M}   & {\Huge 198.0G / +26.0G\phantom{.}}  \\
{\Huge Glad\cite{glad} / + UCF}         & {\Huge 8.79 / +2.07}   & {\Huge 3.96 / +0.37}  & {\Huge \phantom{.}\phantom{.}1.3B / +43.8M}   & {\Huge \phantom{.}$>$2.2T / +32.7G\phantom{.}} \\
{\Huge HVQ-Trans\cite{hvqtrans} / + UCF}     & {\Huge 4.78 / +0.94}   & {\Huge 0.05 / +0.07}  & {\Huge \phantom{.}18.0M / +43.0M}  & {\Huge \phantom{.}\phantom{.}7.4G / +26.0G\phantom{.}}     \\
{\Huge AnomalDF\cite{anomalydino} / + UCF}       & {\Huge 3.25 / +0.82}   & {\Huge 0.31 / +0.32} & {\Huge \phantom{.}21.0M / +43.8M}  & {\Huge \phantom{.}\phantom{.}4.9G / +32.7G\phantom{.}}  \\
{\Huge Dinomaly\cite{dinomaly} / + UCF}   & {\Huge 4.32 / +1.11}   & {\Huge 0.11 / +0.05}  & {\Huge 132.8M / +43.6M} & {\Huge 104.7G / +14.3G\phantom{.}}    \\
\midrule
{\Huge M3DM\cite{m3dm} / + UCF}  & {\Huge 6.52 / +1.17}   & {\Huge 14.9 / +2.60} & {\Huge 122.9M / +44.1M} & {\Huge 794.5G / +73.7G\phantom{.}}   \\
{\Huge CFM\cite{cfm} / + UCF}      & {\Huge 4.20 / +1.01}   & {\Huge 0.23 / +0.17}  & {\Huge 112.6M / +43.4M}  & {\Huge 431.1G / +106.2G} \\
\midrule
{\Huge AprilGAN\cite{aprilgan} / + UCF}      & {\Huge 3.70 / +0.26}   & {\Huge 0.19 / +0.01}    & {\Huge 202.1M / +43.3M}  & {\Huge 276.6G / +23.8G\phantom{.}}  \\
{\Huge AdaCLIP\cite{adaclip} / + UCF}      & {\Huge 3.26 / +0.15}   & {\Huge 0.31 / +0.04}  & {\Huge 404.1M / +43.3M}  & {\Huge 1.111T / +23.8G\phantom{.}}  \\
{\Huge AnomalyCLIP\cite{anomalyclip} / + UCF}   & {\Huge 3.38 / +0.26}   & {\Huge 0.14 / +0.15}  & {\Huge 286.0M / +43.3M} & {\Huge 434.7G / +6.51G\phantom{.}}    \\
\bottomrule
\end{tabular}}
\vspace{-4mm}
\end{table}

\subsubsection{Time and Memory Efficiency}

Table~\ref{efficiency} reports memory usage, per-image inference latency, parameter size, and FLOPs on an A100-40GB GPU (batch size = 1), with and without our method. Memory and time overheads are marginal. For unimodal RGB UAD, memory increases by $+0.56$ to $+2.07$~GB and inference latency by $+0.04$ to $+0.37$~s/img. For RGB–3D UAD, costs remain bounded (M3DM: $+1.17$~GB, $+2.60$~s/img; CFM: $+1.01$~GB, $+0.17$~s/img), with most slowdown attributable to multimodal feature matching. For RGB–Text UAD, the overhead also remains small. Parameter growth is nearly constant at $+43$M, since the filtering head has fixed capacity and attaches without modifying the frozen baseline backbone; minor variations stem from projecting cost volumes of different channel sizes (e.g., 196, 768, 1024) into a unified 96-dimensional space. FLOPs overhead is modest for RGB UAD and even smaller for RGB–Text UAD, whereas RGB–3D UAD incurs higher additions due to dense anomaly cost volumes. In summary, our method delivers consistent performance gains with bounded memory growth and limited runtime cost, highlighting its effectiveness, efficiency, generality, and deployment readiness for unified anomaly detection.

\begin{figure}
\setlength{\abovecaptionskip}{2pt}  
\setlength{\belowcaptionskip}{0pt} 
\begin{center}
\centerline{\includegraphics[width=0.5\textwidth]{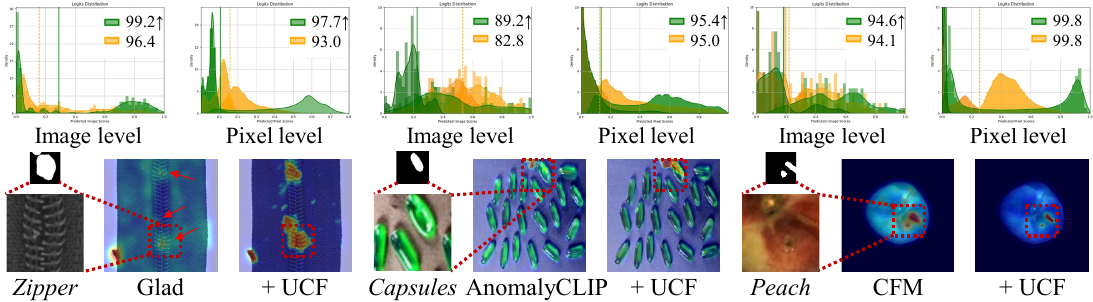}}
\vspace{-1mm}
\caption{Joint visualization of logit distributions and anomaly heatmaps for different pipelines: unimodal RGB (GLAD~\cite{glad}), multimodal RGB–Text (AnomalyCLIP~\cite{anomalyclip}), and multimodal RGB–3D (CFM~\cite{cfm}). Compared to baselines (\textcolor{myyellow}{yellow}), our method (\textcolor{mygreen}{green}) compacts the normal manifold, shifts anomalies toward higher logits, and reduces overlap, while concurrently generating more accurate anomaly localization results across these pipelines.}
\label{heatlogits}
\end{center}
\vspace{-6mm}
\end{figure}

\subsubsection{Coupling Heatmaps with Logit Distributions}

We link anomaly detection distributions with spatial evidence by jointly examining logits and heatmaps in Fig.~\ref{heatlogits}. For unimodal and multimodal settings, we plot KDE curves~\cite{kde} of image and pixel logits alongside heatmaps, zoom-ins, and ground-truth masks. In all cases, our method (green curves) shows clearer normal versus anomaly separation with markedly reduced overlap. Spatially, baselines exhibit false negatives, irrelevant highlights, and fragmented responses, whereas our filter yields sharper boundaries and compact activations confined to the defect region. Taken together, these distributional and spatial improvements indicate that the filter denoises the cost volume, compacts the normal manifold, and strengthens anomaly discrimination at both image and pixel levels, consistent with the quantitative gains across modalities.

\begin{table}[!t]
     \centering
    \renewcommand{\arraystretch}{1.3}  
    \caption{Few-shot exploration for \textbf{multimodal RGB–Text UAD} on VisA, comparing WinCLIP~\cite{winclip}, AprilGAN~\cite{aprilgan}, and AprilGAN+UCF.}

    \label{fewshotvisa}%
    \vspace{-1mm}
    \resizebox{1.0\linewidth}{!}{
	\setlength{\tabcolsep}{2.95pt}  
	\huge 
        \begin{tabular}{c|c|ccc|cccc}
    \toprule
    \multirow{2}[2]{*}{Shots} & 
    \multicolumn{1}{|c|}{\multirow{2}[2]{*}{  Method}} & 
    \multicolumn{3}{c|}{  Image-level} & 
    \multicolumn{4}{c}{  Pixel-level} \\
    \cmidrule{3-9}          
    &  &    AUROC  &    AP  &    F1-max  &    AUROC  &    AP  &    F1-max  &    AUPRO   \\
    \midrule

			\multirow{3}[1]{*}  {0-shot}   &    WinCLIP\cite{winclip}  &    78.1   &    81.2   &    79.0   &    79.6   &    -   &    14.8   &    56.8  \\
			&    AprilGAN\cite{aprilgan}  &      77.5    &    80.9   &      78.6    &      94.2    &   25.8       &      32.3    &     86.6    \\
     
            & {\cellcolor{lightblue}\cite{aprilgan} + UCF}  
            & {\cellcolor{lightblue}\textbf{83.9}}    
            & {\cellcolor{lightblue}\textbf{87.0}}   
            & {\cellcolor{lightblue}\textbf{80.9}}    
            & {\cellcolor{lightblue}\textbf{94.8}}   
            & {\cellcolor{lightblue}\textbf{27.1}}    
            & {\cellcolor{lightblue}\textbf{33.2}}    
            & {\cellcolor{lightblue}\textbf{88.1}} \\

			\midrule
			\multirow{3}[1]{*}     {1-shot}   &    WinCLIP\cite{winclip}  &    83.8$_{\text{\LARGE ±4.0}}$   &    85.1$_{\text{\LARGE ±4.0}}$   &    83.1$_{\text{\LARGE ±1.7}}$   &    96.4$_{\text{\LARGE ±0.4}}$   &    -   &    41.3$_{\text{\LARGE ±2.3}}$   &    85.1$_{\text{\LARGE ±2.1}}$  \\
            &    AprilGAN\cite{aprilgan}  &    91.2$_{\text{\LARGE ±0.8}}$   &    93.3$_{\text{\LARGE ±0.8}}$   &    86.9$_{\text{\LARGE ±0.6}}$   &    96.0$_{\text{\LARGE ±0.0}}$   &    30.9$_{\text{\LARGE ±0.3}}$   &    38.5$_{\text{\LARGE ±0.3}}$   &    90.1$_{\text{\LARGE ±0.1}}$  \\

            & {\cellcolor{lightblue}\cite{aprilgan} + UCF} 
            & {\cellcolor{lightblue}\textbf{93.2$_{\text{\LARGE ±0.9}}$}}   
            & {\cellcolor{lightblue}\textbf{94.7$_{\text{\LARGE ±0.7}}$}}   
            & {\cellcolor{lightblue}\textbf{88.8$_{\text{\LARGE ±0.6}}$}}   
            & {\cellcolor{lightblue}\textbf{97.5$_{\text{\LARGE ±0.1}}$}}   
            & {\cellcolor{lightblue}\textbf{37.1$_{\text{\LARGE ±0.2}}$}}   
            & {\cellcolor{lightblue}\textbf{43.9$_{\text{\LARGE ±0.1}}$}}   
            & {\cellcolor{lightblue}\textbf{92.9$_{\text{\LARGE ±0.1}}$}} \\

            \midrule
            
            \multirow{3}[1]{*}     {2-shot}   &   WinCLIP\cite{winclip}  &  84.6$_{\text{\LARGE ±2.4}}$   &    85.8$_{\text{\LARGE ±2.7}}$   &    83.0$_{\text{\LARGE ±1.4}}$   &    96.8$_{\text{\LARGE ±0.3}}$   &    -   &    43.5$_{\text{\LARGE ±3.3}}$   &    86.2$_{\text{\LARGE ±1.4}}$  \\
            &    AprilGAN\cite{aprilgan}  &    92.2$_{\text{\LARGE ±0.3}}$   &    94.2$_{\text{\LARGE ±0.3}}$   &    87.7$_{\text{\LARGE ±0.3}}$   &    96.2$_{\text{\LARGE ±0.0}}$   &    31.6$_{\text{\LARGE ±0.3}}$   &    39.3$_{\text{\LARGE ±0.2}}$   &    90.1$_{\text{\LARGE ±0.1}}$  \\
             
            & {\cellcolor{lightblue}\cite{aprilgan} + UCF} 
            & {\cellcolor{lightblue}\textbf{93.9$_{\text{\LARGE ±0.3}}$}}   
            & {\cellcolor{lightblue}\textbf{95.3$_{\text{\LARGE ±0.3}}$}}   
            & {\cellcolor{lightblue}\textbf{89.4$_{\text{\LARGE ±0.4}}$}}   
            & {\cellcolor{lightblue}\textbf{97.6$_{\text{\LARGE ±0.1}}$}}   
            & {\cellcolor{lightblue}\textbf{38.9$_{\text{\LARGE ±1.7}}$}}   
            & {\cellcolor{lightblue}\textbf{45.4$_{\text{\LARGE ±1.4}}$}}   
            & {\cellcolor{lightblue}\textbf{93.2$_{\text{\LARGE ±0.1}}$}} \\

            \midrule
            
            \multirow{3}[1]{*}     {4-shot}   &    WinCLIP\cite{winclip}  &    87.3$_{\text{\LARGE ±1.8}}$   &    88.8$_{\text{\LARGE ±1.8}}$   &    84.2$_{\text{\LARGE ±1.6}}$   &    97.2$_{\text{\LARGE ±0.2}}$   &    -   &    \textbf{47.0$_{\text{\LARGE ±3.0}}$}   &    87.6$_{\text{\LARGE ±0.9}}$  \\
            &    AprilGAN\cite{aprilgan}  &    92.6$_{\text{\LARGE ±0.4}}$   &    94.5$_{\text{\LARGE ±0.3}}$   &    88.4$_{\text{\LARGE ±0.5}}$   &    96.2$_{\text{\LARGE ±0.0}}$   &    32.2$_{\text{\LARGE ±0.1}}$   &    40.0$_{\text{\LARGE ±0.1}}$   &    90.2$_{\text{\LARGE ±0.1}}$  \\
            
            & {\cellcolor{lightblue}\cite{aprilgan} + UCF}  
            & {\cellcolor{lightblue}\textbf{94.7$_{\text{\LARGE ±0.2}}$}}   
            & {\cellcolor{lightblue}\textbf{95.9$_{\text{\LARGE ±0.1}}$}}   
            & {\cellcolor{lightblue}\textbf{90.4$_{\text{\LARGE ±0.5}}$}}   
            & {\cellcolor{lightblue}\textbf{97.8$_{\text{\LARGE ±0.1}}$}}   
            & {\cellcolor{lightblue}\textbf{40.3$_{\text{\LARGE ±0.8}}$}}   
            & {\cellcolor{lightblue}46.9$_{\text{\LARGE ±0.9}}$}   
            & {\cellcolor{lightblue}\textbf{93.5$_{\text{\LARGE ±0.2}}$}} 
            \\
            
			\bottomrule
		\end{tabular}%
	}
       \vspace{-4mm}
\end{table}%

\subsubsection{Few-shot Exploration on Multimodal RGB-Text UAD}
\label{sec_rgbtextfew_uad}

Following AprilGAN~\cite{aprilgan}, we reuse the zero-shot model without additional training and provide a small number of randomly selected normal samples from the target category as references. For each shot level, we repeat the sampling under five random seeds and report the mean and standard deviation.

Few-shot results on \textbf{VisA}. Table~\ref{fewshotvisa} reports the results of 0-, 1-, 2-, and 4-shots. Our method consistently enhances AprilGAN~\cite{aprilgan}, a representative successor to WinCLIP~\cite{winclip}. Under the 1-shot configuration, it achieves 93.2\% I-AUROC and 92.9\% AUPRO, delivering about a 2\% gain over AprilGAN. Similar improvements are observed at 2- and 4-shots, with margins up to 6.9\% in P-F1-max and 3.3\% in AUPRO. These results demonstrate that our design substantially strengthens few-shot performance and scales effectively with minimal reference. Notable gains are also achieved on \textbf{MVTec-AD}, with further details provided in the Appendix.

\begin{table}[!t]
\centering
 \normalsize
\caption{Compatibility exploration with various templates, comparing reconstruction-, embedding-, or hybrid-based templates.  }
\label{hybrid}
\vspace{-1mm}
\setlength{\tabcolsep}{5pt}
  \renewcommand{\arraystretch}{1.1}  
\resizebox{0.9\linewidth}{!}{
\begin{tabular}{cc|cc|cc|cc}
\toprule
\multicolumn{2}{c|}{Train} 
& \multicolumn{2}{c|}{Test} 
& \multicolumn{2}{c|}{MVTec-AD} 
& \multicolumn{2}{c}{VisA} \\
\cmidrule{1-8}
Recon. & Embed.
& Recon. & Embed. 
& Image & Pixel 
& Image & Pixel \\
\midrule

\checkmark &           &           & \checkmark & 97.5 & 97.1 & 92.6 & 98.0 \\
\checkmark &           & \checkmark &           & {\cellcolor{lightblue}{98.7}} & {\cellcolor{lightblue}{\textbf{98.2}}} & {\cellcolor{lightblue}{\textbf{93.2}}} & {\cellcolor{lightblue}{98.1}} \\
  \checkmark       & \checkmark &    \checkmark    &  & {\cellcolor{cvprpurple}{\textbf{98.8}}} & {\cellcolor{cvprpurple}{98.1}} & {\cellcolor{cvprpurple}{93.1}} & {\cellcolor{cvprpurple}{\textbf{98.2}}} \\   \midrule
          & \checkmark & \checkmark &           & 94.5 & 98.0 & 85.6 & 96.9 \\
& \checkmark & &     \checkmark       & {\cellcolor{lightblue}{98.5}} & {\cellcolor{lightblue}{98.8}} & {\cellcolor{lightblue}{\textbf{94.3}}} & {\cellcolor{lightblue}{99.2}} \\
     
 \checkmark & \checkmark &           & \checkmark & {\cellcolor{cvprpurple}{\textbf{98.6}}} & {\cellcolor{cvprpurple}{\textbf{98.9}}} & {\cellcolor{cvprpurple}{92.9}} & {\cellcolor{cvprpurple}{\textbf{99.3}}} \\\bottomrule
\end{tabular}
}
       \vspace{-3mm}
\end{table}

\subsubsection{Compatibility Exploration with Different Template Types}

Fundamentally, UCF is agnostic to how anomaly cost volumes are constructed: it accommodates both reconstruction- and embedding-based features as well as arbitrary modality combinations. For fairness, all experiments adhere to each baseline’s original design. In unimodal RGB UAD, for instance, volumes are constructed either from reconstruction matchings (Recon.) or from embedding matchings (Embed.), with the blue rows in Table~\ref{hybrid} reporting the main results. To assess complementarity, we train a unified model (Hybrid) using anomaly volumes alternately sourced from both reconstruction- and embedding-based templates. The purple rows show consistent gains over single-template pipelines across most metrics, indicating that the two cost-volume types provide complementary cues that our filter exploits without architectural modifications. Although demonstrated on RGB UAD, the same principle naturally extends to multimodal integration, with unified volumes formed from RGB, 3D, and text matchings through either concatenation or alternation, highlighting a promising direction.

\begin{figure}[t]
\setlength{\abovecaptionskip}{2pt}  
\setlength{\belowcaptionskip}{0pt} 
\begin{center}
\centerline{\includegraphics[width=0.5\textwidth]{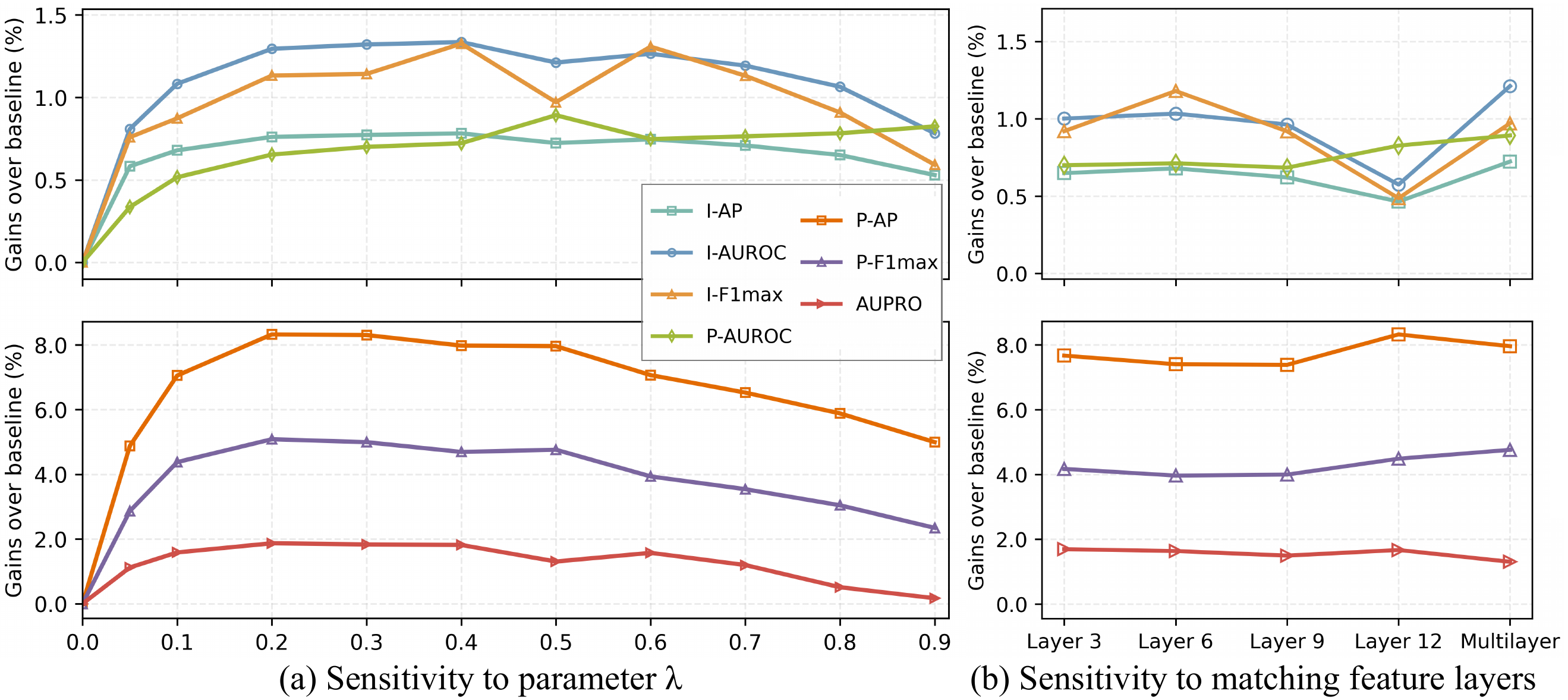}}
\vspace{-1mm}
\caption{Sensitivity analysis on MVTec-AD (RGB UAD with GLAD+UCF), assessing $\lambda$ and DINOv2 patch token layer choice for anomaly cost volume construction. The vertical axis reports gains over the baseline across a comprehensive set of image- and pixel-level metrics.}
\label{lamda}
\end{center}
\vspace{-7mm}
\end{figure}

\subsubsection{Hyperparameter Sensitivity of $\lambda$ and Patch-token Layers}
Fig.~\ref{lamda}(a) shows the sensitivity to $\lambda$ on MVTec-AD with RGB UAD (GLAD+UCF), where the vertical axis shows absolute gains over the baseline. At $\lambda=0$, the method collapses to the baseline($97.5\%/97.3\%$ image/pixel AUROC). For all $\lambda>0$, performance improves across all seven metrics, with steady gains that peak around $\lambda=0.4$ to $0.6$. Pixel-level metrics (P-AP, P-F1-max, AUPRO) benefit most, with up to 8\% relative gain, reflecting stronger localization. We adopt $\lambda=0.5$ by default in this configuration.
Fig.~\ref{lamda}(b) studies the DINOv2 patch-token layer used for anomaly cost volume construction. Any single layer yields consistent image- and pixel-level improvements. 
Constructing a multi-layer cost volume from all four layers offers the overall trade-off, yielding the best image-level results and robust pixel-level gains, indicating complementary cues from shallow (boundary/detail) and deep (semantic) tokens.

\subsubsection{Analysis of Failure Cases}

\begin{figure}
\setlength{\abovecaptionskip}{2pt}  
\setlength{\belowcaptionskip}{0pt} 
\begin{center}
\centerline{\includegraphics[width=0.5\textwidth]{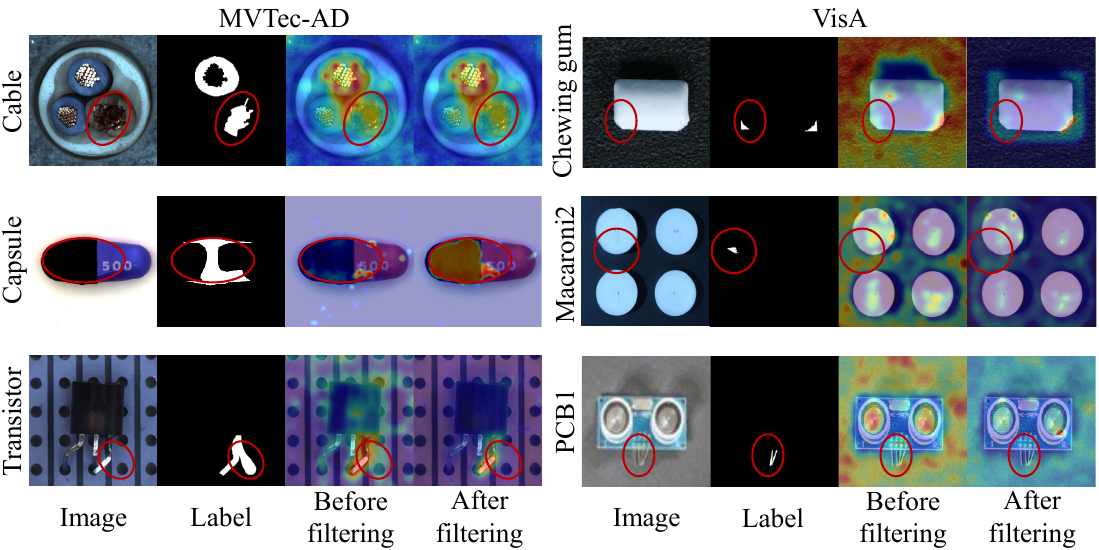}}
\vspace{-1mm}
\caption{Failure cases on MVTec-AD and VisA. Our method suppresses matching noise, yet subtle anomalies may still be missed or inaccurately localized, often due to limited representation within the anomaly cost volume.}
\label{fig_failure}
\end{center}
\vspace{-7mm}
\end{figure}

We analyze representative failures to delineate current limits. Fig.~\ref{fig_failure} shows six categories from MVTec-AD and VisA with outputs before and after filtering. While our filter suppresses matching noise, performance still depends on anomaly-relevant evidence in the cost volume. Low-resolution inputs or weak feature extraction can under-represent anomalies, for which we inject input-image features as auxiliary guidance. Nevertheless, highly subtle defects, weak normal/abnormal contrast (e.g., \textit{macaroni2}), and even unseen categories (e.g., \textit{capsule}) remain challenging. These cases mainly reflect the dependence on upstream feature quality: when anomaly cues are weakly encoded, the filter has less to leverage, suggesting room for improvement with stronger backbones, higher resolution, or additional cues.

\section{Conclusion}

We present Unified Cost Filtering (UCF), a unified framework for unsupervised anomaly detection across unimodal and multimodal scenarios. 
By reformulating UAD as a three-stage pipeline of feature extraction, cost volume construction, and cost volume filtering, UCF addresses intrinsic matching noise through matching cost filtering and serves as a generic post-hoc plug-in for both reconstruction- and embedding-based methods. 
Guided by multi-layer two-stream attention, it refines anomaly cost volumes by suppressing noise from ``identical shortcut'' artifacts or feature misalignment, while highlighting subtle anomalies. 
Extensive evaluations on 22 industrial and medical benchmarks show that integrating UCF into 10 diverse baselines consistently achieves state-of-the-art results at low computational cost, establishing it as a robust, unifying, and deployable solution for real-world anomaly detection.

\textbf{Future work.} A promising direction is to advance UCF by developing hybrid cost volume filtering that unifies correspondences across RGB, 3D, and text modalities, integrates multi-view representations, and leverages features from advanced foundation models and reconstruction frameworks. We will further explore broader applications, including logical and video anomaly detection, few-/zero-shot point-cloud anomaly detection, and open-vocabulary anomaly classification and segmentation, thereby broadening the applicability of our method across diverse and challenging scenarios.

\section*{Acknowledgment}
This work is supported in part by the Science and Technology Major Project of Liaoning Province under Grant No. 2024JH1/11700048, in part by the Research Program of the Liaoning Liaohe Laboratory under Grant No. LLL23ZZ-05-01, in part by the Natural Science Foundation of Liaoning Province under Grant No. 2024-MSBA-42, and in part by the Key Research and Development Program of Liaoning Province under Grant No. 2023JH26/10200011, and in part by the Program of China Scholarship Council 202306080142.

\clearpage
\section*{\centering \Large Supplementary Material}

\renewcommand{\thesection}{S\arabic{section}}
\setcounter{section}{0}

\section{Overview}
\phantomsection
\label{supp:overview}
We propose a generic method, UCF, for unified unsupervised anomaly detection (UAD), applicable to both unimodal RGB and multimodal RGB-3D and RGB-Text scenarios. We reveal the critical, yet often overlooked, issue of matching noise, which can reduce accuracy. To mitigate this, we reformulate UAD into a three-step pipeline: feature extraction, anomaly cost volume construction, and matching cost filtering. 
This material complements the main paper and is organized as follows.

\noindent\textbf{Sec.~\ref{supp:overview} Overview}

\noindent\textbf{Sec.~\ref{suppsec15} Additional motivation examples and analysis}

\noindent\textbf{Sec.~\ref{suppsec2} Details of Baselines and Benchmarks}
\begin{itemize}
  \item Sec.~\ref{suppsec2a} Baseline Methods for Integration
  \item Sec.~\ref{suppsec2b} Evaluation Benchmarks
\end{itemize}

\noindent\textbf{Sec.~\ref{suppsec3} More Experimental Details}
\begin{itemize}
  \item Sec.~\ref{suppsec3a} Training Setup
  \item Sec.~\ref{suppsec3b} Unimodal RGB UAD Details
  \item Sec.~\ref{suppsec3c} Multimodal RGB-3D UAD Details
  \item Sec.~\ref{suppsec3d} Multimodal RGB-Text UAD Details
  \item Sec.~\ref{suppsec3e} Other Details: RCSA, Visualization, and More
\end{itemize}

\noindent\textbf{Sec.~\ref{suppsec4} Further Experimental Analyses}
\begin{itemize}
  \item Sec.~\ref{suppsec4a} Validation of Single-Class Compatibility
  \item Sec.~\ref{suppsec4b} Few-Shot on MVTec-AD for RGB-Text UAD
  \item Sec.~\ref{suppsec4c} Additional Metrics in RGB-Text UAD
  \item Sec.~\ref{suppsec4d} Progressive Noise Denoising Visualization
  \item Sec.~\ref{suppsec4e} Resolution and Baseline Protocols in RGB UAD
  \item Sec.~\ref{suppsec4f} Analysis of Shortcut Issue in Reconstruction
\end{itemize}

\noindent\textbf{Sec.~\ref{suppsec5} Comprehensive Per-Class Quantitative Results}
\begin{itemize}
  \item Sec.~\ref{suppsec5a} RGB UAD: Per-Class Quantitative Results
  \item Sec.~\ref{suppsec5b} RGB-3D UAD: Per-Class Quantitative Results
  \item Sec.~\ref{suppsec5c} RGB-Text UAD: Per-Class Quantitative Results
\end{itemize}

\noindent\textbf{Sec.~\ref{suppsec6} Comprehensive Per-Class Qualitative Visualization}
\begin{itemize}
  \item Sec.~\ref{suppsec6a} RGB UAD: Per-Class Qualitative Results
  \item Sec.~\ref{suppsec6b} RGB-3D UAD: Per-Class Qualitative Results
  \item Sec.~\ref{suppsec6c} RGB-Text UAD: Per-Class Qualitative Results
\end{itemize}

\noindent\textbf{Sec.~\ref{suppsec7} Per-Class KDE Analysis of Logits}
\begin{itemize}
  \item Sec.~\ref{suppsec7a} RGB UAD: KDE Analysis of Logits
  \item Sec.~\ref{suppsec7b} RGB-3D UAD: KDE Analysis of Logits
  \item Sec.~\ref{suppsec7c} RGB-Text UAD: KDE Analysis of Logits
\end{itemize}

\begin{figure*}
\setlength{\abovecaptionskip}{2pt}  
\setlength{\belowcaptionskip}{0pt} 
\begin{center}
\centerline{\includegraphics[width=1.0\textwidth]{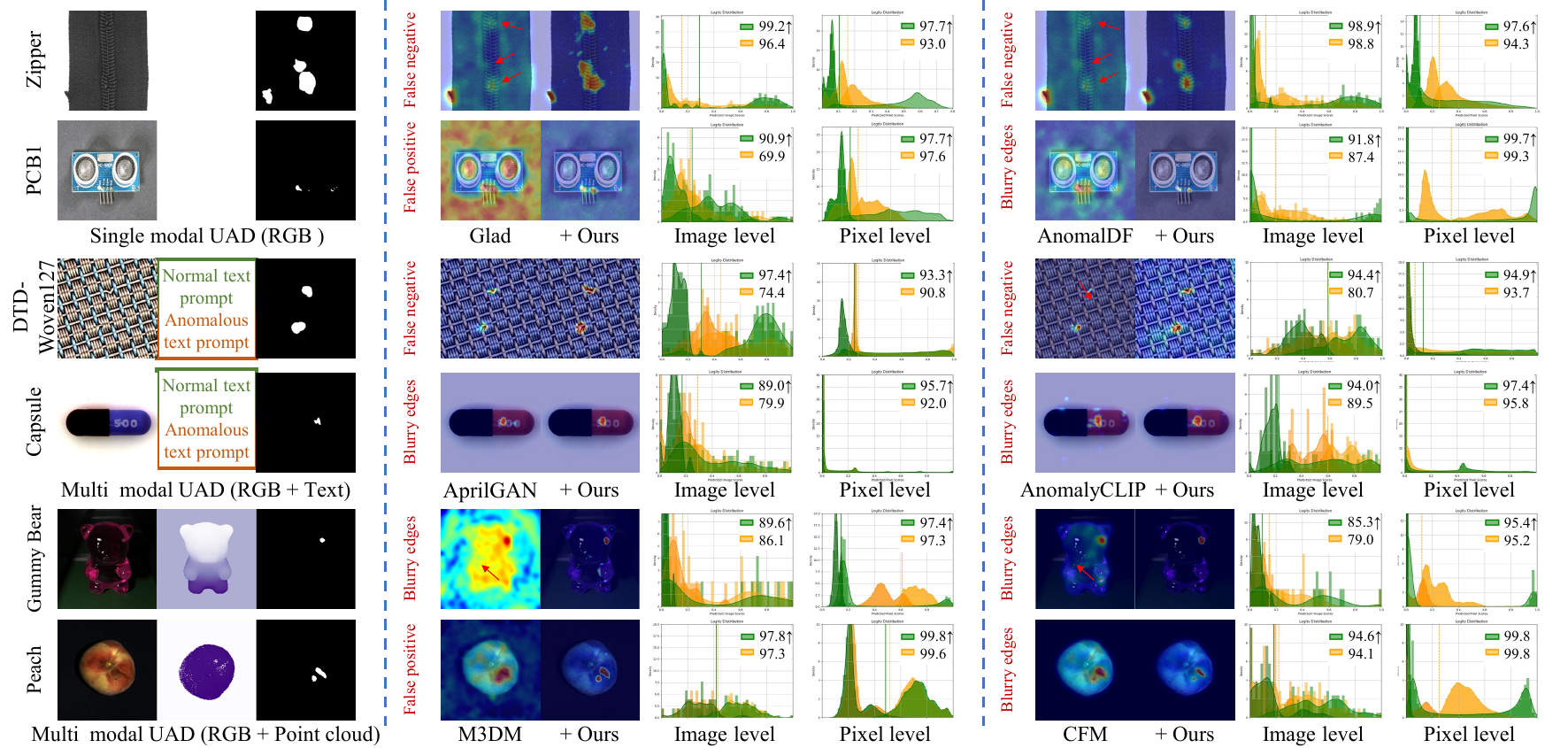}}    
\caption{Comparison of unimodal and multimodal UAD results. We present the visualization results and kernel density estimation curves (KDE) \cite{kde} of image- and pixel-level logits. Baseline results are highlighted in \textcolor{myyellow}{yellow}, while ours are shown in \textcolor{mygreen}{green}. Our model achieves superior performance by detecting anomalies with mitigated matching noise and providing a clearer distinction between normal and abnormal logits.}
\label{suppfig1}
\end{center}
\end{figure*}

\section{Additional motivation examples and analysis}
\phantomsection
\label{suppsec15}
Fig.~\ref{suppfig1} provides additional examples and analysis from different modality scenarios to illustrate the motivation of the proposed matching cost filtering for UAD, complementing Fig.~1 in the main paper. Each row displays the input, the ground truth mask, baseline anomaly detection heatmaps from GLAD~\cite{glad}, AnomalDF~\cite{anomalydino} (RGB UAD), M3DM~\cite{m3dm}, CFM~\cite{cfm} (RGB-3D UAD), AprilGAN~\cite{aprilgan}, and AnomalyCLIP~\cite{anomalyclip} (RGB-Text UAD), along with the results after integrating our method. In practice, anomaly maps in both unimodal and multimodal UAD pipelines are typically computed using direct L2 norm distances or cosine similarity, often followed by a simple Gaussian filter, which smooths the maps but leaves intrinsic matching noise largely uncontrolled.

This matching noise manifests itself as spurious activations on background or texture, blurry defect boundaries, and missed small or low-contrast anomalies, especially under conditions like cross-modal misalignment or prompt shifts. These observations motivate our reformulation. We first construct an anomaly cost volume by matching the input to modality-appropriate references, combining evidence from multiple sources, such as different template locations. The cost volume is then refined using an input-guided filtering network, where attention mechanisms enable dynamic selection or suppression based on the input, thus mitigating the matching noise.

Inspired by works in stereo matching~\cite{stereo}, depth estimation~\cite{depth}, flow estimation~\cite{gudovskiy2022cflow}, and light field rendering~\cite{lightfield}, our anomaly cost volume can be viewed as a representation of the ``energy landscape"~\cite{energy}, aiming to minimize the energy associated with false matches and background noise while maximizing the energy of true anomaly signals. The green (ours) KDE~\cite{kde} curves in Fig.~\ref{suppfig1}, representing image- and pixel-level anomaly detection logits, exhibit a markedly greater separation between normal and abnormal distributions than the yellow curves (baselines), with reduced overlap, indicating enhanced separability. By suppressing noisy correspondences and preserving edge structures and subtle signals, our method generates more accurate heatmaps and decisive decision boundaries. These improvements are consistent across unimodal RGB and multimodal RGB–3D and RGB–Text UAD, demonstrating that cost volume filtering effectively mitigates inherent matching noise and generalizes across categories and datasets without requiring task-specific tuning.

\begin{table*}[!t]
\caption{Overview of unimodal and multimodal baselines for validation, with our method integrated into each. Baseline configurations are organized by modality, data regime (full-/zero-/few-shot), learning paradigm, feature extractor, and input resolution, while our experiments additionally specify the benchmarks, the shapes of the constructed cost volumes, and the generated anomaly maps.
}
\label{supptab:baseline_overview}
\centering
\tabcolsep=3pt
\renewcommand{\arraystretch}{1.2}
\Large
\resizebox{1\textwidth}{!}{
\begin{tabular}{@{}l|ccc|ccc|c|c|c|c|c|c@{}}
\toprule
\multirow{2}{*}{Baseline}
& \multicolumn{3}{c|}{Modality} 
& \multicolumn{3}{c|}{Shot}
& \multirow{2}{*}[-0.6ex]{\shortstack{Paradigm\\(-based)}}
& \multirow{2}{*}[-0.6ex]{\shortstack{Image\\Resize}}
& \multirow{2}{*}{Feature Extractor}
& \multirow{2}{*}{Benchmarks} 
& \multirow{2}{*}[-0.6ex]{\shortstack{Cost Volume\\Shape}}
& \multirow{2}{*}[-0.6ex]{\shortstack{Anomaly Map \\ Shape}} \\
\cmidrule(lr){2-4}\cmidrule(lr){5-7}
& RGB & 3D & Text & Full & Zero & Few &  &  &  &  &  \\
\midrule
\multicolumn{12}{@{}l}{\textit{\textbf{Unimodal RGB UAD}}} \\
UniAD~\cite{uniad}        & $\checkmark$ & $\times$ & $\times$ & $\checkmark$ & $\times$ & $\times$ & Recons. & 224 $\times$ 224 & EfficientNet+ViT          & 4 Industrial datasets & 196$\times$4$\times$64$\times$64 & 2$\times$64$\times$64 \\
HVQ-Trans~\cite{hvqtrans}    & $\checkmark$ & $\times$ & $\times$ & $\checkmark$ & $\times$ & $\times$ & Recons.& 224 $\times$ 224 & EfficientNet+ViT          & 4 Industrial datasets & 196$\times$4$\times$64$\times$64 & 2$\times$64$\times$64 \\
GLAD~\cite{glad}         & $\checkmark$ & $\times$ & $\times$ & $\checkmark$ & $\times$ & $\times$ & Recons.& 256 $\times$ 256 & Diffusion+DINOv2          & 4 Industrial datasets & 1024$\times$4$\times$64$\times$64 & 2$\times$64$\times$64 \\
AnomalDF~\cite{anomalydino}     & $\checkmark$ & $\times$ & $\times$ & $\checkmark$ & $\times$ & $\times$ & Embed. & 256 $\times$ 256& DINOv2                      & 4 Industrial datasets & 1024$\times$4$\times$64$\times$64 & 2$\times$64$\times$64 \\
Dinomaly~\cite{dinomaly}     & $\checkmark$ & $\times$ & $\times$ & $\checkmark$ & $\times$ & $\times$ & Recons. & 448 $\times$ 448& DINOv2+ViT                          & 4 Industrial datasets & 768$\times$2$\times$64$\times$64 & 2$\times$64$\times$64 \\
\midrule
\multicolumn{12}{@{}l}{\textit{\textbf{Multimodal RGB-3D UAD}}} \\
M3DM~\cite{m3dm}         & $\checkmark$ & $\checkmark$ & $\times$ & $\checkmark$ & $\times$ & $\times$ & Embed.& 224 $\times$ 224 & PointMAE+DINOv2           & MVTec 3D-AD, Eyecandies    & 3$\times$1$\times$224$\times$224    & 2$\times$224$\times$224 \\
CFM~\cite{cfm}          & $\checkmark$ & $\checkmark$ & $\times$ & $\checkmark$ & $\times$ & $\times$ & Recons.& 224 $\times$ 224 & PointMAE+DINOv2+MLP      & MVTec 3D-AD, Eyecandies    & 3$\times$1$\times$224$\times$224     & 2$\times$224$\times$224 \\
\midrule
\multicolumn{12}{@{}l}{\textit{\textbf{Multimodal RGB-Text UAD}}} \\
AprilGAN~\cite{aprilgan}     & $\checkmark$ & $\times$ & $\checkmark$ & $\times$ & $\checkmark$ & $\times$ & Embed.& 518 $\times$ 518 & CLIP+linear layers         & 9 Industrial+7 Medical datasets        & 3$\times$4$\times$64$\times$64      & 2$\times$64$\times$64 \\
AprilGAN~\cite{aprilgan}     & $\checkmark$ & $\times$ & $\checkmark$ & $\times$ & $\times$ & $\checkmark$ & Embed.& 518 $\times$ 518 & CLIP+linear layers         & MVTec-AD, VisA        & 3$\times$4$\times$64$\times$64      & 2$\times$64$\times$64 \\
AdaCLIP~\cite{adaclip}      & $\checkmark$ & $\times$ & $\checkmark$ & $\times$ & $\checkmark$ & $\times$ & Embed.& 518 $\times$ 518 & CLIP+prompt-learning & 9 Industrial+7 Medical datasets & 3$\times$4$\times$64$\times$64   & 2$\times$64$\times$64 \\
AnomalyCLIP~\cite{anomalyclip}  & $\checkmark$ & $\times$ & $\checkmark$ & $\times$ & $\checkmark$ & $\times$ & Embed.& 518 $\times$ 518 & CLIP+prompt-learning  & 9 Industrial+7 Medical datasets & 3$\times$1$\times$64$\times$64   & 2$\times$64$\times$64 \\
\bottomrule
\end{tabular}
}
\end{table*}

\begin{table*}[!t]
\caption{Benchmark overview across domain (industrial or medical), category count, detection type, acquisition modality, and label granularity, together with the numbers of normal/abnormal test samples and the applicability to UAD scenarios.}
\label{supptab:dataset_overview}
\centering
\tabcolsep=3pt
\renewcommand{\arraystretch}{1.2}
\resizebox{1\textwidth}{!}{
\begin{tabular}{@{}l|c|c|c|c|cc|c|ccc@{}}
\toprule
\multirow{2}{*}{Benchmark} & \multirow{2}{*}{Domain} & \multirow{2}{*}{$| \mathbb{C} |$} & \multirow{2}{*}{Detection Type} & \multirow{2}{*}[-0.6ex]{\shortstack{Acquisition\\Modality}} & \multicolumn{2}{c|}{Labels} & \multicolumn{1}{c|}{Sample number} & \multicolumn{3}{c}{Applicable UAD Scenarios} \\
\cmidrule(lr){6-7}\cmidrule(lr){8-8}\cmidrule(lr){9-11}
& & & & & Image & Pixel & Normal / Abnormal & RGB & RGB-3D & RGB-Text \\
\midrule
\multicolumn{11}{@{}l}{\textit{\textbf{Industrial RGB (image- and pixel-level labels)}}} \\
MVTec-AD~\cite{mvtec}      & Industrial & 15 & Object, Texture defect & Photography        & $\checkmark$ & $\checkmark$ & 467 / 1258   & $\checkmark$ & $\times$ & $\checkmark$ \\
VisA~\cite{visa}          & Industrial & 12 & Object defect          & Photography        & $\checkmark$ & $\checkmark$ & 962 / 1200   & $\checkmark$ & $\times$ & $\checkmark$ \\
MPDD~\cite{mpdd}          & Industrial & 6  & Object defect          & Photography        & $\checkmark$ & $\checkmark$ & 176 / 282    & $\checkmark$ & $\times$ & $\checkmark$ \\
BTAD~\cite{btad}          & Industrial & 3  & Object defect          & Photography        & $\checkmark$ & $\checkmark$ & 451 / 290     & $\checkmark$ & $\times$ & $\checkmark$ \\
SDD~\cite{ksdd}           & Industrial & 1  & Object defect          & Photography        & $\checkmark$ & $\checkmark$ & 181 / 74     & $\times$ & $\times$ & $\checkmark$ \\
DAGM~\cite{dagm}          & Industrial & 10 & Texture defect         & Photography        & $\checkmark$ & $\checkmark$ & 6996 / 1054  & $\times$ & $\times$ & $\checkmark$ \\
DTD~\cite{dtd}           & Industrial & 12 & Texture defect        & Photography        & $\checkmark$ & $\checkmark$     & 357 / 947    & $\times$ & $\times$ & $\checkmark$ \\
\midrule
\multicolumn{11}{@{}l}{\textit{\textbf{Industrial RGB and 3D (image- and pixel-level labels)}}} \\
MVTec 3D-AD~\cite{mvtec3d}   & Industrial & 10 & Object defect          & Real       & $\checkmark$ & $\checkmark$ & 249 / 948   & $\times$ & $\checkmark$ & $\times$ \\
Eyecandies~\cite{eyecan}    & Industrial & 10 & Object defect          & Synthetic    & $\checkmark$ & $\checkmark$ & 2000 / 2000 & $\times$ & $\checkmark$ & $\times$ \\
\midrule
\multicolumn{11}{@{}l}{\textit{\textbf{Medical RGB (image-level labels only)}}} \\
HeadCT~\cite{headct}        & Medical    & 1  & Brain tumor           & Radiology (CT)  & $\checkmark$ & $\times$  & 100 / 100   & $\times$ & $\times$ & $\checkmark$ \\
BrainMRI~\cite{brainmri}      & Medical    & 1  & Brain tumor           & Radiology (MRI) & $\checkmark$ & $\times$  & 98 / 155    & $\times$ & $\times$ & $\checkmark$ \\
Br35H~\cite{br35h}         & Medical    & 1  & Brain tumor           & Radiology (MRI) & $\checkmark$ & $\times$  & 1500 / 1500 & $\times$ & $\times$ & $\checkmark$ \\
\midrule
\multicolumn{11}{@{}l}{\textit{\textbf{Medical RGB (pixel-level labels only)}}} \\
ISIC~\cite{isic}          & Medical    & 1  & Skin cancer            & Photography        & $\times$     & $\checkmark$ & 0 / 379      & $\times$ & $\times$ & $\checkmark$ \\
CVC-ColonDB~\cite{clodb}       & Medical    & 1  & Colon polyp           & Endoscopy    & $\times$     & $\checkmark$ & 0 / 612      & $\times$ & $\times$ & $\checkmark$ \\
CVC-ClinicDB~\cite{clidb}      & Medical    & 1  & Colon polyp           & Endoscopy    & $\times$     & $\checkmark$ & 0 / 380      & $\times$ & $\times$ & $\checkmark$ \\
Kvasir~\cite{kvasir}        & Medical    & 1  & Colon polyp           & Endoscopy    & $\times$     & $\checkmark$ & 0 / 1000     & $\times$ & $\times$ & $\checkmark$ \\
Endo~\cite{endo}          & Medical    & 1  & Colon polyp           & Endoscopy    & $\times$     & $\checkmark$ & 0 / 200      & $\times$ & $\times$ & $\checkmark$ \\
TN3K~\cite{tn3k}          & Medical    & 1  &  Thyroid nodule         & Ultrasound   & $\times$     & $\checkmark$ & 0 / 614      & $\times$ & $\times$ & $\checkmark$ \\
\bottomrule
\end{tabular}
}
\end{table*}

\section{Details of Baselines and Benchmarks}
\phantomsection
\label{suppsec2}

\subsection{Baseline Methods for Integration}
\phantomsection
\label{suppsec2a}
We integrate our method into 10 representative UAD baselines spanning unimodal and multimodal scenarios, covering both reconstruction- and embedding-based paradigms. To ensure fair comparison, we preserve the official configurations of each baseline whenever possible when integrating UCF. Table~\ref{supptab:baseline_overview} summarizes the core properties of each baseline along with our cost volume shapes, facilitating transparent comparison across paradigms and baselines.

\textit{\textbf{Unimodal RGB UAD.}} 
UniAD~\cite{uniad}, HVQ-Trans~\cite{hvqtrans}, and Dinomaly~\cite{dinomaly} adopt reconstruction pipelines based on transformer networks, whereas GLAD~\cite{glad} employs diffusion-based reconstruction. AnomalDF, the full-shot variant of AnomalDINO~\cite{anomalydino}, constructs feature memory banks using DINOv2~\cite{dino}. All these methods quantify anomaly evidence from residuals or similarity measures in the latent feature space.

\textit{\textbf{Multimodal RGB-3D UAD.}}  
M3DM~\cite{m3dm} is embedding-based, mapping features extracted by PointMAE~\cite{pointmae} and DINOv2~\cite{dino}. CFM~\cite{cfm} leverages cross-modal mapping to align RGB and 3D representations for anomaly detection.

\textit{\textbf{Multimodal RGB-Text UAD.}} AprilGAN~\cite{aprilgan}, AnomalyCLIP~\cite{anomalyclip}, and AdaCLIP~\cite{adaclip} adopt CLIP-style image–text feature matching, employing prompt learning or linear adapters to enhance cross-modal anomaly detection.

Collectively, these baselines span a wide range of reconstruction- and embedding-based paradigms across multiple modalities, providing robust references for evaluating the effectiveness and generality of our method.

\subsection{Evaluation Benchmarks}
\phantomsection
\label{suppsec2b}

We evaluated our method on 22 benchmarks spanning industrial inspection and medical diagnosis, including 4 RGB UAD, 2 RGB–3D UAD, and 16 RGB–Text UAD datasets, enabling validation across both unimodal and multimodal scenarios. Table~\ref{supptab:dataset_overview} summarizes each dataset in terms of domain, number of categories, detection type, acquisition modality, label granularity (image- or pixel-level), and applicable UAD scenarios. For unimodal RGB UAD and multimodal RGB–3D UAD, the training sets consist exclusively of normal images, whereas some multimodal RGB–Text UAD datasets are available only as test sets. Accordingly, Table~\ref{supptab:dataset_overview} reports the counts of normal and abnormal samples in the test sets, while the numbers of training images for datasets with training sets are provided in the main paper. We follow the official dataset splits and baseline protocols to ensure reproducibility and comparability.

\textit{\textbf{Industrial datasets.}} MVTec-AD~\cite{mvtec} contains 15 categories of objects and textures, while VisA~\cite{visa} includes 12 object classes. Additional industrial UAD datasets comprise MPDD~\cite{mpdd}, BTAD~\cite{btad}, SDD~\cite{ksdd}, DAGM~\cite{dagm}, and DTD~\cite{dtd}. Multimodal RGB-3D datasets include MVTec 3D-AD~\cite{mvtec3d} and Eyecandies~\cite{hvqtrans}. All datasets provide image- and pixel-level annotations suitable for localization evaluation.

\textit{\textbf{Medical datasets.}}
Since these datasets consist solely of anomalous images, ISIC~\cite{isic}, CVC-ColonDB~\cite{clodb}, CVC-ClinicDB~\cite{clidb}, Kvasir~\cite{kvasir}, Endo~\cite{endo}, and TN3K~\cite{tn3k} provide pixel-level labels only. We therefore report pixel-level results exclusively. In contrast, HeadCT~\cite{headct}, BrainMRI~\cite{brainmri}, and Br35H~\cite{br35h} provide image-level labels only, so image-level results are reported for these datasets.

\section{More Experimental Details}
\phantomsection
\label{suppsec3}
\subsection{Training Setup} 
\phantomsection
\label{suppsec3a}

We use a batch size of $B=8$ and a ReduceLROnPlateau scheduler, which halves the learning rate when the loss plateaus to promote stable convergence. In Eq.~8 of the main paper, $\mathcal{M}$ denotes the pixel-level anomaly score map generated by our method, while $\mathcal{M}_s$ represents our synthesized anomaly mask used to train the proposed cost filtering network. $\mathcal{L}_\text{Focal}$ refers to Focal Loss~\cite{focalloss}, with parameter $\gamma$ controlling the emphasis on hard-to-detect samples. For Focal Loss, the class-aware adaptor is configured with the initial $\gamma_0 = 3$. $\mathcal{L}_\text{Soft-IoU}$ denotes Soft Intersection-over-Union Loss~\cite{iouloss}, refining anomaly localization through IoU optimization. $\mathcal{L}_\text{SSIM}$ corresponds to the structural similarity index loss~\cite{ssimloss}, ensuring spatial structural consistency, and $\mathcal{L}_\text{CE}$ denotes cross-entropy loss~\cite{celoss}, enhancing multi-class classification by mitigating entropy-based uncertainty. For both training and testing, the generated anomaly maps are upsampled via interpolation to match the resolution of the input image and corresponding pixel-level masks.

As presented in Sec.~III-C of the main paper, feature tensors are represented generically as $C \times H' \times W'$ to unify the description across modalities, given that different baselines adopt distinct feature dimensions depending on their configurations. These features form the basis for constructing anomaly cost volumes. In the fourth layer of the filtering network decoder, a 3D convolution along the matching dimension reduces the cost volume from $L$ to $1$, producing the feature used to generate the normal/abnormal score map. We employ 3D convolution for its effectiveness in aggregating evidence across templates, yet the backbone is not limited to this design. Transformer-based or Mamba-like architectures could also be adopted. Exploring such alternatives lies beyond the present scope and represents a promising direction for future research. Moreover, given the diversity of backbones (e.g., DINOv2~\cite{dino}, SAM~\cite{sam}, CLIP encoders~\cite{clipac}, diffusion models~\cite{diffusion}, and ViTs~\cite{vit}), we do not enforce a unified feature extractor or identical operations across the 10 baselines. Instead, we present matching cost filtering as a flexible plug-in for any UAD methods that offers a unified perspective. Integrated seamlessly into diverse unimodal and multimodal pipelines, it consistently delivers performance gains with minimal overhead, forming the core of our method.

\subsection{Unimodal RGB UAD Details}
\phantomsection
\label{suppsec3b}
\subsubsection{Integration with Baselines}
\phantomsection
\label{suppsec3b1}

To ensure fair comparison, we follow the official feature configurations of each baseline when constructing cost volumes for anomaly matching. For GLAD~\cite{glad}+UCF and AnomalDF~\cite{anomalydino}+UCF, features are extracted from the 3rd, 6th, 9th, and 12th layers of a pre-trained DINO model~\cite{dino}. For UniAD~\cite{uniad}+UCF and HVQ-Trans~\cite{hvqtrans}+UCF, we use the 1st, 5th, 9th, and 21st decoder layers of the released pre-trained models. Dinomaly~\cite{dinomaly}+UCF follows its pre-trained decoder to form low- and high-level semantic feature groups for cost volume construction. As for the anomaly cost volume construction strategy, for reconstruction-based baselines (HVQ-Trans, UniAD, Dinomaly), we compute cost volumes directly in latent space between input and reconstruction, rather than decoding back to the image domain, since their decoders already yield semantically meaningful features. In contrast, GLAD+UCF and AnomalDF+UCF rely on external pre-trained Dino encoders for feature extraction, consistent with their original settings.

\subsubsection{Additional Remarks}
\phantomsection
\label{suppsec3b2}

We refer to the full-shot variant of AnomalDINO~\cite{anomalydino} as AnomalDF, where ``F” indicates the full-shot setting. AnomalDINO studies few-shot and full-shot regimes. The few-shot regime fixes a small set of normal templates per category, while the full-shot regime builds a large memory bank from all normal training samples per category, incurring higher storage cost. In this paper, we reduce storage by coupling global feature matching with denoising and by using a limited number of templates. 

\textbf{During training}, AnomalDF+UCF randomly samples $N{=}3$ templates per input from the full training set, rather than using a fixed template pool as in the original few-shot configuration~\cite{anomalydino}. This dynamic sampling covers the full training distribution and is therefore categorized as full-shot while being memory efficient. \textbf{During testing}, we evaluate AnomalDF (+UCF) under the same random sampling protocol for fairness, which is reflected in the reported results.

\subsubsection{Anomaly Synthesis for Unimodal RGB UAD} 
Synthetic RGB anomalies are generated by applying Perlin-noise~\cite{perlin} masks to guide the insertion of external textures or local structural perturbations, following the protocols of GLAD and DRAEM~\cite{glad,draem}. Texture sources are randomly sampled from the Describable Textures Dataset (DTD)~\cite{dtdglad} (note that it is distinct from the DTD-Synthetic benchmark~\cite{dtd} mentioned in the main paper), while structural anomalies are created by grid-wise shuffling of input images to disrupt local consistency. The Perlin mask defines the spatial extent of anomalies, and a blending factor $\beta \in [0,1]$ controls the mixture between original and perturbed regions: small $\beta$ values emphasize external textures, whereas larger $\beta$ values preserve more original content. This pipeline yields visually diverse anomalies with binary masks directly used for our filtering network supervision. All hyperparameters, including Perlin scales, thresholds, and blending factors, strictly follow GLAD~\cite{glad} configurations to ensure fair comparison. By leveraging such synthetic data, the proposed cost filtering model generalizes effectively to real-world anomalies, as validated in extensive quantitative and qualitative evaluations.

\subsection{Multimodal RGB-3D UAD Details}
\phantomsection
\label{suppsec3c}
\subsubsection{Integration with Baselines}
\phantomsection
\label{suppsec3c1}

We integrate our method with two multimodal RGB–3D UAD baselines, M3DM~\cite{m3dm} and CFM~\cite{cfm}, where M3DM extends PatchCore~\cite{patchcore} to the RGB–3D setting. RGB images are resized to $224 \times 224$ and encoded by a frozen DINO ViT-B/8 pretrained on ImageNet to obtain patch tokens. Point clouds are pre-processed by fitting a background plane with RANSAC~\cite{btf,ast,pami2} (inlier threshold 0.005) and encoded with a Point Transformer~\cite{pointmae} pretrained on ShapeNet~\cite{shapenet}. Farthest-point sampling~\cite{m3dm} generates $M$ point groups, whose features are interpolated back to all points and projected onto the image plane using camera parameters, producing a 2D feature map at RGB resolution. This map is average-pooled to the ViT patch grid, yielding token-wise alignment between 3D and RGB features for subsequent matching and cost volume construction. Table~\ref{suppeyeclsabbr} lists the category abbreviations of the Eyecandies dataset~\cite{eyecan}, as used in Table VI of the main paper, together with their full names.

\begin{table}[!t]
\centering
\caption{Full category names and abbreviations of Eyecandies~\cite{eyecan}.}
\label{suppeyeclsabbr}
\setlength\tabcolsep{12pt} 
\renewcommand{\arraystretch}{1.2}
\begin{tabular}{c|c}
\toprule
\textbf{Abbreviation} & \textbf{Full Category Name} \\
\hline
Can. C. & Candy Cane \\
Cho. C. & Chocolate Cookie \\
Cho. P. & Chocolate Praline \\
Conf. & Confetto \\
Gum. B. & Gummy Bear \\
Haz. T. & Hazelnut Truffle \\
Lic. S. & Licorice Sandwich \\
Lollip. & Lollipop \\
Marsh. & Marshmallow \\
Pep. C. & Peppermint Candy \\
\bottomrule
\end{tabular}
\end{table}

For M3DM+UCF, we follow the original M3DM~\cite{m3dm} setup. Point cloud features from the 3rd, 7th, and 11th layers of the PointMAE Point Transformer~\cite{pointmae} are aggregated into a single 3D representation, while the image branch uses last-layer features from DINO ViT-B/8~\cite{dino}. Intra-modal matching is then performed on RGB, 3D, and fused RGB–3D features of MVTec 3D-AD~\cite{mvtec3d}, producing a cost volume with $L=1$. Since M3DM does not include Eyecandies~\cite{eyecan}, we reproduce results by following the CFM~\cite{cfm} protocol, which matches RGB and 3D features only. For CFM+UCF, we use the CFM configuration and conduct cross-modal matching: last-layer PointMAE features are paired with reconstructed RGB features, and DINO ViT-B/8 features are paired with reconstructed 3D features, also yielding $L=1$. All matched pairs are shape-aligned to the token grid, and their matching cost maps are concatenated along the channel dimension to construct the anomaly cost volume. Baseline hyperparameters remain unchanged, and no additional tuning is applied beyond our filtering module. The cost volume filtering networks are trained from scratch for 40 epochs with a batch size $8$, using the Adam optimizer with a ReduceLROnPlateau scheduler to enhance training stability. The default loss weight is $\alpha=0.1$. This protocol ensures fairness and isolates the contribution of our matching cost volume filtering.

\subsubsection{Additional Remarks}
\phantomsection
\label{suppsec3c2}

Following the original baselines, we adopt their 3D feature propagation schemes: for M3DM~\cite{m3dm}, inverse distance weighting propagates group features to all points. For CFM~\cite{cfm}, three-nearest-neighbor interpolation is applied, followed by a $3\times3$ smoothing convolution. Consistent with M3DM, we train a single multi-class model per dataset, while consistent with CFM, we train one model per class following its released protocol. In the modality-missing ablation (Table~X in the main paper), to ensure fair comparison and reproducibility, we preserve the two-modality cost volume shape by duplicating the available modality’s matching cost along the channel dimension to substitute for the missing one. In Table~V of the main paper, the results of BTF~\cite{btf} and AST~\cite{ast} are unavailable for certain metrics, as they were not reported in their original or subsequent studies, and are thus indicated by ``–”.

\subsubsection{Anomaly Synthesis for Multimodal RGB-3D UAD}

For RGB–3D UAD, synthetic anomalies are constructed in a paired manner, ensuring one-to-one correspondence between RGB images and point clouds. RGB anomaly synthesis follows the procedure used in unimodal RGB UAD. For point clouds, 3D anomalies are generated by projecting 2D anomaly masks onto organized point clouds, following the representations in CFM~\cite{cfm}, M3DM~\cite{m3dm}, and M3DM-NR~\cite{pami2}, and perturbing the affected regions. Three strategies are employed: (i) Gaussian noise injection to simulate surface roughness or sensor errors; (ii) local point shuffling to disrupt geometric continuity; and (iii) interpolation-based filling to replace masked regions with interpolated points, mimicking missing or deformed structures. These perturbations yield 3D anomalies resembling real defects such as dents, scratches, or deformations, thereby fostering generalization to unseen cases. The paired point clouds and masks provide supervision at both pixel and point levels.

\subsection{Multimodal RGB-Text UAD Details}
\phantomsection
\label{suppsec3d}
\subsubsection{Integration with Baselines}
\phantomsection
\label{suppsec3d1}

RGB–Text UAD methods are typically implemented by fine-tuning pre-trained CLIP models and thus belong to embedding-based approaches. Following AdaCLIP~\cite{adaclip} and AprilGAN~\cite{aprilgan}, we adopt object-aware prompt designs to derive text and image features. Multiple normal and abnormal descriptions are encoded, averaged separately, and normalized into normal/abnormal text representations, which are then matched with image patch tokens. The resulting normal- and abnormal-similarity maps are stacked along the matching dimension to construct the anomaly cost volume. In addition, following AnomalyCLIP~\cite{anomalyclip}, we incorporate pre-trained object-agnostic normal and abnormal text embeddings for matching with image patch tokens.

For each baseline, we freeze the CLIP backbones and baseline-specific fine-tuned layers during cross-modal feature extraction. Following AprilGAN~\cite{aprilgan} and AdaCLIP~\cite{adaclip}, we extract patch-token features from the 6th, 12th, 18th, and 24th layers of the CLIP image encoder, pairing them with text features to construct cost volumes with $L=4$. For AnomalyCLIP~\cite{anomalyclip}, we follow its configuration and use only the patch-token features from the final (24th) layer, yielding $L=1$. The training protocol for our cost filtering networks is identical to that used in unimodal RGB UAD and multimodal RGB–3D UAD.

\begin{table*}[!t]
\centering
\caption{Class-specific text prompts for multimodal RGB–Text UAD: normal and anomalous prompts used with AdaCLIP/+UCF, where [cls] denotes the class name.}
\label{suppprompttable}
\setlength\tabcolsep{6pt}
\renewcommand{\arraystretch}{1.2}
\begin{tabular}{c|c}
\hline
\textbf{Type} & \textbf{Prompted Sentence} \\
 \midrule
Normal Text & a bad photo of a [cls], a low resolution photo of the [cls], a cropped photo of the [cls], \\
            & a bad photo of a flawless [cls], a low resolution photo of the flawless [cls], a cropped photo of the flawless [cls], \\
            & a bad photo of a perfect [cls], a low resolution photo of the perfect [cls], a cropped photo of the perfect [cls], \\
            & a bad photo of an unblemished [cls], a low resolution photo of the unblemished [cls], a cropped photo of the unblemished [cls], \\
            & a bad photo of a [cls] without flaw, a low resolution photo of the [cls] without flaw, a cropped photo of the [cls] without flaw, \\
            & a bad photo of a [cls] without defect, a low resolution photo of the [cls] without defect, a cropped photo of the [cls] without defect, \\
            & a bad photo of a [cls] without damage, a low resolution photo of the [cls] without damage, a cropped photo of the [cls] without damage. \\
\hline
Anomalous Text & a bad photo of a damaged [cls], a low resolution photo of the damaged [cls], a cropped photo of the damaged [cls], \\
               & a bad photo of a broken [cls], a low resolution photo of the broken [cls], a cropped photo of the broken [cls], \\
               & a bad photo of a [cls] with flaw, a low resolution photo of the [cls] with flaw, a cropped photo of the [cls] with flaw, \\
               & a bad photo of a [cls] with defect, a low resolution photo of the [cls] with defect, a cropped photo of the [cls] with defect, \\
               & a bad photo of a [cls] with damage, a low resolution photo of the [cls] with damage, a cropped photo of the [cls] with damage. \\
\bottomrule
\end{tabular}
\end{table*}

\subsubsection{Additional Remarks}
\phantomsection
\label{suppsec3d2}

In the RGB–Text ablation (Table~XI of the main paper), image patch tokens are matched with either normal or abnormal text embeddings. When only one text type is available, similarity scores are normalized with a sigmoid function. To ensure fair comparison and preserve the shape of cost volume, we keep the channel dimension identical to the two-text type setting by duplicating the available text-based matching cost along the matching dimension to substitute for the missing counterpart. In Fig.~5 of the main paper, an approximate mask for the HeadCT dataset~\cite{headct} is generated and outlined in red for visual clarity, since the HeadCT dataset provides only image-level labels without pixel-level ground truth, thereby enabling intuitive qualitative comparison. In Sec.~III-A of the main paper, we define the text embeddings as 
$T_{\mathrm{text}} \in \mathbb{R}^{N \times E}$ 
(prompt-based embeddings of dimension $E$), while in Sec.~III-C, the text feature is represented as 
$f_{T,\mathrm{nor}} \in \mathbb{R}^{L \times C \times (H'W')}$.
Since $E = H'W'$, this unified notation allows us to describe text features in the same format as other modalities. 
In practice, \(L=C=1\) for text features, but the general form facilitates a consistent representation across modalities. In Table VII of the main paper, since the CoOp~\cite{coop} method does not provide results on the DTD-Synthetic dataset~\cite{dtd}, we denote the missing entries as “–/–”. The corresponding mean value is therefore computed over the remaining six datasets, whereas the means of other methods are calculated across all seven datasets reported in the table.

Note that we do not use baseline-generated image-level anomaly scores (obtained by matching CLIP-derived image cls tokens with text features). Instead, consistent with unimodal RGB UAD and multimodal RGB–3D UAD, we derive image-level logits from the top-$250$ values of the predicted pixel-level anomaly maps, following GLAD~\cite{glad}. This design is guided by the intuition that if pixel-level anomalies exist in an input, the input is likely anomalous. The choice of $250$ reduces instability by preventing some extreme pixel values from dominating the decision at the image level.

\subsubsection{Anomaly Synthesis for Multimodal RGB-Text UAD}

RGB anomaly synthesis follows the same procedure as in unimodal RGB UAD, while textual templates are constructed to represent normal and abnormal states. For auxiliary datasets containing both normal and abnormal samples, additional anomalies are synthesized to ensure the unification. RGB–Text UAD is evaluated under \textit{zero-shot} and \textit{few-shot} settings, where neither target test images nor anomaly labels are available during training, rendering the task unsupervised as defined in the main paper. Extensive experiments further demonstrate the \textit{cross-domain generalization} of our method, where ``domain” encompasses both dataset-level shifts (e.g., non-overlapping anomaly categories across training and testing datasets) and broader feature shifts across industrial and medical scenarios.

Building on this formulation, we define text prompts according to the settings of each baseline. For AprilGAN/+UCF, we adopt a prompt ensemble strategy with state-level and template-level prompts. At the state level, normal and abnormal objects are described with generic terms (e.g., “flawless,” “damaged”), avoiding excessive details such as ``chip around edge and corner." At the template level, we use 85 CLIP templates from ImageNet~\cite{imagenet}, discarding those unsuitable for anomaly detection (e.g., “a photo of the weird [obj.]”). For AnomalyCLIP/+UCF, we adopt object-agnostic prompt templates~\cite{anomalyclip}, replacing class names with “[object]” to suppress class-specific semantics, formulated as:
\begin{itemize}
\item Normal prompt: $[V_1][V_2] \dots [V_E][object]$,
\item Abnormal prompt: $[W_1][W_2] \dots [W_E][damaged][object]$,
\end{itemize}
where $V_1, V_2, \dots, V_E$ denote normal state embeddings and $W_1, W_2, \dots, W_E$ denote anomalous state embeddings. This design encourages the model to learn shared patterns across diverse anomalies. For AdaCLIP/+UCF, anomalies are detected by computing similarities in the CLIP embedding space between images and text captions for normal and abnormal states, as detailed in Table~\ref{suppprompttable}.

\begin{figure*}
\setlength{\abovecaptionskip}{2pt}  
\setlength{\belowcaptionskip}{0pt} 
\begin{center}
\centerline{\includegraphics[width=\textwidth]{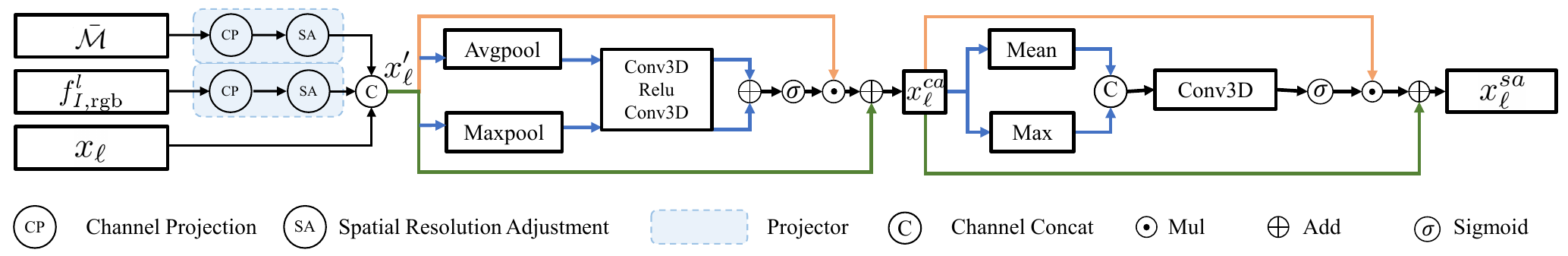}}
\caption{Design of the Residual Channel–Spatial Attention (RCSA) module with dual-stream feature guidance. Symbols: $x_{\ell}$ denotes the anomaly cost volume feature at encoder layer $\ell$ of the filtering network; $\bar{\mathcal{M}}$ is the initial anomaly map (matching guidance); $f_{I,{\mathrm{rgb}}}^{l}$ is the input image feature (spatial guidance), where $l$ indexes the layer feature extracted by embedding- or reconstruction-based models for anomaly cost volume construction; $x_{\ell}'$ is the concatenated feature combining cost volume, matching, and spatial guidance; $x_{\ell}^{ca}$ and $x_{\ell}^{sa}$ are the channel- and spatial-attended features, respectively.}
\label{suppfig4}
\end{center}
\end{figure*}

\subsection{Other Details: RCSA, Visualization, and More}
\phantomsection
\label{suppsec3e}
\textbf{Residual Channel-Spatial Attention (RCSA) module}. Fig. \ref{suppfig4} illustrates the RCSA module (Eq.~7 in the main paper), which generates two attention tensors: a channel attention tensor of shape $(B, C', 1, 1, 1)$ and a spatial attention tensor of shape $(B, 1, D', H', W')$, where $D'$ represents the depth of features at each layer. These tensors refine feature representations across both matching and spatial dimensions, being well-suited for dynamic selection or suppression since they adaptively adjust to the input. Residual channel attention strengthens global feature matching, while residual spatial attention enhances pixel-level anomaly localization. The residual connections preserve anomaly-relevant information, enabling progressive denoising and precise anomaly detection.

\textbf{Details for qualitative visualization.}
We adopt standard practice for qualitative visualization~\cite{hvqtrans,dinomaly}, presenting anomaly map logits on a single representative image to enable clear side-by-side comparison. Each anomaly map is normalized to $[0,1]$ using its minimum and maximum, rendered as a pseudo-color heatmap using \texttt{cv2.COLORMAP\_JET}, and blended with the reverse-normalized image using weighted fusion, and we report the fused visualization. In addition, as shown in Fig.~4 in the main paper, and Fig.~\ref{suppkeshihuamvtec3drgb3d}, Fig.~\ref{suppkeshihuaeyecanrgb3d} in this material, the RGB–point cloud visualizations may appear misaligned due to the 2D projection of point clouds. Nevertheless, the underlying feature maps remain \textbf{pixel-registered}, consistent with the protocols of M3DM~\cite{m3dm} and CFM~\cite{cfm}. In Fig. 6 of the main paper, panels (a), (b), and (c) illustrate examples from the VisA, BTAD, and Eyecandies datasets, respectively, while Fig. 7 presents cases from MVTec-AD (zipper), VisA (capsules), and MVTec 3D-AD (peach).

\textbf{Implementation details for KDE analysis.}
For quantitative evaluation, we generate KDE logit curves using Seaborn (\texttt{sns.kdeplot}) to visualize the separation between normal and abnormal samples. The normal–abnormal map logits are normalized at both image and pixel levels across all images within each category, providing a comprehensive presentation of the separability. As an auxiliary, histograms are additionally plotted via Matplotlib (\texttt{plt.hist}), with the optimal AUROC threshold indicated by a vertical line to aid the analysis of the distribution between normal and anomalies.

\textbf{Details of initial anomaly map generation.}
To generate the initial anomaly map for attention guidance (after Eq.~6 in the main paper), we apply global min pooling along the channel dimension of the cost volume for unimodal RGB UAD, following~\cite{glad}. This step selects the normal template most similar to each patch token, enabling self-attention to focus on discriminative feature patterns without interference from depth information. For multimodal UAD, we instead adopt average pooling~\cite{adaclip,cfm}, which aggregates information uniformly across modalities. This strategy preserves the complementary nature of each modality and facilitates more robust multimodal fusion.

\textbf{Inference details for anomaly detection and localization.}
Our cost filtering network generates pixel-level anomaly score maps for localization. For image-level anomaly detection, the anomaly score is computed by averaging the top 250 values (following Glad~\cite{glad}) in the anomaly map for unimodal and multimodal anomaly detection. This approach is based on the intuition that accurate anomaly localization correlates with reliable image-level classification. Using the top 250 values effectively mitigates the instability caused by relying on a single top (max) anomaly point.

\textbf{Details of Metric Calculations.}
For quantitative evaluation, we employ several widely used metrics to assess performance at both image and pixel levels.
(i) The Area Under the Receiver Operating Characteristic Curve (AUROC) is computed for both image- and pixel-level evaluations using the \texttt{roc\_auc\_score} function from \texttt{scikit-learn}.
(ii) To evaluate the precision-recall trade-off, the precision-recall curve is calculated using the \texttt{precision\_recall\_curve} function from \texttt{scikit-learn}. The F1-score is then derived from the precision and recall values with the formula:
\[
\text{F1-score} = \frac{2 \times \text{precision} \times \text{recall}}{\text{precision} + \text{recall}}.
\]
The F1-max score is obtained by selecting the maximum value from the computed F1-scores at all thresholds.
(iii) The Area Under the Precision-Recall Curve (AP, also referred to as AUPRC), is calculated at both the image and pixel levels using the \texttt{average\_precision\_score} function from \texttt{scikit-learn}.
(iv) For anomaly localization of unimodal RGB and multimodal RGB-Text scenarios, we compute the AUPRO (Area Under the Per-Region Overlap Curve) using the standard function \texttt{compute\_pro}, a common function in unimodal RGB and multimodal RGB-Text UAD methods~\cite{glad,dinomaly,anomalydino,adaclip,anomalyclip,aprilgan}.
(v) Finally, for the RGB-3D UAD task, we calculate region-level AUPRO metrics at multiple thresholds (30\%, 10\%, 5\%, and 1\%) using the \texttt{calculate\_au\_pro} function, which is widely adopted in multimodal RGB-3D UAD methods~\cite{cfm,m3dm,pami2,ast,btf}. For instance, AUPRO@30\% is calculated as the area under the PRO curve, integrated up to FPR = 0.3. This evaluation measures the Area Under the Per-Region Overlap Curve (AUPRO) at different sensitivity levels based on the top predicted anomaly scores, thereby enabling a comprehensive assessment of performance across thresholds.

\section{Further Experimental Analyses}
\phantomsection
\label{suppsec4}

\begin{table}[!t]
     \centering
    \renewcommand{\arraystretch}{1.3}  
	\caption{Quantitative results of \textbf{one-model-per-category} in \textbf{unimodal RGB UAD}, obtained with our \textbf{unified multi-class filtering model}. The table corresponds to Sec.~\ref{suppsec4a}.}
    \label{suppsinglclass}%
    \resizebox{1.0\linewidth}{!}{
	\setlength{\tabcolsep}{2.95pt}  
	\huge 
	\resizebox{1\linewidth}{!}{
       \begin{tabular}{c|cc|ccc|cccc}
    \toprule
    \multirow{2}[2]{*}{\Huge Datasets} & 
    \multicolumn{2}{c|}{\Huge Method} & 
    \multicolumn{3}{c|}{\Huge Image-level} & 
    \multicolumn{4}{c}{\Huge Pixel-level} \\
    \cmidrule{2-10}
    & \Huge Glad & \Huge +UCF & \Huge AUROC & \Huge AP & \Huge F1-max & \Huge AUROC & \Huge AP & \Huge F1-max & \Huge AUPRO \\
    \midrule

			\multirow{2}[1]{*}{{\Huge MVTec.}} & \checkmark &  & {\Huge 99.0}  & {\Huge \textbf{99.7}}  & {\Huge 98.2}  & {\Huge 98.7}  & {\Huge 63.8}  & {\Huge 63.7}  & {\Huge 95.2} \\
            
            & {\cellcolor{lightblue}{\checkmark}} & {\cellcolor{lightblue}{\checkmark}}
            & {\cellcolor{lightblue}{\Huge \textbf{99.3}}}  
            & {\cellcolor{lightblue}{\Huge \textbf{99.7}}}  
            & {\cellcolor{lightblue}{\Huge \textbf{98.3}}}  
            & {\cellcolor{lightblue}{\Huge \textbf{98.9}}}  
            & {\cellcolor{lightblue}{\Huge \textbf{66.2}}}  
            & {\cellcolor{lightblue}{\Huge \textbf{65.0}}}  
            & {\cellcolor{lightblue}{\Huge \textbf{96.4}}} \\

			\midrule
			\multirow{2}[1]{*}{{\Huge VisA}} & \checkmark &  & {\Huge 99.3}  & {\Huge 99.6}  & {\Huge 97.6}  & {\Huge 98.3}  & {\Huge 35.8}  & {\Huge 42.4}  & {\Huge 94.1} \\
            & {\cellcolor{lightblue}{\checkmark}} & {\cellcolor{lightblue}{\checkmark}}
            & {\cellcolor{lightblue}{\Huge \textbf{99.5}}}  
            & {\cellcolor{lightblue}{\Huge \textbf{99.7}}}  
            & {\cellcolor{lightblue}{\Huge \textbf{98.1}}}  
            & {\cellcolor{lightblue}{\Huge \textbf{98.6}}} 
            & {\cellcolor{lightblue}{\Huge \textbf{37.3}}}  
            & {\cellcolor{lightblue}{\Huge \textbf{45.3}}}  
            & {\cellcolor{lightblue}{\Huge \textbf{94.5}}}
             \\
			\bottomrule
		\end{tabular}}%
	}
\end{table}%

\subsection{Validation of Single-Class Compatibility}
\phantomsection
\label{suppsec4a}

In unimodal RGB UAD, a common alternative is to train one model per category~\cite{glad,liu2023simplenet,pami1} rather than a unified model for multiple classes. To assess generalization, we apply our unified multi-class model to filter anomaly volumes using features extracted from the reconstructions of category-specific diffusion models in GLAD~\cite{glad}. We adopt the per-category diffusion models at $256 \times 256$ resolution and keep our unified model fixed without additional fine-tuning or per-class training. As reported in Table~\ref{suppsinglclass}, this plug-in setup consistently improves class-wise mean performance for image-level detection and pixel-level localization, validating its compatibility with the ``one model per category'' paradigm for unimodal RGB UAD.

\begin{table}[!t]
     \centering
    \renewcommand{\arraystretch}{1.4}  
	\caption{\textbf{Few-shot} exploration for \textbf{multimodal RGB–Text UAD} on Mvtec-AD, comparing WinCLIP \cite{winclip}, AprilGAN \cite{aprilgan}, and AprilGAN+UCF.}
    \label{suppfewshotmvtec}%
    \resizebox{1.0\linewidth}{!}{
	\setlength{\tabcolsep}{2.95pt}  
	\huge 
	\resizebox{1\linewidth}{!}{
        \begin{tabular}{c|c|ccc|cccc}
    \toprule
    \multirow{2}[2]{*}{Shots} & 
    \multicolumn{1}{|c|}{\multirow{2}[2]{*}{  Method}} & 
    \multicolumn{3}{c|}{  Image-level} & 
    \multicolumn{4}{c}{  Pixel-level} \\
    \cmidrule{3-9}          
    &  &    AUROC  &    AP  &    F1-max  &    AUROC  &    AP  &    F1-max  &    AUPRO   \\
    \midrule

			\multirow{3}[1]{*}  {0-shot}   &    WinCLIP\cite{winclip}  &    \textbf{91.8}   &    \textbf{96.5}   &    \textbf{92.9}   &    85.1   &    -   &    31.7   &    \textbf{64.6}  \\
			&    AprilGAN\cite{aprilgan}  &      86.2    &    93.6   &      90.4    &      87.6    &      40.8    &      43.3    &      44.0   \\
            
            & {\cellcolor{lightblue}\cite{aprilgan} + UCF}  
            & {\cellcolor{lightblue}91.1}  
            & {\cellcolor{lightblue}95.7}  
            & {\cellcolor{lightblue}92.0}  
            & {\cellcolor{lightblue}\textbf{89.3}}  
            & {\cellcolor{lightblue}\textbf{41.0}}  
            & {\cellcolor{lightblue}\textbf{43.5}}  
            & {\cellcolor{lightblue}54.0} \\

			\midrule

            \multirow{3}[1]{*}     {1-shot}   &   WinCLIP\cite{winclip}  &  93.1$_{\text{\LARGE ±2.0}}$   &    96.5$_{\text{\LARGE ±0.9}}$   &    93.7$_{\text{\LARGE ±1.1}}$   &    95.2$_{\text{\LARGE ±0.5}}$   &    -   &    55.9$_{\text{\LARGE ±2.7}}$   &    87.1$_{\text{\LARGE ±1.2}}$  \\
            &    AprilGAN\cite{aprilgan}  &    92.0$_{\text{\LARGE ±0.3}}$   &    95.8$_{\text{\LARGE ±0.2}}$   &    92.4$_{\text{\LARGE ±0.2}}$   &    95.1$_{\text{\LARGE ±0.1}}$   &    51.8$_{\text{\LARGE ±0.1}}$   &    54.2$_{\text{\LARGE ±0.0}}$   &    90.6$_{\text{\LARGE ±0.2}}$  \\
    
            & {\cellcolor{lightblue}\cite{aprilgan} + UCF} 
            & {\cellcolor{lightblue}\textbf{97.7$_{\text{\LARGE ±0.1}}$}} 
            & {\cellcolor{lightblue}\textbf{98.6$_{\text{\LARGE ±0.1}}$}} 
            & {\cellcolor{lightblue}\textbf{96.5$_{\text{\LARGE ±0.1}}$}} 
            & {\cellcolor{lightblue}\textbf{96.7$_{\text{\LARGE ±0.1}}$}} 
            & {\cellcolor{lightblue}\textbf{66.9$_{\text{\LARGE ±0.3}}$}} 
            & {\cellcolor{lightblue}\textbf{65.7$_{\text{\LARGE ±0.3}}$}} 
            & {\cellcolor{lightblue}\textbf{93.6$_{\text{\LARGE ±0.1}}$}} \\ \midrule

	\multirow{3}[1]{*}     {2-shot}   &    WinCLIP\cite{winclip}  &    94.4$_{\text{\LARGE ±1.3}}$   &    94.4$_{\text{\LARGE ±0.8}}$   &    \textbf{97.0$_{\text{\LARGE ±0.7}}$ }  &    96.0$_{\text{\LARGE ±0.3}}$   &    -   &    58.4$_{\text{\LARGE ±1.7}}$   &    88.4$_{\text{\LARGE ±0.9}}$  \\
            &    AprilGAN\cite{aprilgan}  &    92.4$_{\text{\LARGE ±0.3}}$   &    96.0$_{\text{\LARGE ±0.2}}$   &    92.6$_{\text{\LARGE ±0.1}}$   &    95.5$_{\text{\LARGE ±0.0}}$   &    53.4$_{\text{\LARGE ±0.4}}$   &    55.9$_{\text{\LARGE ±0.5}}$   &    91.3$_{\text{\LARGE ±0.1}}$  \\

            & {\cellcolor{lightblue}\cite{aprilgan} + UCF}  
            & {\cellcolor{lightblue}\textbf{98.2$_{\text{\LARGE ±0.2}}$}}  
            & {\cellcolor{lightblue}\textbf{98.9$_{\text{\LARGE ±0.2}}$}}  
            & {\cellcolor{lightblue}96.9$_{\text{\LARGE ±0.3}}$}  
            & {\cellcolor{lightblue}\textbf{97.0$_{\text{\LARGE ±0.0}}$}}  
            & {\cellcolor{lightblue}\textbf{68.0$_{\text{\LARGE ±0.3}}$}}  
            & {\cellcolor{lightblue}\textbf{66.5$_{\text{\LARGE ±0.2}}$}}  
            & {\cellcolor{lightblue}\textbf{94.1$_{\text{\LARGE ±0.1}}$}} \\ \midrule

            \multirow{3}[1]{*}     {4-shot}   &    WinCLIP\cite{winclip}  &    95.2$_{\text{\LARGE ±1.3}}$   &    97.3$_{\text{\LARGE ±0.6}}$   &    94.7$_{\text{\LARGE ±0.8}}$   &    96.2$_{\text{\LARGE ±0.3}}$   &    -   &    59.5$_{\text{\LARGE ±1.8}}$   &    89.0$_{\text{\LARGE ±0.8}}$  \\
            &    AprilGAN\cite{aprilgan}  &    92.8$_{\text{\LARGE ±0.2}}$   &    96.3$_{\text{\LARGE ±0.1}}$   &    92.8$_{\text{\LARGE ±0.1}}$   &    95.9$_{\text{\LARGE ±0.0}}$   &    54.5$_{\text{\LARGE ±0.2}}$   &    56.9$_{\text{\LARGE ±0.1}}$   &    91.8$_{\text{\LARGE ±0.1}}$  \\

            & {\cellcolor{lightblue}\cite{aprilgan} + UCF}  
            & {\cellcolor{lightblue}\textbf{98.5$_{\text{\LARGE ±0.0}}$}}  
            & {\cellcolor{lightblue}\textbf{99.0$_{\text{\LARGE ±0.1}}$}}  
            & {\cellcolor{lightblue}\textbf{97.2$_{\text{\LARGE ±0.1}}$}}  
            & {\cellcolor{lightblue}\textbf{97.2$_{\text{\LARGE ±0.1}}$}}  
            & {\cellcolor{lightblue}\textbf{68.9$_{\text{\LARGE ±0.2}}$}}  
            & {\cellcolor{lightblue}\textbf{67.1$_{\text{\LARGE ±0.2}}$}}  
            & {\cellcolor{lightblue}\textbf{94.5$_{\text{\LARGE ±0.1}}$}} \\

			\bottomrule
		\end{tabular}}%
	}
       \vspace{1mm}
\end{table}%

\subsection{Few-Shot on MVTec-AD for RGB-Text UAD}
\phantomsection
\label{suppsec4b}

Table~\ref{suppfewshotmvtec} reports consistent few-shot gains, with category-wise means and standard deviations computed across 5 random seeds for sampling few-shot normal templates on MVTec-AD. In the 0-shot case, pixel-level AUROC improves from 87.6\% to 89.3\%, indicating refined anomaly localization. With only a few normal references at the inference stage (not during training), the margins grow: at 1-shot, image-level AUROC and AP reach 97.7\% and 98.6\%, respectively, and AUPRO rises from 90.6\% to 93.6\%. At 4-shots, I-AUROC attains 98.5\% and AUPRO reaches 94.5\%, exceeding AprilGAN by over 5 percentage points. Overall, the few-shot configuration yields robust improvements with low seed variance.
Per-category results on MVTec-AD are reported in Table~\ref{suppfewshot_mvtec1} (1-shot), Table~\ref{suppfewshot_mvtec2} (2-shots), and Table~\ref{suppfewshot_mvtec4} (4-shots). Corresponding per-category results on VisA are presented in Table~\ref{suppfewshot_visa1} (1-shot), Table~\ref{suppfewshot_visa2} (2-shot), and Table~\ref{suppfewshot_visa4} (4-shot) in this material, and category-wise means are summarized in Table XIII of the main paper.

\begin{table*}[!t]
\centering
\caption{Quantitative comparison for \textbf{multimodal RGB–Text UAD}. Supplementary to Table~VII in the main paper, zero-shot anomaly detection and localization results are reported on seven \textbf{industrial-domain} datasets using \textbf{I-F1-max}/\textbf{P-F1-max}/\textbf{P-AP}. \\ \textbf{Best} results are in bold and \underline{runner-ups} are underlined.
}

\label{suppzsadindus}
\setlength\tabcolsep{3pt} 
\renewcommand{\arraystretch}{1.2}
\begin{tabular}{c|c>{\columncolor{lightblue}}c|c>{\columncolor{lightblue}}c|c>{\columncolor{lightblue}}c}
\toprule
Datasets  & AprilGan \cite{aprilgan} & \shortstack{ + UCF} & AadCLIP \cite{adaclip} & \shortstack{ + UCF} & AnomalyCLIP \cite{anomalyclip} & \shortstack{ + UCF}  \\ \hline

MVTec-AD &    90.4 / 43.3 / 40.8    &    92.0 / 43.5 / \textbf{41.0}    &    92.0 / \textbf{44.0} / 41.6    &    92.7 / 41.1 / 39.1    &    92.7 / 39.1 / 34.5    &    \textbf{93.2} / 43.2 / 39.2   \\           
VisA  &    78.6 / 32.3 / 25.8    &    80.9 / 33.2 / 27.1    &    83.5 / 37.1 / 31.5    &    \textbf{83.8} / \textbf{39.7} / \textbf{32.1}    &    80.4 / 28.3 / 21.3    &    81.6 / 33.3 / 26.4   \\
MPDD  &    80.4 / 31.3 / 26.6    &    81.0 / 30.9 / 26.7    &    78.7 / 32.8 / 29.8    &    79.0 / 29.7 / 26.6    &    80.4 / 34.2 / 28.9    &    \textbf{82.4} / \textbf{36.9} / \textbf{31.7}   \\
BTAD  &    68.1 / 40.6 / 36.5    &    76.5 / 40.8 / 38.4    &    \textbf{89.1} / \textbf{52.2} / \textbf{48.2}    &    87.7 / 49.8 / 47.4    &    83.8 / 49.7 / 45.5    &    86.4 / 49.4 / 44.6   \\           
SDD  &    89.8 / 44.4 / 36.8    &    82.5 / 42.5 / 37.2    &    82.4 / 51.3 / 45.5    &    83.5 / 52.3 / 45.6    &    89.7 / 56.5 / 51.9    &    \textbf{91.6} / \textbf{59.8} / \textbf{52.7}   \\
DAGM  &    91.0 / 44.5 / 38.0    &    92.3 / 41.5 / 35.8    &    94.5 / \textbf{66.4} / \textbf{64.4}    &    93.2 / 61.5 / 59.8    &    95.9 / 62.0 / 61.3    &    \textbf{97.0} / 62.1 / 62.7   \\
DTD-Synthetic   &    89.0 / 67.4 / 66.9    &    93.9 / 67.9 / 70.2    &    92.4 / 69.0 / 72.1    &    94.1 / \textbf{69.2} / \textbf{71.5}    &    93.6 / 62.2 / 62.6    &    \textbf{96.0} / 68.4 / 68.9   \\    \hline        
\textbf{Mean}   &    83.9 / 43.4 /38.8    &    85.6 / 42.9 / 39.5    &    87.5 / \textbf{50.4} / \textbf{47.6}    &    87.7 / \underline{49.0} / 46.0    &    \underline{88.1} / 47.4 / 43.7    &    \textbf{89.7} / \textbf{50.4} / \underline{46.6}   \\

\bottomrule
\end{tabular}%
\end{table*}

\begin{table}[!t]
\centering
\caption{Quantitative comparison for \textbf{multimodal RGB–Text UAD}. Supplementary to Table~VIII in the main paper, zero-shot \textbf{image-level F1-max} results on three \textbf{medical-domain} datasets. \textbf{Best} and \underline{runner-ups} are highlighted.}

\label{suppmedical2f1max}
\setlength\tabcolsep{2.2pt} 
\renewcommand{\arraystretch}{1.2}
\begin{tabular}{c|c>{\columncolor{lightblue}}c|c>{\columncolor{lightblue}}c|c>{\columncolor{lightblue}}c}
\toprule
Datasets  & AprilGan & \shortstack{ + UCF} & AnomalyCLIP & \shortstack{ + UCF} & AadCLIP & \shortstack{ + UCF}  \\ \hline
HeadCT   &    81.2    &    83.3    &    88.4    &    93.9    &    93.2    &    \textbf{95.6}   \\           
BrainMRI   &    91.0    &    94.0    &    86.5    &    93.4    &    94.4    &    \textbf{94.6}   \\
Br35H   &    85.6    &    90.7    &    86.8    &    93.1    &    94.9    &    \textbf{95.2}   \\ \hline
\textbf{Mean}   &    85.9    &    89.3    &    87.2    &    93.5    &    \underline{94.2}    &    \textbf{95.1} \\

\bottomrule
\end{tabular}%
\end{table}

\begin{table}[!t]
\centering
\caption{Quantitative comparison for \textbf{multimodal RGB–Text UAD}. Supplementary to Table~VIII in the main paper, zero-shot \textbf{pixel-level F1-max/AP} results on six \textbf{medical-domain} datasets. \textbf{Best} and \underline{runner-ups} are highlighted.}

\label{suppmedical3pap}
\setlength\tabcolsep{1.6pt} 
\renewcommand{\arraystretch}{1.2}
\begin{tabular}{c|c>{\columncolor{lightblue}}c|c>{\columncolor{lightblue}}c|c>{\columncolor{lightblue}}c}
\toprule
Datasets  & AprilGan & \shortstack{ + UCF} & AnomalyCLIP & \shortstack{ + UCF} & AadCLIP & \shortstack{+ UCF}  \\ \hline
ISIC  &    72.6/79.6   &    77.1/83.2    &    71.6/76.1    &    \textbf{77.5}/\textbf{83.4}    &    72.9/76.7    &    72.5/75.4    \\
ColonDB   &    29.4/21.4    &    31.1/22.9    &    37.5/31.7    &    38.1/32.9    &    56.6/60.0    &    \textbf{57.7}/\textbf{61.3}    \\
ClinicDB    &    36.9/29.1     &    38.6/30.9    &    40.9/34.0    &    44.5/38.2    &    63.6/68.0    &    \textbf{66.6}/\textbf{72.8}   \\
Kvasir    &    40.0/32.2    &    46.5/36.3    &    46.2/39.6    &    49.5/43.8    &    77.0/84.5    &    \textbf{77.8}/\textbf{85.1}    \\
Endo    &    44.8/38.6    &    51.3/42.8    &    50.3/46.6    &    54.7/51.3    &    80.2/\textbf{87.7}    &    \textbf{81.9}/87.4    \\
TN3K  &    35.1/32.8    &    39.7/35.5    &    47.8/45.7    &    \textbf{49.2}/\textbf{48.2}    &    44.8/39.1    &    46.5/40.3   \\  \hline
\textbf{Mean}    &    43.1/39.0    &    47.4/41.9    &    49.1/45.6    &    52.3/49.6    &    \underline{65.8}/\underline{69.3}    &    \textbf{67.1}/\textbf{70.4}   \\

\bottomrule
\end{tabular}%
\end{table}

\begin{figure}
\setlength{\abovecaptionskip}{2pt}  
\setlength{\belowcaptionskip}{0pt} 
\begin{center}
\centerline{\includegraphics[width=0.5\textwidth]{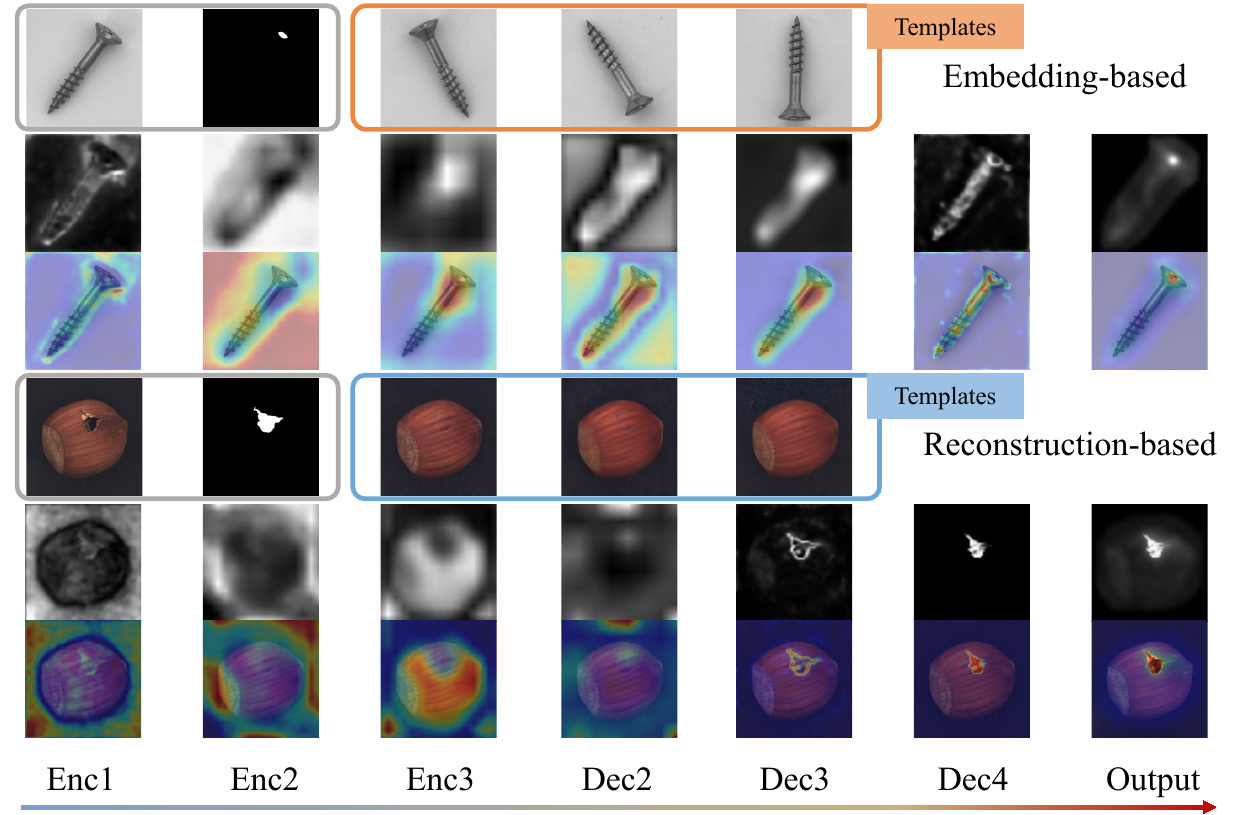}}
\caption{Visualization of progressive matching-noise suppression. Anomaly features extracted by our RCSA modules across encoder (Enc1–3) and decoder (Dec2–4) layers are shown for the embedding-based paradigm (AnomalDF+UCF, example: \textit{screw}) and reconstruction-based (GLAD+UCF, example: \textit{metal nut}) paradigm. At each layer, the most anomaly-indicative channel is aggregated into a score map and upsampled to heatmaps. Across both cases, our method progressively refines from Enc1 to Dec4, suppressing template-induced noise, focusing activations on true defects, and yielding accurate anomaly maps.}
\label{suppdenoise1}
\end{center}
\vspace{-2mm}
\end{figure}

\subsection{Additional Metrics in RGB-Text UAD}
\phantomsection
\label{suppsec4c}

As a complement to Table~VII in the main paper, Table~\ref{suppzsadindus} reports additional dataset-level results for zero-shot anomaly detection on seven industrial datasets (MVTec-AD~\cite{mvtec}, VisA~\cite{visa}, MPDD~\cite{mpdd}, BTAD~\cite{btad}, SDD~\cite{ksdd}, DAGM~\cite{dagm}, and DTD-Synthetic~\cite{dtd}) using I-F1-max, P-F1-max, and P-AP. 
Across most metrics, our plug-in consistently improves the mean performance of all three baselines. For example, AprilGAN+UCF raises the mean I-F1-max/P-F1-max/P-AP from 83.9\%/43.4\%/38.8\% to 85.6\%/42.9\%/39.5\%, with notable gains on MVTec-AD (I-F1-max 90.4\% $\rightarrow$ 92.0\%) and DTD-Synthetic (P-AP 66.9\% $\rightarrow$ 70.2\%). AnomalyCLIP+UCF also benefits, with mean I-F1-max increasing from 88.1\% to 89.7\% and P-AP from 43.7\% to 46.6\%. These results demonstrate that cost volume filtering consistently enhances anomaly detection accuracy and yields a more balanced trade-off between precision and recall on industrial benchmarks.

Supplementing Table VIII in the main paper, Table~\ref{suppmedical2f1max} presents the I-F1-max results in the medical domain datasets. It shows that image-level F1-max improves by 3.4\% percentage points for AprilGAN~\cite{aprilgan} and 6.3\% percentage points for AnomalyCLIP~\cite{anomalyclip}, while AdaCLIP+UCF attains the best mean of 95.1\% and sets a new state of the art on HeadCT~\cite{headct} (95.6\%), BrainMRI~\cite{brainmri} (94.6\%), and Br35H~\cite{br35h} (95.2\%). Pixel-level results in Table~\ref{suppmedical3pap} are consistent. Across six datasets, our plug-in raises both F1-max and AP, with AdaCLIP+UCF achieving the highest means of 67.1\% and 70.4\%. Gains are particularly strong on CVC-ClinicDB~\cite{clidb} and Kvasir~\cite{endo}, and AnomalyCLIP+UCF performs best on TN3K~\cite{tn3k}, achieving 49.2\% and 48.2\%.

\begin{table*}[!t]
\centering
\caption{Quantitative comparison of \textbf{unimodal RGB UAD} on MVTec-AD and VisA, comparing AnomalDF~\cite{anomalydino} and AnomalDF+UCF under a unified, fair evaluation protocol across varied resize and template settings.}
\label{supptab:adf_ours_comparison}
\renewcommand{\arraystretch}{1.0}
\resizebox{\linewidth}{!}{
\begin{tabular}{c|c|c|c|c|ccc|cccc}
\toprule
ID & Dataset & Method & Resize & Templates & I-AUROC & I-AP & I-F1-max & P-AUROC & P-AP & P-F1-max & AUPRO \\
\midrule
1  & \multirow{8}{*}{\centering MVTec-AD} & AnomalDF     & 256 & 3    & 96.8 & 98.6 & 97.1 & 98.1 & 61.3 & 60.8 & 93.6 \\
2  &          & {\cellcolor{lightblue}+UCF}   & {\cellcolor{lightblue}256} & {\cellcolor{lightblue}3}    & {\cellcolor{lightblue}98.5} & {\cellcolor{lightblue}99.4} & {\cellcolor{lightblue}97.8} & {\cellcolor{lightblue}98.8} & {\cellcolor{lightblue}67.8} & {\cellcolor{lightblue}64.9} & {\cellcolor{lightblue}94.2} \\
\cmidrule(lr){3-12}
3  &          & AnomalDF     & 256 & Full & 99.0 & 99.3 & 98.4 & 97.5 & --   & 58.7 & 91.7 \\
4  &          & {\cellcolor{lightblue}+UCF}   & {\cellcolor{lightblue}256} & {\cellcolor{lightblue}Full} & {\cellcolor{lightblue}99.3} & {\cellcolor{lightblue}99.8} & {\cellcolor{lightblue}98.6} & {\cellcolor{lightblue}98.9} & {\cellcolor{lightblue}68.7} & {\cellcolor{lightblue}65.5} & {\cellcolor{lightblue}96.6} \\
\cmidrule(lr){3-12}
5  &          & AnomalDF     & 448 & Full & 99.3 & 99.7 & 98.8 & 97.9 & --   & 61.8 & 92.9 \\
6  &          & {\cellcolor{lightblue}+UCF}   & {\cellcolor{lightblue}448} & {\cellcolor{lightblue}Full} & {\cellcolor{lightblue}99.5} & {\cellcolor{lightblue}99.8} & {\cellcolor{lightblue}98.9} & {\cellcolor{lightblue}99.0} & {\cellcolor{lightblue}72.4} & {\cellcolor{lightblue}68.4} & {\cellcolor{lightblue}95.4} \\
\cmidrule(lr){3-12}
7  &          & AnomalDF     & 672 & Full & 99.5 & 99.8 & 99.0 & 98.2 & --   & 64.3 & 95.0 \\
8  &          & {\cellcolor{lightblue}+UCF}   & {\cellcolor{lightblue}672} & {\cellcolor{lightblue}Full} & {\cellcolor{lightblue}99.6} & {\cellcolor{lightblue}99.9} & {\cellcolor{lightblue}99.0} & {\cellcolor{lightblue}99.1} & {\cellcolor{lightblue}74.4} & {\cellcolor{lightblue}69.7} & {\cellcolor{lightblue}96.3} \\
\midrule
9  & \multirow{8}{*}{\centering VisA}     & AnomalDF     & 256 & 3    & 90.5 & 91.4 & 86.2 & 97.4 & 39.6 & 40.4 & 86.3 \\
10 &          & {\cellcolor{lightblue}+UCF}   & {\cellcolor{lightblue}256} & {\cellcolor{lightblue}3}    & {\cellcolor{lightblue}94.3} & {\cellcolor{lightblue}95.1} & {\cellcolor{lightblue}90.6} & {\cellcolor{lightblue}99.2} & {\cellcolor{lightblue}44.6} & {\cellcolor{lightblue}45.5} & {\cellcolor{lightblue}84.5} \\
\cmidrule(lr){3-12}
11 &          & AnomalDF     & 256 & Full & 94.6 & 95.7 & 90.9 & 98.3 & --   & 44.3 & 86.7 \\
12 &          & {\cellcolor{lightblue}+UCF}   & {\cellcolor{lightblue}256} & {\cellcolor{lightblue}Full} & {\cellcolor{lightblue}95.5} & {\cellcolor{lightblue}96.3} & {\cellcolor{lightblue}91.5} & {\cellcolor{lightblue}99.4} & {\cellcolor{lightblue}45.9} & {\cellcolor{lightblue}46.6} & {\cellcolor{lightblue}87.0} \\
\cmidrule(lr){3-12}
13 &          & AnomalDF     & 448 & Full & 97.2 & 97.6 & 93.7 & 98.7 & --   & 50.5 & 95.0 \\
14 &          & {\cellcolor{lightblue}+UCF}   & {\cellcolor{lightblue}448} & {\cellcolor{lightblue}Full} & {\cellcolor{lightblue}97.4} & {\cellcolor{lightblue}97.7} & {\cellcolor{lightblue}93.8} & {\cellcolor{lightblue}99.4} & {\cellcolor{lightblue}42.2} & {\cellcolor{lightblue}53.6} & {\cellcolor{lightblue}95.2} \\
\cmidrule(lr){3-12}
15 &          & AnomalDF     & 672 & Full & 97.6 & 97.2 & 94.3 & 98.9 & --   & 53.8 & 96.1 \\
16 &          & {\cellcolor{lightblue}+UCF}   & {\cellcolor{lightblue}672} & {\cellcolor{lightblue}Full} & {\cellcolor{lightblue}97.8} & {\cellcolor{lightblue}98.0} & {\cellcolor{lightblue}94.6} & {\cellcolor{lightblue}99.4} & {\cellcolor{lightblue}47.6} & {\cellcolor{lightblue}54.5} & {\cellcolor{lightblue}96.4} \\
\bottomrule
\end{tabular}}
\end{table*}

\begin{figure}
\setlength{\abovecaptionskip}{2pt}  
\setlength{\belowcaptionskip}{0pt} 
\begin{center}
\centerline{\includegraphics[width=0.5\textwidth]{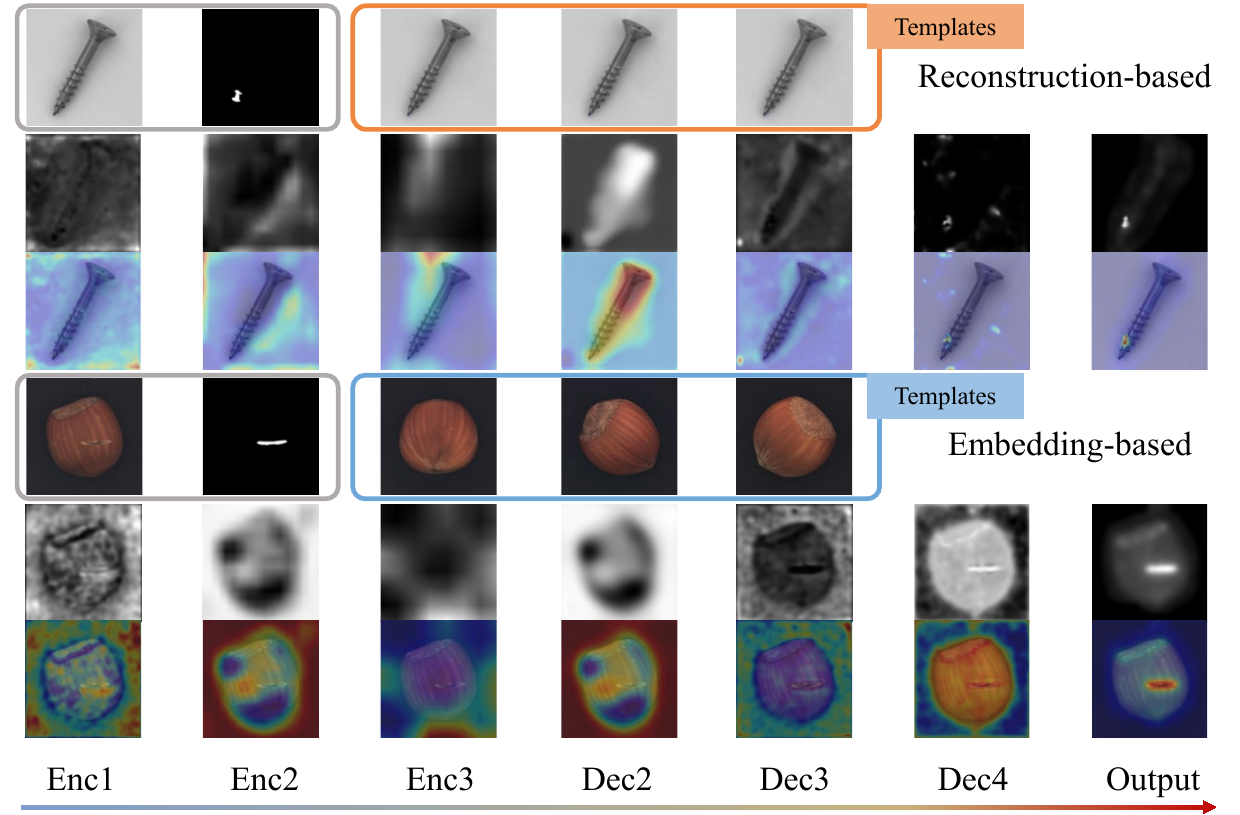}}
\caption{Visualization of progressive matching-noise suppression. Anomaly features extracted by our RCSA modules across different layers are shown for the embedding-based paradigm (AnomalDF+UCF, example: \textit{metal nut}) and the reconstruction-based paradigm (GLAD+UCF, example: \textit{screw}). Our method refines anomaly localization from global to local scales by suppressing spurious correspondences and consolidating anomaly cues.
}
\label{suppdenoise2}
\end{center}
\vspace{-2mm}
\end{figure}

\subsection{Progressive Noise Denoising Visualization}
\phantomsection
\label{suppsec4d}

Complementing the coarse-to-fine anomaly refinement illustrated in the bottom-right panel of Fig.~2 in the main paper, Fig.~\ref{suppdenoise1} and Fig.~\ref{suppdenoise2} visualize anomaly features extracted by the Residual Channel–Spatial Attention (RCSA) modules across the first three encoder and last three decoder layers. For each layer, we select the channel with the highest attention weight, most indicative of anomalies, and aggregate its activations into an attention score map of resolution $H' \times W'$, which is then upsampled to generate layer-wise heatmaps. As illustrated for the \textit{screw} and \textit{metal nut} categories, our method progressively refines anomaly localization for matching results between input images and either randomly sampled or multi-step reconstructed templates. The refinement evolves from global to local scales, suppressing spurious correspondences while consolidating anomaly-relevant evidence, thereby validating the effectiveness of progressive matching noise suppression.

\subsection{Resolution and Baseline Protocols in RGB UAD}
\phantomsection
\label{suppsec4e}
In the main paper, AnomalDF and AnomalDF+UCF were trained with $N=3$ randomly sampled reference templates per input and a resolution of $256 \times 256$, and evaluated using a similarly limited number of templates, offering a trade-off between template diversity and memory efficiency. Exp. IDs 1, 2, 9, and 10 in Table~\ref{supptab:adf_ours_comparison} report the corresponding results.
By contrast, the original full-shot setting of AnomalyDINO \cite{anomalydino} utilizes the entire training set as reference templates and resizes images to a larger resolution. 
To ensure a fair and thorough comparison, we further conducted evaluations under the original full-shot setting of AnomalyDINO. Exp. IDs 3–8 and Exp. IDs 11–16 in Table~\ref{supptab:adf_ours_comparison} report results on MVTec-AD and VisA, respectively.
Notably, to mitigate storage and compute overhead, we directly reuse the models trained in Exp. ID 2 and 10, and test them under different resolutions and template amounts, without additional retraining. The shape of the anomaly cost volumes is adapted via interpolation to match the input shape required by our models. Despite this constraint, UCF consistently improved the performance of AnomalyDINO across various resolutions and datasets.
Remarkably, our method at lower resolution (e.g., $448 \times 448$) can match or outperform the original AnomalDF baseline at higher resolution (e.g., $672 \times 672$), demonstrating its effectiveness. 
These results highlight the scalability and generalization capability of our plug-in method, even under varied operational constraints.

\begin{figure*}
\setlength{\abovecaptionskip}{2pt}  
\setlength{\belowcaptionskip}{0pt} 
\begin{center}
\centerline{\includegraphics[width=\textwidth]{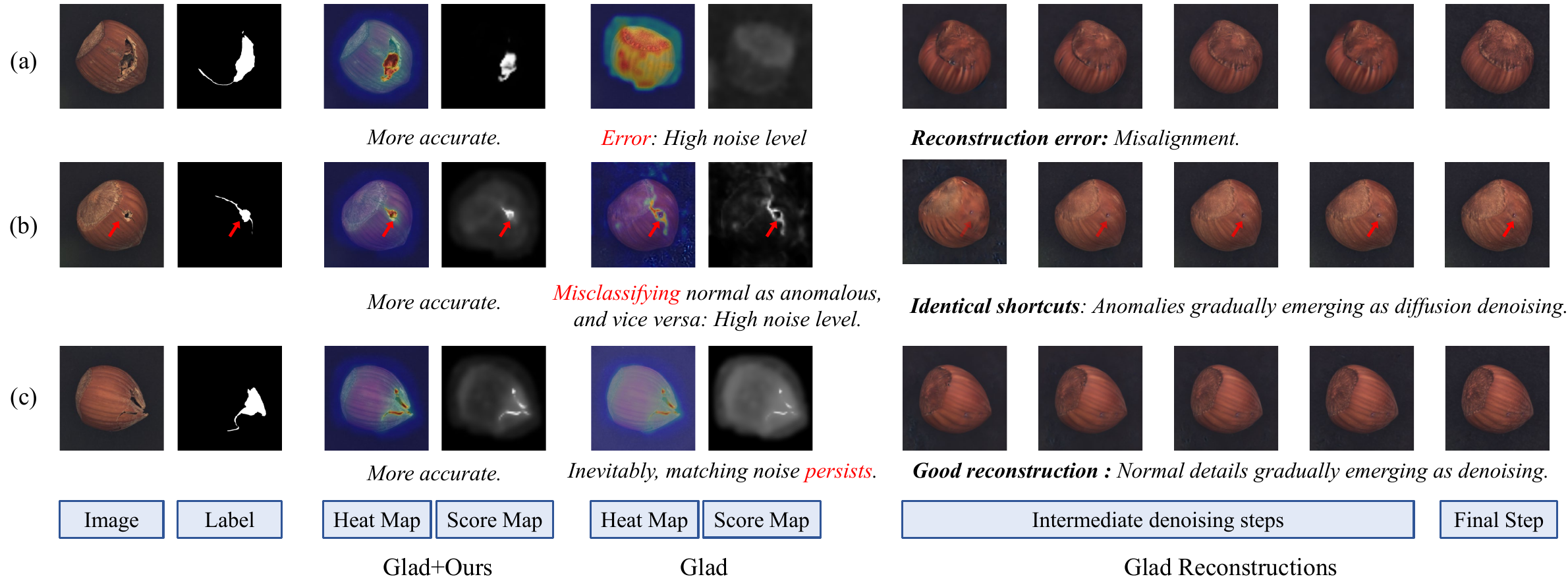}}
\caption{Visualization of challenges in the diffusion-based reconstruction model GLAD \cite{glad} and advancements enabled by our approach.}
\label{suppfig5}
\end{center}
\vspace{-6mm}
\end{figure*}

\subsection{Analysis of Shortcut Issue in Reconstruction}
\phantomsection
\label{suppsec4f}

As discussed in the main paper, the final anomaly score map in unsupervised anomaly detection is typically derived by matching inputs against normal templates. Despite recent advances, anomaly regions often retain substantial matching noise. We use GLAD~\cite{glad} as an example to examine three key challenges encountered by reconstruction-based methods.

(i) Asymmetry in multi-class reconstruction introduces significant matching noise artifacts. As shown in Fig.~\ref{suppfig5}(a), a unified multi-class model must reconstruct diverse anomalies and often distorts shape, texture, or orientation, leading to asymmetric feature matching and pronounced artifacts. Beyond reconstruction-based designs, analogous challenges have been acknowledged in embedding-based methods~\cite{patchcore, anomalydino}. Our approach mitigates it by using input feature guidance within dual-stream feature attention, which focuses the filtering model on spatial structures and preserves edge details to improve localization, thereby ensuring compatibility with both reconstruction- and embedding-based models.

(ii) The ``identical shortcut" issue weakens residual signals. Reconstruction-based methods~\cite{glad} may leak anomaly information such that anomalies persist in outputs, commonly referred to as the ``identical shortcut" issue. As shown in Fig. \ref{suppfig5} (b), a hazelnut's small-hole anomaly remains visible after reconstruction, weakening residual anomaly signals and hindering detection. In contrast, the intermediate outputs of the multi-step diffusion denoising process primarily reconstruct low-frequency features \cite{freq} (e.g., normal structures) at earlier stages. As the reconstruction progresses, this issue may gradually reintroduce anomalous information, making anomalies more apparent in later steps. Motivated by this observation, our matching cost volume and filtering network integrate multi-step reconstruction results as templates, substantially improving anomaly detection.

(iii) Noise interference is common during template–image matching. As shown in Fig.~\ref{suppfig5}(c), the matching process may introduce noise even when anomalies are reconstructed in a predominantly normal style. Our method effectively suppresses these potential artifacts, enhancing anomaly localization.

\section{Comprehensive Per-Class Quantitative Results}
\phantomsection
\label{suppsec5}
We replicated multiple baselines across diverse benchmarks and integrated our plug-in into them, yielding comprehensive results. All model weights and codes will be publicly released at \url{https://github.com/ZHE-SAPI/CostFilter-AD}. For clarity, results are reported to one decimal place, and category-wise means are computed from full-precision values. As broadly recognized in the community, overall mean improvements are of primary interest. While a few category-specific metrics exhibit minor declines, these reflect the inherent trade-offs of pursuing a unified model that generalizes across multi-class anomalies and extends naturally to cross-dataset zero-/few-shot scenarios. Overall, our method delivers strong and consistent improvements across various categories and metrics, underscoring its robustness, effectiveness, and reliability.

\subsection{RGB UAD: Per-Class Quantitative Results}
\phantomsection
\label{suppsec5a}

This subsection compiles per-category results for unimodal RGB UAD under the multi-class protocol with explicit table pointers for direct navigation. For MVTec-AD~\cite{mvtec}, Tables~\ref{suppmvtectable1}--\ref{suppmvtectable4} report I-AUROC/P-AUROC, I-AP/P-AP, I-F1-max/P-F1-max, and AURPO. For VisA~\cite{visa}, Tables~\ref{suppvisatable1}--\ref{suppvisatable4} present I-AUROC/P-AUROC, I-AP/P-AP, I-F1-max/P-F1-max, and AUPRO. For BTAD~\cite{btad}, Tables~\ref{suppbtadtable1} and \ref{suppbtadtable2} summarize I-AUROC/I-AP/I-F1-max and P-AUROC/P-AP/P-F1-max/AUPRO. For MPDD~\cite{mpdd}, Tables~\ref{suppmpddtable1} and \ref{suppmpddtable2} report I-AUROC/I-AP/I-F1-max and P-AUROC/P-AP/P-F1-max/AUPRO. All tables provide per-category entries, best results are bold, and runners-up are underlined. These results complement Table~I, Table~III, as well as Table~IV in the main paper.

Across MVTec-AD, VisA, BTAD, and MPDD, our method consistently improves image-level detection and pixel-level localization under the multi-class setting. Gains are observed on I-AUROC, I-AP, and I-F1-max with concurrent improvements on P-AUROC, P-AP, P-F1-max, and AUPRO. The trend holds for both object and texture categories in MVTec-AD and for multi-instance, structurally complex scenes in VisA, indicating robustness to background clutter and structural variation. On BTAD and MPDD, where defects often span large regions or reflect subtle material changes, gains on AUPRO and P-F1-max further confirm accurate localization with stable precision–recall behavior. These per-category results align with the average mean improvements, showing that the proposed cost volume filtering suppresses matching noise and transfers across datasets without task-specific tuning.

\subsection{RGB-3D UAD: Per-Class Quantitative Results}
\phantomsection
\label{suppsec5b}

Per-class results on MVTec 3D-AD~\cite{mvtec3d} and Eyecandies~\cite{eyecan} have been summarized in Tables~V and VI of the main paper, covering image-level and pixel-level performance with I-AUROC, P-AUROC, and AUPRO@1\%/5\%/10\%/30\%. Across both datasets, our method yields consistent category-wise gains, confirming the effectiveness and cross-dataset applicability of the proposed cost volume filtering.

It is noted that we compute AUPRO as the area under the PRO curve up to a false-positive-rate (FPR) cap. We report AUPRO@30\% (FPR limit $0.30$) and, to reflect stricter industrial requirements, also AUPRO@@1\%/5\%/10\% (FPR limit $0.01$/$0.05$/$0.10$), and AUPRO@1\% is the strictest and most discriminative.

\subsection{RGB-Text UAD: Per-Class Quantitative Results}
\phantomsection
\label{suppsec5c}
This subsection compiles per-category results for multimodal RGB–Text UAD in the zero-shot anomaly detection (ZSAD) setting, with direct table references for navigation. For each industrial dataset, two tables are provided: the first reports I-AUROC/I-AP/P-AUROC/AUPRO, corresponding to the tables in the main paper; the second reports additional metrics, including I-F1-max/P-AP/P-F1-max. Specifically:
MVTec-AD~\cite{mvtec} (Tables~\ref{suppmvtectexttable1}, \ref{suppmvtectexttable2});
VisA~\cite{visa} (Tables~\ref{suppvisatexttable1}, \ref{suppvisatexttable2});
MPDD~\cite{mpdd} (Tables~\ref{suppmpddtexttable1}, \ref{suppmpddtexttable2});
BTAD~\cite{btad} (Tables~\ref{suppbtadtexttable1}, \ref{suppbtadtexttable2});
SDD~\cite{ksdd} (Tables~\ref{suppsddtexttable1}, \ref{suppsddtexttable2});
DAGM~\cite{dagm} (Tables~\ref{suppdagmtexttable1}, \ref{suppdagmtexttable2});
and DTD-Synthetic~\cite{dtd} (Tables~\ref{suppdtdtexttable1}, \ref{suppdtdtexttable2}).

For the medical domain, Table~\ref{suppmedicaltexttable1} reports the image-level results per category in three datasets (HeadCT~\cite{headct}, BrainMRI~\cite{brainmri}, and Br35H~\cite{br35h}), as these datasets provide only image-level labels. Table~\ref{suppmedicaltexttable2} reports the pixel-level results per category in six datasets (ISIC~\cite{isic}, CVC-ClinicDB~\cite{clidb}, CVC-ColonDB~\cite{clodb}, Kvasir~\cite{kvasir}, TN3K~\cite{tn3k}, and Endo~\cite{endo}), as all images in these datasets are abnormal. Across all tables, best results are bold and runner-ups are underlined, complementing Table~VII-VIII in the main paper.

In addition, per-category few-shot results on MVTec-AD are reported in Table~\ref{suppfewshot_mvtec1}, \ref{suppfewshot_mvtec2}, and \ref{suppfewshot_mvtec4}, and on VisA in Table~\ref{suppfewshot_visa1}, \ref{suppfewshot_visa2}, and \ref{suppfewshot_visa4}. Our method consistently improves image-level detection and pixel-level localization. Gains appear on I-AUROC, I-AP, and I-F1-max together with concurrent improvements on P-AUROC, P-AP, P-F1-max, and AUPRO, indicating stronger cross-modal alignment and more reliable delineation of anomalous regions. The trend holds for single-instance and multi-instance categories, texture and object types, and scenes with substantial structural variation, suggesting robustness to background clutter and prompt shift. These per-category results are consistent with the dataset-level averages reported in the main paper, specifically Tables~VII and VIII for the zero-shot case and Table~XIII for the few-shot case, and with Table~\ref{suppfewshotmvtec} in this supplementary material for the few-shot case, indicating that the proposed cost volume filtering generalizes across datasets and domains for multimodal anomaly detection.

\section{Comprehensive Per-Class Qualitative Visualization}
\phantomsection
\label{suppsec6}
\subsection{RGB UAD: Per-Class Qualitative Results}
\phantomsection
\label{suppsec6a}

\begin{figure*}
\setlength{\abovecaptionskip}{2pt}  
\setlength{\belowcaptionskip}{0pt} 
\begin{center}
\centerline{\includegraphics[width=\textwidth]{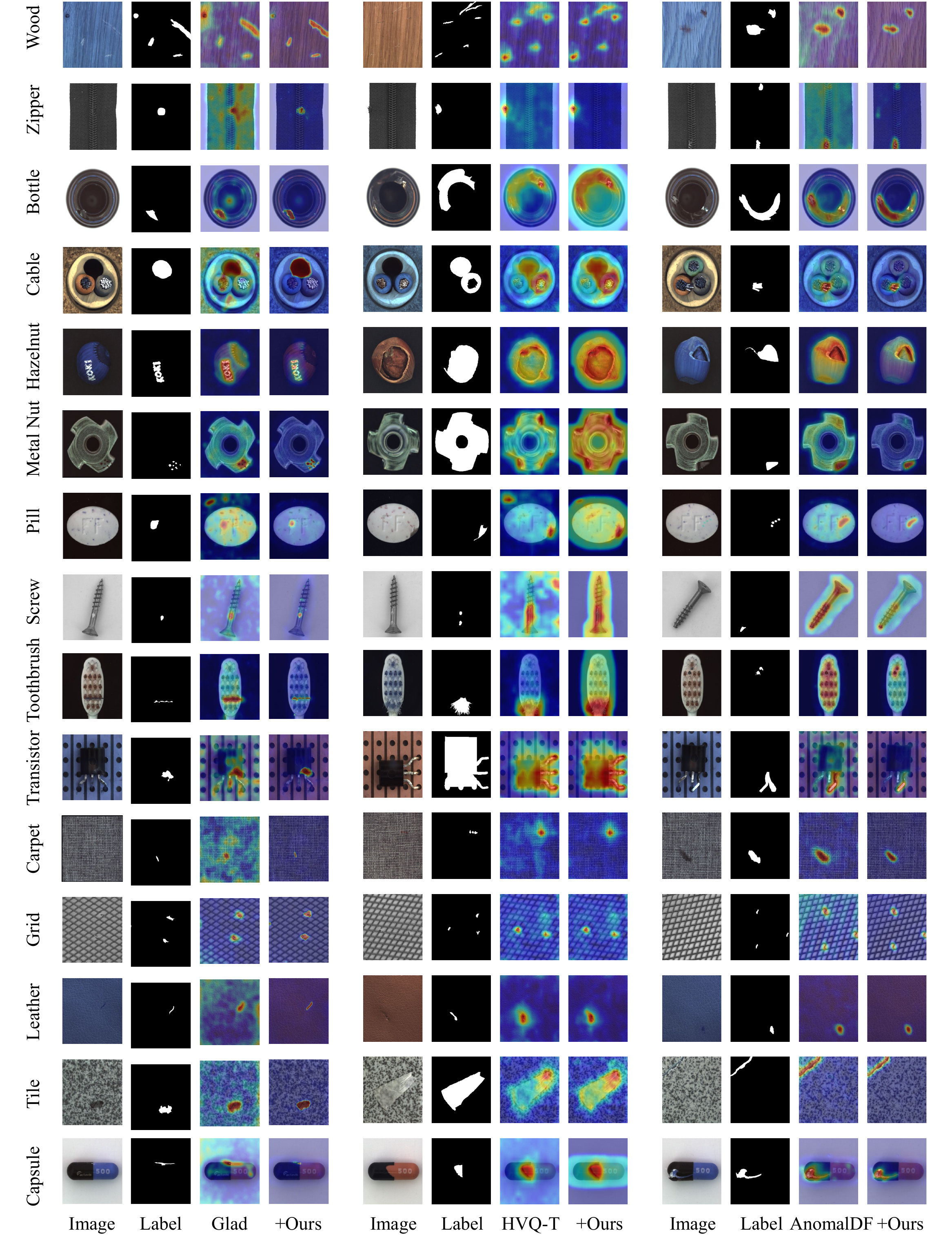}}
\caption{Qualitative examples of \textbf{unimodal RGB UAD} anomaly localization on \textbf{MVTec-AD}~\cite{mvtec}, reporting per-category results. The comparison is divided into three groups, each following the same left-to-right order: input anomaly, ground truth mask, anomaly map predicted by GLAD~\cite{glad}, HVQ-Trans~\cite{hvqtrans}, or AnomalDF~\cite{anomalydino}, and the anomaly map obtained with our method integrated. This figure corresponds to Sec.\ref{suppsec6a}, and Tables~\ref{suppmvtectable1},~\ref{suppmvtectable2},~\ref{suppmvtectable3},~\ref{suppmvtectable4}.
}
\label{suppkeshihuamvtecrgb}
\end{center}
\end{figure*}

\begin{figure*}
\setlength{\abovecaptionskip}{2pt}  
\setlength{\belowcaptionskip}{0pt} 
\begin{center}
\centerline{\includegraphics[width=\textwidth]{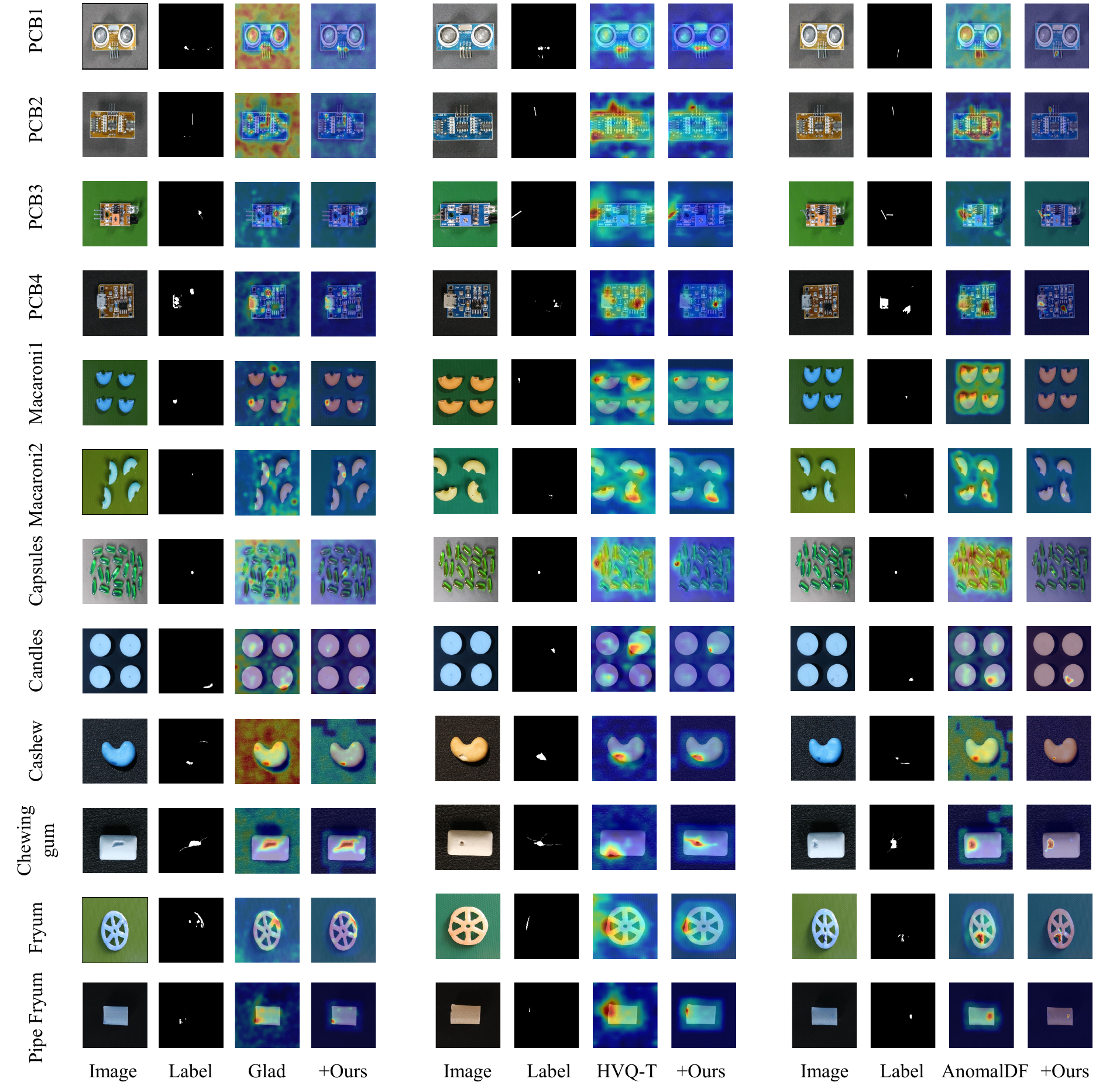}}
\caption{Qualitative examples of \textbf{unimodal RGB UAD} anomaly localization on \textbf{VisA}~\cite{visa}, reporting per-category results. The comparison is divided into three groups, each following the same left-to-right order: input anomaly, ground truth mask, anomaly map predicted by GLAD~\cite{glad}, HVQ-Trans~\cite{hvqtrans}, or AnomalDF~\cite{anomalydino}, and the anomaly map obtained with our method integrated. This figure corresponds to Sec.\ref{suppsec6a}, and Tables~\ref{suppvisatable1},~\ref{suppvisatable2},~\ref{suppvisatable3},~\ref{suppvisatable4}.
}
\label{suppkeshihuavisargb}
\end{center}
\end{figure*}

We present per-category qualitative comparisons for unimodal RGB UAD with comprehensive visualizations. Fig.~\ref{suppkeshihuamvtecrgb} illustrates results on MVTec-AD~\cite{mvtec}, while Fig.~\ref{suppkeshihuavisargb} reports results on VisA~\cite{visa}. In each panel, from left to right, we display the input image, the ground-truth mask, the anomaly maps produced by GLAD~\cite{glad}, HVQ-Trans~\cite{hvqtrans}, and AnomalDF~\cite{anomalydino}, and the anomaly maps obtained after integrating our method.

Across both datasets, the integrated maps suppress background clutter and template-induced noise, concentrate responses on true defects, and improve boundary adherence. Relative to GLAD, HVQ-Trans, and AnomalDF, our integration reduces false positives in texture categories and recovers fine structures in object categories. The heatmaps are more refined and accurate, and align better with the masks. These observations are consistent with the quantitative gains reported in the main paper and Sec~\ref{suppsec5a}.

\subsection{RGB-3D UAD: Per-Class Qualitative Results}
\phantomsection
\label{suppsec6b}

\begin{figure*}
\setlength{\abovecaptionskip}{2pt}  
\setlength{\belowcaptionskip}{0pt} 
\begin{center}
\centerline{\includegraphics[width=\textwidth]{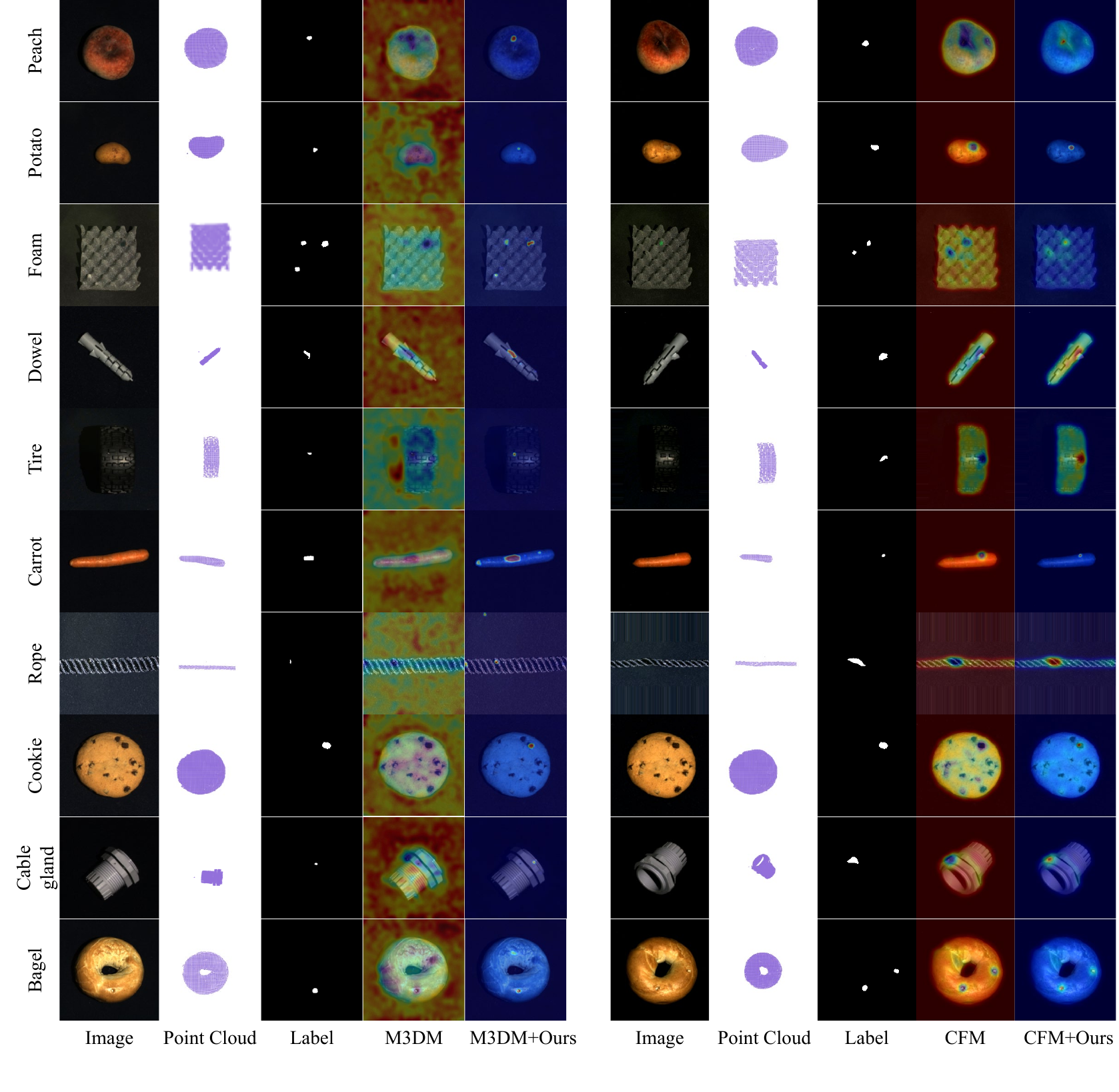}}
\caption{Qualitative examples of \textbf{multimodal RGB-3D UAD} anomaly localization on \textbf{MVTec 3D-AD}~\cite{mvtec3d}, reporting per-category results. The comparison is divided into two groups, each following the same left-to-right order: input anomaly image and point cloud (projected onto 2D for visualization), ground truth mask, anomaly map predicted by M3DM~\cite{m3dm} or CFM~\cite{cfm}, and the anomaly map obtained with our method integrated. Note that although the RGB–point cloud visualizations exhibit misalignment caused by 2D projection, their feature maps are \textbf{pixel-registered}, following M3DM~\cite{m3dm} and CFM~\cite{cfm}. This figure corresponds to Sec.\ref{suppsec6b}, and Table V in the main paper.
}
\label{suppkeshihuamvtec3drgb3d}
\end{center}
\end{figure*}

\begin{figure*}
\setlength{\abovecaptionskip}{2pt}  
\setlength{\belowcaptionskip}{0pt} 
\begin{center}
\centerline{\includegraphics[width=\textwidth]{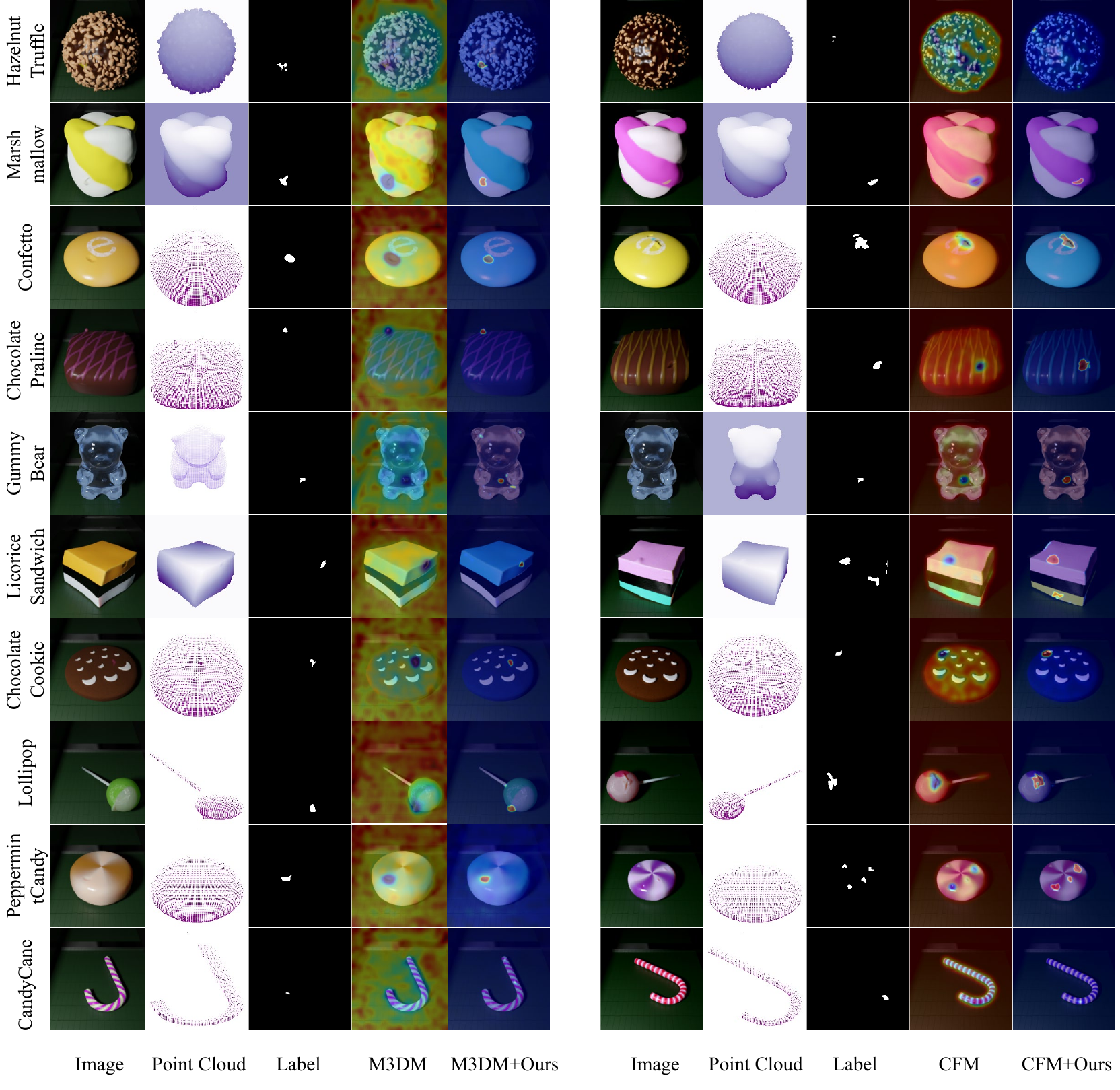}}
\caption{Qualitative examples of \textbf{multimodal RGB-3D UAD} anomaly localization on \textbf{Eyecandies}~\cite{eyecan}, reporting per-category results. The comparison is divided into two groups, each following the same left-to-right order: input anomaly image and point cloud (projected onto 2D for visualization), ground truth mask, anomaly map predicted by M3DM~\cite{m3dm} or CFM~\cite{cfm}, and the anomaly map obtained with our method integrated. Note that although the RGB–point cloud visualizations exhibit misalignment caused by 2D projection, their feature maps are \textbf{pixel-registered}, following M3DM~\cite{m3dm} and CFM~\cite{cfm}. This figure corresponds to Sec.\ref{suppsec6b}, and Table VI in the main paper.
}
\label{suppkeshihuaeyecanrgb3d}
\end{center}
\end{figure*}

We provide per-category qualitative comparisons for multimodal RGB–3D UAD. Fig.~\ref{suppkeshihuamvtec3drgb3d} presents MVTec 3D-AD~\cite{mvtec3d} and Fig.~\ref{suppkeshihuaeyecanrgb3d} presents Eyecandies~\cite{eyecan}. Each panel follows an identical layout that includes the input image, the ground truth mask, the anomaly maps from M3DM~\cite{m3dm} and CFM~\cite{cfm}, and the anomaly map obtained with our method integrated.

Across both datasets, our anomaly maps suppress cross-modal mismatches and background clutter, concentrate responses on defective surfaces and boundaries, and refine edge adherence. Relative to M3DM and CFM, our integration reduces false positives on specular highlights and repetitive textures and preserves fine structures in small defects. The method remains stable under geometry variation and viewpoint changes and shows consistent depth and appearance agreement across modalities. These qualitative trends agree with the comprehensive improvements summarized in Tables~V and VI of the main paper and Sec~\ref{suppsec5b} of this appendix.

\subsection{RGB-Text UAD: Per-Class Qualitative Results}
\phantomsection
\label{suppsec6c}

\begin{figure*}
\setlength{\abovecaptionskip}{2pt}  
\setlength{\belowcaptionskip}{0pt} 
\begin{center}
\centerline{\includegraphics[width=\textwidth]{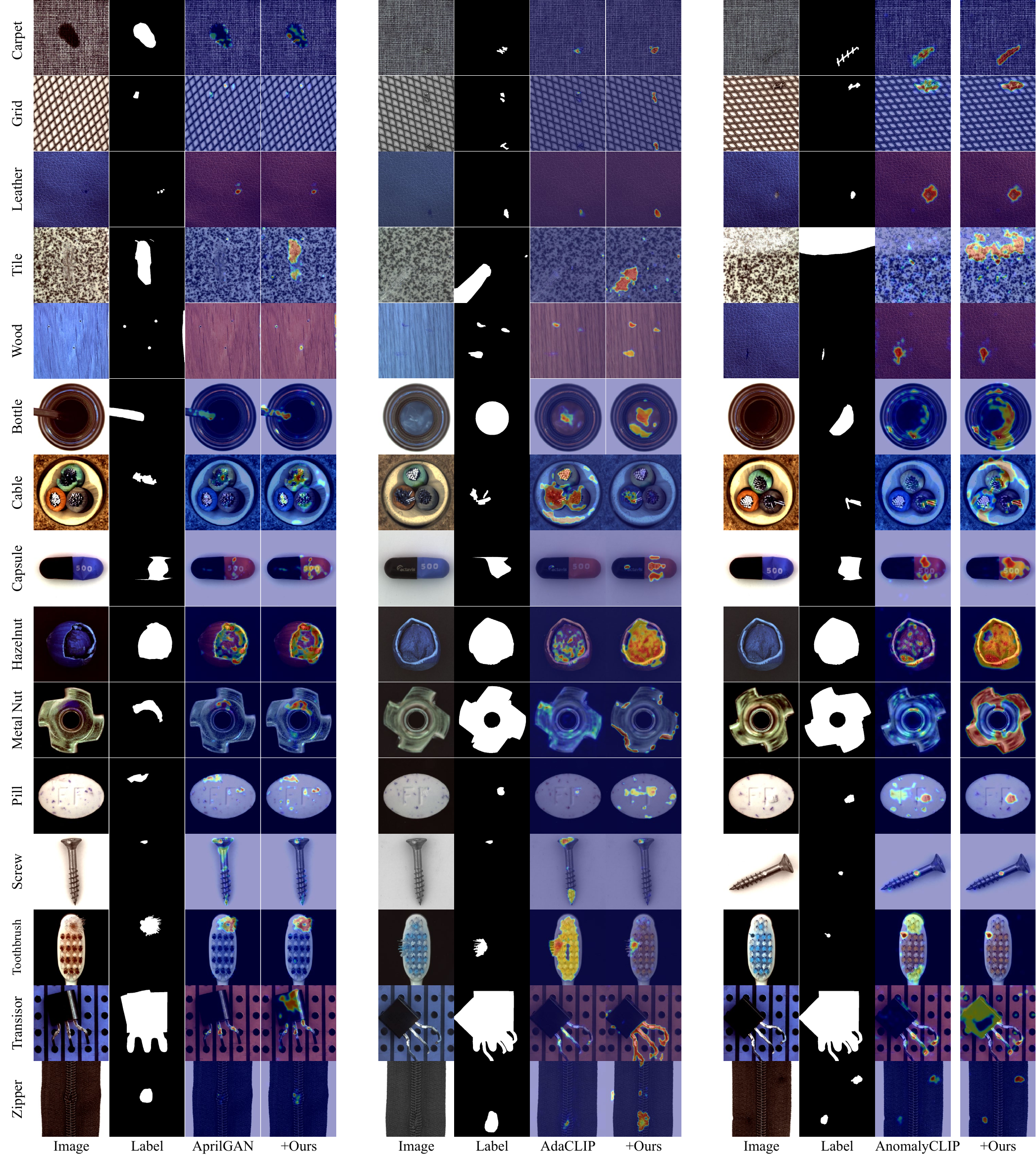}}
\caption{Qualitative examples of \textbf{multimodal RGB-Text UAD} anomaly localization on \textbf{MVTec-AD}~\cite{mvtec}, reporting per-category results. The comparison is divided into three groups, each following the same left-to-right order: input anomaly, ground truth mask, anomaly map predicted by AprilGAN~\cite{aprilgan}, AdaCLIP~\cite{adaclip} or AnomalyCLIP~\cite{anomalyclip}, and the anomaly map obtained with our method integrated. This figure corresponds to Sec.\ref{suppsec6c}, and Tables~\ref{suppmvtectexttable1},~\ref{suppmvtectexttable2}. 
}
\label{suppkeshihuamvtecrgbtext}
\end{center}
\end{figure*}

\begin{figure*}
\setlength{\abovecaptionskip}{2pt}  
\setlength{\belowcaptionskip}{0pt} 
\begin{center}
\centerline{\includegraphics[width=\textwidth]{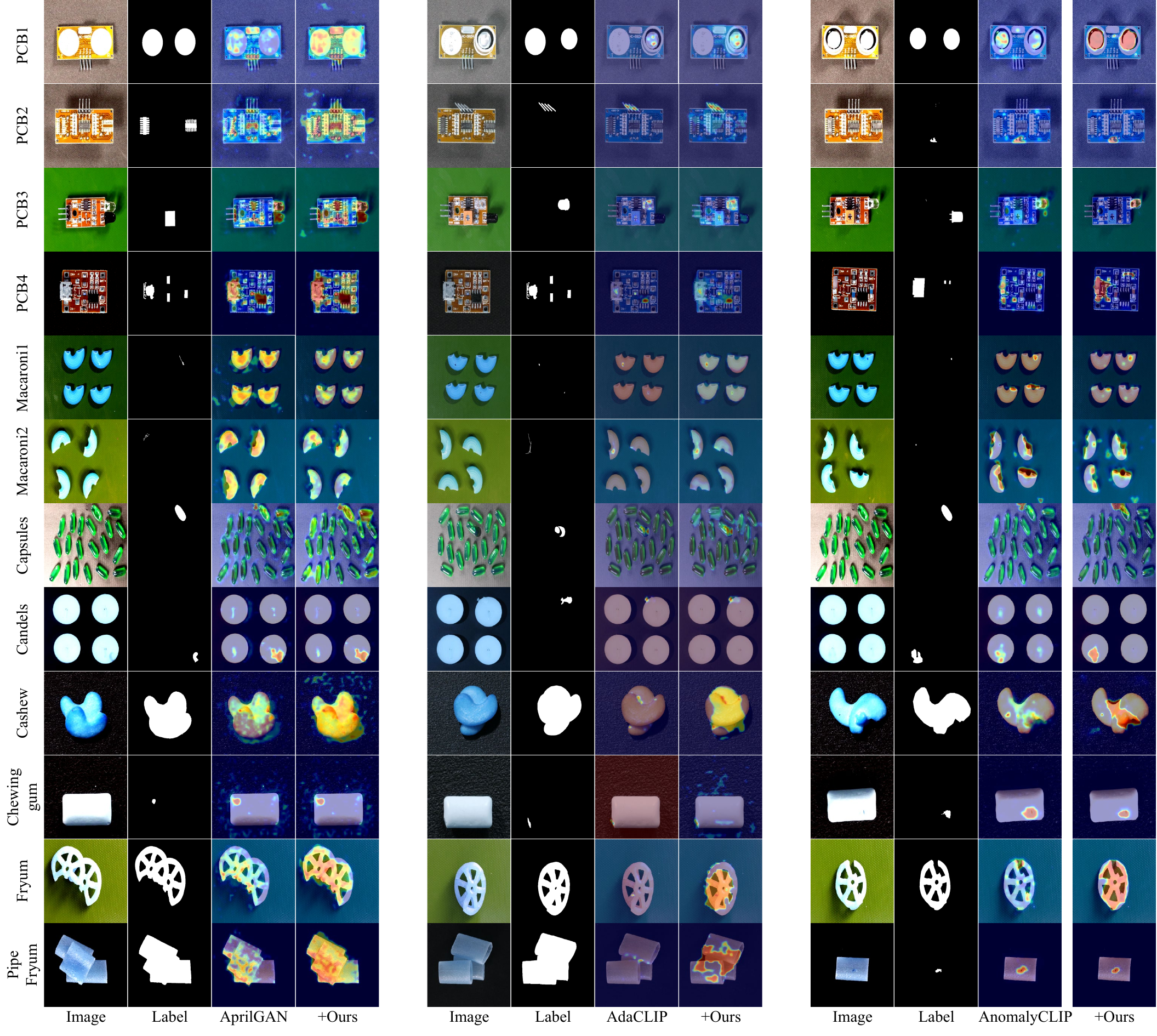}}
\caption{Qualitative examples of \textbf{multimodal RGB-Text UAD} anomaly localization on \textbf{VisA}~\cite{visa}, reporting per-category results. The comparison is divided into three groups, each following the same left-to-right order: input anomaly, ground truth mask, anomaly map predicted by AprilGAN~\cite{aprilgan}, AdaCLIP~\cite{adaclip} or AnomalyCLIP~\cite{anomalyclip}, and the anomaly map obtained with our method integrated. This figure corresponds to Sec.\ref{suppsec6c}, and Tables~\ref{suppvisatexttable1},~\ref{suppvisatexttable2}. 
}
\label{suppkeshihuavisargbtext}
\end{center}
\end{figure*}

\begin{figure*}
\setlength{\abovecaptionskip}{2pt}  
\setlength{\belowcaptionskip}{0pt} 
\begin{center}
\centerline{\includegraphics[width=\textwidth]{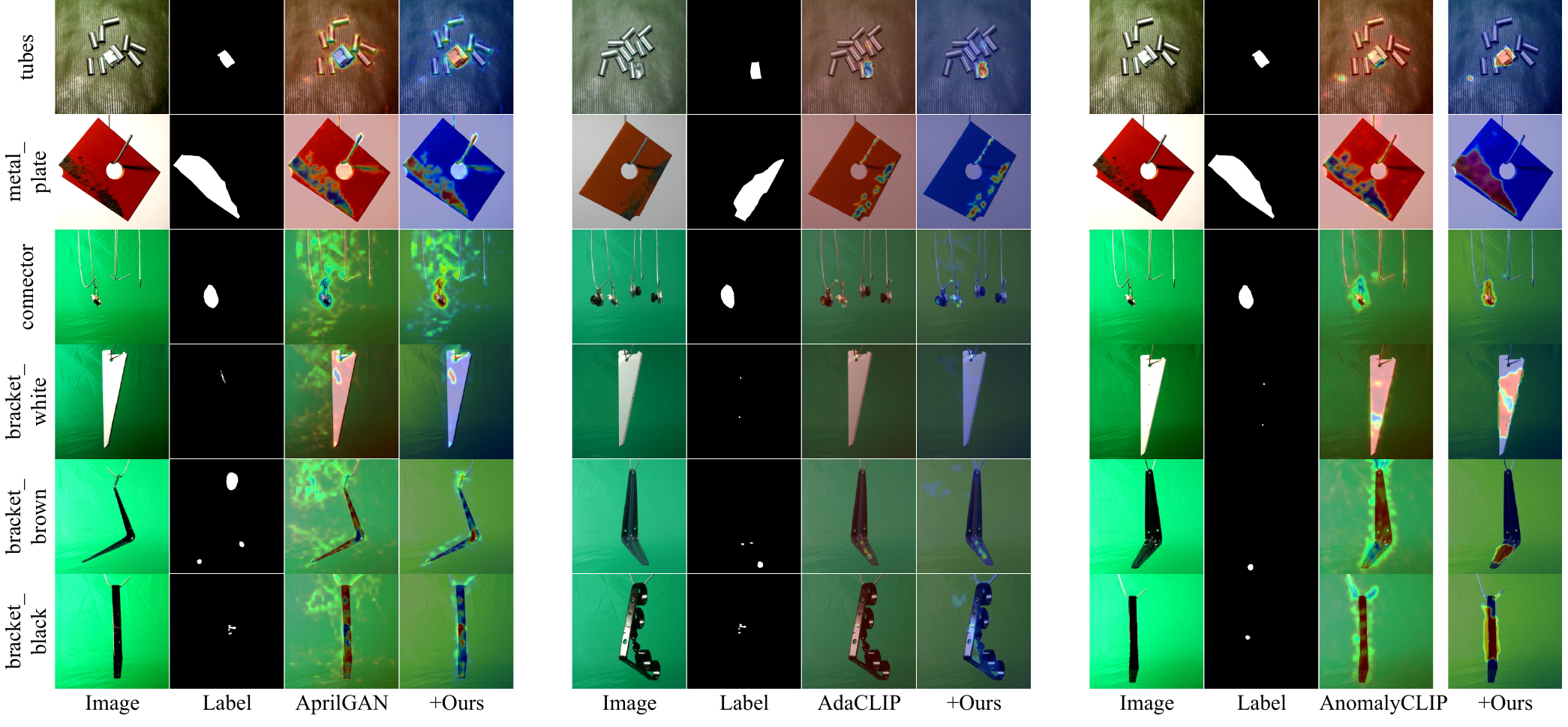}}
\caption{Qualitative examples of \textbf{multimodal RGB-Text UAD} anomaly localization on \textbf{MPDD}~\cite{mpdd}, reporting per-category results. The comparison is divided into three groups, each following the same left-to-right order: input anomaly, ground truth mask, anomaly map predicted by AprilGAN~\cite{aprilgan}, AdaCLIP~\cite{adaclip} or AnomalyCLIP~\cite{anomalyclip}, and the anomaly map obtained with our method integrated. This figure corresponds to Sec.\ref{suppsec6c}, and Tables~\ref{suppmpddtexttable1},~\ref{suppmpddtexttable2}.  
}
\label{suppkeshihuampddrgbtext}
\end{center}
\end{figure*}

\begin{figure*}
\setlength{\abovecaptionskip}{2pt}  
\setlength{\belowcaptionskip}{0pt} 
\begin{center}
\centerline{\includegraphics[width=\textwidth]{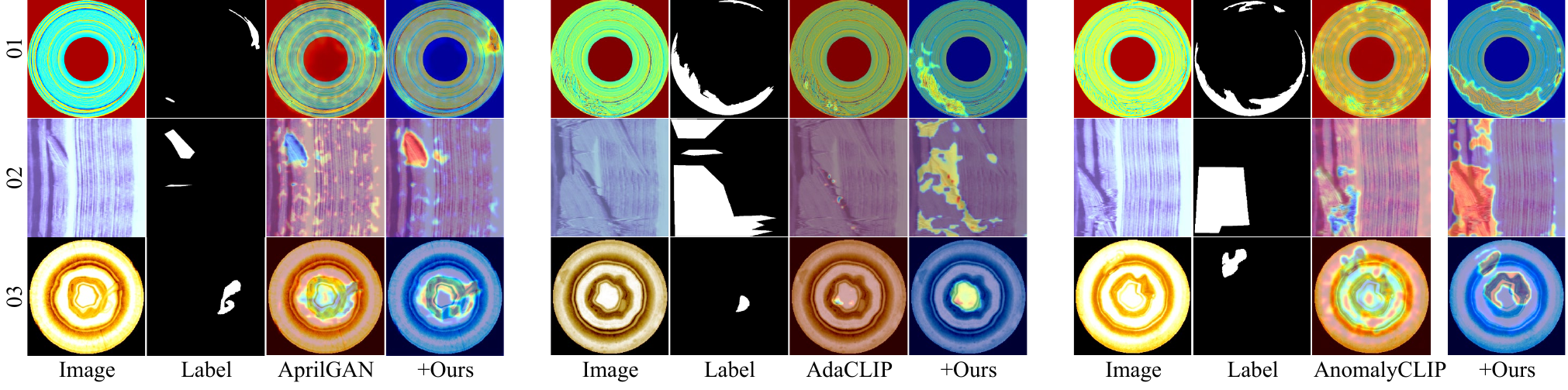}}
\caption{Qualitative examples of \textbf{multimodal RGB-Text UAD} anomaly localization on \textbf{BTAD}~\cite{btad}, reporting per-category results. The comparison is divided into three groups, each following the same left-to-right order: input anomaly, ground truth mask, anomaly map predicted by AprilGAN~\cite{aprilgan}, AdaCLIP~\cite{adaclip} or AnomalyCLIP~\cite{anomalyclip}, and the anomaly map obtained with our method integrated. This figure corresponds to Sec.\ref{suppsec6c}, and Tables~\ref{suppbtadtexttable1},~\ref{suppbtadtexttable2}.  
}
\label{suppkeshihuabtadrgbtext}
\end{center}
\end{figure*}

\begin{figure*}
\setlength{\abovecaptionskip}{2pt}  
\setlength{\belowcaptionskip}{0pt} 
\begin{center}
\centerline{\includegraphics[width=\textwidth]{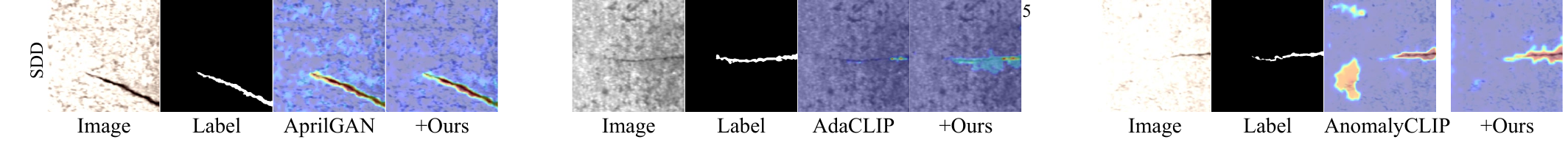}}
\caption{Qualitative examples of \textbf{multimodal RGB-Text UAD} anomaly localization on \textbf{SDD}~\cite{ksdd}, reporting per-category results. The comparison is divided into three groups, each following the same left-to-right order: input anomaly, ground truth mask, anomaly map predicted by AprilGAN~\cite{aprilgan}, AdaCLIP~\cite{adaclip} or AnomalyCLIP~\cite{anomalyclip}, and the anomaly map obtained with our method integrated. This figure corresponds to Sec.\ref{suppsec6c}, and Tables~\ref{suppsddtexttable1},~\ref{suppsddtexttable2}.  
}
\label{suppkeshihuasddrgbtext}
\end{center}
\end{figure*}

\begin{figure*}
\setlength{\abovecaptionskip}{2pt}  
\setlength{\belowcaptionskip}{0pt} 
\begin{center}
\centerline{\includegraphics[width=\textwidth]{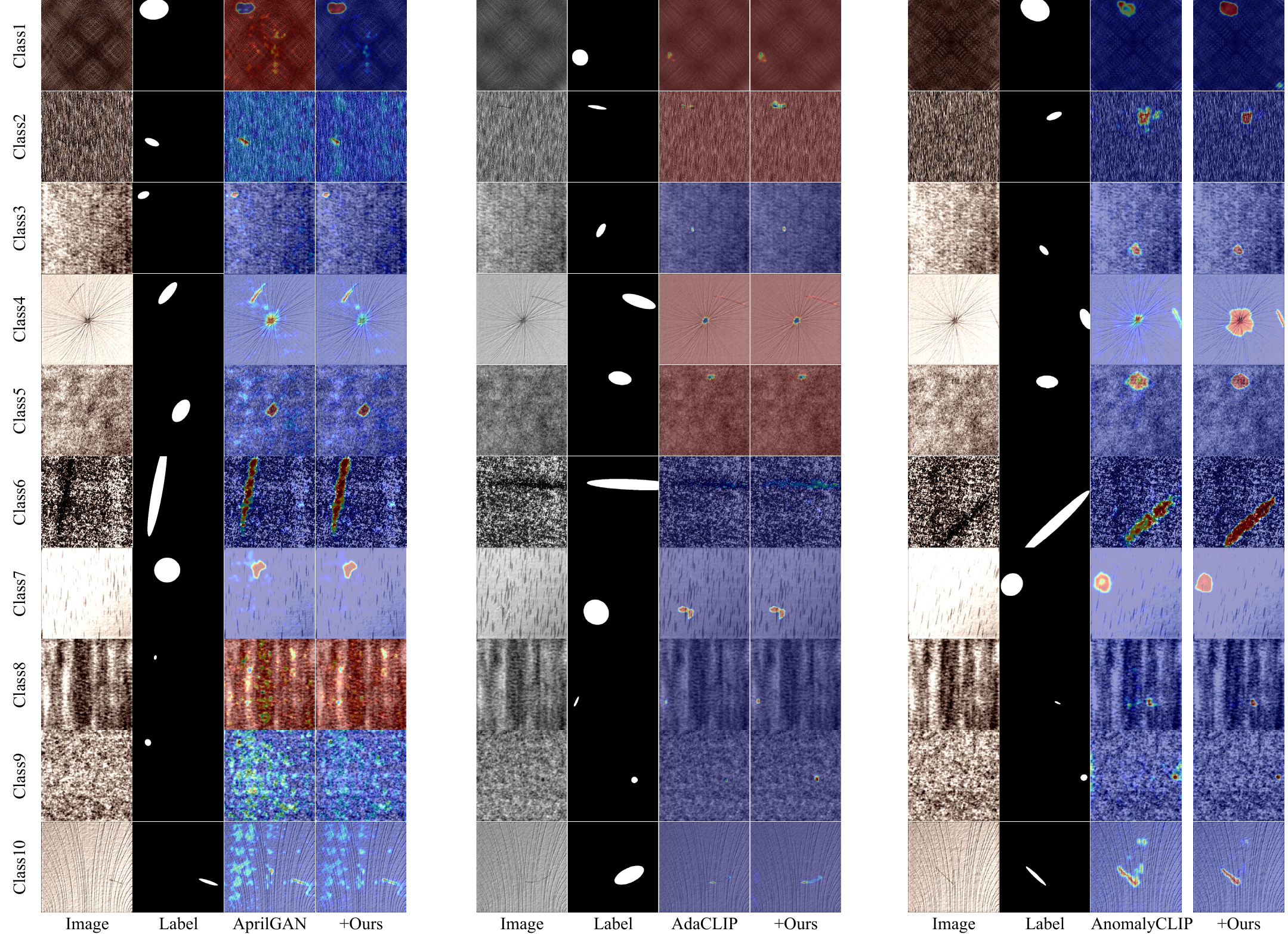}}
\caption{Qualitative examples of \textbf{multimodal RGB-Text UAD} anomaly localization on \textbf{DAGM}~\cite{dagm}, reporting per-category results. The comparison is divided into three groups, each following the same left-to-right order: input anomaly, ground truth mask, anomaly map predicted by AprilGAN~\cite{aprilgan}, AdaCLIP~\cite{adaclip} or AnomalyCLIP~\cite{anomalyclip}, and the anomaly map obtained with our method integrated. This figure corresponds to Sec.\ref{suppsec6c}, and Tables~\ref{suppdagmtexttable1},~\ref{suppdagmtexttable2}.  
}
\label{suppkeshihuadagmrgbtext}
\end{center}
\end{figure*}

\begin{figure*}
\setlength{\abovecaptionskip}{2pt}  
\setlength{\belowcaptionskip}{0pt} 
\begin{center}
\centerline{\includegraphics[width=\textwidth]{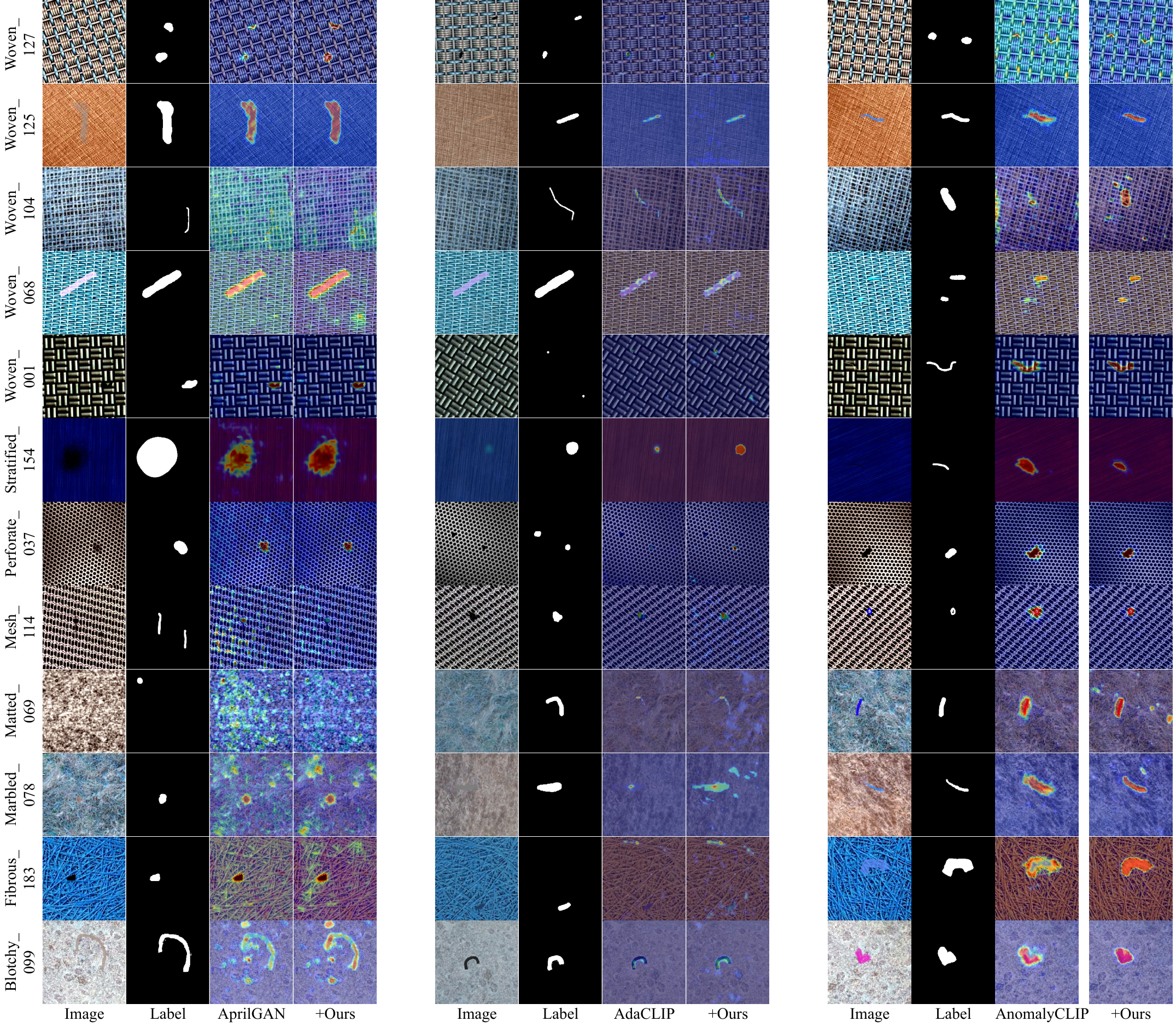}}
\caption{Qualitative examples of \textbf{multimodal RGB-Text UAD} anomaly localization on \textbf{DTD}~\cite{dtd}, reporting per-category results. The comparison is divided into three groups, each following the same left-to-right order: input anomaly, ground truth mask, anomaly map predicted by AprilGAN~\cite{aprilgan}, AdaCLIP~\cite{adaclip} or AnomalyCLIP~\cite{anomalyclip}, and the anomaly map obtained with our method integrated. This figure corresponds to Sec.\ref{suppsec6c}, and Tables~\ref{suppdtdtexttable1},~\ref{suppdtdtexttable2}.  
}
\label{suppkeshihuadtdrgbtext}
\end{center}
\end{figure*}

\begin{figure*}
\setlength{\abovecaptionskip}{2pt}  
\setlength{\belowcaptionskip}{0pt} 
\begin{center}
\centerline{\includegraphics[width=\textwidth]{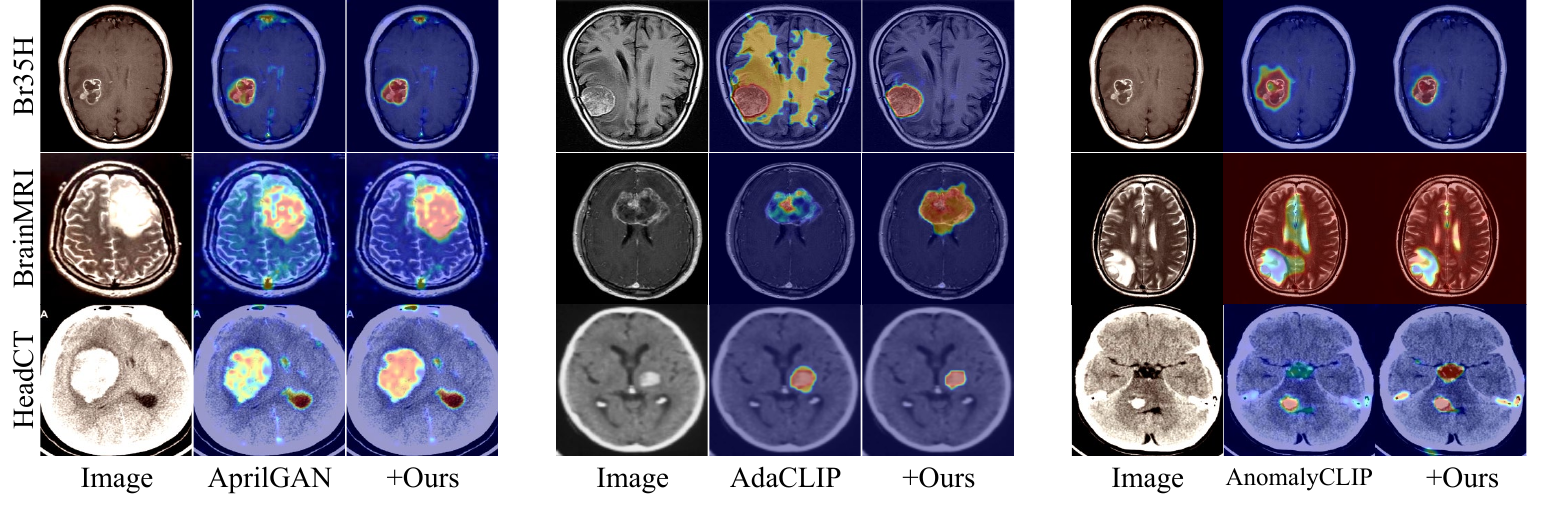}}
\caption{Qualitative examples of \textbf{multimodal RGB-Text UAD} anomaly localization on \textbf{HeatCT, BrainmRI, and Br35h}~\cite{headct,brainmri,br35h}, reporting per-category results. Note that these datasets do not provide pixel-level ground truth masks. The comparison is divided into three groups, each following the same left-to-right order: input anomaly, anomaly map predicted by AprilGAN~\cite{aprilgan}, AdaCLIP~\cite{adaclip} or AnomalyCLIP~\cite{anomalyclip}, and the anomaly map obtained with our method integrated. This figure corresponds to Sec.\ref{suppsec6c}, and Table~\ref{suppmedicaltexttable1}.  
}
\label{suppkeshihuamedical3rgbtext}
\end{center}
\end{figure*}

\begin{figure*}
\setlength{\abovecaptionskip}{2pt}  
\setlength{\belowcaptionskip}{0pt} 
\begin{center}
\centerline{\includegraphics[width=\textwidth]{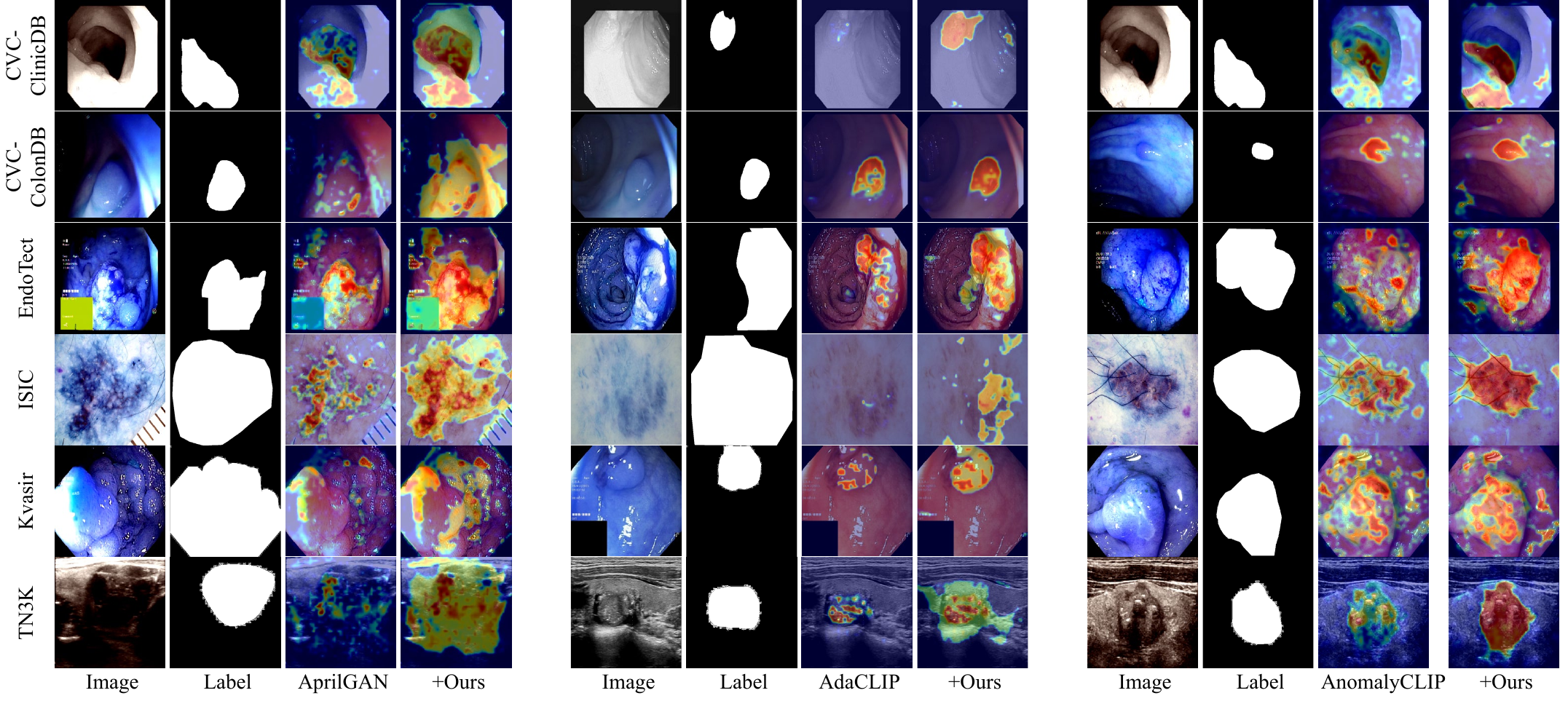}}
\caption{Qualitative examples of \textbf{multimodal RGB-Text UAD} anomaly localization on \textbf{CVC-ClinicDB~\cite{clidb}}, \textbf{CVC-ColonDB~\cite{clodb}}, \textbf{Endo~\cite{endo}}, \textbf{ISIC~\cite{isic}}, \textbf{Kvasir~\cite{kvasir}}, and \textbf{TN3K~\cite{tn3k}}, reporting per-category results. The comparison is divided into three groups, each following the same left-to-right order: input anomaly, anomaly map predicted by AprilGAN~\cite{aprilgan}, AdaCLIP~\cite{adaclip} or AnomalyCLIP~\cite{anomalyclip}, and the anomaly map obtained with our method integrated. This figure corresponds to Sec.\ref{suppsec6c}, and Table~\ref{suppmedicaltexttable2}.  
}
\label{suppkeshihuamedical6rgbtext}
\end{center}
\end{figure*}

We assemble per-category qualitative comparisons for multimodal RGB–Text UAD with comprehensive figures. Industrial datasets are organized as follows. Fig.~\ref{suppkeshihuamvtecrgbtext} presents MVTec-AD~\cite{mvtec}. Figures~\ref{suppkeshihuavisargbtext}, \ref{suppkeshihuampddrgbtext}, \ref{suppkeshihuabtadrgbtext}, \ref{suppkeshihuasddrgbtext}, \ref{suppkeshihuadagmrgbtext}, and \ref{suppkeshihuadtdrgbtext} present VisA~\cite{visa}, MPDD~\cite{mpdd}, BTAD~\cite{btad}, SDD~\cite{ksdd}, DAGM~\cite{dagm}, and DTD~\cite{dtd}. Medical visualizations appear in Fig.~\ref{suppkeshihuamedical3rgbtext} for HeadCT, BrainMRI, and Br35H~\cite{headct,brainmri,br35h}, and Fig.~\ref{suppkeshihuamedical6rgbtext} for ISIC, CVC-ColonDB, CVC-ClinicDB, Kvasir, Endo, and TN3K~\cite{isic,clodb,clidb,kvasir,endo,tn3k}. Within each panel, we follow the same left-to-right order with the input image, the ground truth mask when available, the anomaly maps from AprilGAN~\cite{aprilgan}, AnomalyCLIP~\cite{anomalyclip}, and AdaCLIP~\cite{adaclip}, and the anomaly map generated by our method.

Across industrial and medical datasets, our anomaly localization results reduce language–vision matching noise and background clutter, concentrate responses within defect regions, and improve boundary adherence. Relative to AprilGAN, AdaCLIP, and AnomalyCLIP, our integration produces sharper and more compact maps and remains stable across categories. The qualitative evidence mirrors the gains on I-AUROC, I-AP, P-AUROC, P-AP, I-F1-max, P-F1-max, and AUPRO reported in the corresponding tables and indicates that the proposed cost volume filtering transfers across datasets and domains.

\begin{table*}[htbp]
  \centering
  \caption{Quantitative comparison for \textbf{unimodal RGB UAD}, reporting per-category results. Multi-class anomaly detection/localization results on \textbf{MVTec-AD}~\cite{mvtec} using \textbf{I-AUROC/P-AUROC}. \textbf{Best} results are in bold and \underline{runners-up} are underlined. This table corresponds to Sec.\ref{suppsec5a} and complements Table~I and Table~IV in the main paper.
}
  \label{suppmvtectable1}%
  \renewcommand{\arraystretch}{1.4}
  \resizebox{1\linewidth}{!}{
    \begin{tabular}{cc|c|>{\columncolor{lightblue}}c|c|>{\columncolor{lightblue}}c|c|>{\columncolor{lightblue}}c|c|>{\columncolor{lightblue}}c|c|>{\columncolor{lightblue}}c}
    \toprule
    \multicolumn{2}{c}{Method~$\rightarrow$} & UniAD & UniAD+UCF & GLAD & GLAD+UCF & HVQ-Trans & HVQ-Trans+UCF & AnomalDF & AnomalDF+UCF & Dinomaly & Dinomaly+UCF \\
    \cline{1-2}
    \multicolumn{2}{c}{Category~$\downarrow$} & NeurIPS'22 & Ours & ECCV'24 & Ours & NeurIPS'23 & Ours & WACV'25 & Ours & CVPR'25 & Ours\\
    \hline
    \multicolumn{1}{c}{\multirow{10}[1]{*}{\begin{turn}{90}Objects\end{turn}}} 
    & Bottle & 99.7 / 98.0 & \textbf{100.0} / 98.0 &  \textbf{100.0} / 98.4 & \underline{99.8} / 97.8 & \textbf{100.0} / 98.3 & \textbf{100.0} / 98.8 & \textbf{100.0} / 99.3 & \textbf{100.0}/ \textbf{99.1} & \textbf{100.0} / 99.1 & \textbf{100.0} / \underline{99.4}\\
    & Cable & 95.2 / 97.4 & 99.2 / 97.2 & 98.7 / 93.4 & 98.0 / 96.3 & 99.0 / 98.1 & \underline{99.8} / 98.2 & 99.6 / \underline{98.3} & 99.3 / 98.2 & \textbf{100.0} / 98.2 & \textbf{100.0} / \textbf{98.7} \\
    & Capsule & 93.4 / 98.7 & 96.3 / 98.7 & 96.5 / \underline{99.1} & 94.3 / \textbf{99.2} & 95.4 / 98.8 & 96.4 / 98.9 & 89.7 / \underline{99.1} & 96.1 / \textbf{99.2} & \underline{97.9} / 98.7 & \textbf{98.2} / 98.8\\
    & Hazelnut & \textbf{100.0} / 98.1 & \textbf{100.0} / 98.5 & 97.0 / 98.9 & 99.4 / 99.1 & \textbf{100.0} / 98.8 & \textbf{100.0} / 99.2 & \underline{99.9} / \textbf{99.6} & \textbf{100.0} / \underline{99.5} & \textbf{100.0} / 99.4 & \textbf{100.0} / \textbf{99.6}\\
    & Metal Nut & 99.5 / 93.7 & 99.6 / 94.6 & \underline{99.9} / 97.3 & \textbf{100.0} / \textbf{99.2} & \underline{99.9} / 96.3 & \textbf{100.0} / 97.9 & \textbf{100.0} / 96.7 & \textbf{100.0} / \underline{99.0} & \textbf{100.0} / 97.0 & \textbf{100.0} / 98.2\\
    & Pill & 94.8 / 96.2 & 96.8 / 97.1 & 94.4 / 97.9 & 97.9 / 97.8 & 95.8 / 97.1 & 96.9 / 96.5 & 97.2 / 98.1 & \underline{98.9} / \textbf{98.4} & \textbf{99.2} / 97.8 & \textbf{99.2} / \underline{98.1}\\
    & Screw & 91.7 / 98.8 & 95.1 / 98.7 & 93.4 / \underline{99.6} & 95.4 / \underline{99.6} & 95.6 / 98.9 & 95.3 / 99.0 & 74.3 / 97.6 & 88.5 / 99.0 & \underline{98.4} / \underline{99.6} & \textbf{98.6} / \textbf{99.7}\\
    & Toothbrush & 92.8 / 98.4 & 98.9 / 98.9 & \underline{99.7} / \textbf{99.2} & \underline{99.7} / \underline{99.1} & 93.6 / 98.6 & \textbf{100.0} / 98.9 & \underline{99.7} / \textbf{99.2} & \underline{99.7} / \textbf{99.2} & \textbf{100.0} / 98.9 & \textbf{100.0} / \textbf{99.2} \\
    & Transistor & 99.5 / \underline{98.0} & \underline{99.8} / \underline{98.0} & 99.4 / 90.9 & 99.5 / 91.6 & 99.7 / \textbf{99.1} & 99.7 / \textbf{99.2} & 96.5 / 95.8 & 97.8 / 97.5 & 99.1 / 93.2 & 99.1 / 93.2\\
    & Zipper & 98.2 / 97.7 & \underline{99.9} / 97.7 & 96.4 / 93.0 & 99.2 / 97.7 & 97.9 / 97.5 & 98.9 / 98.3 & 98.8 / 94.3 & 98.9 / 96.7 & \textbf{100.0} / \underline{99.4} & \textbf{100.0} / \textbf{99.3}\\
    \hline
    \multicolumn{1}{c}{\multirow{5}[1]{*}{\begin{turn}{90}Textures\end{turn}}} 
    & Carpet & 99.8 / 98.4 & \underline{99.9} / 98.4 & 97.2 / 98.9 & \textbf{100.0} / 99.1 & \underline{99.9} / 98.7 & \textbf{100.0} / 98.5 & \underline{99.9} / 99.4 & \underline{99.9} / \textbf{99.6} & 99.8 / 99.3 & \underline{99.9} / \underline{99.5}\\
    & Grid & 98.7 / 97.3 & \underline{99.9} / 98.7 & 95.1 / 98.2 & \textbf{100.0} / \textbf{99.5} & 97.0 / 97.0 & 99.3 / 98.3 & 98.7 / 97.8 & \textbf{100.0} / \textbf{99.5} & 99.7 / \underline{99.4} & 99.8 / \textbf{99.5}\\
    & Leather & \textbf{100.0} / 98.7 & \textbf{100.0} / 99.4 & \underline{99.5} / 99.7 & \textbf{100.0} / \underline{99.6} & \textbf{100.0} / 98.8 & \textbf{100.0} / 99.3 & \textbf{100.0} / \textbf{99.7} & \textbf{100.0} / \textbf{99.7} & \textbf{100.0} / 99.3 & \textbf{100.0} / 99.5\\
    & Tile & \underline{99.5} / 91.8 & \textbf{100.0} / 95.3 & \textbf{100.0} / 97.8 & \textbf{100.0} / \underline{99.4} & 99.2 / 92.2 & \textbf{100.0} / 95.0 & \textbf{100.0} / 98.5 & \textbf{100.0} / \textbf{99.6} & \textbf{100.0} / 98.1 & \textbf{100.0} / 99.0\\
    & Wood & 98.5 / 93.1 & 98.9 / 94.0 & 95.4 / 96.8 & 97.4 / 97.4 & 97.2 / 92.4 & 98.5 / 94.3 & 97.9 / \underline{97.6} & \underline{98.9} / \textbf{98.2} & \textbf{99.9} / 97.6 & \textbf{99.9} / 97.7\\
    \hline
    \multicolumn{2}{c}{Mean} & 97.5 / 96.9 & 99.0 / 97.5 & 97.5 / 97.3 & 98.7 / 98.2 & 98.0 / 97.3 & 99.0 / 98.0 & 96.8 / 98.1 & 98.5 / \textbf{98.8} & \underline{99.6} / 98.3 & \textbf{99.7}/ \underline{98.7}\\
    \bottomrule
    \end{tabular}%
  }
\end{table*}

\section{Per-Class KDE Analysis of Logits}
\phantomsection
\label{suppsec7}
\subsection{RGB UAD: KDE Analysis of Logits}
\phantomsection
\label{suppsec7a}

\begin{figure*}
\setlength{\abovecaptionskip}{2pt}  
\setlength{\belowcaptionskip}{0pt} 
\begin{center}
\centerline{\includegraphics[width=0.85\textwidth]{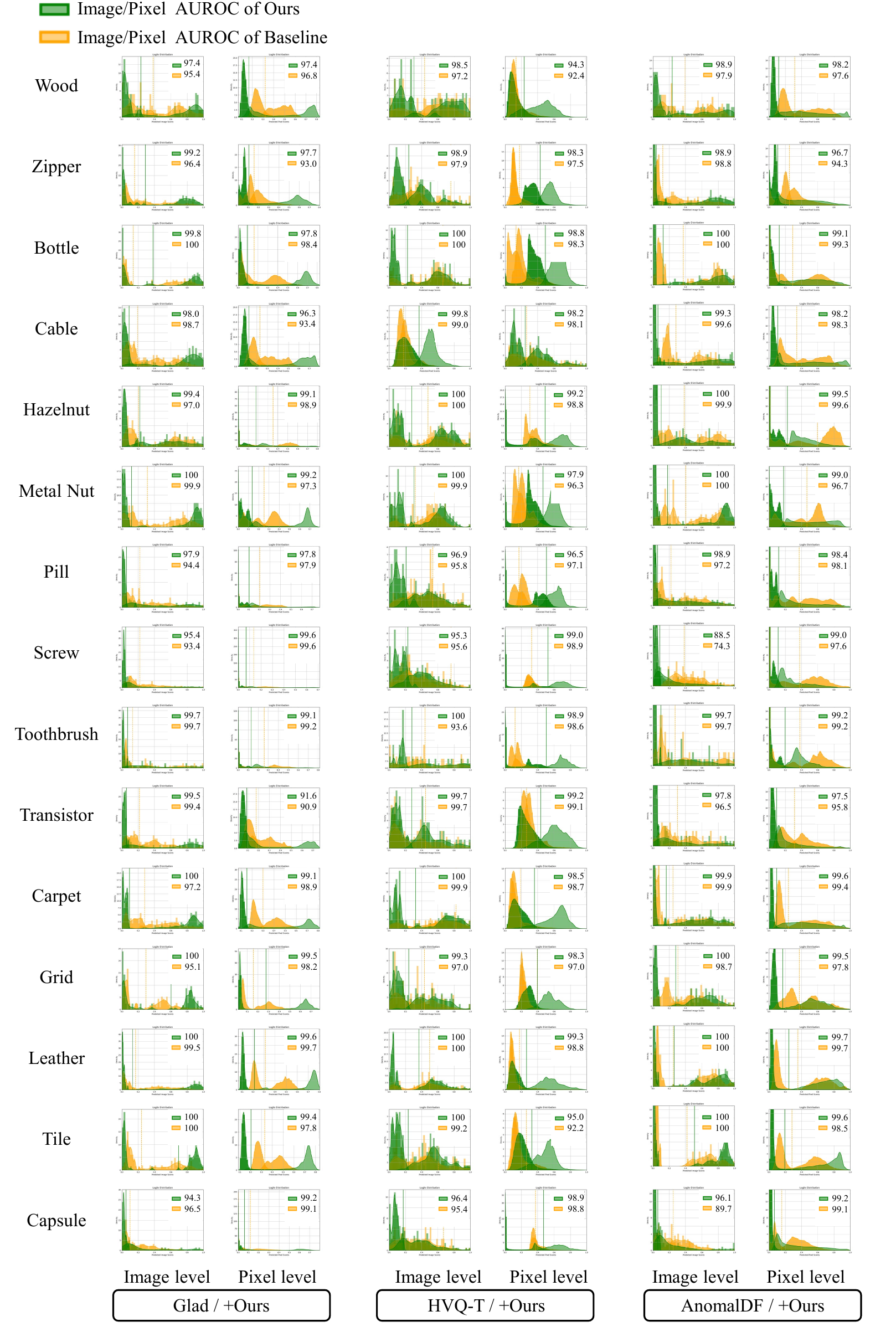}}
\caption{Quantitative comparison on \textbf{unimodal RGB UAD} for \textbf{MVTec-AD}~\cite{mvtec} using KDE curves of image- and pixel-level anomaly logits. Each two-column pair (from left to right) compares GLAD~\cite{glad}, HVQ-Trans~\cite{hvqtrans}, and AnomalDF~\cite{anomalydino} with our method, where the first and second columns show image- and pixel-level APROC, respectively. This figure corresponds to Sec.\ref{suppsec7a}, Tables~\ref{suppmvtectable1}, \ref{suppmvtectable2}, \ref{suppmvtectable3}, \ref{suppmvtectable4}, and Fig.~\ref{suppkeshihuamvtecrgb}.}
\label{suppkdemvtecrgb}
\end{center}
\end{figure*}

\begin{figure*}
\setlength{\abovecaptionskip}{2pt}  
\setlength{\belowcaptionskip}{0pt} 
\begin{center}
\centerline{\includegraphics[width=0.85\textwidth]{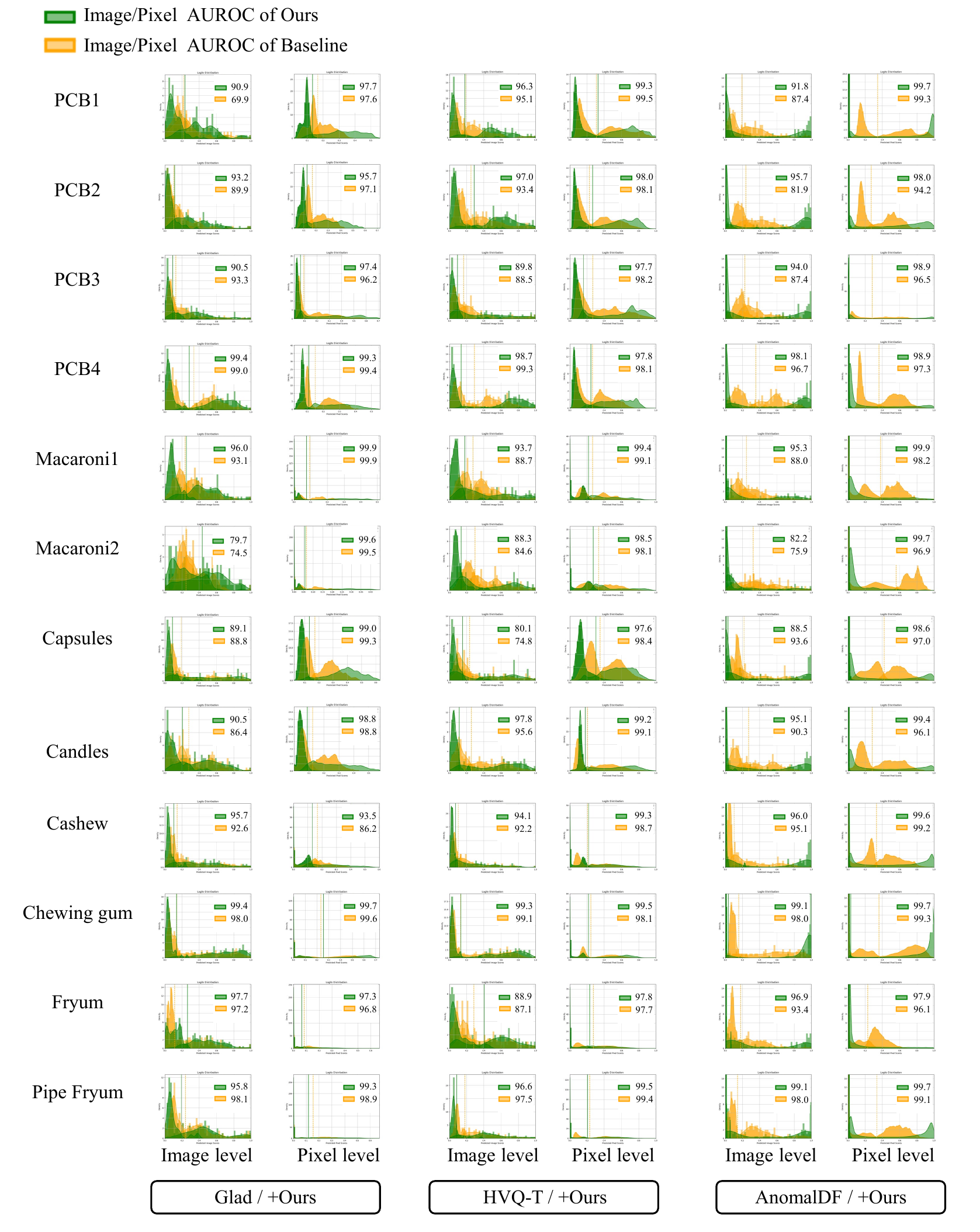}}
\caption{Quantitative comparison on \textbf{unimodal RGB UAD} for \textbf{VisA}~\cite{visa} using KDE curves of image- and pixel-level anomaly logits. Each two-column pair (from left to right) compares GLAD~\cite{glad}, HVQ-Trans~\cite{hvqtrans}, and AnomalDF~\cite{anomalydino} with our method, where the first and second columns show image- and pixel-level APROC, respectively. This figure corresponds to Sec.\ref{suppsec7a}, Tables~\ref{suppvisatable1}, \ref{suppvisatable2}, \ref{suppvisatable3}, \ref{suppvisatable4}, and Fig.~\ref{suppkeshihuavisargb}.}
\label{suppkdevisargb}
\end{center}
\end{figure*}

We present category-wise KDE analyses~\cite{kde} for unimodal RGB UAD. For MVTec-AD~\cite{mvtec}, Fig.~\ref{suppkdemvtecrgb} plots the KDEs of image-level and pixel-level anomaly logits; each two-column group compares GLAD~\cite{glad}, HVQ-Trans~\cite{hvqtrans}, and AnomalDF~\cite{anomalydino} with our integration. VisA~\cite{visa} is shown in Fig.~\ref{suppkdevisargb} with the same layout.

Across both datasets, our curves show larger separation between normal and abnormal modes with reduced overlap at both levels, indicating stronger separability. The distributions exhibit attenuated matching noise and cleaner decision margins, consistent with gains in image- and pixel-level metrics.

\subsection{RGB-3D UAD: KDE Analysis of Logits}
\phantomsection
\label{suppsec7b}

\begin{figure*}
\setlength{\abovecaptionskip}{2pt}  
\setlength{\belowcaptionskip}{0pt} 
\begin{center}
\centerline{\includegraphics[width=0.65\textwidth]{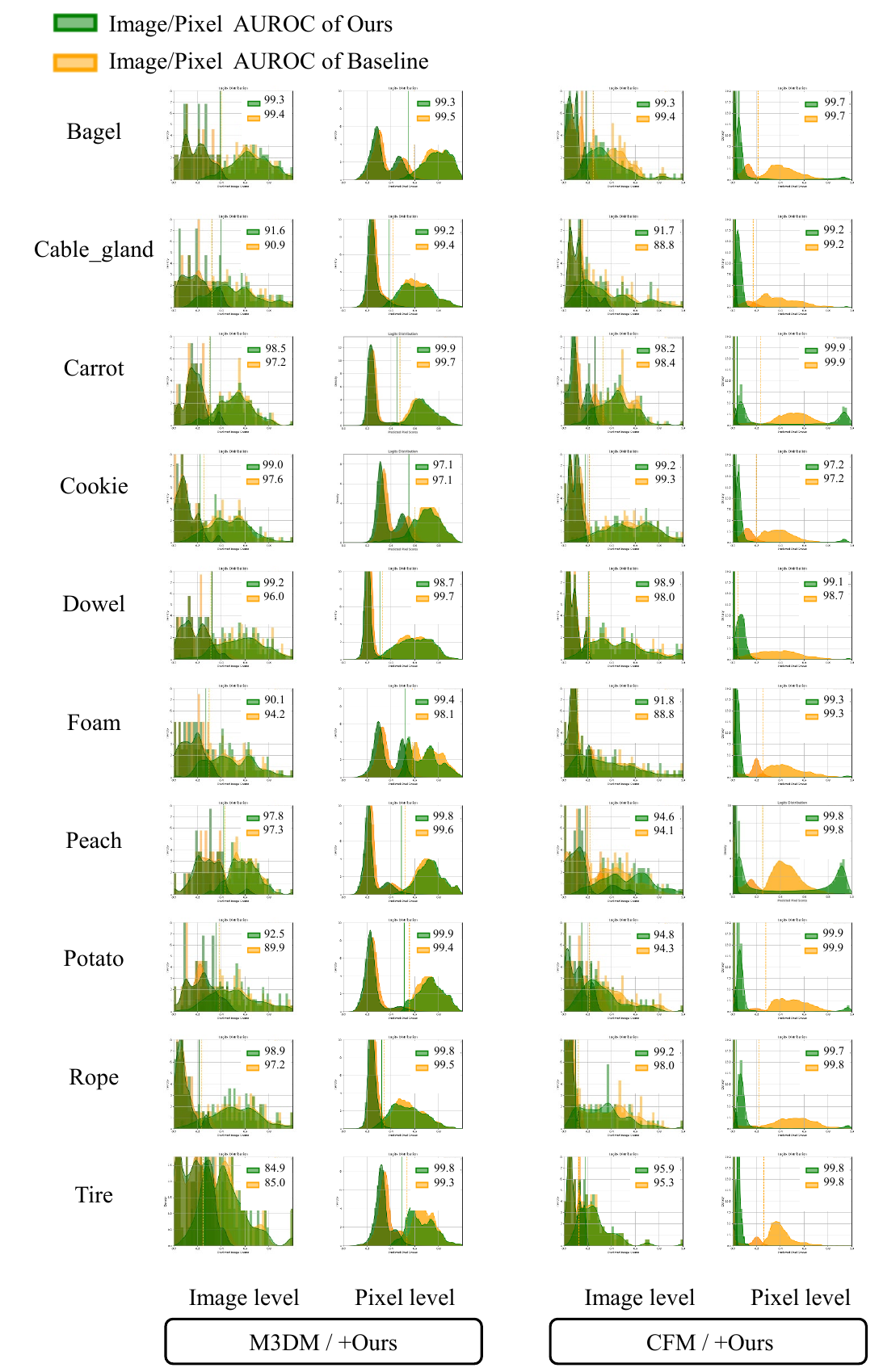}}
\caption{Quantitative comparison on \textbf{multimodal RGB-3D UAD} for \textbf{MVTec 3D-AD}~\cite{mvtec3d} using KDE curves of image- and pixel-level anomaly logits. Each two-column pair (from left to right) compares M3DM~\cite{m3dm} and CFM~\cite{cfm} with our method, where the first and second columns show image- and pixel-level APROC, respectively. This figure corresponds to Sec.\ref{suppsec7b}, Fig.~\ref{suppkeshihuamvtec3drgb3d} in this material, and Table~V in the main paper.}
\label{suppkdemvtec3d}
\end{center}
\end{figure*}

\begin{figure*}
\setlength{\abovecaptionskip}{2pt}  
\setlength{\belowcaptionskip}{0pt} 
\begin{center}
\centerline{\includegraphics[width=0.65\textwidth]{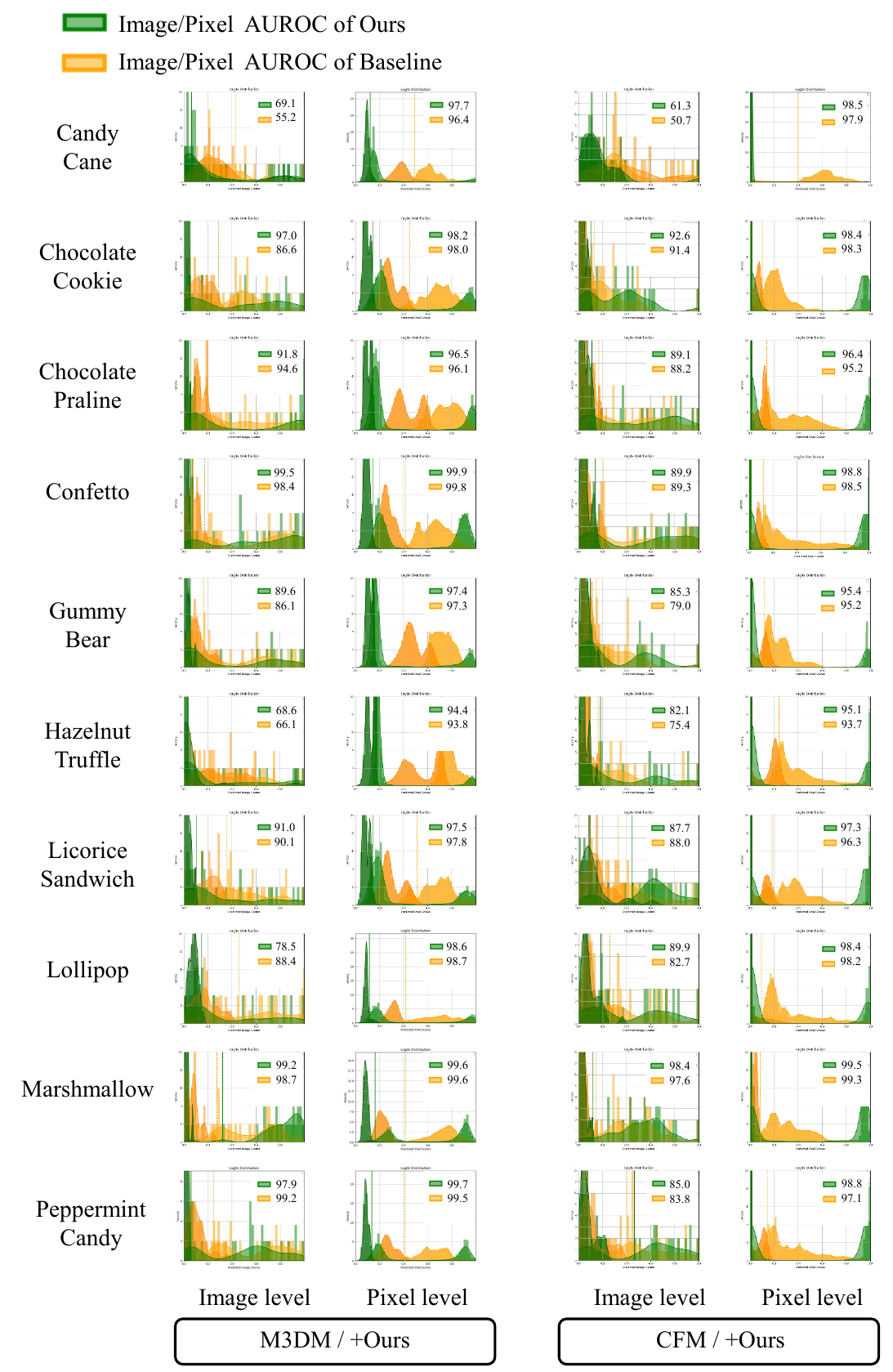}}
\caption{Quantitative comparison on \textbf{multimodal RGB-3D UAD} for \textbf{Eyecandies}~\cite{eyecan} using KDE curves of image- and pixel-level anomaly logits. Each two-column pair (from left to right) compares M3DM~\cite{m3dm} and CFM~\cite{cfm} with our method, where the first and second columns show image- and pixel-level APROC, respectively. This figure corresponds to Sec.\ref{suppsec7b}, Fig.~\ref{suppkeshihuaeyecanrgb3d} in this material, and Tables~VI in the main paper.}
\label{suppkdeeye}
\end{center}
\end{figure*}

We present category-wise KDE analyses~\cite{kde} for multimodal RGB–3D UAD. For MVTec 3D-AD~\cite{mvtec3d}, Fig.~\ref{suppkdemvtec3d} shows KDE curves of image-level anomaly-detection logits and pixel-level localization logits, comparing M3DM~\cite{m3dm} and CFM~\cite{cfm} with and without our integration; corresponding results for Eyecandies~\cite{eyecan} appear in Fig.~\ref{suppkdeeye}.

Our method produces wider inter-class margins and smaller overlaps, reflecting reduced cross-modal matching noise and more decisive boundaries. These trends are consistent with the quantitative improvements summarized in the tables and with the qualitative anomaly-heatmap visualizations.

\subsection{RGB-Text UAD: KDE Analysis of Logits}
\phantomsection
\label{suppsec7c}

\begin{figure*}
\setlength{\abovecaptionskip}{2pt}  
\setlength{\belowcaptionskip}{0pt} 
\begin{center}
\centerline{\includegraphics[width=0.85\textwidth]{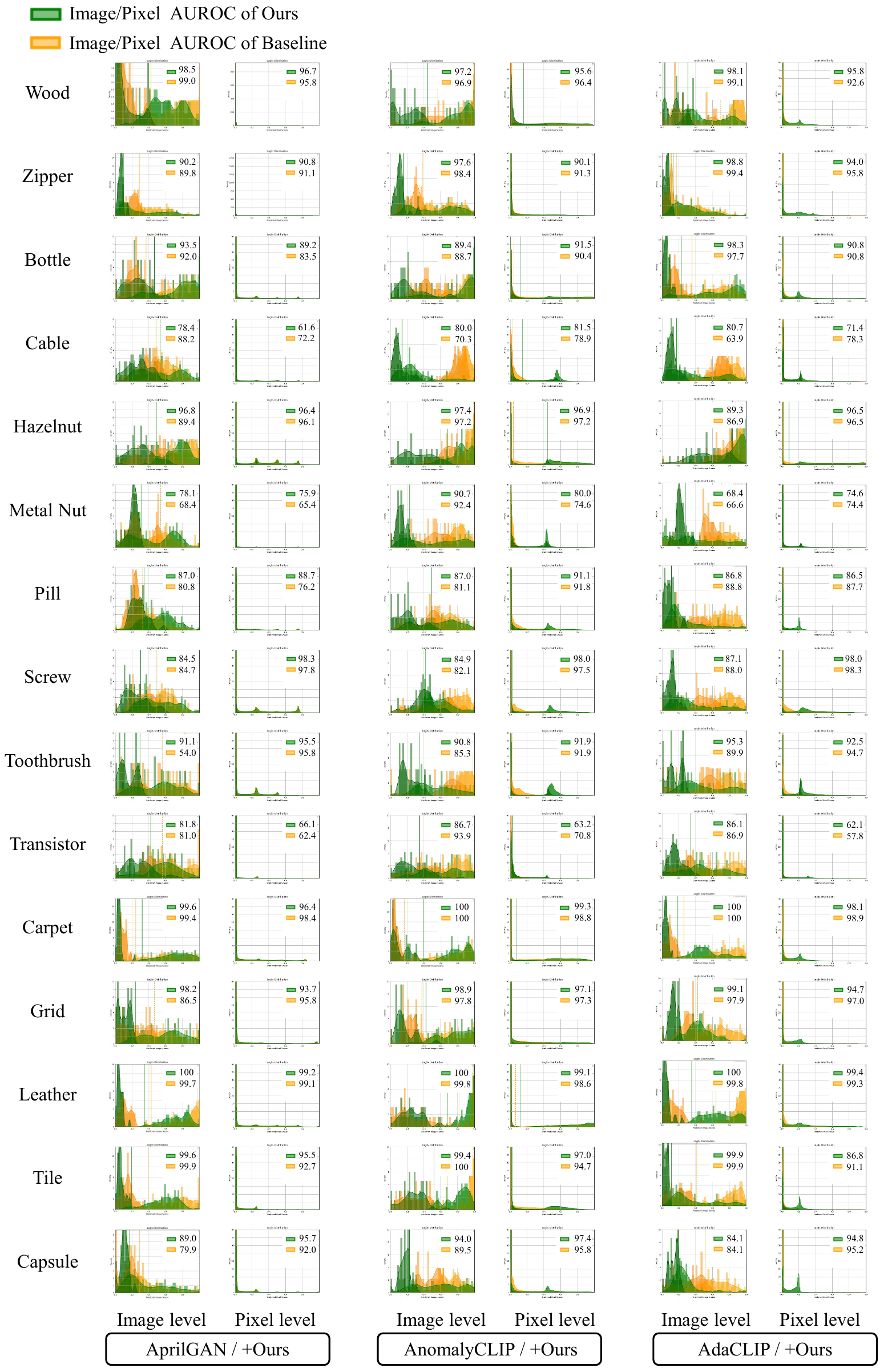}}
\caption{Quantitative comparison on \textbf{multimodal RGB-TEXT UAD} for \textbf{MVTec-AD}~\cite{mvtec} using KDE curves of image- and pixel-level anomaly logits. Each two-column pair (from left to right) compares AprilGAN~\cite{aprilgan}, AnomalyCLIP~\cite{anomalyclip}, and AdaCLIP~\cite{adaclip} with our method, where the first and second columns show image- and pixel-level APROC, respectively. This figure corresponds to Sec.\ref{suppsec7c}, Tables~\ref{suppmvtectexttable1}, \ref{suppmvtectexttable2}, and Fig.~\ref{suppkeshihuamvtecrgbtext}.}
\label{suppkdemvtectext}
\end{center}
\end{figure*}

\begin{figure*}
\setlength{\abovecaptionskip}{2pt}  
\setlength{\belowcaptionskip}{0pt} 
\begin{center}
\centerline{\includegraphics[width=0.85\textwidth]{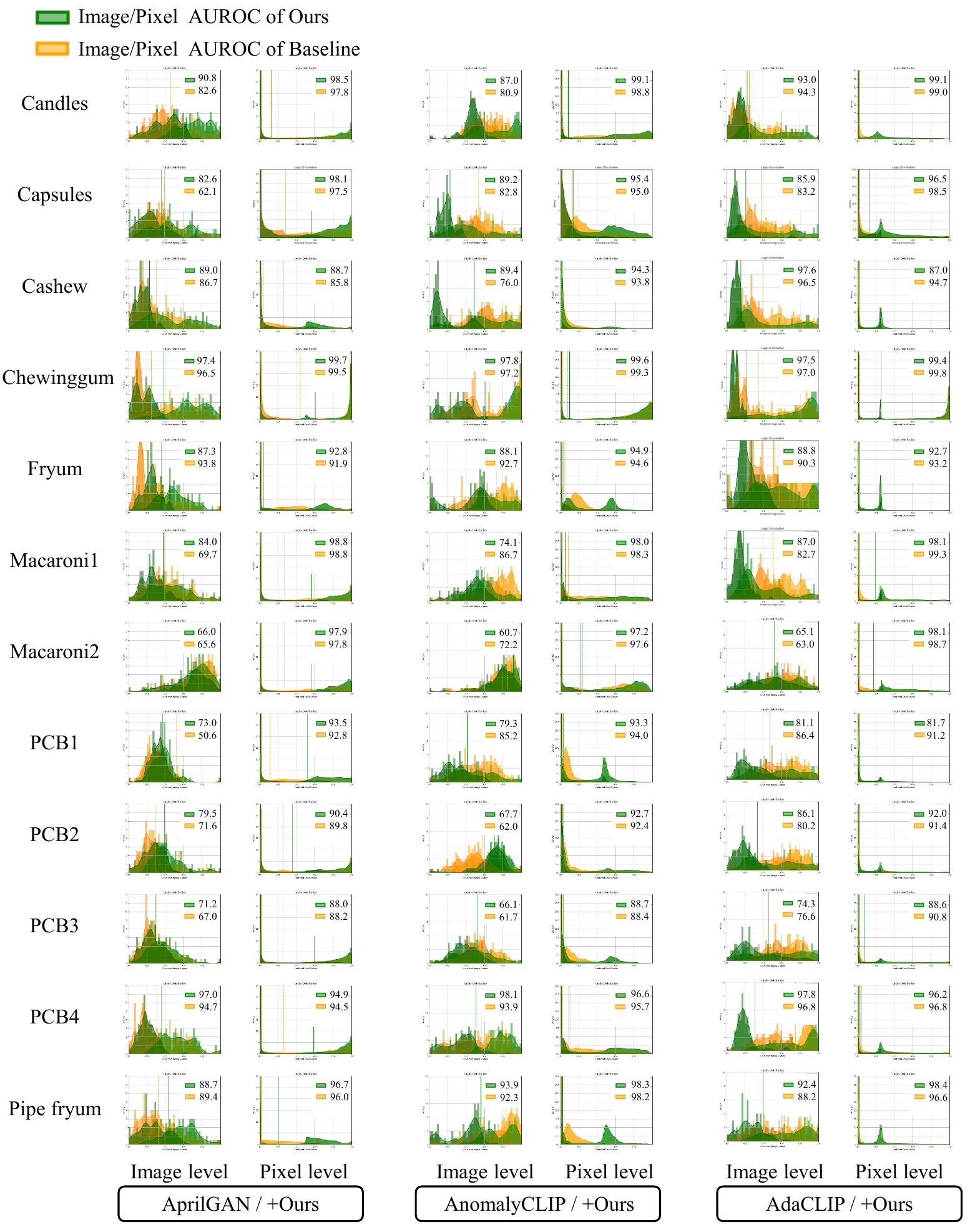}}
\caption{Quantitative comparison on \textbf{multimodal RGB-TEXT UAD} for \textbf{VisA}~\cite{visa} using KDE curves of image- and pixel-level anomaly logits. Each two-column pair (from left to right) compares AprilGAN~\cite{aprilgan}, AnomalyCLIP~\cite{anomalyclip}, and AdaCLIP~\cite{adaclip} with our method, where the first and second columns show image- and pixel-level APROC, respectively. This figure corresponds to Sec.\ref{suppsec7c}, Tables~\ref{suppvisatexttable1}, \ref{suppvisatexttable2}, and Fig.~\ref{suppkeshihuavisargbtext}.}
\label{suppkdevisatext}
\end{center}
\end{figure*}

\begin{figure*}
\setlength{\abovecaptionskip}{2pt}  
\setlength{\belowcaptionskip}{0pt} 
\begin{center}
\centerline{\includegraphics[width=0.85\textwidth]{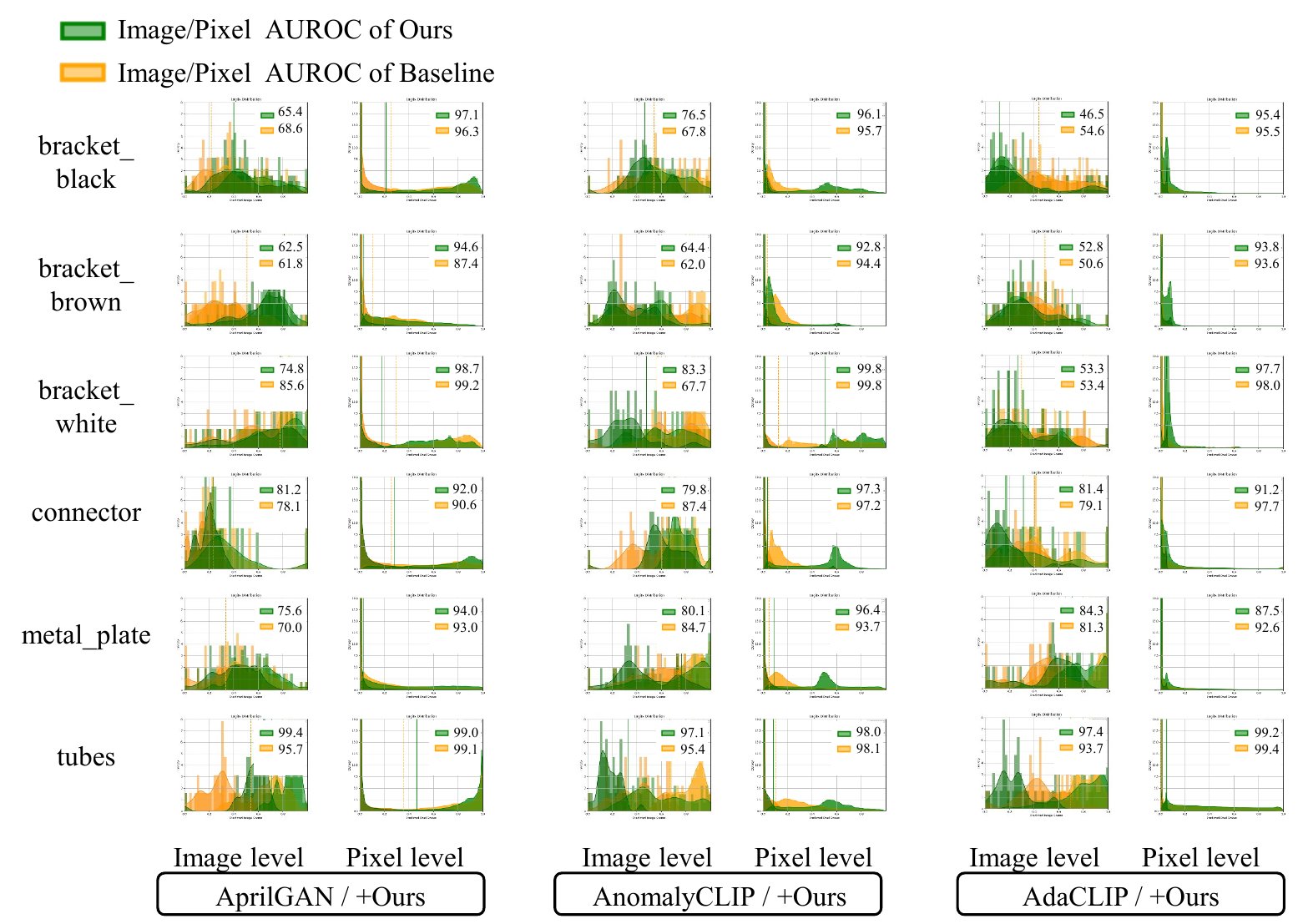}}
\caption{Quantitative comparison on \textbf{multimodal RGB-TEXT UAD} for \textbf{MPDD}~\cite{mpdd} using KDE curves of image- and pixel-level anomaly logits. Each two-column pair (from left to right) compares AprilGAN~\cite{aprilgan}, AnomalyCLIP~\cite{anomalyclip}, and AdaCLIP~\cite{adaclip} with our method, where the first and second columns show image- and pixel-level APROC, respectively. This figure corresponds to Sec.\ref{suppsec7c}, Tables~\ref{suppmpddtexttable1}, \ref{suppmpddtexttable2}, and Fig.~\ref{suppkeshihuampddrgbtext}.}
\label{suppkdempddtext}
\end{center}
\end{figure*}

\begin{figure*}
\setlength{\abovecaptionskip}{2pt}  
\setlength{\belowcaptionskip}{0pt} 
\begin{center}
\centerline{\includegraphics[width=0.85\textwidth]{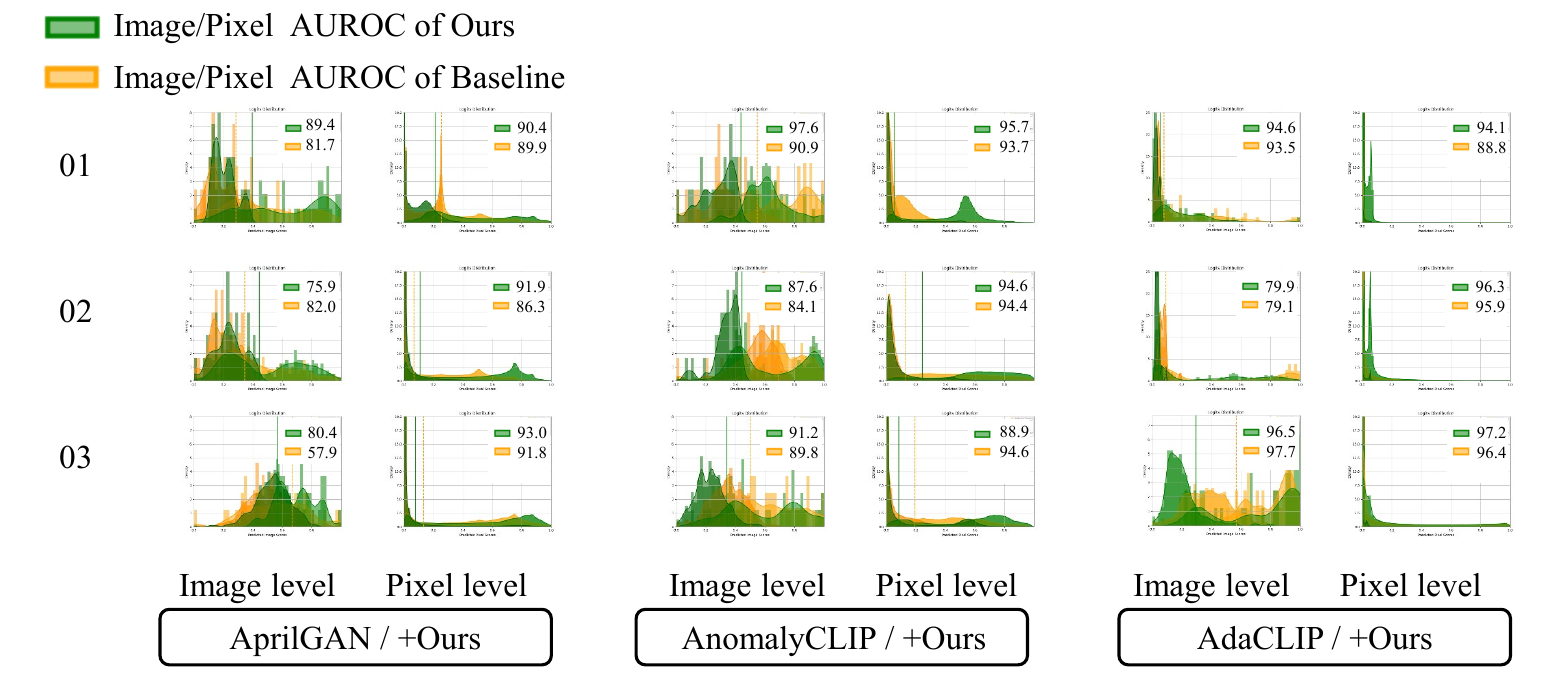}}
\caption{Quantitative comparison on \textbf{multimodal RGB-TEXT UAD} for \textbf{BTAD}~\cite{btad} using KDE curves of image- and pixel-level anomaly logits. Each two-column pair (from left to right) compares AprilGAN~\cite{aprilgan}, AnomalyCLIP~\cite{anomalyclip}, and AdaCLIP~\cite{adaclip} with our method, where the first and second columns show image- and pixel-level APROC, respectively. This figure corresponds to Sec.\ref{suppsec7c}, Tables~\ref{suppbtadtexttable1}, \ref{suppbtadtexttable2}, and Fig.~\ref{suppkeshihuabtadrgbtext}.}
\label{suppkdebtadtext}
\end{center}
\end{figure*}

\begin{figure*}
\setlength{\abovecaptionskip}{2pt}  
\setlength{\belowcaptionskip}{0pt} 
\begin{center}
\centerline{\includegraphics[width=0.85\textwidth]{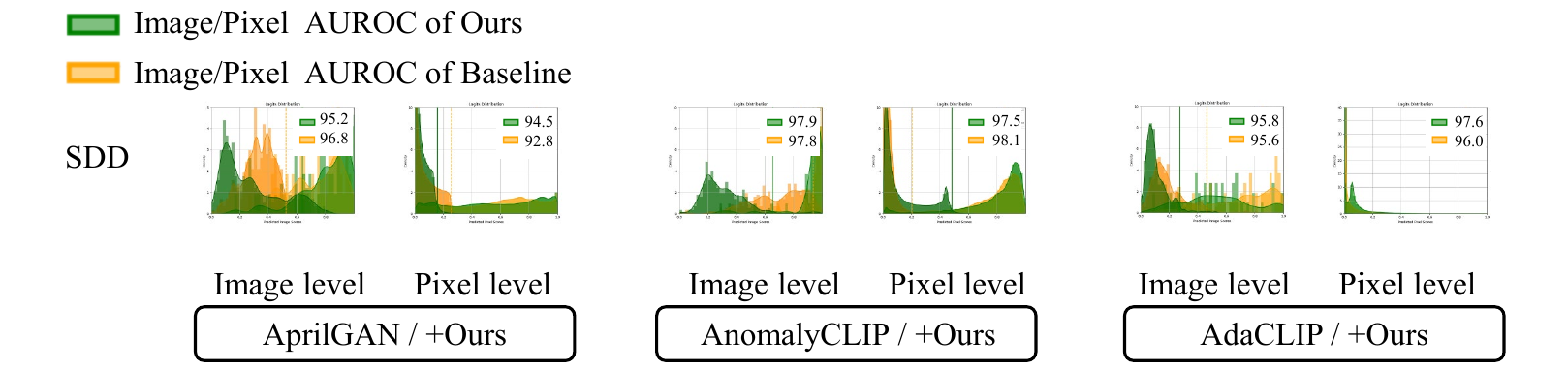}}
\caption{Quantitative comparison on \textbf{multimodal RGB-TEXT UAD} for \textbf{SDD}~\cite{ksdd} using KDE curves of image- and pixel-level anomaly logits. Each two-column pair (from left to right) compares AprilGAN~\cite{aprilgan}, AnomalyCLIP~\cite{anomalyclip}, and AdaCLIP~\cite{adaclip} with our method, where the first and second columns show image- and pixel-level APROC, respectively. This figure corresponds to Sec.\ref{suppsec7c}, Tables~\ref{suppsddtexttable1}, \ref{suppsddtexttable2}, and Fig.~\ref{suppkeshihuasddrgbtext}.}
\label{suppkdesdd}
\end{center}
\end{figure*}

\begin{figure*}
\setlength{\abovecaptionskip}{2pt}  
\setlength{\belowcaptionskip}{0pt} 
\begin{center}
\centerline{\includegraphics[width=0.85\textwidth]{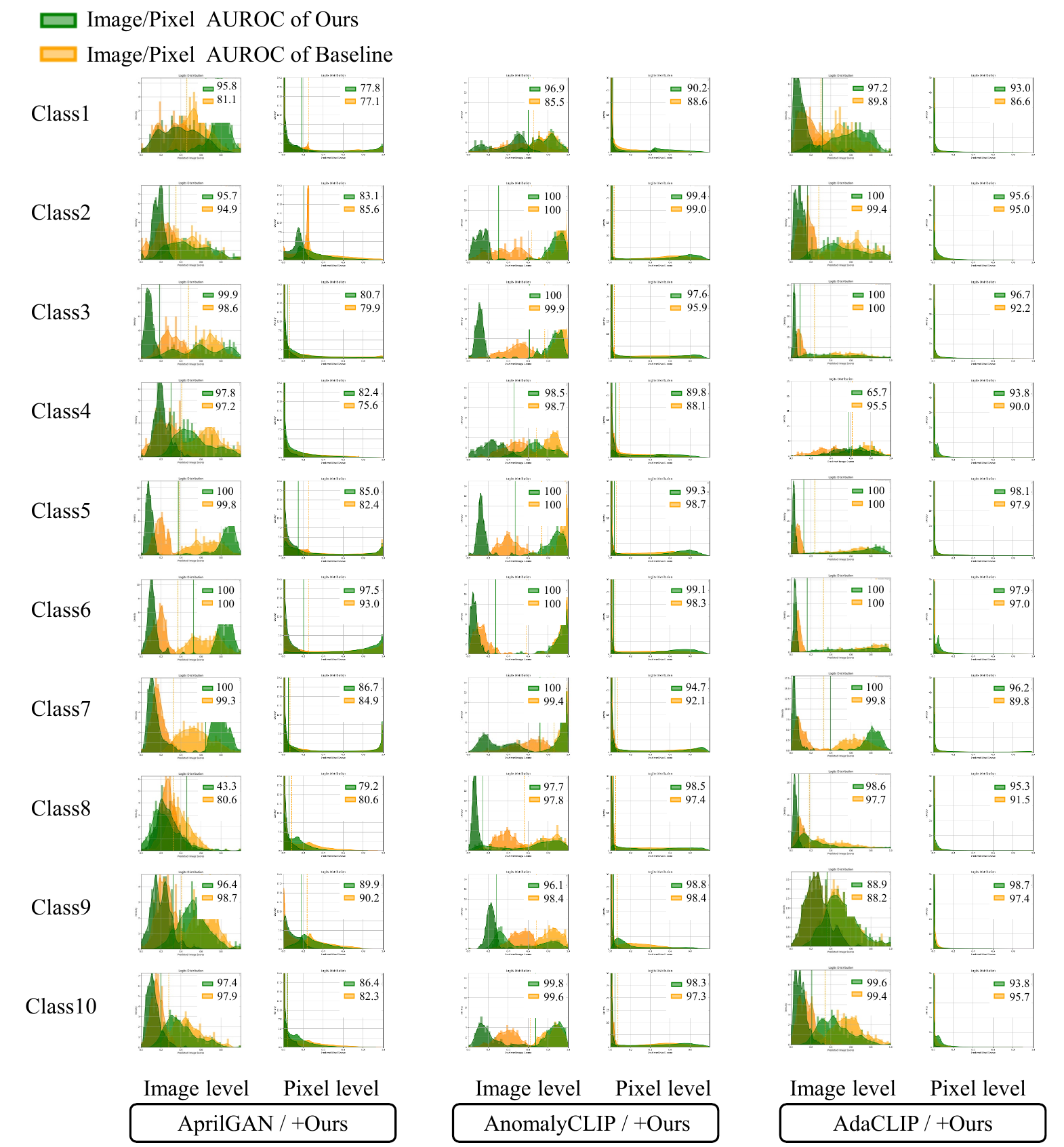}}
\caption{Quantitative comparison on \textbf{multimodal RGB-TEXT UAD} for \textbf{DAGM}~\cite{dagm} using KDE curves of image- and pixel-level anomaly logits. Each two-column pair (from left to right) compares AprilGAN~\cite{aprilgan}, AnomalyCLIP~\cite{anomalyclip}, and AdaCLIP~\cite{adaclip} with our method, where the first and second columns show image- and pixel-level APROC, respectively. This figure corresponds to Sec.\ref{suppsec7c}, Tables~\ref{suppdagmtexttable1}, \ref{suppdagmtexttable2}, and Fig.~\ref{suppkeshihuadagmrgbtext}.}
\label{suppkdedagm}
\end{center}
\end{figure*}

\begin{figure*}
\setlength{\abovecaptionskip}{2pt}  
\setlength{\belowcaptionskip}{0pt} 
\begin{center}
\centerline{\includegraphics[width=0.85\textwidth]{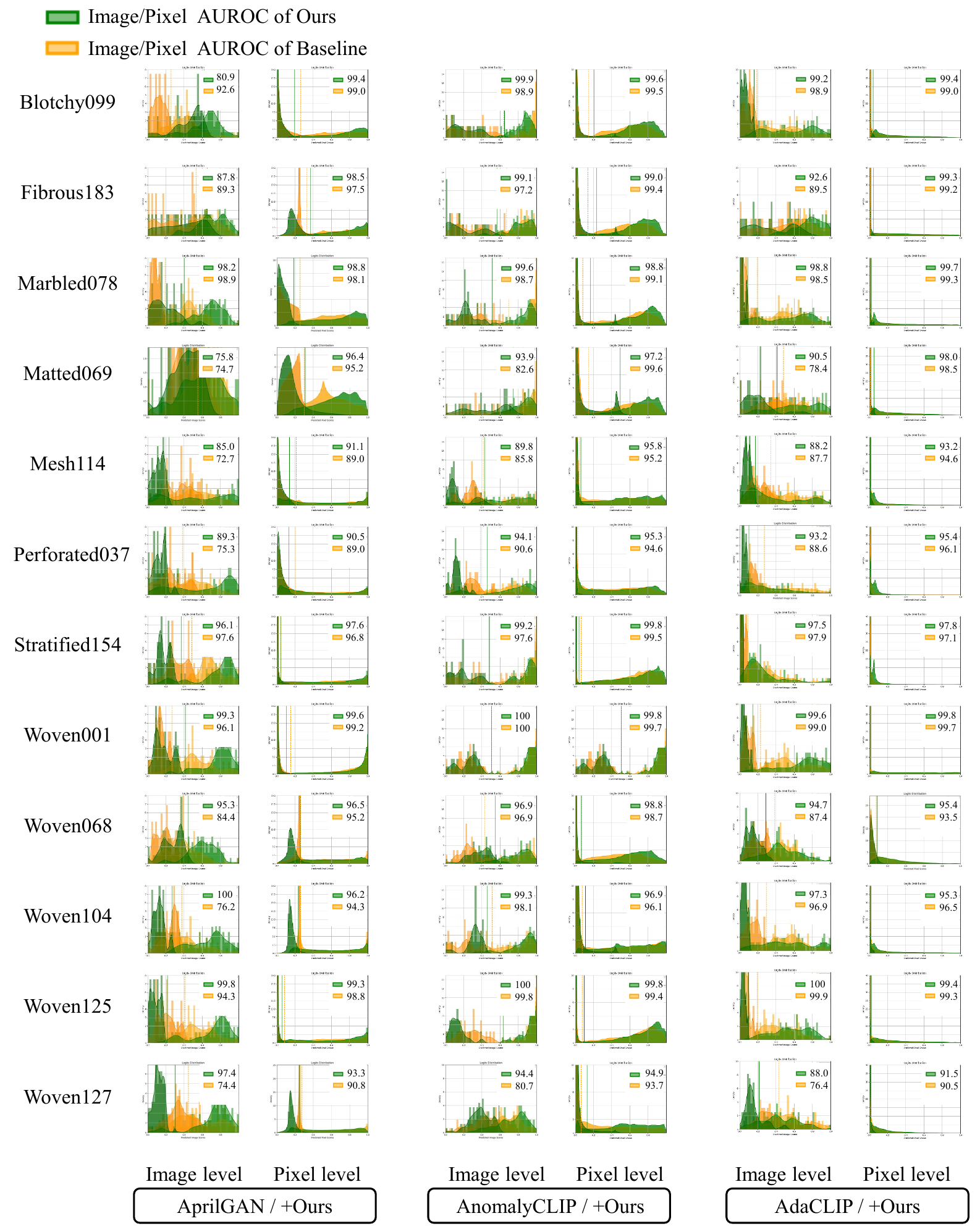}}
\caption{Quantitative comparison on \textbf{multimodal RGB-TEXT UAD} for \textbf{DTD}~\cite{dtd} using KDE curves of image- and pixel-level anomaly logits. Each two-column pair (from left to right) compares AprilGAN~\cite{aprilgan}, AnomalyCLIP~\cite{anomalyclip}, and AdaCLIP~\cite{adaclip} with our method, where the first and second columns show image- and pixel-level APROC, respectively. This figure corresponds to Sec.\ref{suppsec7c}, Tables~\ref{suppdtdtexttable1}, \ref{suppdtdtexttable2}, and Fig.~\ref{suppkeshihuadtdrgbtext}.}
\label{suppkdedtd}
\end{center}
\end{figure*}

\begin{figure*}
\setlength{\abovecaptionskip}{2pt}  
\setlength{\belowcaptionskip}{0pt} 
\begin{center}
\centerline{\includegraphics[width=0.85\textwidth]{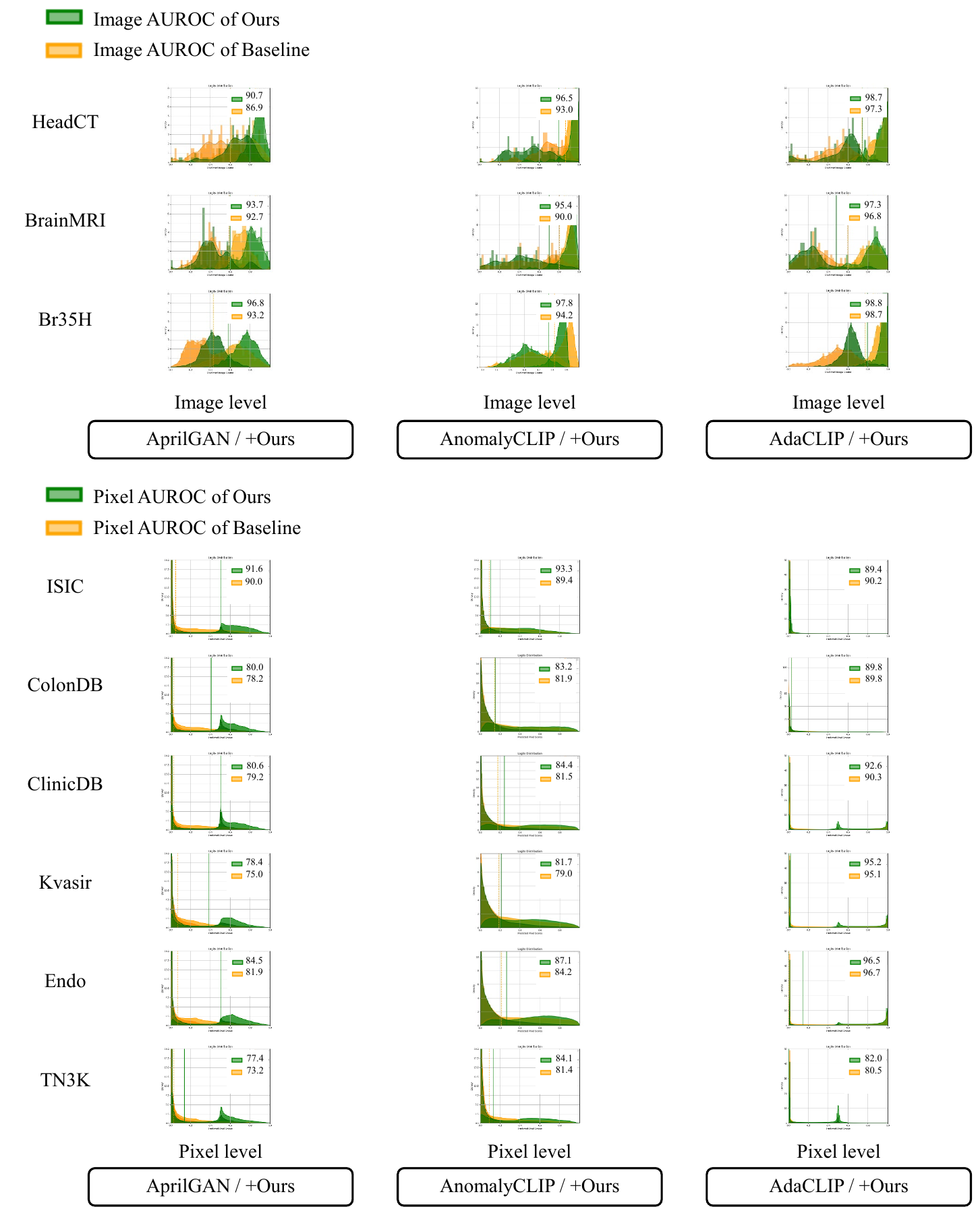}}
\caption{Quantitative comparison on \textbf{multimodal RGB-TEXT UAD} for \textbf{medical datasets}~\cite{headct,brainmri,br35h,isic,clodb,clidb,kvasir,endo,tn3k} using KDE curves of image- or pixel-level anomaly logits. Each column (from left to right) compares AprilGAN~\cite{aprilgan}, AnomalyCLIP~\cite{anomalyclip}, and AdaCLIP~\cite{adaclip} with our method. This figure corresponds to Sec.\ref{suppsec7c}, Tables~\ref{suppmedicaltexttable1}, \ref{suppmedicaltexttable2}, and Figs.~\ref{suppkeshihuamedical3rgbtext}, \ref{suppkeshihuamedical6rgbtext}.}
\label{suppkdemedicaltext}
\end{center}
\end{figure*}

We compile category-wise KDE analyses~\cite{kde} for multimodal RGB–Text UAD. Results on MVTec-AD~\cite{mvtec} are shown in Fig.~\ref{suppkdemvtectext}, and VisA~\cite{visa} appears in Fig.~\ref{suppkdevisatext}. Results on MPDD~\cite{mpdd}, BTAD~\cite{btad}, SDD~\cite{ksdd}, DAGM~\cite{dagm}, and DTD~\cite{dtd} are reported in Figures~\ref{suppkdempddtext}, \ref{suppkdebtadtext}, \ref{suppkdesdd}, \ref{suppkdedagm}, and \ref{suppkdedtd}. Medical datasets~\cite{headct,brainmri,br35h,isic,clodb,clidb,kvasir,endo,tn3k} are summarized in Fig.~\ref{suppkdemedicaltext}. Each two-column group plots image-level and pixel-level KDEs and compares AprilGAN~\cite{aprilgan}, AnomalyCLIP~\cite{anomalyclip}, and AdaCLIP~\cite{adaclip} with the proposed method.

Across industrial and medical domains, our curves exhibit larger mode separation and reduced overlap at both levels, indicating stronger cross-modal alignment and more reliable region delineation. The distributional evidence corroborates the quantitative gains and qualitative visualizations, underscoring the generality and effectiveness of the proposed cost volume filtering method for anomaly detection.

\begin{table*}[htbp]
  \centering
  \caption{Quantitative comparison for \textbf{unimodal RGB UAD}, reporting per-category results. Multi-class anomaly detection/localization results on \textbf{MVTec-AD}~\cite{mvtec} using \textbf{I-AP/P-AP}. \textbf{Best} results are in bold and \underline{runners-up} are underlined. This table corresponds to Sec.\ref{suppsec5a} and complements Table~I and Table~IV in the main paper.}
    \label{suppmvtectable2}%
  \renewcommand{\arraystretch}{1.4}
  \resizebox{1\linewidth}{!}{
\begin{tabular}{c>{\columncolor{lightblue}}c|c>{\columncolor{lightblue}}c|c>{\columncolor{lightblue}}c|c>{\columncolor{lightblue}}c|c>{\columncolor{lightblue}}c|c>{\columncolor{lightblue}}c}
    \toprule
    \multicolumn{2}{c}{Method~$\rightarrow$} & UniAD & UniAD+UCF & GLAD & GLAD+UCF & HVQ-Trans & HVQ-Trans+UCF & AnomalDF & AnomalDF+UCF & Dinomaly & Dinomaly+UCF \\
    \cline{1-2}
    \multicolumn{2}{c}{Category~$\downarrow$} & NeurIPS'22 & Ours & ECCV'24 & Ours & NeurIPS'23 & Ours & WACV'25 & Ours & CVPR'25 & Ours\\
    \hline
    \multicolumn{1}{c}{\multirow{10}[1]{*}{\begin{turn}{90}Objects\end{turn}}} 
    & \multicolumn{1}{c}{Bottle} & 
    \textbf{100.0} / 66.4 & 
    \textbf{100.0} / 81.2 & 
    \textbf{100.0} / 80.9 & 
    \textbf{100.0} / 79.2 & 
    \textbf{100.0} / 73.9 & 
    \textbf{100.0} / 81.4 & 
    \textbf{100.0} / 87.3 & 
    \textbf{100.0}/ 85.9 & 
    \textbf{100.0} / \underline{88.3} & 
    \textbf{100.0} / \textbf{91.7}\\
    \multicolumn{1}{c}{} & \multicolumn{1}{c}{ Cable} &  97.3 / 47.6 &  99.5 / 57.2 &  99.3 / 51.4 &  98.8 / 64.1 &  99.7 / 54.2 &  \underline{99.8} / 58.0 &  \underline{99.8} / 69.3 &  99.6 / \underline{72.3} &  \textbf{100.0} / 66.7 &  \textbf{100.0} / \textbf{75.4}  \\
    \multicolumn{1}{c}{} & \multicolumn{1}{c}{Capsule}  & 98.4 / 44.5 & 
    99.2 / 51.4 & 
    99.2 / 49.1 & 
    98.8 / \underline{53.1} & 
    99.0 / 44.0 & 
    99.2 / 49.2 &
    97.3 / 45.9 & 
    99.1 / 45.4 & 
    \underline{99.6} / \textbf{60.7 }& 
    \textbf{99.7} / \textbf{64.3}\\
    \multicolumn{1}{c}{} & \multicolumn{1}{c}{ Hazelnut} &  \textbf{100.0} / 54.6 &  \textbf{100.0} / 70.4  &  98.2 / 68.0 &  99.6 / 75.8 &  \textbf{100.0} / 63.1 &  \textbf{100.0} / 72.4 &  \underline{99.9} / \underline{79.0} & 
     \textbf{100.0} / 77.6 &
     \textbf{100.0} / \textbf{81.9} & 
      \textbf{100.0} / \textbf{89.6} \\
\bottomrule
    \end{tabular}%
  }
\end{table*}

\begin{table*}[htbp]
  \centering
  \caption{Quantitative comparison for \textbf{unimodal RGB UAD}, reporting per-category results. Multi-class anomaly detection/localization results on \textbf{MVTec-AD}~\cite{mvtec} using \textbf{I-F1-max/P-F1-max}. \textbf{Best} results are in bold and \underline{runners-up} are underlined. This table corresponds to Sec.\ref{suppsec5a} and complements Table~I and Table~IV in the main paper.}
    \label{suppmvtectable3}%
  \renewcommand{\arraystretch}{1.4}
  \resizebox{1\linewidth}{!}{
    \begin{tabular}{c>{\columncolor{lightblue}}c|c>{\columncolor{lightblue}}c|c>{\columncolor{lightblue}}c|c>{\columncolor{lightblue}}c|c>{\columncolor{lightblue}}c|c>{\columncolor{lightblue}}c}
    \toprule
    \multicolumn{2}{c}{Method~$\rightarrow$} & UniAD & UniAD+UCF & GLAD & GLAD+UCF & HVQ-Trans & HVQ-Trans+UCF & AnomalDF & AnomalDF+UCF & Dinomaly & Dinomaly+UCF \\
    \cline{1-2}
    \multicolumn{2}{c}{Category~$\downarrow$} & NeurIPS'22 & Ours & ECCV'24 & Ours & NeurIPS'23 & Ours & WACV'25 & Ours & CVPR'25 & Ours\\
    \hline
    \multicolumn{1}{c}{\multirow{10}[1]{*}{\begin{turn}{90}Objects\end{turn}}} 
    & \multicolumn{1}{c}{Bottle} & 
    \textbf{100.0} / 68.9 & 
    \textbf{100.0} / 73.7 & 
    \textbf{100.0} / 75.5  & 
    \underline{99.2} / 72.6 & 
    \textbf{100.0} / 71.6 & 
    \textbf{100.0} / 77.6  & 
    \textbf{100.0} / 80.2 & 
    \textbf{100.0}/ 77.0 & 
    \textbf{100.0} / \underline{83.8} & 
    \textbf{100.0} / \textbf{86.2}\\
    \multicolumn{1}{c}{} & \multicolumn{1}{c}{ Cable} &  89.9 / 55.4 & 
     97.3 / 59.8 &
     97.3 / 53.4 &  95.2 / 60.5 &  97.8 / 60.8  &  98.4 / 63.0  &  97.8 / 67.0  &  97.9 / 66.2 &  \underline{99.5} / \underline{69.1} &  \textbf{99.6} / \textbf{74.7}  \\
    \multicolumn{1}{c}{} & \multicolumn{1}{c}{Capsule}  & 95.5 / 48.2 & 
    95.6 / \underline{53.2} & 
    \underline{96.8} / 51.2 & 
    95.5 / 51.6 & 
    96.8 / 48.3 & 
    96.4 / 53.0 & 
    94.3 / 48.9  & 
    96.4 / 50.3 & 
    \textbf{97.3} / \textbf{60.5} & 
    \textbf{97.3} / \textbf{61.2}\\
    \multicolumn{1}{c}{} & \multicolumn{1}{c}{ Hazelnut} &  \textbf{100.0} / 55.9 &  \textbf{100.0} / 67.8  &  94.4 / 63.8 &  97.2 / 69.4  &  \underline{99.3} / 63.1 &  \textbf{100.0} / 70.8 &  \underline{99.3} / \underline{75.5} & 
     \textbf{100.0} / 69.9 &
     \textbf{100.0} / \textbf{76.8} & 
      \textbf{100.0} / \textbf{82.7} \\
\bottomrule
    \end{tabular}%
  }
\end{table*}

\begin{table*}[htbp]
  \centering
  \caption{Quantitative comparison for \textbf{unimodal RGB UAD}, reporting per-category results. Multi-class anomaly localization results on \textbf{MVTec-AD}~\cite{mvtec} using \textbf{P-AURPO}. \textbf{Best} results are in bold and \underline{runners-up} are underlined. This table corresponds to Sec.\ref{suppsec5a} and complements Table~I and Table~IV in the main paper.}
    \label{suppmvtectable4}%
  \renewcommand{\arraystretch}{1.4}
  \resizebox{1\linewidth}{!}{
    \begin{tabular}{cc|c>{\columncolor{lightblue}}c|c>{\columncolor{lightblue}}c|c>{\columncolor{lightblue}}c|c>{\columncolor{lightblue}}c|c>{\columncolor{lightblue}}c}
    \toprule
    \multicolumn{2}{c}{Method~$\rightarrow$} & UniAD & UniAD+UCF & GLAD & GLAD+UCF & HVQ-Trans & HVQ-Trans+UCF & AnomalDF & AnomalDF+UCF & Dinomaly & Dinomaly+UCF \\
     \cline{1-2}
    \multicolumn{2}{c}{Category~$\downarrow$} & NeurIPS'22 & Ours & ECCV'24 & Ours & NeurIPS'23 & Ours & WACV'25 & Ours & CVPR'25 & Ours\\
    \hline
    \multicolumn{1}{c}{\multirow{10}[1]{*}{\begin{turn}{90}Objects\end{turn}}} 
    & \multicolumn{1}{c}{Bottle} & 
    93.2 & 
    94.3 & 
     96.1  & 
     90.8 & 
    94.4 & 
    96.1  & 
    97.5  & 
    95.3 & 
    \underline{96.6} & 
    \textbf{97.0}\\
    \multicolumn{1}{c}{} & \multicolumn{1}{c}{ Cable} &  86.0 & 
     88.8  &
     89.6  & 
      89.8  & 
     89.6  & 
     91.7 & 
     \textbf{94.2 }& 
     92.5 & 
     \underline{93.7} &  \textbf{94.2}  \\
    \multicolumn{1}{c}{} & \multicolumn{1}{c}{Capsule}  & 91.1 & 
    90.0 & 
    96.1 & 
    93.1 & 
     89.9  & 
    92.9 & 
    95.8  & 
    93.5 & 
    \textbf{97.3}& 
    \underline{97.1}\\
    \multicolumn{1}{c}{} & \multicolumn{1}{c}{ Hazelnut} &  92.8 & 
     92.2  & 
     90.8  & 
     93.0 & 
     93.8  & 
     92.9&
     92.5  & 
     92.8 &
     \underline{96.9} & 
      \textbf{97.7} \\
    \multicolumn{1}{c}{} & \multicolumn{1}{c}{Metal Nut}  & 82.4 & 
    87.2  & 
     94.2 & 
    96.3 & 
    90.6 & 
    93.3  & 
    94.7 & 
    96.2 & 
    \textbf{97.5} & 
    \underline{96.5}\\
    \multicolumn{1}{c}{} & \multicolumn{1}{c}{ Pill}  &  95.3 & 
     95.4  & 
      94.3 & 
     96.8 & 
      94.9& 
     96.1  & 
     96.7 & 
     95.8 & 
     \underline{97.5} &  \textbf{98.0}\\
\bottomrule
    \end{tabular}%
  }
\end{table*}

\begin{table*}[htbp]
  \centering
  \caption{Quantitative comparison for \textbf{unimodal RGB UAD}, reporting per-category results. Multi-class anomaly detection/localization results on \textbf{VisA}~\cite{visa} using \textbf{I-AUROC/P-AUROC}. \textbf{Best} results are in bold and \underline{runners-up} are underlined. This table corresponds to Sec.\ref{suppsec5a} and complements Table~III and Table~IV in the main paper.}
    \label{suppvisatable1}%
  \renewcommand{\arraystretch}{1.4}
  \resizebox{1\linewidth}{!}{
    \begin{tabular}{cc|c>{\columncolor{lightblue}}c|c>{\columncolor{lightblue}}c|c>{\columncolor{lightblue}}c|c>{\columncolor{lightblue}}c|c>{\columncolor{lightblue}}c}
    \toprule
    \multicolumn{2}{c}{Method~$\rightarrow$} & UniAD & UniAD+UCF & GLAD & GLAD+UCF & HVQ-Trans & HVQ-Trans+UCF & AnomalDF & AnomalDF+UCF & Dinomaly & Dinomaly+UCF \\
    \cline{1-2}
    \multicolumn{2}{c}{Category~$\downarrow$} & NeurIPS'22 & Ours & ECCV'24 & Ours & NeurIPS'23 & Ours & WACV'25 & Ours & CVPR'25 & Ours\\
    \hline
    \multicolumn{1}{c}{\multirow{4}[1]{*}{\begin{turn}{90}{\shortstack{Complex\\Structure}} \end{turn}}}
    & \multicolumn{1}{c}{PCB1} & 
    95.1 / 98.9 & 
    95.4 / 99.4 & 
    69.9 / 97.6  & 
    90.9 / 97.7 & 
    95.1 / \underline{99.5} & 
    96.3 / 99.3 & 
    87.4 / 99.3  & 
    91.8 / \textbf{99.7} & 
    \underline{99.0} / \underline{99.5} &  
    \textbf{99.1} / 99.4\\
    \multicolumn{1}{c}{} & \multicolumn{1}{c}{ PCB2} &
     93.0 / 96.7 &  93.2 / 97.9 &  89.9 / 97.1 &  93.2 / 95.7 &  93.4 / \textbf{98.1} &  97.0 / \underline{98.0}  &  81.9 / 94.2 &  95.7 / \underline{98.0}  & 
     \underline{99.2} / \underline{98.0} &  \textbf{99.6} /\textbf{ 98.8}\\
    \multicolumn{1}{c}{} & \multicolumn{1}{c}{PCB3} & 
    88.4 / 96.5 & 
    89.6 / 98.3 & 
    93.3 / 96.2 &
    90.5 / 97.4 & 
    88.5 / 98.2 & 
    89.8 / 97.7  & 
    87.4 / 96.5 & \underline{94.0} / \textbf{98.9} & 
    \textbf{98.8} / \underline{98.4} & \textbf{99.2} / 98.3 \\
    \multicolumn{1}{c}{} & \multicolumn{1}{c}{ PCB4} &  98.7 / 98.1 &  99.3 / 97.8 &  99.0 / \textbf{99.4} &  99.4 / \underline{99.3} &  99.3 / 98.1 &  98.7 / 97.8 &  96.7 / 97.3 &  98.1 / 98.9 &  \underline{99.7} / 98.7 &  \textbf{99.8} / 98.4 \\
\bottomrule
    \end{tabular}%
  }
\end{table*}

\begin{table*}[htbp]
  \centering
  \caption{Quantitative comparison for \textbf{unimodal RGB UAD}, reporting per-category results. Multi-class anomaly detection/localization results on \textbf{VisA}~\cite{visa} using \textbf{I-AP/P-AP}. \textbf{Best} results are in bold and \underline{runners-up} are underlined. This table corresponds to Sec.\ref{suppsec5a} and complements Table~III and Table~IV in the main paper.}
    \label{suppvisatable2}%
  \renewcommand{\arraystretch}{1.4}
  \Large
  \resizebox{1\linewidth}{!}{
    \begin{tabular}{c>{\columncolor{lightblue}}c|c>{\columncolor{lightblue}}c|c>{\columncolor{lightblue}}c|c>{\columncolor{lightblue}}c|c>{\columncolor{lightblue}}c|c>{\columncolor{lightblue}}c}
    \toprule
    \multicolumn{2}{c}{Method~$\rightarrow$} & UniAD & UniAD+UCF & GLAD & GLAD+UCF & HVQ-Trans & HVQ-Trans+UCF & AnomalDF & AnomalDF+UCF & Dinomaly & Dinomaly+UCF \\
    \cline{1-2}
    \multicolumn{2}{c}{Category~$\downarrow$} & NeurIPS'22 & Ours & ECCV'24 & Ours & NeurIPS'23 & Ours & WACV'25 & Ours & CVPR'25 & Ours\\
    \hline
    \multicolumn{1}{c}{\multirow{4}[1]{*}{\begin{turn}{90}{\shortstack{Complex\\Structure}} \end{turn}}}
    & \multicolumn{1}{c}{PCB1} & 
    93.8 / 51.5 &
    94.4 / 63.3 & 
    72.5 / 38.0 &
    88.7 / 64.5 & 
    94.5 / 71.6 &
    \underline{95.4} / 71.6 & 
    84.6 / 81.3 & 
    90.6 / 82.9 & 
    \textbf{98.9} /\underline{ 87.8} & \textbf{98.9} / \textbf{88.5}\\
    \multicolumn{1}{c}{} & \multicolumn{1}{c}{ PCB2} &  93.2 / 10.6 &  93.9 / 9.3 &  88.9 / 6.4 &  92.0 / 6.5 &  94.2 / 9.5 &  97.1 / 12.3 &  81.1 / 12.0 &  96.2 / 13.1 &
      \underline{99.2} / \underline{45.6} &  \textbf{99.4 }/ \textbf{46.5} \\
    \multicolumn{1}{c}{} & \multicolumn{1}{c}{PCB3} & 
    89.5 / 23.8 &
    90.2 / 18.7 & 
    94.0 / 25.0 & 
    90.7 / 22.4  & 
    89.4 / 18.0 & 
    90.4 / 25.5 & 
    90.2 / 23.3 & 
   94.5 / 30.7 &
    \underline{98.8} / \underline{41.0} & \textbf{99.0} / \textbf{38.5}\\
    \multicolumn{1}{c}{} & \multicolumn{1}{c}{ PCB4} &  98.6 / 35.2 &  99.2 / 33.2 &  98.2 / 52.6 &  99.4 / 47.5 &  99.2 / 31.9 &  98.4 / 38.2 &  96.3 / 37.4 &  97.7 / 32.8 &  \underline{99.7} / \underline{50.1} & \textbf{100.0} / \textbf{53.8}\\
\bottomrule
    \end{tabular}%
  }
\end{table*}

\begin{table*}[htbp]
  \centering
  \caption{Quantitative comparison for \textbf{unimodal RGB UAD}, reporting per-category results. Multi-class anomaly detection/localization results on \textbf{VisA}~\cite{visa} using \textbf{I-F1-max/P-F1-max}. \textbf{Best} results are in bold and \underline{runners-up} are underlined. This table corresponds to Sec.\ref{suppsec5a} and complements Table~III and Table~IV in the main paper.}
    \label{suppvisatable3}%
  \renewcommand{\arraystretch}{1.4}
  \resizebox{1\linewidth}{!}{
    \begin{tabular}{cc|c>{\columncolor{lightblue}}c|c>{\columncolor{lightblue}}c|c>{\columncolor{lightblue}}c|c>{\columncolor{lightblue}}c|c>{\columncolor{lightblue}}c}
    \toprule
    \multicolumn{2}{c}{Method~$\rightarrow$} & UniAD & UniAD+UCF & GLAD & GLAD+UCF & HVQ-Trans & HVQ-Trans+UCF & AnomalDF & AnomalDF+UCF & Dinomaly & Dinomaly+UCF \\
    \cline{1-2}
    \multicolumn{2}{c}{Category~$\downarrow$} & NeurIPS'22 & Ours & ECCV'24 & Ours & NeurIPS'23 & Ours & WACV'25 & Ours & CVPR'25 & Ours\\
    \hline
   \multicolumn{1}{c}{\multirow{4}[1]{*}{\begin{turn}{90}{\shortstack{Complex\\Structure}} \end{turn}}}
    & \multicolumn{1}{c}{PCB1} & 91.4 / 55.9 & 93.2 / 62.3 &  70.1 / 44.4 & 86.1 / 60.4 & 91.6 / 63.7 & 91.6 / 66.9 & 82.2 / 68.0 &  86.9 / 75.5 &\underline{96.6} / \underline{80.2} & \textbf{96.6} / \textbf{82.0}\\
    \multicolumn{1}{c}{} & \multicolumn{1}{c}{ PCB2} &  86.1 / 20.4 &  88.2 / 16.6 &  83.3 / 14.4&  87.9 / 14.1 &  88.8 / 16.5 &   92.0 / 21.1&  76.2 / 23.7&  90.1 / 25.7 &  \underline{97.0} / \underline{48.8} &  \textbf{98.5} / \textbf{46.2}\\
    \multicolumn{1}{c}{} & \multicolumn{1}{c}{PCB3} & 81.5 / 29.1 & 83.1 / 25.0 & 87.6 / 27.7 & 84.5 / 27.0& 81.1 / 22.7 &  82.1 / 28.6  & 80.2 / 38.8& 88.7 / 34.9 & \underline{95.6} / \underline{45.6} & \textbf{96.6} / \textbf{40.8}\\
    \multicolumn{1}{c}{} & \multicolumn{1}{c}{ PCB4} &  \textbf{98.5} / 38.7 &  97.6 / 35.6 &  98.0 / 52.0 &   97.0 / 50.1&  97.0 / 36.0&  97.1 / 39.4&  91.0 / 30.8 &  95.1 / 35.2 &  \underline{98.0} / \underline{52.8} &  \underline{98.0} / \textbf{55.9}\\
    \hline
    \multicolumn{1}{c}{\multirow{4}[1]{*}{\begin{turn}{90}{\shortstack{Multiple\\Instances}} \end{turn}}}
    & \multicolumn{1}{c}{Macaroni1} & 
    90.5 / 25.0 & 
    87.6 / 17.5 & 
    85.4 / 19.2 & 
    89.8 / 26.3 & 
    86.1 / 19.9 & 
    86.7 / 19.7 & 
    79.5 / 17.3 & 
    88.8 / 23.3 & 
    \underline{94.5} / \underline{38.9} & \textbf{95.5} / \textbf{45.8}\\
\bottomrule
    \end{tabular}%
  }
\end{table*}

\begin{table*}[htbp]
  \centering
  \caption{Quantitative comparison for \textbf{unimodal RGB UAD}, reporting per-category results. Multi-class anomaly localization results on \textbf{VisA}~\cite{visa} using \textbf{AUPRO}. \textbf{Best} results are in bold and \underline{runners-up} are underlined. This table corresponds to Sec.\ref{suppsec5a} and complements Table~III and Table~IV in the main paper.}
    \label{suppvisatable4}%
  \renewcommand{\arraystretch}{1.4}
 
  \resizebox{1\linewidth}{!}{
    \begin{tabular}{cc|c>{\columncolor{lightblue}}c|c>{\columncolor{lightblue}}c|c>{\columncolor{lightblue}}c|c>{\columncolor{lightblue}}c|c>{\columncolor{lightblue}}c}
    \toprule
    \multicolumn{2}{c}{Method~$\rightarrow$} & UniAD & UniAD+UCF & GLAD & GLAD+UCF & HVQ-Trans & HVQ-Trans+UCF & AnomalDF & AnomalDF+UCF & Dinomaly & Dinomaly+UCF \\
    \cline{1-2}
    \multicolumn{2}{c}{Category~$\downarrow$} & NeurIPS'22 & Ours & ECCV'24 & Ours & NeurIPS'23 & Ours & WACV'25 & Ours & CVPR'25 & Ours\\
    \hline
    \multicolumn{1}{c}{\multirow{4}[1]{*}{\begin{turn}{90}\shortstack{Complex\\Structure}\end{turn}}}
   & \multicolumn{1}{c}{PCB1} & 82.5 & 89.5 & 88.3 & 82.7 & \underline{90.4} & 88.3 & 82.8 &  77.9 & \textbf{95.2} & \textbf{95.2}\\
    \multicolumn{1}{c}{} & \multicolumn{1}{c}{ PCB2} &   61.4 &  82.8 &  \textbf{91.7} &  84.7 &  84.1 &   85.4 &  77.7 &  82.7 &  91.3 &  \underline{91.5} \\
    \multicolumn{1}{c}{} & \multicolumn{1}{c}{PCB3}  & 45.7 & 79.3 & 94.2 & 92.2 & 79.9 & 75.6 &  79.7 & 74.5 & \underline{94.7} & \textbf{94.8}\\
    \multicolumn{1}{c}{} & \multicolumn{1}{c}{ PCB4} &  84.8 &  83.9 &  \textbf{94.9} &  \underline{94.7} &  84.8 &  84.4 &  83.1 &  79.7 &   94.1 &   \underline{94.7}\\
    \hline
    \multicolumn{1}{c}{\multirow{4}[1]{*}{\begin{turn}{90}\shortstack{Multiple\\Instances}\end{turn}}}
    & \multicolumn{1}{c}{Macaroni1}  & 96.1 & 95.7 & \textbf{99.1} & \underline{99.0} & 93.9 & 96.4 & 90.2 & 91.9 & 96.4 & 96.8\\
    \multicolumn{1}{c}{} & \multicolumn{1}{c}{ Macaroni2}  &  79.5 &  89.9 &  97.2 &  98.3 &  91.9 &   94.2 &  84.8 &  97.9 &  \underline{98.6} &  \textbf{98.8}\\
\bottomrule
    \end{tabular}%
  }
\end{table*}

  \begin{table*}[htbp]
    \centering
    \caption{Quantitative comparison for \textbf{unimodal RGB UAD}, reporting per-category results. Multi-class anomaly detection results on \textbf{BTAD}~\cite{btad} using \textbf{I-AUROC/I-AP/I-F1-max}. \textbf{Best} results are in bold and \underline{runners-up} are underlined. This table corresponds to Sec.\ref{suppsec5a} and complements Table~IV in the main paper.}
    \label{suppbtadtable1} 
    \renewcommand{\arraystretch}{1.4}
    {
    {\scriptsize
      \begin{tabular}{cc>{\columncolor{lightblue}}c|c>{\columncolor{lightblue}}c}
      \toprule
      \multicolumn{1}{c}{Method~$\rightarrow$} & Dinomaly & Dinomaly+UCF  & HVQ-Trans & HVQ-Trans+UCF \\
      \cline{1-1}
      \multicolumn{1}{c}{Category~$\downarrow$} & CVPR'25 & Ours & NeurIPS'23 & Ours \\
      \hline
      01 & 
      96.8 / 98.8 / 94.9 & 
      \underline{98.1} / \underline{99.3} / \textbf{96.9} & 
      96.9 / 98.8 / 94.9 & 
      \textbf{98.3} / \textbf{99.4} / \textbf{96.9} \\
      02 & 
       \textbf{89.7} / \textbf{98.4} / \textbf{93.9} & 
       \textbf{90.7} / \textbf{98.5} / \textbf{95.4} &
       75.8 / 95.9 / \underline{92.8} & 
       \underline{81.7} / \underline{97.0} / \underline{92.8}  \\
      03 & 
      \textbf{99.9} / 98.4 / \underline{97.6} & 
      \underline{99.8} / 98.0 / 96.5 & 
      \textbf{99.9} / \underline{98.8} / 96.8 & 
      \textbf{99.9} / \textbf{99.6} / \textbf{98.4}  \\
      \hline
       Mean & 
       \underline{95.4} / \underline{98.5} / 95.5 & 
       \textbf{96.2} / \textbf{98.6} / \textbf{96.3} & 
       90.9 / 97.8 / 94.8 & 
       93.3 / \textbf{98.6} /\underline{96.0}  \\
      \bottomrule
      \end{tabular}%
    }
   }
  \end{table*}

  \begin{table*}[htbp]
    \centering
    \caption{Quantitative comparison for \textbf{unimodal RGB UAD}, reporting per-category results. Multi-class anomaly localization results on \textbf{BTAD}~\cite{btad} using \textbf{P-AUROC/P-AP/P-F1-max/AUPRO}. \textbf{Best} results are in bold and \underline{runners-up} are underlined. This table corresponds to Sec.\ref{suppsec5a} and complements Table~IV in the main paper.
    }
    \label{suppbtadtable2}
    \renewcommand{\arraystretch}{1.4}
    {\scriptsize
    {
      \begin{tabular}{cc>{\columncolor{lightblue}}c|c>{\columncolor{lightblue}}c}
      \toprule
      \multicolumn{1}{c}{Method~$\rightarrow$} & Dinomaly & Dinomaly+UCF  & HVQ-Trans & HVQ-Trans+UCF \\
      \cline{1-1}
      \multicolumn{1}{c}{Category~$\downarrow$} & CVPR'25 & Ours & NeurIPS'23 & Ours \\
      \hline
      01 & 
      \underline{97.1} / \underline{62.9} / \underline{64.5} / 72.4 & 
      \textbf{97.8} / \textbf{67.9} / \textbf{65.8} / \textbf{82.1} & 
      96.4 / 46.8 / 50.9 / 75.6 & 
      \underline{97.1} / 51.3 / 53.8 / \underline{76.1} \\
       02 & 
       \underline{96.8} / \underline{72.7} / \underline{68.0} / \underline{59.4} & 
       \textbf{97.1} / \textbf{77.6} / \textbf{70.7} / \textbf{61.5} & 
       94.6 / 48.8 / 55.2 / 55.8 & 
       95.0 / 44.1 / 51.2 / 56.3 \\
      03 & 
      \underline{99.7} / \underline{74.6} / \underline{71.6} / \underline{97.8} & 
      \textbf{99.9} / \textbf{79.1} / \textbf{73.4} / \textbf{99.2} & 
      99.0 / 34.1 / 39.9 / 95.4 & 
      \textbf{99.9} / 45.5 / 45.6 / 96.3 \\
      \hline
       Mean & 
       \underline{97.9} / \underline{70.1} / \underline{68.0} / \underline{76.5} & 
       \textbf{98.2} / \textbf{74.8} / \textbf{70.0} / \textbf{81.0} & 
       96.7 / 43.2 / 48.7 / 75.6 & 
       97.3 / 47.0 / 50.2 / 76.2 \\
      \bottomrule
      \end{tabular}%
    }
    }
  \end{table*}

  \begin{table*}[!t]
    \centering
    \caption{
    Quantitative comparison for \textbf{unimodal RGB UAD}, reporting per-category results. Multi-class anomaly detection results on \textbf{MPDD}~\cite{mpdd} using \textbf{I-AUROC/I-AP/I-F1-max}. \textbf{Best} results are in bold and \underline{runners-up} are underlined. This table corresponds to Sec.\ref{suppsec5a} and complements Table~IV in the main paper.}
    \label{suppmpddtable1}
    \renewcommand{\arraystretch}{1.4}
    {\scriptsize
      \begin{tabular}{cc>{\columncolor{lightblue}}c|c>{\columncolor{lightblue}}c}
      
      \toprule
      \multicolumn{1}{c}{Method~$\rightarrow$} & Dinomaly & Dinomaly+UCF  & HVQ-Trans & HVQ-Trans+UCF \\
      \cline{1-1}
      \multicolumn{1}{c}{Category~$\downarrow$} & CVPR'25 & Ours & NeurIPS'23 & Ours \\
      \hline
      Bracket black &  
      \underline{93.4} / \underline{96.3} / 87.9    & 
      \textbf{93.8} / \textbf{96.5} / \textbf{90.1}    & 
      92.8 / 95.6 / \underline{89.8} & 
      91.3 / 93.9 / 89.3 \\
      
       Bracket brown &  
       \textbf{95.3} / \textbf{96.9} / \textbf{95.3}      & 
       \textbf{95.3} / \textbf{96.9} / \textbf{95.3}    & 
       89.4 / 93.4 / 90.3 &
       \underline{93.7} / \underline{96.2} / \underline{94.4} \\
      Bracket white &  
      \textbf{99.0} / \textbf{99.1} / \underline{94.7}    & 
      \textbf{99.0} / \textbf{99.1} / \textbf{94.9}    & 
      79.2 / 82.6 / 74.7 & 
      \underline{92.1} / \underline{93.0} / 85.2 \\
       Connector     &  
       \textbf{100.0} / \textbf{100.0} / \textbf{100.0} & 
       \textbf{100.0} / \textbf{100.0} / \textbf{100.0} & 
       89.3 / 69.3 / 81.3 & 
       \underline{97.9} / \underline{96.0} / \underline{90.3} \\
      Metal plate   &  
      \textbf{100.0} / \textbf{100.0} / \textbf{100.0} & 
      \textbf{100.0} / \textbf{100.0} / \textbf{100.0} & 
      97.5 / 99.1 / 95.8 &
      \underline{98.6} / \underline{99.5} / \underline{97.3} \\
       Tubes         &  
       \underline{95.9} / \textbf{98.5} / \textbf{95.5}    & 
       \textbf{96.1} / \textbf{98.5} / \textbf{95.5}    & 
       70.6 / 87.2 / 81.7 & 
       84.6 / \underline{93.7} / \underline{85.3} \\
      \hline
      Mean          &  
      \underline{97.3} / \textbf{98.5} / \underline{95.6}    & 
      \textbf{97.4} / \textbf{98.5} / \textbf{96.0}    & 
      86.5 / 87.9 / 85.6 & 
      93.1 / \underline{95.4} / 90.3 \\
      \bottomrule
      \end{tabular}%
      
    }
\end{table*}

\begin{table*}[!t]
    \centering
    \caption{Quantitative comparison for \textbf{unimodal RGB UAD}, reporting per-category results. Multi-class anomaly localization results on \textbf{MPDD}~\cite{mpdd} using \textbf{P-AUROC/P-AP/P-F1-max/AUPRO}. \textbf{Best} results are in bold and \underline{runners-up} are underlined. This table corresponds to Sec.\ref{suppsec5a} and complements Table~IV in the main paper.
    }
    \label{suppmpddtable2}
    \renewcommand{\arraystretch}{1.4}
    {\scriptsize
    
      \begin{tabular}{cc>{\columncolor{lightblue}}c|c>{\columncolor{lightblue}}c}
      \toprule
      \multicolumn{1}{c}{Method~$\rightarrow$} & Dinomaly & Dinomaly+UCF  & HVQ-Trans & HVQ-Trans+UCF \\
      \cline{1-1}
      \multicolumn{1}{c}{Category~$\downarrow$} & CVPR'25 & Ours & NeurIPS'23 & Ours \\
      \hline
      Bracket black & 
      \textbf{99.4} / \textbf{37.5} / \textbf{47.1} / \textbf{98.3} & 
      \textbf{99.4} / \underline{37.3} / \underline{46.9} / \textbf{98.3} & 
      97.0 / 1.5 / 3.5 / \underline{90.1}   & 
      \underline{97.2} / 3.1 / 7.7 / 87.3 \\
       Bracket brown & 
       \underline{98.2} / \textbf{50.3} / \underline{48.5} / \textbf{96.7} & 
       \underline{98.2} / \underline{50.2} / \textbf{48.8} / \underline{96.5} & 
       \textbf{98.3} / 31.5 / 36.7 / 88.6 & 
       97.3 / 31.1 / 34.4 / 63.0 \\
      Bracket white & 
      \textbf{99.4} / \underline{18.3} / \textbf{25.1} / \underline{93.4} & 
      \textbf{99.4} / \textbf{19.0} / \textbf{25.1} / \textbf{93.7} & 
      95.2 / 0.7 / 2.6 / 84.2   & 
      \underline{98.0} / 6.8 / \underline{15.5} / 88.6 \\
       Connector     & 
       \textbf{99.3} / \textbf{74.6} / \textbf{69.3} / \underline{97.5} &
       \textbf{99.3} / \underline{74.5} / \underline{69.2} / \textbf{97.6} & 
       97.5 / 16.6 / 27.1 / 91.1 &
       \underline{97.7} / 28.8 / 33.8 / 86.9 \\
      Metal plate   & 
      \textbf{99.6} / \textbf{97.7} / \underline{92.3} / \textbf{97.7} & 
      \textbf{99.5} / \textbf{97.7} / \textbf{92.4} / \textbf{97.7} & 
      96.6 / 73.6 / 74.4 / \underline{86.6} & 
      \underline{98.4} / \underline{88.1} / 82.1 / 84.6 \\
       Tubes         & 
       \textbf{99.1} / \underline{81.8} / \underline{76.5} / \textbf{96.5} & 
       \textbf{99.1} / \textbf{82.3} / \textbf{76.7} / \textbf{96.5} & 
       \underline{96.7} / 34.4 / 38.5 / \underline{87.5} &
       96.2 / 47.0 / 48.6 / 86.9 \\
\bottomrule
      \end{tabular}
    }
\end{table*}

\begin{table*}[htbp]
  \centering
  \caption{Quantitative comparison for \textbf{multimodal RGB–Text UAD}, reporting per-category results. Zero-shot anomaly detection and localization results on \textbf{MVTec-AD}~\cite{mvtec} using \textbf{I-AUROC/I-AP/P-AUROC/AUPRO}. \textbf{Best} results are in bold and \underline{runner-ups} are underlined. This table corresponds to Sec.\ref{suppsec5c} and complements Table~VII in the main paper.}
    \label{suppmvtectexttable1}%
  \renewcommand{\arraystretch}{1.6}
  \resizebox{1\linewidth}{!}{
    \begin{tabular}{cc|c|>{\columncolor{lightblue}}c|c|>{\columncolor{lightblue}}c|c|>{\columncolor{lightblue}}c}
    \toprule
    \multicolumn{2}{c}{Method~$\rightarrow$} & AprilGan & AprilGan+UCF & AdaCLIP & AdaCLIP+UCF & AnomalyCLIP & AnomalyCLIP+UCF \\
    \cline{1-2}
    \multicolumn{2}{c}{Category~$\downarrow$} & CVPRW'23 & Ours & ECCV'24 & Ours & ICLR'24& Ours\\
    \hline
    \multicolumn{1}{c}{\multirow{10}[1]{*}{\begin{turn}{90}Objects\end{turn}}} 
    & Bottle & 92.0 / 97.7 / 83.5 / 45.6 & 93.5 / 98.0 / 89.2 / 58.2 & \underline{97.7} / \underline{99.3} / \underline{90.8} / 57.6 & \textbf{98.3} / \textbf{99.5} / \underline{90.8} / 74.1 & 88.7 / 96.8 / 90.4 / \underline{80.8} & 89.4 / 96.9 / \textbf{91.5} / \textbf{84.6} \\
    & Cable & \textbf{88.2} / \textbf{92.9} / 72.2 / 25.7 & 78.4 / 87.5 / 61.6 / 35.7 & 63.9 / 79.1 / 78.3 / 35.3 & \underline{80.7} / \underline{89.2} / 71.4 / 60.1 & 70.3 / 81.7 / \underline{78.9} / \underline{64.0} & 80.0 / 86.2 / \textbf{81.5} / \textbf{69.0} \\
    & Capsule & 79.9 / 95.4 / 92.0 / 51.3 & 89.0 / 97.7 / 95.7 / 66.4 & 84.1 / 96.5 / 95.2 / 18.0 & 84.1 / 96.5 / 94.8 / 74.5 & \underline{89.5} / \underline{97.8} / 95.8 / \underline{87.6} & \textbf{94.0} / \textbf{98.8} / 97.4 / \textbf{91.5} \\
    & Hazelnut & 89.4 / 94.6 / 96.1 / 70.2 & 96.8 / 98.4 / 96.4 / 73.7 & 86.9 / 93.0 / 96.5 / 59.2 & 89.3 / 94.6 / 96.5 / 80.9 & \underline{97.2} / \underline{98.5} / \textbf{97.2} / \underline{92.5} & \textbf{97.4} / \textbf{98.6} / \underline{96.9} / \textbf{93.0} \\
    & Metal Nut & 68.4 / 91.8 / 65.4 / 38.4 & 78.1 / 95.0 / 75.9 / 51.5 & 66.6 / 92.1 / 74.4 / 62.4 & 68.4 / 92.6 / 74.6 / 70.1 & \textbf{92.4} / \textbf{98.2} / \underline{74.6} / \underline{71.1} & \underline{90.7} / \underline{97.8} / \textbf{80.0} / \textbf{77.9} \\
\bottomrule
    \end{tabular}%
  }
\end{table*}

\begin{table*}[htbp]
  \centering
  \caption{
  Quantitative comparison for \textbf{multimodal RGB–Text UAD}, reporting per-category results. Zero-shot anomaly detection and localization results on \textbf{MVTec-AD}~\cite{mvtec} using \textbf{I-F1-max/P-AP/P-F1-max}. \textbf{Best} results are in bold and \underline{runner-ups} are underlined. This table corresponds to Sec.\ref{suppsec5c} and complements Table~VII in the main paper.}
    \label{suppmvtectexttable2}%
  \renewcommand{\arraystretch}{1.2}
  \resizebox{1\linewidth}{!}{
    \begin{tabular}{cc|c|>{\columncolor{lightblue}}c|c|>{\columncolor{lightblue}}c|c|>{\columncolor{lightblue}}c}
    \toprule
    \multicolumn{2}{c}{Method~$\rightarrow$} & AprilGan & AprilGan+UCF & AdaCLIP & AdaCLIP+UCF & AnomalyCLIP & AnomalyCLIP+UCF \\
    \cline{1-2}
    \multicolumn{2}{c}{Category~$\downarrow$} & CVPRW'23 & Ours & ECCV'24 & Ours & ICLR'24& Ours\\
    \hline
    \multicolumn{1}{c}{\multirow{10}[1]{*}{\begin{turn}{90}Objects\end{turn}}} 
    & Bottle & 
    92.1 / 53.0 / \underline{53.4} & 
    92.4 / 53.1 / 53.2 & 
    \textbf{96.1} / \textbf{62.5} / \textbf{60.8} & 
    \underline{96.0} / 54.9 / 51.2 & 
    90.9 / 55.3 / 51.6 & 
    90.6 / \underline{57.4} / 53.3 \\
    & Cable & 
    \textbf{84.8} / \underline{18.2} / 23.9 & 
    79.6 / 17.7 / 24.6 & 
    76.0 / \textbf{20.2} / \textbf{26.6} & 
    80.9 / 17.8 / \underline{25.5} & 
    77.4 / 12.3 / 18.9 & 
    \underline{82.1} / 12.5 / 20.2 \\
    & Capsule  & 
    \underline{92.0} / 29.6 / 33.1 & 
    91.5 / 29.8 / 33.3 & 
    91.9 / \underline{30.7} / \underline{35.9} & 
    91.0 / 25.2 / 27.2 & 
    91.7 / 27.7 / 31.0 & 
    \textbf{94.3} / \textbf{37.7} / \textbf{44.8} \\
    & Hazelnut & 
    87.0 / \underline{49.6} / 50.5 & 
    \textbf{95.1} / \textbf{51.6} / \textbf{52.0} & 
    85.1 / 42.5 / 42.1 & 
    87.0 / 42.4 / 42.2 & 
    92.6 / 43.4 / 47.6 & 
    \underline{93.7} / 48.9 / \underline{51.0} \\
    & Metal Nut  & 
    89.4 / 25.9 / 28.0 & 
    89.4 / \underline{28.2} / 34.3 & 
    89.4 / 24.5 / 35.3 & 
    89.4 / 24.5 / \textbf{37.0} & 
    \textbf{93.7} / 26.5 / 33.1 & 
    \underline{92.6} / \textbf{32.2 }/ \underline{36.8} \\
    & Pill  & 
    91.6 / 23.6 / 27.7 & 
    92.4 / 27.6 / 28.3 & 
    \textbf{93.9} / 32.2 / 35.7 & 
    93.3 / 31.6 / 34.0 & 
    92.1 / \textbf{34.1} / \underline{35.5} & 
    \underline{93.7} / \underline{33.7} / \textbf{38.1} \\
\bottomrule
    \end{tabular}%
  }
\end{table*}

\begin{table*}[htbp]
  \centering
  \caption{Quantitative comparison for \textbf{multimodal RGB–Text UAD}, reporting per-category results. Zero-shot anomaly detection and localization results on \textbf{VisA}~\cite{visa} using \textbf{I-AUROC/I-AP/P-AUROC/AUPRO}. \textbf{Best} results are in bold and \underline{runner-ups} are underlined. This table corresponds to Sec.\ref{suppsec5c} and complements Table~VII in the main paper.
  }
  \label{suppvisatexttable1}%
  \renewcommand{\arraystretch}{1.6}
  \resizebox{1\linewidth}{!}{
    \begin{tabular}{cc|c|>{\columncolor{lightblue}}c|c|>{\columncolor{lightblue}}c|c|>{\columncolor{lightblue}}c}
    \toprule
    \multicolumn{2}{c}{Method~$\rightarrow$} & AprilGan & AprilGan+UCF & AdaCLIP & AdaCLIP+UCF & AnomalyCLIP & AnomalyCLIP+UCF \\
    \cline{1-2}
    \multicolumn{2}{c}{Category~$\downarrow$} & CVPRW'23 & Ours & ECCV'24 & Ours & ICLR'24& Ours\\
    \hline
    \multicolumn{1}{c}{\multirow{4}[1]{*}{\begin{turn}{90}{\shortstack{Complex\\Structure}} \end{turn}}}
    & PCB1 & 50.6 / 55.0 / 92.8 / \textbf{87.3} & 73.0 / 75.1 / \underline{93.5} / \underline{86.8} & \textbf{86.4} / \textbf{87.3} / 91.2 / 36.0 & 81.1 / 82.3 / 81.7 / 68.2 & \underline{85.2} / \underline{86.7} / \textbf{94.0} / 80.7 & 79.3 / 82.1 / 93.3 / 69.5 \\
    & PCB2 & 71.6 / 73.7 / 89.8 / 75.4 & 79.5 / 81.0 / 90.4 / 76.0 & \underline{80.2} / \underline{81.2} / 91.4 / 40.3 & \textbf{86.1} / \textbf{87.9} / 92.0 / 73.2 & 62.0 / 64.4 / \underline{92.4} / \textbf{78.9} & 67.7 / 69.6 / \textbf{92.7} / \underline{78.4} \\
    & PCB3 & 67.0 / 70.6 / 88.2 / \underline{77.2} & 71.2 / 73.1 / 88.0 / \textbf{78.1} & \textbf{76.6} / \textbf{78.4} / \textbf{90.8} / 38.5 & \underline{74.3} / \underline{77.4} / 88.6 / 63.9 & 61.7 / 69.4 / 88.4 / 76.8 & 66.1 / 70.6 / \underline{88.7} / 77.0 \\
    & PCB4 & 94.7 / 94.8 / 94.5 / 86.6 & 97.0 / 96.9 / 94.9 / 87.2 & 96.8 / 96.9 / \textbf{96.8} / 52.8 & \underline{97.8} / \underline{97.7} / 96.2 / 83.5 & 93.9 / 94.3 / 95.7 / \underline{89.4} & \textbf{98.1} / \textbf{97.8} / \underline{96.6} / \textbf{90.6} \\
    \hline
    \multicolumn{1}{c}{\multirow{4}[1]{*}{\begin{turn}{90}{\shortstack{Multiple\\Instances}} \end{turn}}}
    & Macaroni1 & 69.7 / 67.7 / \underline{98.8} / \underline{93.0} & 84.0 / \underline{86.4} / \underline{98.8} / \textbf{93.7} & 82.7 / 79.4 / \textbf{99.3} / 61.9 & \textbf{87.0} / \textbf{88.0} / 98.1 / 92.2 & \underline{86.7} / 85.5 / 98.3 / 89.8 & 74.1 / 77.5 / 98.0 / 88.4 \\
\bottomrule
    \end{tabular}
  }
\end{table*}

\begin{table*}[htbp]
  \centering
  \caption{Quantitative comparison for \textbf{multimodal RGB–Text UAD}, reporting per-category results. Zero-shot anomaly detection and localization results on \textbf{VisA}~\cite{visa} using \textbf{I-F1-max/P-AP/P-F1-max}. \textbf{Best} results are in bold and \underline{runner-ups} are underlined. This table corresponds to Sec.\ref{suppsec5c} and complements Table~VII in the main paper.
  }
  \label{suppvisatexttable2}%
  \renewcommand{\arraystretch}{1.2}
  \resizebox{1\linewidth}{!}{
    \begin{tabular}{cc|c|>{\columncolor{lightblue}}c|c|>{\columncolor{lightblue}}c|c|>{\columncolor{lightblue}}c}
    \toprule
    \multicolumn{2}{c}{Method~$\rightarrow$} & AprilGan & AprilGan+UCF & AdaCLIP & AdaCLIP+UCF & AnomalyCLIP & AnomalyCLIP+UCF \\
    \cline{1-2}
    \multicolumn{2}{c}{Category~$\downarrow$} & CVPRW'23 & Ours & ECCV'24 & Ours & ICLR'24& Ours\\
    \hline
    \multirow{4}[1]{*}{\begin{turn}{90}{\shortstack{Complex\\Structure}} \end{turn}}
    & PCB1 & 66.9 / 8.3 / 12.2 & 69.5 / 8.4 / 12.2 & \underline{78.5} / \textbf{13.2} / 21.5 & 75.0 / \underline{11.2} / 20.8 & \textbf{78.8} / 8.6 / 12.7 & 74.2 / 10.8 / 18.9 \\
    & PCB2 & 69.1 / 15.2 / 23.3 & 73.4 / 15.5 / 23.5 & \underline{75.5} / \textbf{21.9} / \underline{31.9} & \textbf{78.3} / \underline{21.7} / \textbf{32.2} & 67.8 / 9.1 / 15.8 & 71.2 / 10.4 / 18.0 \\
    & PCB3 & 66.7 / 14.5 / 21.9 & 70.6 / 14.1 / 21.4 & \textbf{74.7} / \textbf{24.1} / \textbf{32.2} & \underline{71.2} / \underline{23.2} / \underline{32.1} & 66.4 / 4.3 / 9.3 & 67.4 / 5.0 / 11.7 \\
    & PCB4 & 87.1 / 24.8 / 31.0 & 91.9 / 26.6 / 32.6 & 90.4 / \textbf{38.7} / \underline{42.7} & \underline{93.7} / \underline{38.1} / \textbf{43.2} & 87.8 / 30.6 / 34.7 & \textbf{95.2} / 36.3 / 40.9 \\
    \hline
    \multirow{4}[1]{*}{\begin{turn}{90}{\shortstack{Multiple\\Instances}} \end{turn}}
    & Macaroni1 & 70.8 / \underline{24.9} / \underline{35.3} & 76.3 / \textbf{26.3} / \textbf{37.0} & 78.5 / 23.3 / 33.3 & \textbf{81.0} / 22.5 / 33.0 & \underline{80.4} / 14.9 / 23.7 & 70.5 / 15.2 / 25.4 \\
    & Macaroni2 & \underline{69.3} / \textbf{7.0} / \textbf{14.0} & 69.2 / 5.4 / \underline{12.4} & 67.8 / \underline{6.8} / 12.0 & 68.8 / 6.0 / 11.2 & \textbf{71.2} / 1.5 / 5.1 & 68.6 / 1.2 / 3.9 \\
\bottomrule
    \end{tabular}
  }
\end{table*}

\begin{table*}[!t]
    \centering
    \caption{Quantitative comparison for \textbf{multimodal RGB–Text UAD}, reporting per-category results. Zero-shot anomaly detection and localization results on \textbf{MPDD}~\cite{mpdd} using \textbf{I-AUROC/I-AP/P-AUROC/AUPRO}. \textbf{Best} results are in bold and \underline{runner-ups} are underlined. This table corresponds to Sec.\ref{suppsec5c} and complements Table~VII in the main paper.
    }
    \label{suppmpddtexttable1}
    \renewcommand{\arraystretch}{1.6}
    \resizebox{1\linewidth}{!}{
      \begin{tabular}{c|c|>{\columncolor{lightblue}}c|c|>{\columncolor{lightblue}}c|c|>{\columncolor{lightblue}}c}
      \toprule
      \multicolumn{1}{c}{Method~$\rightarrow$} & AprilGan & AprilGan+UCF  & AdaCLIP & AdaCLIP+UCF & AnomalyCLIP & AnomalyCLIP+UCF \\
      \cline{1-1}
      \multicolumn{1}{c}{Category~$\downarrow$} & CVPRW'23 & Ours & ECCV'24 & Ours & ICLR'24& Ours \\
      \hline
      Bracket black & 
      \underline{68.6} / 72.3 / \underline{96.3} / \underline{90.6} & 
      65.4 / 72.5 / \textbf{97.1} / \textbf{92.3} & 
      54.6 / 64.4 / 95.5 / 16.1 & 
      46.5 / 56.3 / 95.4 / 59.0 &
      67.8 / \underline{73.4} / 95.7 / 85.2 &
      \textbf{76.5} / \textbf{84.4} / 96.1 / 88.9 \\
      Bracket brown & 
      61.8 / 78.1 / 87.4 / 72.6 & 
      \underline{62.5} / \underline{81.1} / \textbf{94.6} / \textbf{83.5} & 
      50.6 / 71.1 / 93.6 / 9.3 & 
      52.8 / 71.0 / 93.8 / 46.7 &
      62.0 / 80.4 / \underline{94.4} / \underline{77.8} &
      \textbf{64.4} / \textbf{81.3} / 92.8 / 75.9 \\
      Bracket white & 
      \textbf{85.6} / \textbf{88.2} / \underline{99.2} / \underline{94.0} & 
      74.8 / 73.6 / 98.7 / 92.5 & 
      53.4 / 50.9 / 98.0 / 7.3 & 
      53.3 / 51.9 / 97.7 / 72.3 &
      67.7 / 71.6 / \textbf{99.8} / \textbf{98.8} &
      \underline{83.3} / \underline{81.7} / \textbf{99.8} / \textbf{98.8} \\
      Connector     & 
      78.1 / 72.7 / 90.6 / 74.4 &
      81.2 / \underline{75.8} / 92.0 / 75.7 & 
      79.1 / 69.8 / \textbf{97.7} / 34.9 &
      \underline{81.4} / 70.9 / 91.2 / 62.0 &
      \textbf{87.4} / \textbf{77.0} / 97.2 / \underline{89.9} &
      79.8 / 61.8 / \underline{97.3} / \textbf{90.5} \\
      Metal plate   & 
      70.0 / 86.5 / 93.0 / 74.5 & 
      75.6 / 89.8 / \underline{94.0} / 85.0 & 
      81.3 / 92.7 / 92.6 / 30.1 & 
      \underline{84.3} / \underline{94.2} / 87.5 / 48.4 &
      \textbf{84.7} / \textbf{94.4} / 93.7 / \underline{86.8} &
      80.1 / 92.7 / \textbf{96.4} / \textbf{91.9} \\
\bottomrule
      \end{tabular}
    }
\end{table*}

\begin{table*}[!t]
    \centering
    \caption{Quantitative comparison for \textbf{multimodal RGB–Text UAD}, reporting per-category results. Zero-shot anomaly detection and localization results on \textbf{MPDD}~\cite{mpdd} using \textbf{I-F1-MAX/P-AP/P-F1-MAX}. \textbf{Best} results are in bold and \underline{runner-ups} are underlined. This table corresponds to Sec.\ref{suppsec5c} and complements Table~VII in the main paper.
    }
    \label{suppmpddtexttable2}
    \renewcommand{\arraystretch}{1.2}
    \resizebox{1\linewidth}{!}{
      \begin{tabular}{c|c|>{\columncolor{lightblue}}c|c|>{\columncolor{lightblue}}c|c|>{\columncolor{lightblue}}c}
      \toprule
      \multicolumn{1}{c}{Method~$\rightarrow$} & AprilGan & AprilGan+UCF  & AdaCLIP & AdaCLIP+UCF & AnomalyCLIP & AnomalyCLIP+UCF \\
      \cline{1-1}
      \multicolumn{1}{c}{Category~$\downarrow$} & CVPRW'23 & Ours & ECCV'24 & Ours & ICLR'24& Ours \\
      \hline
      Bracket black & 
      \underline{80.7} / 6.3 / 15.8 & 
      77.1 / 5.2 / 14.6 & 
      75.8 / 3.4 / 10.2 & 
      75.2 / 3.0 / 10.8 &
      78.6 / \textbf{17.3} / \textbf{27.2} &
      \textbf{81.6} / \underline{15.6} / \underline{26.6} \\
      Bracket brown & 
      \textbf{81.0} / 3.5 / 8.7 & 
      80.3 / 5.2 / 10.5 & 
      79.7 / \underline{5.8} / \underline{12.4} & 
      \underline{80.7} / 4.3 / 8.9 &
      80.3 / \textbf{7.0} / \textbf{13.1} &
      80.3 / 3.1 / 7.5 \\
      Bracket white & 
      \underline{78.1} / 3.9 / 8.9 & 
      75.3 / 1.9 / 5.2 & 
      68.2 / 1.3 / 3.3 & 
      66.7 / 1.2 / 3.7 &
      69.8 / \underline{15.9} / \underline{22.9} &
      \textbf{80.7} / \textbf{17.3} / \textbf{25.3} \\
      Connector     & 
      64.3 / 15.8 / 22.4 &
      68.6 / 15.0 / 22.3 & 
      66.7 / \textbf{36.4} / \textbf{39.1} &
      68.4 / \underline{28.0} / \underline{37.6} &
      \textbf{73.7} / 16.8 / 27.0 &
      \underline{70.3} / 23.1 / 36.5 \\
      Metal plate   & 
      87.0 / 57.7 / 63.1 & 
      86.8 / 60.4 / \underline{63.9} & 
      \textbf{88.2} / 53.9 / 60.4 & 
      \underline{88.0} / 45.9 / 45.2 &
      87.5 / \underline{64.0} / 61.9 &
      86.7 / \textbf{77.0} / \textbf{73.2} \\
      Tubes         & 
      91.7 / 72.3 / 68.7 & 
      \textbf{97.8} / 72.7 / 69.1 & 
      93.9 / \textbf{77.8} / \underline{71.4} &
      \underline{95.0} / \underline{77.2} / \textbf{71.7} &
      92.3 / 52.5 / 53.3 &
      94.9 / 53.9 / 52.3 \\
\bottomrule
      \end{tabular}
    }
\end{table*}

\begin{table*}[htbp]
    \centering
    \caption{Quantitative comparison for \textbf{multimodal RGB–Text UAD}, reporting per-category results. Zero-shot anomaly detection and localization results on \textbf{BTAD}~\cite{btad} using \textbf{I-AUROC/I-AP/P-AUROC/AUPRO}. \textbf{Best} results are in bold and \underline{runner-ups} are underlined. This table corresponds to Sec.\ref{suppsec5c} and complements Table~VII in the main paper.
    }
    \label{suppbtadtexttable1} 
    \renewcommand{\arraystretch}{1.6}
    \resizebox{1\linewidth}{!}{
      \begin{tabular}{c|c|>{\columncolor{lightblue}}c|c|>{\columncolor{lightblue}}c|c|>{\columncolor{lightblue}}c}
      \toprule
      \multicolumn{1}{c}{Method~$\rightarrow$} & AprilGan & AprilGan+UCF  & AdaCLIP & AdaCLIP+UCF & AnomalyCLIP & AnomalyCLIP+UCF \\
      \cline{1-1}
      \multicolumn{1}{c}{Category~$\downarrow$} & CVPRW'23 & Ours & ECCV'24 & Ours &ICLR'24& Ours \\
      \hline
      01 & 
      81.7 / 92.1 / 89.9 / 72.3 & 
      89.4 / 96.1 / 90.4 / 65.4 &
      93.5 / 97.8 / 88.8 / 1.6 &
      \underline{94.6} / \underline{98.1} / \underline{94.1} / 26.7 & 
      90.9 / 96.6 / 93.7 / \underline{73.0} & 
      \textbf{97.6} / \textbf{99.1} / \textbf{95.7} / \textbf{77.6} \\
      02 & 
      82.0 / 96.8 / 86.3 / 50.3 & 
      75.9 / 95.9 / 91.9 / 54.5 &
      79.1 / 96.6 / \underline{95.9} / 12.0 &
      79.9 / 96.7 / \textbf{96.3} / 27.6 &
      \underline{84.1} / \underline{97.4} / 94.4 / \underline{66.0} & 
      \textbf{87.6} / \textbf{98.1} / 94.6 / \textbf{74.1}  \\
      03 &
      57.9 / 19.7 / 91.8 / \underline{83.6} & 
      80.4 / 47.0 / 93.0 / 82.3 &
      \textbf{97.7} / \textbf{87.8} / \underline{96.4} / 47.1 &
      \underline{96.5} / \underline{82.6} / \textbf{97.2} / 70.6 &
      89.8 / 70.7 / 94.6 / \textbf{87.1} &
      91.2 / 75.6 / 88.9 / 74.6  \\
      \hline
      Mean & 
      73.8 / 69.5 / 89.3 / \underline{68.7} & 
      81.9 / 79.7 / 91.7 / 67.4 & 
      90.1 / \textbf{94.0} / 93.7 / 20.2 &
      \underline{90.3} / \underline{92.5} / \textbf{95.9} / 41.6 &
      88.2 / 88.2 / \underline{94.2} / \textbf{75.4} & 
      \textbf{92.1} / 90.9 / 93.1 / \textbf{75.4}  \\
      \bottomrule
      \end{tabular}%
    }
\end{table*}

\begin{table*}[htbp]
    \centering
    \caption{Quantitative comparison for \textbf{multimodal RGB–Text UAD}, reporting per-category results. Zero-shot anomaly detection and localization results on \textbf{BTAD}~\cite{btad} using \textbf{I-F1-max/P-AP/P-F1-max}. \textbf{Best} results are in bold and \underline{runner-ups} are underlined. This table corresponds to Sec.\ref{suppsec5c} and complements Table~VII in the main paper.
    }
    \label{suppbtadtexttable2} 
    \renewcommand{\arraystretch}{1.2}
    \resizebox{1\linewidth}{!}{
    {
      \begin{tabular}{c|c|>{\columncolor{lightblue}}c|c|>{\columncolor{lightblue}}c|c|>{\columncolor{lightblue}}c}
      \toprule
      \multicolumn{1}{c}{Method~$\rightarrow$} & AprilGan & AprilGan+UCF  & AdaCLIP & AdaCLIP+UCF & AnomalyCLIP & AnomalyCLIP+UCF \\
      \cline{1-1}
      \multicolumn{1}{c}{Category~$\downarrow$} & CVPRW'23 & Ours & ECCV'24 & Ours & ICLR'24& Ours \\
      \hline
      01 & 
      84.3 / 51.3 / 53.3 & 
      88.7 / \underline{52.2} / \underline{54.3} & 
      \underline{94.9} / 49.7 / 54.0 & 
      92.6 / 47.0 / 47.1 &
      89.4 / 47.8 / 52.9 & 
      \textbf{96.8} / \textbf{52.8} / \textbf{55.6} \\
      02 & 
      \textbf{93.5} / 51.9 / 56.7 & 
      \underline{93.4} / 58.6 / 59.1 &
      93.0 / \textbf{65.9} / \textbf{64.0} & 
      93.0 / \underline{65.6} / \underline{63.6} &
      93.3 / 61.0 / 60.0 & 
      93.0 / 61.3 / 62.5 \\
      03 & 
      26.5 / 6.4 / 12.0 & 
      47.5 / 4.5 / 8.8 & 
      \textbf{79.5} / \underline{29.2} / \underline{38.4} & 
      \underline{77.3} / \textbf{29.6} / \textbf{38.8} &
      68.6 / 27.8 / 36.4 & 
      69.2 / 19.8 / 30.2 \\
      \hline
      Mean & 
      68.1 / 36.5 / 40.6 & 
      76.5 / 38.4 / 40.8 & 
      \textbf{89.1} / \textbf{48.2} / \textbf{52.2} & 
      \underline{87.7} / \underline{47.4} / \underline{49.8} &
      83.8 / 45.5 / 49.7 & 
      86.4 / 44.6 / 49.4 \\
      \bottomrule
      \end{tabular}%
    }
   }
\end{table*}

\begin{table*}[htbp]
    \centering
    \caption{Quantitative comparison for \textbf{multimodal RGB–Text UAD}, reporting per-category results. Zero-shot anomaly detection and localization results on \textbf{SDD}~\cite{ksdd} using \textbf{I-AUROC/I-AP/P-AUROC/AUPRO}. \textbf{Best} results are in bold and \underline{runner-ups} are underlined. This table corresponds to Sec.\ref{suppsec5c} and complements Table~VII in the main paper.
    }
    \label{suppsddtexttable1} 
    \renewcommand{\arraystretch}{1.6}
    \resizebox{1\linewidth}{!}{
\begin{tabular}{cc>{\columncolor{lightblue}}c|c>{\columncolor{lightblue}}c|c>{\columncolor{lightblue}}c}
      \toprule
       \multicolumn{1}{c}{Method~$\rightarrow$} & AprilGan & AprilGan+UCF  & AdaCLIP & AdaCLIP+UCF & AnomalyCLIP & AnomalyCLIP+UCF \\
      \cline{1-1}
      \multicolumn{1}{c}{Single Category~$\downarrow$} & CVPRW'23 & Ours & ECCV'24 & Ours & ICLR'24& Ours \\
      \hline
      Results & 
      96.8 / 92.3 / 92.8 / 84.3  & 
      95.2 / 87.2 / 94.5 / 88.7 &
      95.6 / 88.4 / 96.0 / 33.8 &
      95.8 / 89.2 / \underline{97.6} / 60.9 & 
      \underline{97.8} / \textbf{94.2} / \textbf{98.1} / \textbf{94.9} & 
      \textbf{97.9} / \underline{93.5} / 97.5 / \underline{91.8} \\
     
      \bottomrule
      \end{tabular}%
    }
\end{table*}

\begin{table*}[htbp]
    \centering
    \caption{Quantitative comparison for \textbf{multimodal RGB–Text UAD}, reporting per-category results. Zero-shot anomaly detection and localization results on \textbf{SDD}~\cite{ksdd} using \textbf{I-F1-max/P-AP/P-F1-max}. \textbf{Best} results are in bold and \underline{runner-ups} are underlined. This table corresponds to Sec.\ref{suppsec5c} and complements Table~VII in the main paper.}
    \label{suppsddtexttable2} 
    \renewcommand{\arraystretch}{1.2}
    \resizebox{1\linewidth}{!}{
      \begin{tabular}{cc>{\columncolor{lightblue}}c|c>{\columncolor{lightblue}}c|c>{\columncolor{lightblue}}c}
      \toprule
       \multicolumn{1}{c}{Method~$\rightarrow$} & AprilGan & AprilGan+UCF  & AdaCLIP & AdaCLIP+UCF & AnomalyCLIP & AnomalyCLIP+UCF \\
      \cline{1-1}
      \multicolumn{1}{c}{Single Category~$\downarrow$} & CVPRW'23 & Ours & ECCV'24 & Ours & ICLR'24& Ours \\
      \hline
      Results & 
      \underline{89.8} / 36.8 / 44.4 & 
      82.5 / 37.2 / 42.5 &
      82.4 / 45.5 / 51.3 &
      83.5 / 45.6 / 52.3 & 
      89.7 / \underline{51.9} / \underline{56.5} & 
      \textbf{91.6} / \textbf{52.7} / \textbf{59.8} \\
     
      \bottomrule
      \end{tabular}%
    }
\end{table*}

\begin{table*}[htbp]
    \centering
    \caption{Quantitative comparison for \textbf{multimodal RGB–Text UAD}, reporting per-category results. Zero-shot anomaly detection and localization results on \textbf{DAGM}~\cite{dagm} using \textbf{I-AUROC/I-AP/P-AUROC/AUPRO}. \textbf{Best} results are in bold and \underline{runner-ups} are underlined. This table corresponds to Sec.\ref{suppsec5c} and complements Table~VII in the main paper.}
    \label{suppdagmtexttable1} 
    \renewcommand{\arraystretch}{1.6}
    \resizebox{1\linewidth}{!}{
    {
      \begin{tabular}{c|c|>{\columncolor{lightblue}}c|c|>{\columncolor{lightblue}}c|c|>{\columncolor{lightblue}}c}
      \toprule
      \multicolumn{1}{c}{Method~$\rightarrow$} & AprilGan & AprilGan+UCF  & AdaCLIP & AdaCLIP+UCF & AnomalyCLIP & AnomalyCLIP+UCF \\
      \cline{1-1}
      \multicolumn{1}{c}{Category~$\downarrow$} & CVPRW'23 & Ours & ECCV'24 & Ours & ICLR'24& Ours \\
      \hline
      Class1 & 
      81.1 / 78.7 / 77.1 / 58 &
      95.8 / 95.6 / 77.8 / 60.9 & 
      89.8 / 85.6 / 86.6 / 36.1 & 
      \textbf{97.2} / \underline{97.3} / \textbf{93.0} / 49.4 &
      85.5 / 83.7 / 88.6 / \textbf{78.2} & 
      \underline{96.9} / \textbf{97.5} / \underline{90.2} / \underline{73.5} \\
      Class2 & 
      94.9 / 95.8 / 85.6 / 82.1 &
      95.7 / 96.8 / 83.1 / 73.3 &
      \underline{99.4} / \underline{99.2} / 95.0 / 46.7 & 
      \textbf{100.0} / \textbf{100.0} / 95.6 / 66.2 &
      \textbf{100.0} / \textbf{100.0} / \underline{99.0} / \underline{98.1} & 
      \textbf{100.0} / \textbf{100.0} / \textbf{99.4} / \textbf{99.2} \\
      Class3 & 
      98.6 / 98.7 / 79.9 / 65 &
      \underline{99.9} / \underline{99.9} / 80.7 / 67.1 & 
      \textbf{100.0} / \textbf{100.0} / 92.2 / 29.0 & 
      \textbf{100.0} / \textbf{100.0} / \underline{96.7} / 36.7 &
      \underline{99.9} / \underline{99.9} / 95.9 / \underline{93.6} & 
      \textbf{100.0} / \textbf{100.0} / \textbf{97.6} / \textbf{95.5} \\
      Class4 & 
      97.2 / 97.3 / 75.6 / 44.8 &
      97.8 / \underline{97.9} / 82.4 / 53.5 &
      95.5 / 95.8 / \underline{90.0} / 11.3 &
      65.7 / 67.7 / \textbf{93.8} / 31.7 &
      \textbf{98.7} / \textbf{98.9} / 88.1 / \textbf{73.5} &
      \underline{98.5} / \textbf{98.9} / 89.8 / \underline{68.8} \\
\bottomrule
      \end{tabular}%
    }
   }
\end{table*}

\begin{table*}[htbp]
    \centering
    \caption{Quantitative comparison for \textbf{multimodal RGB–Text UAD}, reporting per-category results. Zero-shot anomaly detection and localization results on \textbf{DAGM}~\cite{dagm} using \textbf{I-F1-max/P-AP/P-F1-max}. \textbf{Best} results are in bold and \underline{runner-ups} are underlined. This table corresponds to Sec.\ref{suppsec5c} and complements Table~VII in the main paper.
   }
    \label{suppdagmtexttable2} 
    \renewcommand{\arraystretch}{1.2}
    \resizebox{1\linewidth}{!}{
    {
      \begin{tabular}{c|c|>{\columncolor{lightblue}}c|c|>{\columncolor{lightblue}}c|c|>{\columncolor{lightblue}}c}
      \toprule
      \multicolumn{1}{c}{Method~$\rightarrow$} & AprilGan & AprilGan+UCF  & AdaCLIP & AdaCLIP+UCF & AnomalyCLIP & AnomalyCLIP+UCF \\
      \cline{1-1}
      \multicolumn{1}{c}{Category~$\downarrow$} & CVPRW'23 & Ours & ECCV'24 & Ours & ICLR'24& Ours \\
      \hline
      Class1 & 
      77.8 / 35.4 / 40.9 & 
      90.7 / 34.2 / 38.9 & 
      85.4 / 55.4 / \underline{57.9} & 
      \underline{92.3} / \underline{56.7} / 52.6 &
      80.8 / \textbf{58.2} / \textbf{59.4} & 
      \textbf{93.2} / 45.7 / 50.0 \\
      Class2 & 
      88.1 / 51.6 / 55.6 & 
      90.2 / 31.1 / 39.8 &
      98.7 / \textbf{68.2} / \textbf{67.8} & 
      99.0 / 55.5 / 60.6 &
      \textbf{100.0} / 61.5 / 62.4 & 
      \underline{99.3} / \underline{66.0} / \underline{65.0} \\
      Class3 & 
      95.0 / 35.3 / 42.8 & 
      99.0 / 36.5 / 43.2 & 
      \textbf{100.0} / \underline{70.9} / \textbf{72.9} & 
      \textbf{100.0} / 70.5 / \underline{71.7} &
      98.7 / 70.5 / 70.2 & 
      \underline{99.7} / \textbf{75.7} / \underline{71.7} \\
      Class4 & 
      92.6 / 8 / 15.4 & 
      93.2 / 12.4 / 17.7 & 
      90.0 / 16.5 / \underline{33.7} & 
      67.9 / 17.4 / 31.5 &
      \underline{94.9} / \textbf{42.1} / \textbf{45.6} & 
      \textbf{95.2} / \underline{23.8} / 29.6 \\
      Class5 & 
      \underline{99.7} / 51 / 58.5 & 
      \underline{99.7} / 53.2 / 60.9 & 
      \textbf{100.0} / \underline{82.5} / \textbf{77.9} & 
      \textbf{100.0} / 77.0 / 71.9 &
      \textbf{100.0} / 74.9 / 72.9 & 
      \textbf{100.0} / \textbf{83.0} / \underline{77.5} \\
      Class6 & 
      \textbf{100.0} / 76.5 / 77.4 & 
      \textbf{100.0} / \underline{86.8} / \underline{82.2} & 
      \textbf{100.0} / 84.4 / 81.2 & 
      \textbf{100.0} / 83.4 / 75.8 &
      \textbf{100.0} / 81.9 / 76.7 & 
      \textbf{100.0} / \textbf{89.4} / \textbf{82.3} \\
\bottomrule
      \end{tabular}%
    }
   }
\end{table*}

\begin{table*}[htbp]
    \centering
    \caption{Quantitative comparison for \textbf{multimodal RGB–Text UAD}, reporting per-category results. Zero-shot anomaly detection and localization results on \textbf{DTD-Synthetic}~\cite{dtd} using \textbf{I-AUROC/I-AP/P-AUROC/AUPRO}. \textbf{Best} results are in bold and \underline{runner-ups} are underlined. This table corresponds to Sec.\ref{suppsec5c} and complements Table~VII in the main paper.
    }
    \label{suppdtdtexttable1} 
    \renewcommand{\arraystretch}{1.6}
    \resizebox{1\linewidth}{!}{
      \begin{tabular}{c|c|>{\columncolor{lightblue}}c|c|>{\columncolor{lightblue}}c|c|>{\columncolor{lightblue}}c}
      \toprule
      \multicolumn{1}{c}{Method~$\rightarrow$} & AprilGan & AprilGan+UCF  & AdaCLIP & AdaCLIP+UCF & AnomalyCLIP & AnomalyCLIP+UCF \\
      \cline{1-1}
      \multicolumn{1}{c}{Category~$\downarrow$} & CVPRW'23 & Ours & ECCV'24 & Ours & ICLR'24& Ours \\
      \hline
      Blotchy099 & 
      92.6 / 98.2 / 99.0 / 90.7 & 
      80.9 / 95.4 / 99.4 / 94 & 
      98.9 / \underline{99.8} / 99.0 / 80.0 & 
      \underline{99.2} / \underline{99.8} / 99.4 / 90.2 &
      98.9 / 99.7 / \underline{99.5} / 96.2 &
      \textbf{99.9} / \textbf{100.0} / \textbf{99.6} / 95.4 \\
      Fibrous183 & 
      89.3 / 97.2 / 97.5 / 90.4 & 
      87.8 / 96.7 / 98.5 / 95.3 &
      89.5 / 97.3 / \underline{99.2} / 92.1 & 
      92.6 / 98.1 / \underline{99.3} / 97.2 &
      \underline{97.2} / \underline{99.3} / \textbf{99.4} / \textbf{98.2} &
      \textbf{99.1} / \textbf{99.8} / 99.0 / \underline{97.8} \\
      Marbled078 & 
      \underline{98.9} / \underline{99.7} / 98.1 / 95.5 & 
      98.2 / 99.5 / 98.8 / 96.7 & 
      98.5 / \underline{99.7} / \textbf{99.3} / 70.3 & 
      98.8 / \underline{99.7} / \underline{99.7} / 93.3 &
      98.7 / \underline{99.7} / \underline{99.1} / \textbf{97.1} &
      \textbf{99.6} / \textbf{99.9} / 98.8 / \underline{96.8} \\
      Matted069 &
      74.7 / 92.4 / 95.2 / 81.8 & 
      75.8 / 93.6 / 96.4 / 83 &
      78.4 / 92.0 / \underline{98.5} / 81.0 & 
      \underline{90.5} / \underline{97.3} / 98.0 / \underline{83.7} &
      82.6 / 95.2 / \textbf{99.6} / \textbf{84.8} &
      \textbf{93.9} / \textbf{98.5} / 97.2 / 79.1 \\
      Mesh114 &
      72.7 / 87.7 / 89.0 / 73.0 & 
      85.0 / 94.3 / 91.1 / 75.1 &
      87.7 / 94.9 / 94.6 / 38.7 & 
      \underline{88.2} / \underline{95.3} / 93.2 / 72.5 &
      85.8 / 94.5 / \underline{95.2} / \underline{77.0} &
      \textbf{89.8} / \textbf{96.1} / \textbf{95.8} / \textbf{80.5} \\
\bottomrule
      \end{tabular}%
    }
\end{table*}

\begin{table*}[htbp]
    \centering
    \caption{Quantitative comparison for \textbf{multimodal RGB–Text UAD}, reporting per-category results. Zero-shot anomaly detection and localization results on \textbf{DTD-Synthetic}~\cite{dtd} using \textbf{I-F1-max/P-AP/P-F1-max}. \textbf{Best} results are in bold and \underline{runner-ups} are underlined. This table corresponds to Sec.\ref{suppsec5c} and complements Table~VII in the main paper.
    }
    \label{suppdtdtexttable2} 
    \renewcommand{\arraystretch}{1.2}
    \resizebox{1\linewidth}{!}{
      \begin{tabular}{c|c|>{\columncolor{lightblue}}c|c|>{\columncolor{lightblue}}c|c|>{\columncolor{lightblue}}c}
      \toprule
      \multicolumn{1}{c}{Method~$\rightarrow$} & AprilGan & AprilGan+UCF  & AdaCLIP & AdaCLIP+UCF & AnomalyCLIP & AnomalyCLIP+UCF \\
      \cline{1-1}
      \multicolumn{1}{c}{Category~$\downarrow$} & CVPRW'23 & Ours & ECCV'24 & Ours & ICLR'24& Ours \\
      \hline
      Blotchy099 & 
      92.0 / 71.0 / 68.5 & 
      89.4 / 76.1 / 71.5 & 
      98.7 / 79.8 / 75.2 & 
      98.1 / \textbf{82.6} / \underline{76.6} &
      \underline{98.8} / 69.7 / 67.5 & 
      \textbf{99.4} / \underline{82.4} / \textbf{79.6} \\
      Fibrous183 & 
      92.8 / 56.1 / 55.7 & 
      93.0 / 67.7 / 62.2 &
      91.9 / \underline{81.2} / 74.3 & 
      94.6 / \textbf{81.5} / \textbf{74.7} &
      \underline{95.7} / 72.9 / 69.2 & 
      \textbf{97.4} / 76.3 / \underline{74.6} \\
      Marbled078 & 
      \underline{98.1} / 77.4 / 73.3 & 
      97.5 / 80.9 / \underline{74.0} & 
      \underline{98.1} / \underline{81.7} / 73.3 & 
      98.2 / \textbf{82.4} / \textbf{75.9} &
      97.5 / 59.0 / 62.0 & 
      \textbf{98.7} / 70.5 / 71.7 \\
      Matted069 & 
      88.8 / 44.1 / 45.1 & 
      89.3 / 47.2 / 49.1 &
      \underline{92.9} / \textbf{70.6} / 65.7 &
      \underline{92.9} / 69.4 / 66.0 &
      91.2 / \underline{69.8} / \underline{66.7} &
      \textbf{93.3} / 69.6 / \textbf{69.2} \\
      Mesh114 & 
      81.6 / 60.6 / \textbf{66.4} & 
      84.0 / 60.8 / \underline{65.5} &
      \underline{87.8} / \underline{62.9} / 65.0 &
      87.5 / 56.0 / 64.0 &
      84.4 / 53.4 / 56.5 &
      \textbf{88.3} / \textbf{63.1} / 63.7 \\
      Perforated037 & 
      88.9 / 61.0 / \textbf{68.1} & 
      90.3 / 59.3 / 62.4 &
      89.9 / \textbf{67.9} / \underline{67.5} &
      92.1 / 59.6 / 58.0 &
      \underline{92.5} / 63.4 / 63.1 &
      \textbf{93.8} / \underline{67.8} / 66.1 \\
\bottomrule
      \end{tabular}%
    }
\end{table*}

\begin{table*}[htbp]
    \centering
    \caption{Quantitative comparison for \textbf{multimodal RGB–Text UAD}, reporting per-category results. Zero-shot anomaly detection results on three \textbf{medical-domain} datasets~\cite{headct,brainmri,br35h} using \textbf{I-AUROC/I-AP/I-F1max}. \textbf{Best} results are in bold and \underline{runner-ups} are underlined. This table corresponds to Sec.\ref{suppsec5c} and complements Table~VIII in the main paper.
    }
    \label{suppmedicaltexttable1} 
    \renewcommand{\arraystretch}{1.4}
    \resizebox{1\linewidth}{!}{
      \begin{tabular}{cc>{\columncolor{lightblue}}c|c>{\columncolor{lightblue}}c|c>{\columncolor{lightblue}}c}
      \toprule
      \multicolumn{1}{c}{Method~$\rightarrow$} & AprilGan & AprilGan+UCF  & AdaCLIP & AdaCLIP+UCF & AnomalyCLIP & AnomalyCLIP+UCF \\
      \cline{1-1}
       \multicolumn{1}{c}{Datasets~$\downarrow$} & CVPRW'23 & Ours & ECCV'24 & Ours & ICLR'24& Ours \\
      \hline
      HeadCT & 
      86.9 / 87.8 / 81.2  &    90.7 / 91.1 / 83.3     &    \underline{97.3} / \underline{97.4} / 93.2  &    \textbf{98.7} / \textbf{98.8} / \textbf{95.6} &    93.0 / 91.1 / 88.4   &    96.5 / 96.2 / 93.9\\
     BrainMRI & 
      92.7 / 93.7 / 91.0   &    93.7 / 95.5 / 94.0   &    \underline{96.8} / \underline{97.3} / 94.4   &    \textbf{97.3} / \textbf{98.2} / 94.6 &    90.0 / 92.1 / 86.5   &    95.4 / 95.8 / 93.4\\
      Br35H &
      93.2 / 93.9 / 85.6   &    96.8 / \underline{96.9} / 90.7   &    \underline{98.7} / \textbf{98.7} / 94.9    &    \textbf{98.8} / \textbf{98.7}  / \textbf{95.2}   &    94.2 / 94.2  / 86.8 &    97.8 / 97.7 / 93.1 
      \\
      \hline
    Mean & 90.9 / 91.8 / 85.9 & 93.7 / 94.5 / 89.3 & \underline{97.6} / \underline{97.8} / 94.2 & \textbf{98.3} / \textbf{98.6} / 95.1 & 92.4 / 92.5 / 87.2 & 96.6 / 96.6 / 93.5
      \\
      \bottomrule
      \end{tabular}%
    }
\end{table*}

\begin{table*}[htbp]
    \centering
    \caption{Quantitative comparison for \textbf{multimodal RGB–Text UAD}, reporting per-category results. Zero-shot anomaly localization results on six \textbf{medical-domain} datasets~\cite{isic,clidb,clodb,kvasir,endo,tn3k} using \textbf{P-AUROC/AUPRO/P-F1max/P-AP}. \textbf{Best} results are in bold and \underline{runner-ups} are underlined. This table corresponds to Sec.\ref{suppsec5c} and complements Table~VIII in the main paper.
    }
    \label{suppmedicaltexttable2} 
    \renewcommand{\arraystretch}{1.4}
    \resizebox{1\linewidth}{!}{
\begin{tabular}{cc>{\columncolor{lightblue}}c|c>{\columncolor{lightblue}}c|c>{\columncolor{lightblue}}c}
      \toprule
      \multicolumn{1}{c}{Method~$\rightarrow$} & AprilGan & AprilGan+UCF  & AdaCLIP & AdaCLIP+UCF & AnomalyCLIP & AnomalyCLIP+UCF \\
      \cline{1-1}
       \multicolumn{1}{c}{Datasets~$\downarrow$} & CVPRW'23 & Ours & ECCV'24 & Ours & ICLR'24& Ours \\
      \hline
      
      ISIC & 90.0 / 80.2 / 72.6 / 79.6  &    \underline{91.6} / \underline{82.6} / \underline{77.1} / \underline{83.2}   &    90.2 / 18.8 / 72.9 / 76.7    &    89.4 / 45.5 / 72.5 / 75.4    &    89.4 / 78.4 / 71.6 / 76.1   &    \textbf{93.3} / \textbf{85.9} / \textbf{77.5} / \textbf{83.4}
      \\
      CVC-ColonDB &  78.2 / 65.0  / 29.4 / 21.4  &    80.0 / 65.9 / 31.1 / 22.9       &    \textbf{89.8} / \underline{81.0} / \underline{56.6} / \underline{60.0}   &    \textbf{89.8} / \textbf{82.1} / \textbf{57.7} / \textbf{61.3}  &    81.9 / 71.2 / 37.5 / 31.7   &    \underline{83.2} / 75.2 / 38.1 / 32.9 
      \\
      CVC-ClinicDB & 79.2 / 57.0 / 36.9 / 29.1    &    80.6 / 58.6 /  38.6 / 30.9        &    \underline{90.3} / 53.9 / \underline{63.6} / \underline{68.0}   &    \textbf{92.6} / \textbf{72.8} / \textbf{66.6} / \textbf{72.8} &    81.5 / 62.1 / 40.9 / 34.0   &    84.4 / \underline{69.1} / 44.5 / 38.2
      \\
      Kvasir & 75.0 / 36.3 / 40.0 / 32.2   &    78.4 / 39.0 /  46.5 / 36.3       &    \underline{95.1} / 36.2 / \underline{77.0} / \underline{84.5}    &    \textbf{95.2} / \textbf{46.6} / \textbf{77.8} / \textbf{85.1} &    79.0 / \underline{45.4} / 46.2 / 39.6    &    81.7 / 41.9 / 49.5 / 43.8
      \\
      Endo & 81.9 / 54.9 / 44.8 / 38.6   &    84.5 / 61.5 / 51.3 / 42.8       &    \textbf{96.7} / \underline{79.2} / \underline{80.2} / \textbf{87.7}   &    \underline{96.5} / \textbf{87.5} / \textbf{81.9} / \underline{87.4} &    84.2 / 63.4 / 50.3 / 46.6   &    87.1 / 70.0 / 54.7 / 51.3
      \\
\bottomrule
      \end{tabular}%
    }
\end{table*}

\begin{table*}[htbp]
  \centering
  \caption{\textbf{One-shot} evaluation for \textbf{multimodal RGB–Text UAD on MVTec-AD~\cite{mvtec}}, reporting per-category results for \textbf{APRILGAN+UCF}. Results are reported as mean ± standard deviation over five seeds. This table corresponds to Sec.\ref{suppsec4b} and complements Table~\ref{suppfewshotmvtec}.}
    \label{suppfewshot_mvtec1}%
  \renewcommand{\arraystretch}{1.1}
  \resizebox{1\linewidth}{!}{
    \begin{tabular}{c|c|c|c|c|c|c|c|c}
    \toprule
    \multicolumn{2}{r|}{Category~$\downarrow$} & \textbf{I-AUROC} & \textbf{I-AP} & \textbf{I-F1-max} & \textbf{P-AUROC} & \textbf{P-AP} & \textbf{P-F1-max} & \textbf{AUPRO} \\
    \cline{1-2}
    \hline
    \multicolumn{1}{c}{\multirow{10}[1]{*}{\begin{turn}{90}Objects\end{turn}}} 
    & \multicolumn{1}{c|}{Bottle}     & 99.64±0.15 & 99.90±0.07 & 98.72±0.44 & 97.36±0.15 & 81.62±0.59 & 76.10±0.52 & 94.92±0.25 \\
    \multicolumn{1}{c}{} & \multicolumn{1}{c|}{Cable} & 95.30±1.45 & 97.38±0.77 & 91.22±1.14 & 95.14±0.30 & 59.28±0.40 & 59.10±0.46 & 89.12±0.43  \\
    \multicolumn{1}{c}{} & \multicolumn{1}{c|}{Capsule}  &  97.80±0.45 & 99.56±0.09 & 96.46±0.37 & 97.88±0.19 & 47.46±1.49 & 47.34±0.53 & 97.22±0.19\\
    \multicolumn{1}{c}{} & \multicolumn{1}{c|}{Hazelnut} & 99.98±0.04 & 100.00±0.00 & 99.72±0.38 & 98.90±0.19 & 77.28±1.55 & 72.48±1.39 & 97.10±0.37\\
    \multicolumn{1}{c}{} & \multicolumn{1}{c|}{Metal Nut}  & 100.00±0.00 & 100.00±0.00 & 100.00±0.00 & 97.12±0.11 & 84.74±0.67 & 82.10±0.77 & 95.16±0.19\\
    \multicolumn{1}{c}{} & \multicolumn{1}{c|}{Pill}  & 97.42±0.62 & 99.48±0.16 & 97.52±0.45 & 96.54±0.23 & 65.42±1.19 & 65.68±0.70 & 98.12±0.08\\
    \multicolumn{1}{c}{} & \multicolumn{1}{c|}{Screw} & 
    87.96±1.43 & 95.38±0.60 & 90.84±1.20 & 98.76±0.05 & 54.28±2.78 & 57.18±2.21 & 94.74±0.66\\
\bottomrule
    \end{tabular}%
  }
\end{table*}

\begin{table*}[htbp]
  \centering
  \caption{\textbf{Two-shot} evaluation for \textbf{multimodal RGB–Text UAD on MVTec-AD~\cite{mvtec}}, reporting per-category results for \textbf{APRILGAN+UCF}. Results are reported as mean ± standard deviation over five seeds. This table corresponds to Sec.\ref{suppsec4b} and complements Table~\ref{suppfewshotmvtec}.}
  \label{suppfewshot_mvtec2}%
  \renewcommand{\arraystretch}{1.1}
  \resizebox{1\linewidth}{!}{
    \begin{tabular}{c|c|c|c|c|c|c|c|c}
    \toprule
   \multicolumn{2}{r|}{Category~$\downarrow$} & \textbf{I-AUROC} & \textbf{I-AP} & \textbf{I-F1-max} & \textbf{P-AUROC} & \textbf{P-AP} & \textbf{P-F1-max} & \textbf{AUPRO} \\
    \cline{1-2}
    \hline
    \multicolumn{1}{c}{\multirow{10}[1]{*}{\begin{turn}{90}Objects\end{turn}}} 
    & \multicolumn{1}{c|}{Bottle}     & 99.78±0.13 & 99.96±0.05 & 99.04±0.36 & 97.62±0.08 & 82.44±0.38 & 76.80±0.29 & 95.16±0.17 \\
    \multicolumn{1}{c}{} & \multicolumn{1}{c|}{Cable} & 95.72±0.83 & 97.62±0.46 & 91.28±0.95 & 95.88±0.18 & 61.18±0.49 & 60.02±0.29 & 90.18±0.27 \\
    \multicolumn{1}{c}{} & \multicolumn{1}{c|}{Capsule}  & 97.70±0.50 & 99.52±0.08 & 96.66±0.61 & 97.96±0.29 & 47.88±0.84 & 47.08±0.26 & 97.40±0.21 \\
    \multicolumn{1}{c}{} & \multicolumn{1}{c|}{Hazelnut} & 100.00±0.00 & 100.00±0.00 & 99.72±0.38 & 99.04±0.17 & 78.12±1.04 & 73.56±0.80 & 97.24±0.24 \\
    \multicolumn{1}{c}{} & \multicolumn{1}{c|}{Metal Nut}  & 100.00±0.00 & 100.00±0.00 & 100.00±0.00 & 97.44±0.26 & 85.94±0.75 & 83.70±0.85 & 95.56±0.31 \\
    \multicolumn{1}{c}{} & \multicolumn{1}{c|}{Pill}  & 97.88±0.82 & 99.56±0.23 & 97.74±0.22 & 96.72±0.15 & 66.46±0.56 & 66.52±0.33 & 98.20±0.07 \\
    \multicolumn{1}{c}{} & \multicolumn{1}{c|}{Screw} & 91.74±2.61 & 97.00±1.13 & 92.78±1.24 & 98.96±0.17 & 57.18±3.35 & 59.02±2.36 & 95.42±0.50 \\
\bottomrule
    \end{tabular}%
  }
\end{table*}

\begin{table*}[htbp]
  \centering
  \caption{\textbf{Four-shot} evaluation for \textbf{multimodal RGB–Text UAD on MVTec-AD~\cite{mvtec}}, reporting per-category results for \textbf{APRILGAN+UCF}. Results are reported as mean ± standard deviation over five seeds. This table corresponds to Sec.\ref{suppsec4b} and complements Table~\ref{suppfewshotmvtec}.}
  \label{suppfewshot_mvtec4}%
  \renewcommand{\arraystretch}{1.1}
  \resizebox{1\linewidth}{!}{
    \begin{tabular}{c|c|c|c|c|c|c|c|c}
    \toprule
    \multicolumn{2}{r|}{Category~$\downarrow$} & \textbf{I-AUROC} & \textbf{I-AP} & \textbf{I-F1-max} & \textbf{P-AUROC} & \textbf{P-AP} & \textbf{P-F1-max} & \textbf{AUPRO} \\
    \cline{1-2}
    \hline
    \multicolumn{1}{c}{\multirow{10}[1]{*}{\begin{turn}{90}Objects\end{turn}}} 
    & \multicolumn{1}{c|}{Bottle}     & 99.80±0.12 & 99.92±0.04 & 99.20±0.57 & 97.74±0.05 & 82.66±0.21 & 77.12±0.19 & 95.38±0.13 \\
    \multicolumn{1}{c}{} & \multicolumn{1}{c|}{Cable} & 96.28±0.86 & 97.90±0.50 & 92.20±1.27 & 96.44±0.32 & 62.80±0.80 & 61.06±0.39 & 91.16±0.24 \\
    \multicolumn{1}{c}{} & \multicolumn{1}{c|}{Capsule}  & 98.42±0.33 & 99.66±0.05 & 97.16±0.35 & 98.38±0.11 & 50.00±0.58 & 48.64±0.42 & 97.82±0.13 \\
    \multicolumn{1}{c}{} & \multicolumn{1}{c|}{Hazelnut} & 100.00±0.00 & 100.00±0.00 & 100.00±0.00 & 99.20±0.07 & 79.14±0.45 & 74.60±0.59 & 97.62±0.15 \\
    \multicolumn{1}{c}{} & \multicolumn{1}{c|}{Metal Nut}  & 100.00±0.00 & 100.00±0.00 & 100.00±0.00 & 97.72±0.08 & 86.66±0.32 & 84.64±0.64 & 96.04±0.13 \\
    \multicolumn{1}{c}{} & \multicolumn{1}{c|}{Pill}  & 98.10±0.43 & 99.64±0.09 & 98.08±0.16 & 97.04±0.11 & 67.64±0.32 & 67.38±0.19 & 98.30±0.07 \\
    \multicolumn{1}{c}{} & \multicolumn{1}{c|}{Screw} & 92.86±1.07 & 97.44±0.43 & 92.48±1.08 & 99.06±0.05 & 59.30±1.02 & 60.48±0.58 & 95.90±0.20 \\
\bottomrule
    \end{tabular}%
  }
\end{table*}

\begin{table*}[htbp]
  \centering
  \caption{\textbf{One-shot} evaluation for \textbf{multimodal RGB–Text UAD on VisA~\cite{visa}}, reporting per-category results for \textbf{APRILGAN+UCF}. Results are reported as mean ± standard deviation over five seeds. This table corresponds to Sec.\ref{suppsec4b} and complements Table~XIII in the main paper.}
  \label{suppfewshot_visa1}%
  \renewcommand{\arraystretch}{1.1}
  \resizebox{1\linewidth}{!}{
    %
%
  }
\end{table*}

\begin{table*}[htbp]
  \centering
  \caption{\textbf{Two-shot} evaluation for \textbf{multimodal RGB–Text UAD on VisA~\cite{visa}}, reporting per-category results for \textbf{APRILGAN+UCF}. Results are reported as mean ± standard deviation over five seeds. This table corresponds to Sec.\ref{suppsec4b} and complements Table~XIII in the main paper.}
  \label{suppfewshot_visa2}%
  \renewcommand{\arraystretch}{1.1}
  \resizebox{1\linewidth}{!}{
    %
%
  }
\end{table*}

\begin{table*}[htbp]
  \centering
  \caption{\textbf{Four-shot} evaluation for \textbf{multimodal RGB–Text UAD on VisA~\cite{visa}}, reporting per-category results for \textbf{APRILGAN+UCF}. Results are reported as mean ± standard deviation over five seeds. This table corresponds to Sec.\ref{suppsec4b} and complements Table~XIII in the main paper.}
  \label{suppfewshot_visa4}%
  \renewcommand{\arraystretch}{1.1}
  \resizebox{1\linewidth}{!}{
    %
%
  }
\end{table*}

\clearpage

\FloatBarrier

\FloatBarrier

\FloatBarrier
\clearpage
\bibliographystyle{IEEEtran}
\bibliography{egbib}

\begin{IEEEbiography}[{\includegraphics[width=1in,height=1.25in,clip,keepaspectratio]{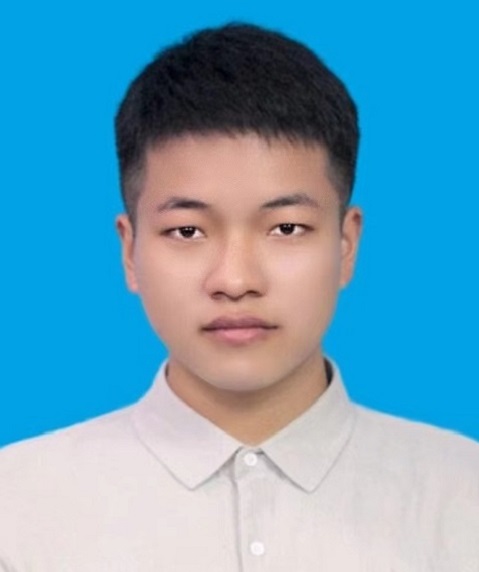}}]{Zhe Zhang}
	received the BS degree in the College of Information Science and Engineering, Northeastern University, China, in 2021. He is currently working toward a Ph.D. degree in the State Key Laboratory of Synthetical Automation for Process Industries, Northeastern University, Shenyang, China. His current research interests include computer vision and deep learning, with a focus on anomaly detection, domain adaptation, video representation, zero-shot and few-shot learning, multi-modal learning, and their applications in dynamic and open environments.
\end{IEEEbiography}

\begin{IEEEbiography}[{\includegraphics[width=1in,height=1.25in,clip,keepaspectratio]{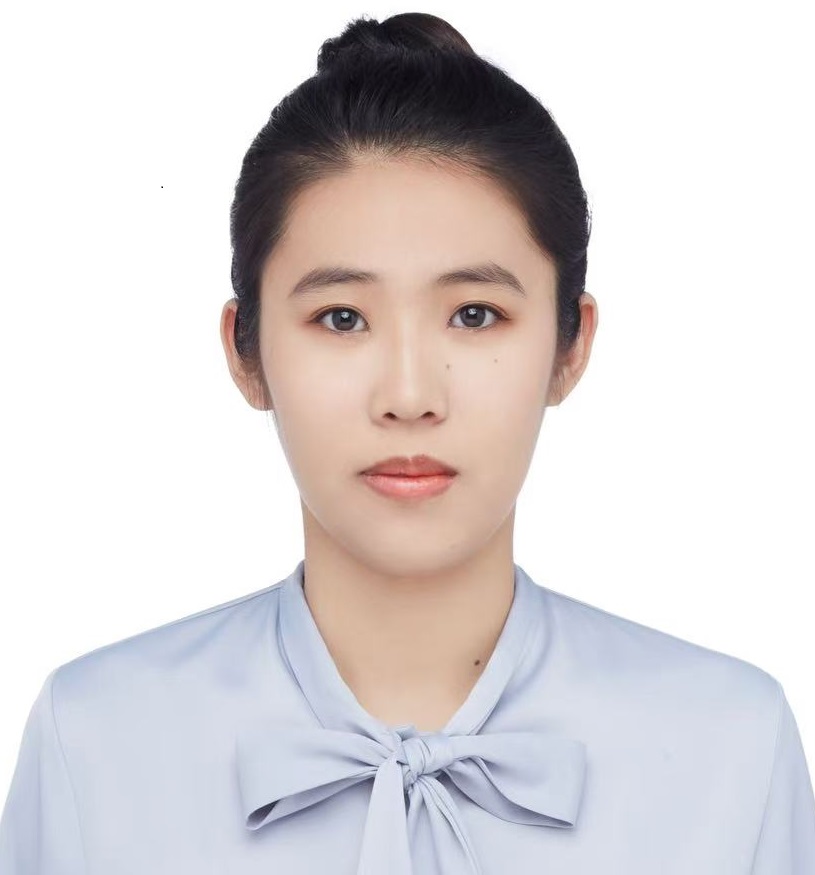}}]{Mingxiu Cai} received the MS degree from the School of Computer Science, Nanjing Audit University, China, in 2024.
	 She is currently working toward a Ph.D. degree in the State Key Laboratory of Synthetical Automation for Process Industries, Northeastern University, Shenyang, China. Her current research interests include computer vision, deep learning, multi-modal learning, and their applications in industrial fields.
\end{IEEEbiography}

\begin{IEEEbiography}[{\includegraphics[width=1in,height=1.25in,clip,keepaspectratio]{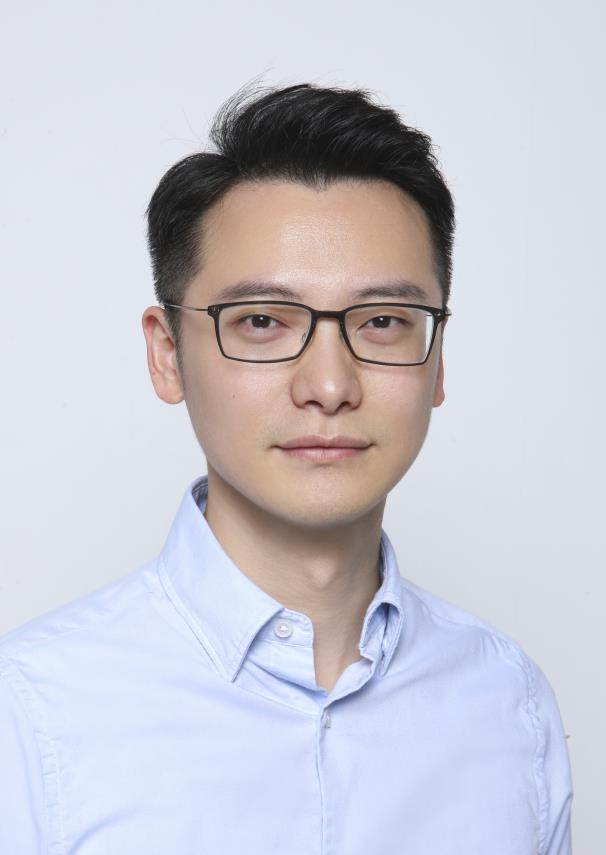}}]{Gaochang Wu}(IEEE Member)
received the BE and MS degrees in mechanical engineering in Northeastern University, Shenyang, China, in 2013 and 2015, respectively, and Ph.D. degree in control theory and control engineering in Northeastern University, Shenyang, China in 2020. He is currently an associate professor in the State Key Laboratory of Synthetical Automation for Process Industries, Northeastern University. He was selected for the 2022-2024 Youth Talent Support Program of the Chinese Association of Automation. His current research interests include multimodal perception and recognition, light field imaging and processing, and computer vision in industrial scenarios.
\end{IEEEbiography}

\begin{IEEEbiography}[{\includegraphics[width=1in,height=1.25in,clip,keepaspectratio]{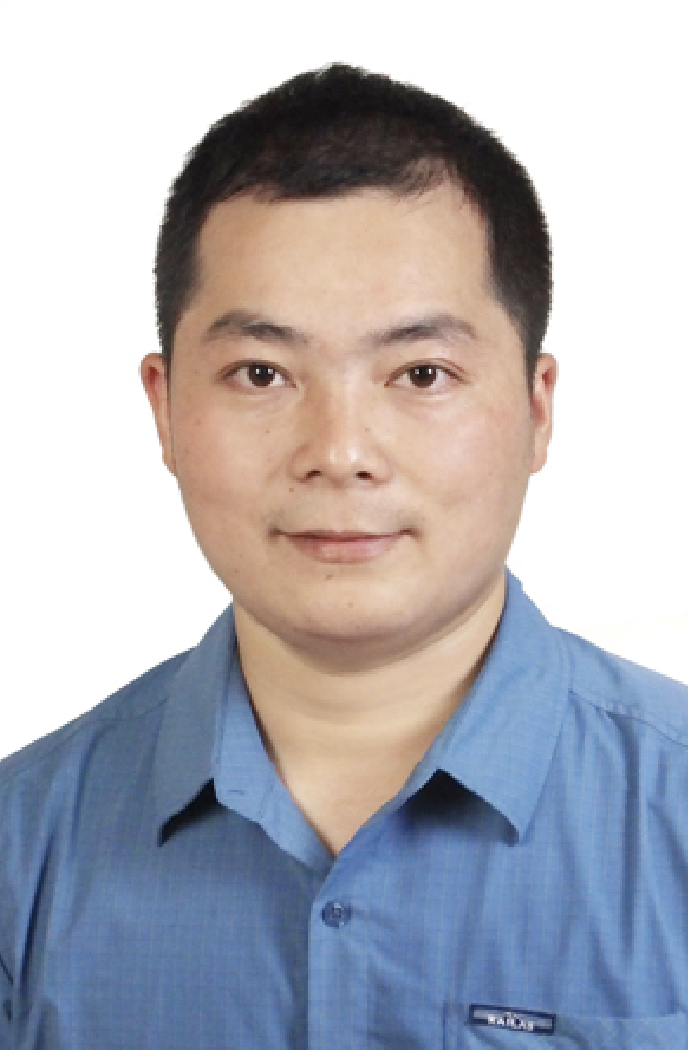}}]{Jing Zhang} (Senior Member, IEEE) is currently a professor at the School of Computer Science, Wuhan University, Wuhan, China. He previously served as a Research Fellow at the School of Computer Science, The University of Sydney. He has published over 100 papers in leading venues such as CVPR, NeurIPS, IEEE TPAMI, and IJCV, with research focused on computer vision and deep learning. He is an Area Chair for NeurIPS and ICLR, a Senior Program Committee member for AAAI and IJCAI, and a guest editor for IEEE TBD, while also regularly reviewing for top-tier journals and conferences.
\end{IEEEbiography}

\begin{IEEEbiography}[{\includegraphics[width=1in,height=1.25in,clip,keepaspectratio]{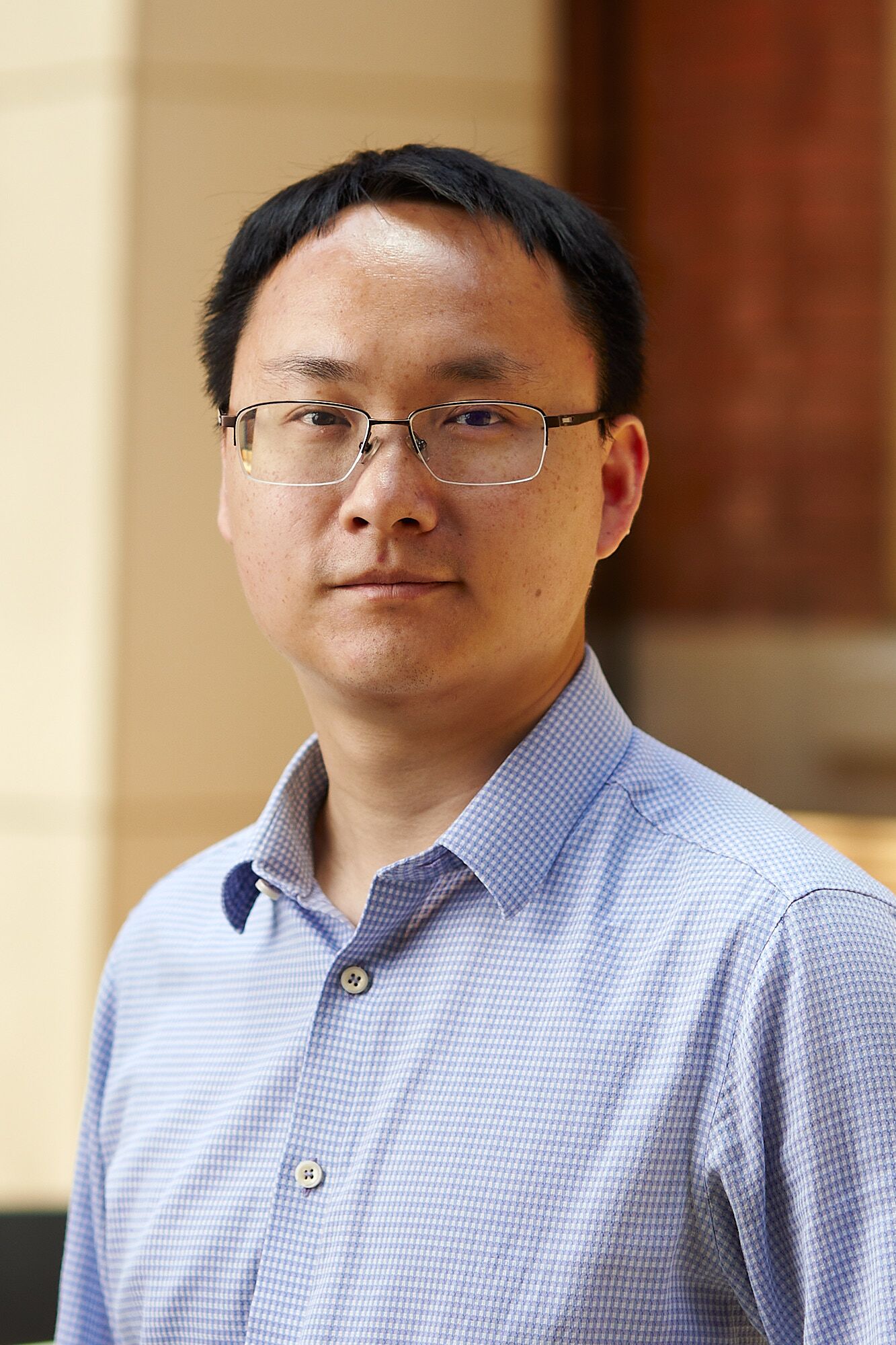}}]{Lingqiao Liu} received the Ph.D. degree from the Australian National University, Canberra, in 2014. He is an associate professor with the University of Adelaide and the Australian Institute for Machine Learning. He was awarded the Discovery Early Career Researcher Award from the Australian Research Council and the University Research Fellow from the University of Adelaide in 2016. His current research interests include compositional zero-shot learning, vision and language, and various topics in computer vision and natural language processing.
\end{IEEEbiography}

\begin{IEEEbiography}[{\includegraphics[width=1in,height=1.25in,clip,keepaspectratio]{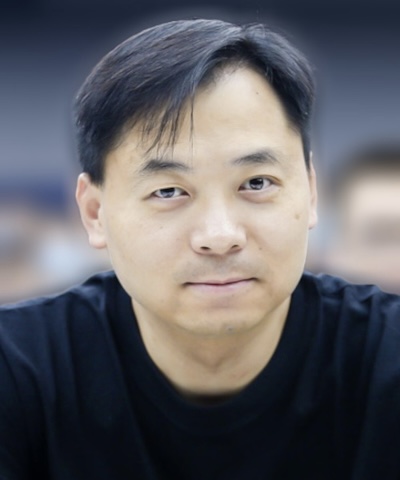}}]{Dacheng Tao} (Fellow, IEEE) is currently a Distinguished University Professor in the College of Computing \& Data Science at Nanyang Technological University. He mainly applies statistics and mathematics to artificial intelligence and data science, and his research is detailed in one monograph and over 200 publications in prestigious journals and proceedings at leading conferences, with best paper awards, best student paper awards, and test-of-time awards. His publications have been cited over 112K times and he has an h-index 160+ in Google Scholar. He received the 2015 and 2020 Australian Eureka Prize, the 2018 IEEE ICDM Research Contributions Award, and the 2021 IEEE Computer Society McCluskey Technical Achievement Award. He is a Fellow of the Australian Academy of Science, AAAS, ACM and IEEE.
\end{IEEEbiography}

\begin{IEEEbiography}[{\includegraphics[width=1in,height=1.25in,clip,keepaspectratio]{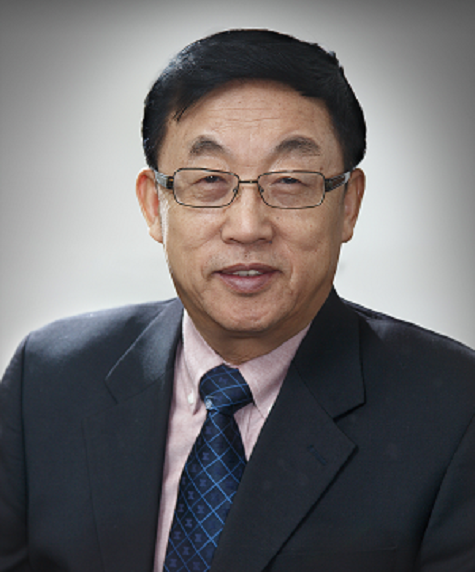}}]{Tianyou Chai}
	(Life Fellow, IEEE) received the Ph.D. degree in control theory and engineering from Northeastern University, Shenyang, China, in 1985. He became a Professor at Northeastern University in 1988. He is the Founder and the Director of the Center of Automation, Northeastern University, which became the National Engineering and Technology Research Center and the State Key Laboratory. He was the Director of the Department of Information Science, National Natural Science Foundation of China, from 2010 to 2018. He has developed control technologies with applications to various industrial processes. He has published more than 320 peer-reviewed international journal articles. His current research interests include modeling, control, optimization, and integrated automation of complex industrial processes.
	
	Dr. Chai is a member of the Chinese Academy of Engineering and a Fellow of International Federation for Automatic Control (IFAC). His paper titled “Hybrid intelligent control for optimal operation of shaft furnace roasting process” was selected as one of the three best papers for the Control Engineering Practice Paper Prize for the term 2011–2013. For his contributions, he has won five prestigious awards of the National Natural Science, the National Science and Technology Progress, and the National Technological Innovation, the 2007 Industry Award for Excellence in Transitional Control Research from IEEE Multi-Conference on Systems and Control, and the 2017 Wook Hyun Kwon Education Award from the Asian Control Association. 
\end{IEEEbiography}

\begin{IEEEbiography}[{\includegraphics[width=1in,height=1.25in,clip,keepaspectratio]{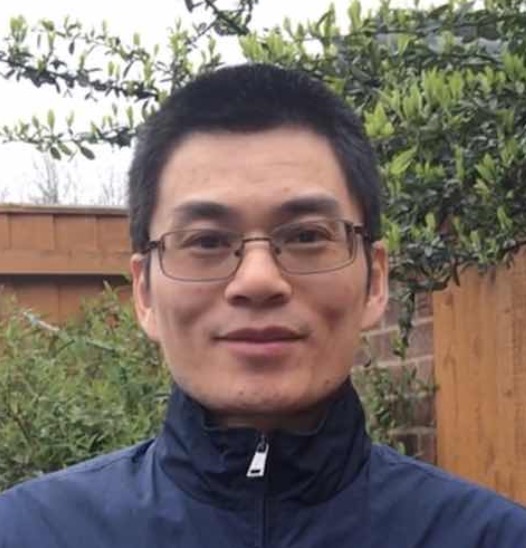}}]{Xiatian Zhu} received the Ph.D. degree from the Queen Mary University of London, London, U.K. He was a Research Scientist with Samsung AI Centre, Cambridge, U.K. He is currently a Senior Lecturer with the Surrey Institute for People-Centred Artificial Intelligence and also with the Centre for Vision, Speech and Signal Processing, University of Surrey, Guildford, U.K. His research interests include computer vision and machine learning. He was the recipient of the Sullivan Doctoral Thesis Prize 2016.
\end{IEEEbiography}

\end{document}